\RequirePackage[l2tabu, orthodox]{nag}

\documentclass[12pt,phd,a4paper,oneside]{ucl_thesis}

\usepackage{blindtext}
\usepackage[english]{babel}
\usepackage{emptypage}

\usepackage{graphicx}
\usepackage{bm}
\usepackage{booktabs}
\usepackage{algorithm} 
\usepackage{algpseudocode}
\usepackage{amsfonts}
\usepackage{xcolor}

\usepackage{float}
\usepackage{rotating}
\usepackage{amsmath}
\usepackage{enumitem}
\usepackage{gensymb}
\usepackage{textcomp}
\usepackage{afterpage}

\usepackage{bm}

\usepackage{setspace}
\usepackage{multirow}

\usepackage{bibentry} 

\usepackage[format=hang,font=small,labelfont=bf]{caption}

\usepackage{etoolbox}
\usepackage{graphicx}

\usepackage{mathptmx}
\usepackage[scaled=.90]{helvet}
\usepackage{courier}

\newcommand{\CalibMetric}{CA}
\newcommand{\IRNNQs}{IRNN$_0$}
\newcommand{\IRNN}{IRNN}

\usepackage{bibentry}
\makeatletter\let\saved@bibitem\@bibitem\makeatother
\usepackage[pdftex,hidelinks]{hyperref}
\makeatletter\let\@bibitem\saved@bibitem\makeatother
\makeatletter
\AtBeginDocument{
    \hypersetup{
        pdfsubject={Thesis Subject},
        pdfkeywords={Thesis Keywords},
        pdfauthor={Author},
        pdftitle={Title},
    }
}
\makeatother

\usepackage[automake]{glossaries-extra}

\newglossary[alg]{acronym}{acr}{acn}{List of Acronyms}
\newglossary[slg]{symbols}{sym}{sbl}{List of Symbols}

\newglossarystyle{acronyms}{%
  \setglossarystyle{list}% base this style on the list style
  \setlength{\itemsep}{0pt}% reduce line spacing
}

\newglossarystyle{symbols}{%
  \setglossarystyle{list}% base this style on the list style
  \setlength{\itemsep}{0pt}% reduce line spacing
}

\newacronym[type=acronym]{rsv}{RSV}{Respiratory Syncytial Virus}
\newacronym[type=acronym]{nn}{NN}{Neural Network}
\newacronym[type=acronym]{ili}{ILI}{Influenza-Like Illness}
\newacronym[type=acronym]{cdc}{CDC}{Centers for Disease Control}
\newacronym[type=acronym]{wili}{wILI}{Weighted Influenza-Like Illness}
\newacronym[type=acronym]{mae}{MAE}{Mean Absolute Error}
\newacronym[type=acronym]{ode}{ODE}{Ordinary Differential Equation}
\newacronym[type=acronym]{ude}{UDE}{Universal Differential Equation}
\newacronym[type=acronym]{rnn}{RNN}{Recurrent Neural Network}
\newacronym[type=acronym]{bnn}{BNN}{Bayesian Neural Network}
\newacronym[type=acronym]{nll}{NLL}{Negative Log-Likelihood}
\newacronym[type=acronym]{lstm}{LSTM}{Long Short-Term Memory}
\newacronym[type=acronym]{gru}{GRU}{Gated Recurrent Unit}
\newacronym[type=acronym]{kl}{KL}{Kullback-Leibler (divergence)}
\newacronym[type=acronym]{elbo}{ELBO}{Evidence Lower Bound}
\newacronym[type=acronym]{mse}{MSE}{Mean Squared Error}
\newacronym[type=acronym]{mad}{MAD}{Mean Absolute Deviation}
\newacronym[type=acronym]{arima}{ARIMA}{Auto-Regressive Integrated Moving Average}
\newacronym[type=acronym]{sarima}{SARIMA}{Seasonal Auto-Regressive Integrated Moving Average}
\newacronym[type=acronym]{mcmc}{MCMC}{Markov Chain Monte Carlo}
\newacronym[type=acronym]{gft}{GFT}{Google Flu Trends}
\newacronym[type=acronym]{gp}{GP}{Gaussian Process}
\newacronym[type=acronym]{rk4}{RK4}{Runge-Kutta Fourth Order}
\newacronym[type=acronym]{crps}{CRPS}{Continuous Ranked Probability Score}
\newacronym[type=acronym]{ci}{CI}{Confidence Interval}
\newacronym[type=acronym]{sir}{SIR}{Susceptible-Infected-Recovered (model)}
\newacronym[type=acronym]{seir}{SEIR}{Susceptible-Exposed-Infected-Recovered (model)}
\newacronym[type=acronym]{n_ode}{N-ODE}{Neural Ordinary Differential Equation}
\newacronym[type=acronym]{vae}{VAE}{Variational Autoencoder}

\newglossaryentry{Fp}{
  type=symbols,
  name={$ \text{Fp} $},
  description={Physical component of a universal differential equation}
}
\newglossaryentry{Fa}{
  type=symbols,
  name={$ \text{Fa} $},
  description={Augmentation component of a universal differential equation}
}

\newacronym[type=acronym]{ODEB}{$\text{ODE}_{\text{B}}$}{Basic neural ODE model used for forecasting}
\newacronym[type=acronym]{ODEBQ}{$\text{ODE}_{\text{B}~\text{Q}}$}{Basic neural ODE model not using web search data, used for forecasting}
\newacronym[type=acronym]{SIRB}{$\text{SIR}_{\text{B}}$}{Basic SIR model used for forecasting}
\newacronym[type=acronym]{SIRadv}{$\text{SIR}_{\text{Adv}}$}{SIR model with parameters estimated by a neural network, used for forecasting}
\newacronym[type=acronym]{SIRadvU}{$\text{SIR}_{\text{Adv}~\text{U}}$}{SIR-based universal differential equation with parameters estimated by a neural network, used for forecasting}
\newacronym[type=acronym]{SIRadvQ}{$\text{SIR}_{\text{Adv}~\text{Q}}$}{SIR model with parameters estimated by a neural network, not using web search data, used for forecasting}
\newacronym[type=acronym]{SEIRadv}{$\text{SEIR}_{\text{Adv}}$}{SEIR model with parameters estimated by a neural network, used for forecasting}
\newacronym[type=acronym]{SEIRadvU}{$\text{SEIR}_{\text{Adv}~\text{U}}$}{SEIR-based universal differential equation with parameters estimated by a neural network, used for forecasting}
\newacronym[type=acronym]{r}{$r$}{Bivariate correlation}

\newacronym[type=acronym]{NNa}{NN$_a$}{Neural network without using web search activity data beyond $t_0$}
\newacronym[type=acronym]{NNb}{NN$_b$}{Neural network without using web search activity data beyond $t_0$, using leave-one-out training}
\newacronym[type=acronym]{IRNN0}{IRNN$_0$}{Iterative Recurrent Neural Network without web search data}
\newacronym[type=acronym]{irnn}{IRNN}{Iterative Recurrent Neural Network}
\newacronym[type=acronym]{irnns}{IRNN$_s$}{Iterative Recurrent Neural Network - sampling}
\newacronym[type=acronym]{ff}{FF}{Feed-Forward (Neural Network)}
\newacronym[type=acronym]{srnn}{SRNN}{Simple Recurrent Neural Network}

\newglossaryentry{gamma}{
  type=symbols,
  name={$\gamma$},
  description={Forecast horizon}
}
\newglossaryentry{y}{
  type=symbols,
  name={$y$},
  description={Output}
}
\newglossaryentry{t}{
  type=symbols,
  name={$t$},
  description={Time}
}
\newglossaryentry{delta}{
  type=symbols,
  name={$\delta$},
  description={Delay in days for ILI data collection}
}
\newglossaryentry{t0}{
  type=symbols,
  name={$t_0$},
  description={Initial time (time of last available ILI rate)}
}
\newglossaryentry{x}{
  type=symbols,
  name={$x$},
  description={Input}
}
\newglossaryentry{N}{
  type=symbols,
  name={$N$},
  description={Size of dataset}
}
\newglossaryentry{n}{
  type=symbols,
  name={$n$},
  description={Number of compartments in a compartmental model}
}
\newglossaryentry{epsilon}{
  type=symbols,
  name={$\epsilon$},
  description={Additive noise}
}
\newglossaryentry{sigma}{
  type=symbols,
  name={$\sigma$},
  description={Standard deviation}
}
\newglossaryentry{hat_y}{
  type=symbols,
  name={$\hat{y}$},
  description={Prediction (of a model)}
}
\newglossaryentry{bmPhi}{
  type=symbols,
  name={$\bm{\Phi}$},
  description={Parameters (of a model)}
}
\newglossaryentry{A}{
  type=symbols,
  name={$\mathbf{A}$},
  description={Design matrix}
}
\newglossaryentry{K}{
  type=symbols,
  name={$K$},
  description={Number of samples (when making a prediction)}
}
\newglossaryentry{zeta}{
  type=symbols,
  name={$\zeta$},
  description={Precision ($1/\sigma$)}
}
\newglossaryentry{iota}{
  type=symbols,
  name={$\iota$},
  description={Precision hyperparameter}
}
\newglossaryentry{D}{
  type=symbols,
  name={$\mathbf{D}$},
  description={Dataset (containing inputs $\mathbf{x}$ and outputs $\mathbf{y}$)}
}
\newglossaryentry{W}{
  type=symbols,
  name={$\mathbf{W}$},
  description={Weights}
}
\newglossaryentry{b}{
  type=symbols,
  name={$\mathbf{b}$},
  description={Biases}
}
\newglossaryentry{g}{
  type=symbols,
  name={$g$},
  description={Activation function}
}
\newglossaryentry{loss}{
  type=symbols,
  name={$\mathcal{L}$},
  description={Loss function}
}
\newglossaryentry{DKL}{
  type=symbols,
  name={$D\text{KL}$},
  description={Kullback-Leibler divergence}
}
\newglossaryentry{lambda}{
  type=symbols,
  name={$\lambda$},
  description={Weighting (in a loss function)}
}
\newglossaryentry{I}{
  type=symbols,
  name={$I$},
  description={Number of units in a neural network layer}
}
\newglossaryentry{theta}{
  type=symbols,
  name={$\mathbf{\theta}$},
  description={Parameters which describe the posterior in variational inference}
}
\newglossaryentry{mu}{
  type=symbols,
  name={$\mu$},
  description={Mean}
}
\newglossaryentry{F_input}{
  type=symbols,
  name={$F$},
  description={ILI rates, used as an input}
}
\newglossaryentry{tau}{
  type=symbols,
  name={$\tau$},
  description={Window size}
}
\newglossaryentry{Q_input}{
  type=symbols,
  name={$Q$},
  description={Query frequencies, used as an input}
}
\newglossaryentry{F}{
  type=symbols,
  name={$F$},
  description={ILI proportions in an input}
}
\newglossaryentry{Q_family}{
  type=symbols,
  name={$\mathcal{Q}$},
  description={Family of distributions for variational inference}
}
\newglossaryentry{ht}{
  type=symbols,
  name={$\mathbf{h}_t$},
  description={Hidden state (in a recurrent neural network)}
}
\newglossaryentry{Ct}{
  type=symbols,
  name={$C_t$},
  description={Memory cell (in a recurrent neural network)}
}
\newglossaryentry{hat_sigma}{
  type=symbols,
  name={$\hat{\sigma}$},
  description={Estimated standard deviation}
}
\newglossaryentry{Lu}{
  type=symbols,
  name={$L$},
  description={Number of units in a neural network layer}
}
\newglossaryentry{m}{
  type=symbols,
  name={$m$},
  description={Number of web searches used in an input}
}
\newglossaryentry{time_step}{
  type=symbols,
  name={$h$},
  description={Time step for ODE model}
}
\newglossaryentry{s_i_r}{
  type=symbols,
  name={$s$, $i$, and $r$},
  description={Susceptible, infected, and recovered fractions in an SIR model}
}
\newglossaryentry{S_I_R}{
  type=symbols,
  name={$S$, $I$, and $R$},
  description={Susceptible, infected, and recovered populations in an SIR model}
}
\newglossaryentry{N_pop}{
  type=symbols,
  name={$N_\text{pop}$},
  description={Population size in a compartmental model}
}
\newglossaryentry{R_e}{
  type=symbols,
  name={$R_e$},
  description={Effective reproductive number. Average number of secondary infections an infected individual will produce before recovering}
}
\newglossaryentry{R_0}{
  type=symbols,
  name={$R_0$},
  description={Basic reproductive number. Number of secondary infections the average infectious person would produce in a fully susceptible population}
}

\newglossaryentry{z}{
  type=symbols,
  name={$z$},
  description={Latent variables in a VAE}
}
\newglossaryentry{beta}{
  type=symbols,
  name={$ \beta $},
  description={In an SIR model, rate at which infected individuals transmit the infection to susceptible individuals per unit time}
}
\newglossaryentry{omega}{
  type=symbols,
  name={$ \omega $},
  description={In an SIR model, probability of an infected individual recovering per unit time}
}
\newglossaryentry{rho}{
  type=symbols,
  name={$ \rho $},
  description={In an SIR model, rate of movement from the exposed population to the infected population}
}
\newglossaryentry{kappa}{
  type=symbols,
  name={$ \kappa $},
  description={Weighting for the norm of the augmentation component in the loss function of a universal differential equation}
}

\makeglossaries

\begin{document}

% Use acronyms in the text

\nobibliography*

% I may change the way this is done in a future version, 
%  but given that some people needed it, if you need a different degree title 
%  (e.g. Master of Science, Master in Science, Master of Arts, etc)
%  uncomment the following 3 lines and set as appropriate (this *has* to be before \maketitle)
% \makeatletter
% \renewcommand {\@degree@string} {Master of Things}
% \makeatother
% \pagenumbering{Roman}
\title{Forecasting infectious disease prevalence with associated uncertainty using neural networks}
\author{Michael Morris}
\department{Department of Computer Science}

\maketitle
% \makedeclaration

\clearpage
I, Michael Morris confirm that the work presented in my thesis is my own. Where information has been derived from other sources, I confirm that this has been indicated in the thesis.

\begin{abstract}
Infectious diseases pose significant human and economic burdens. Accurately forecasting disease incidence can enable public health agencies to respond effectively to existing or emerging diseases. Despite progress in the field, developing accurate forecasting models remains a significant challenge. This thesis proposes two methodological frameworks using neural networks (NNs) with associated uncertainty estimates --- a critical component limiting the application of NNs to epidemic forecasting thus far. We develop our frameworks by forecasting influenza-like illness (ILI) in the United States.
 
    Our first proposed method uses Web search activity data in conjunction with historical ILI rates as observations for training NN architectures. Our models incorporate Bayesian layers to produce uncertainty intervals, positioning themselves as legitimate alternatives to more conventional approaches. The best performing architecture: iterative recurrent neural network (IRNN), reduces mean absolute error by $10.3\%$ and improves Skill by $17.1\%$ on average in forecasting tasks across four flu seasons compared to the state-of-the-art. We build on this method by introducing IRNN$_s$, an architecture which changes the sampling procedure in the IRNN to improve the uncertainty estimation.
 
    Our second framework uses neural ordinary differential equations to bridge the gap between mechanistic compartmental models and NNs; benefiting from the physical constraints that compartmental models provide. 
    We evaluate eight neural ODE models utilising a mixture of ILI rates and Web search activity data to provide forecasts. These are compared with the IRNN and IRNN$_0$ --- the IRNN using only ILI rates. Models trained without Web search activity data outperform the IRNN$_0$ by $16\%$ in terms of Skill. Future work should focus on more effectively using neural ODEs with Web search data to compete with the best performing IRNN.
 
    These frameworks are a step forward in epidemic forecasting, offering accurate predictions with credible uncertainty estimates, and providing useful tools for public health agencies in combating infectious diseases.
\end{abstract}

\chapter*{Impact Statement}
The work presented in this thesis has had or has the potential to have social and academic impacts.

There is a potential for significant social impact; modelling the incidence of an infectious disease enables public health organisations to prepare for and minimise the disease's impact. Work presented in Chapter~\ref{chapterlabel1}  achieves state-of-the-art accuracy in influenza-like-illness forecasting models by incorporating Web search query data with neural networks, thereby avoiding the significant delay that exists in collecting disease rates. In addition, the proposed Bayesian neural network architectures provide associated uncertainty estimates for the forecasts, positioning this methodology as a practical complementary tool for disease surveillance and policy making. Further, the work in Chapter~\ref{chapterlabel1} has had an academic impact, having been published in a respected journal. 

Work presented in Chapter~\ref{chapterlabel2} bridges the gap between existing mechanistic disease models and neural networks. We demonstrate a method that benefits from the advantages of both kinds of models. This has the potential for social impacts by providing more accurate and less data-intensive forecasts. 

\begin{acknowledgements}

I am deeply grateful to Dr Vasileios Lampos for his invaluable support and advice throughout my PhD. I would also like to thank Professor Ingemar J. Cox for his guidance. Thank you to Natasha Branch who supported and believed in me through the highs and lows, and spent countless hours proofreading my thesis and listening to explanations and complaints about my work. Thanks to my friends and family who have provided support, ideas, proofreading, and welcome distractions without which I would no doubt have not made it this far.

\end{acknowledgements}

\setcounter{tocdepth}{2} 
% Setting this higher means you get contents entries for
%  more minor section headers.

\tableofcontents
\listoffigures
\listoftables

\printglossary[type=acronym, style=acronyms]
\printglossary[type=symbols, style=symbols]

\chapter{Introduction}

% 1. Motivate why we want to forecast infectious diseases
% 2. Forecasting can be based on historical data for the thing being forecasted and exogenous variables
% 3. What exogenous data is useful
% Mechanistic vs non-mechanistic models
% 	- mechanistic aren’t as well suited to exogenous data
% 	- mechanistic models facilitate an understanding of the underlying processes …. Conversely ….
% 	- less rambley
% 4. Need an associated uncertainty
% 5. Where uncertainty comes from
% 6. It’s difficult to accurately measure the prevalence of influenza in a population, instead people estimate the prevalence of ILI, defined as … 

% Define other acronyms similarly

\label{chapterlabelIntro}
Infectious diseases are responsible for huge social and economic costs and are among the top causes of human illness and death worldwide. Respiratory infections such as Influenza, Respiratory Syncytial Virus (\gls{rsv}) and Covid-19, and diarrheal diseases including Rotavirus, Escherichia-coli and Cholera are among the most common and dangerous infectious diseases, each with the potential to become epidemic. As humanity becomes more connected and the climate warms, pathogens will become more virulent and common \cite{patz2005impact}. Infectious disease forecasts can be used by public health agencies to decide if and what interventions should be used to limit infections.  Moreover, forecasts serve as a preemptive alert to front-line health workers, ensuring they are better prepared for potential outbreaks.

Forecasts can be based on historical values of the target time series, or on exogenous variables, i.e. variables which are determined outside of the time series being forecasted. Weather data, mobility data, medicine sales data and web search query data are just some of the exogenous variables which modellers use for epidemic forecasting. 
Web search query data records the frequency at which people search for terms in a set period. For example, the frequency of people searching for the term ``flu symptoms'' each day. 
Case counts are often reported with a delay, 
whereas Web search data can be collected immediately so can give more timely indications of an outbreak of a disease. Work on nowcasting influenza using web search query data \cite{lampos2017enhancing} highlighted its utility for improving model accuracy. We build on this work using similar query selection methods for forecasting using neural networks. 

Neural networks are a family of non-mechanistic models, meaning that they learn patterns directly from data without prior knowledge of what generated that data. Another family of models are mechanistic, which explicitly models the physical processes which generate the data they are modelling. For example, Newtonian mechanics can be used to estimate how a spring will compress or extend under an external force. 

Mechanistic and non-mechanistic models have different advantages and disadvantages. Non-mechanistic models tend to be better suited to modelling complex dynamics and using exogenous data. Mechanistic models facilitate an understanding of the underlying process behind a forecast.

Mechanistic models rely on having a thorough understanding of the systems they are representing. However, real-world systems often have complex dynamics which are not understood. Compartmental models use simplifying assumptions --- for example that members of a population mix uniformly --- but these limit how closely the models can fit to data. 
To overcome this lack of flexibility, models can be modified with non-mechanistic components to approximate the error between the model output and the target~\cite{osthus2019dynamic,rackauckas2020universal}. Alternatively, a proxy dataset which is closer to the model output can be used~\cite{shaman2013real}, however, this requires understanding the relationship between the proxy and the true values. 

In epidemiology, non-mechanistic models are generally more flexible than their mechanistic counterparts and exhibit superior forecasting performance~\cite{reich2019collaborative}. Neural networks (\gls{nn}s) are ``universal approximators'', meaning that if they are sufficiently large they can fit any data. They have been shown to be competitive with the state-of-the-art in forecasting tasks~\cite{zhang2005neural, khashei2010artificial, kaastra1996designing} and have been applied to influenza forecasting~\cite{volkova2017forecasting, venna2018novel, aiken2019towards}. NNs application to influenza has been limited in part because estimating uncertainty with NNs can be challenging. 

Uncertainty in forecasting is attributed to two sources~\cite{der2009aleatory}: data uncertainty and model uncertainty. Data uncertainty --- often referred to as aleatoric uncertainty --- is inherent in the data and may be caused by noisy measurements or sampling error. Web search data has inherent uncertainty caused by sampling a subset of users to estimate search activity. We can further separate data uncertainty into heteroscedastic and homoscedastic uncertainty; homoscedastic uncertainty is constant whereas heteroscedastic uncertainty changes over time. In the context of influenza forecasting, we choose to estimate heteroscedastic uncertainty as we find changes in uncertainty across the flu season: there is more uncertainty during the winter when case counts are high and less uncertainty in the summer when case counts are low. 

Model uncertainty --- often referred to as epistemic uncertainty --- is uncertainty within the modelling process. This is often estimated by setting the parameters of the model to distributions, which is equivalent to an ensemble of models that is weighted to fit the training data. 
Increasing the size and diversity of a dataset enables models to be more confident. However, novel scenarios which differ from the training set should result in less confident forecasts. When producing forecasts over several forecast horizons, model uncertainty should increase with the horizon whereas data uncertainty is independent of the forecast horizon. 

It is considered impossible to make informed decisions from forecasts without understanding their uncertainty. A forecast predicting an increase in cases with low confidence should be treated very differently from the same forecast with high confidence. Estimates of uncertainty also enable better ensembling of multiple forecasts. FluSight~\cite{centers2019flusight} --- a competition to forecast influenza-like illness (\gls{ili}) in the US --- highlights the need for uncertainty estimates by requiring them in submitted forecasts.

ILI is a proxy for influenza; it is used in place of case counts because it is very expensive and difficult to measure the prevalence of disease circulating in a population. ILI is defined by the \gls{cdc} as a fever (temperature of $37.8^{\circ}$ or greater) and cough and/or sore throat without a known cause besides influenza. In the US, ILI is monitored through several public health surveillance efforts including the Outpatient Influenza-like Illness Surveillance Network (ILINet) which collects weekly state-level ILI proportions from over $2{,}000$ healthcare providers from all states. The state-level ILI proportions are weighted by population size to report the weighted-ILI (\gls{wili}) at regional and national levels~\cite{cdc_flu}. 
Obtaining the ILI proportions is difficult and introduces a delay.
Consequently, data may be reported up to two weeks after the fact. In all the work recorded here, we are in fact forecasting the ILI (not influenza) proportions, the latter of which is unknown. 

In this thesis, we develop neural network models to forecast ILI with uncertainty. 
Chapter \ref{chapterlabel1} describes our work using Bayesian neural networks, incorporating Web search data to improve forecast accuracy, while Chapter \ref{chapterlabel2} looks at combining neural networks and mechanistic models. 
% In chapter 2 we begin with an introduction to a, b, c
% More detailed structure culminating in 2.3-2.6 that describes our original contribution where we outline our methods results and discussion. 

Chapter~\ref{chapterlabel1} begins with an introduction to uncertainty quantification and examples of model and data uncertainty for a linear model. 
From Sections \ref{sec:ff_nn_intro} to \ref{sec:nn_uncertainty} we give a basic overview of neural networks and describe how they can be modified to estimate model and data uncertainty --- we provide examples using a synthetic dataset.
We review existing non-mechanistic ILI nowcasting and forecasting models in Section~\ref{sec:nonmechILImodels}, where we also discuss how some models have been integrated with Web search activity data. This culminates in Section \ref{sec:nn_methods} which outlines our methods for forecasting ILI using neural networks with Web search activity data. In Sections~\ref{sec:nn_results} and \ref{sec:nn_discussion}, we provide a comparative analysis of our best performing neural network architecture to a state-of-the-art baseline over four flu seasons at a national level in the United States. The proposed framework reduces forecasting error and provides significantly earlier insights about emerging ILI trends compared with existing models.

Section \ref{sec:IRNNs} addresses concerns about the uncertainty estimation in IRNN. We discuss why the model uncertainty does not increase as expected with longer forecast horizons. We propose a modified version of the architecture --- \gls{irnns} and compare it with IRNN. IRNN$_s$ improves the uncertainty estimation at the cost of a higher mean-absolute-error (\gls{mae}), and so we suggest future work to mitigate the change in MAE.

In Chapter \ref{chapterlabel2} we merge neural networks and compartmental models (a family of mechanistic models common in epidemiology)  using neural ordinary differential equations (neural \gls{ode}s). We first provide an overview of neural ODEs and follow with mechanistic models for infectious disease modelling.
In Section \ref{sec:UDEs} we introduce universal-differential-equations (\gls{ude}s), which use a neural network to augment an existing ODE and improve flexibility to reduce forecasting error. 
We provide examples on synthetic and real-world data to show how UDEs can improve compartmental models. Finally, in Section~\ref{sec:ODEmethods} we describe a UDE framework for forecasting ILI, which uses variational auto encoders --- a probabilistic generative neural network architecture. In Sections~\ref{sec:oderesults} and \ref{sec:odediscussion} we compare eight ODE architectures of varying complexity, including two UDEs with the neural networks from Chapter~\ref{chapterlabel1}. We find that UDEs outperform our best NNs when neither model uses Web search activity data. However, our attempts to combine these models with Web search activity data were less successful. Finally, Chapter \ref{chapterlabel4} presents conclusions and suggestions for future work. 

The main contributions of this thesis are two practical frameworks for forecasting infectious diseases. The first uses Bayesian neural networks with uncertainty estimates and Web search activity data, and the second uses compartmental models to give prior knowledge to neural ODEs. The specific contributions are as follows:
\begin{itemize}
    \item{Presentation of existing methods for forecasting infectious diseases, uncertainty estimation with neural networks, and universal differential equations.}
    \item{Development of an uncertainty modelling solution for ILI forecasting using Web search data and Bayesian neural networks.}
    \item{We show our method can be incorporated into common NN architectures, such as feed-forward (\gls{ff}) and recurrent neural networks (\gls{rnn}s).}
    \item{Comparative analysis of our framework with the existing state-of-the-art, improving by $17.1\%$ in terms of skill and $10.3\%$ in terms of mean-absolute-error.}
    \item{Modifications to our published work~\cite{morris2023neural} where we improve the uncertainty estimation further}.
    \item{Development of a UDE framework for forecasting ILI, showing neural networks can be used with two different compartmental models, and laying the groundwork for combining our models with Web search activity data.}
\end{itemize}
\chapter{Forecasting Influenza with Bayesian Recurrent Neural Networks}
\label{chapterlabel1}

This chapter is based on the paper ``Neural network models for influenza forecasting with associated uncertainty using Web search activity trends'' (Morris et al.)~\cite{morris2023neural}, which was published in PLOS Computational Biology in 2023.

\section{Introduction}
In this chapter, we develop Bayesian Neural Network (\gls{bnn}) models to forecast influenza-like illness (ILI) in the United States.
In the US, the CDC organise a competition to forecast the ILI proportion at a state, regional, and national level from one to four weeks ahead\cite{centers2019flusight}. This has resulted in an active research community with baselines for both mechanistic and non-mechanistic models to which we can compare our own models. 

We propose and evaluate the performance of three NN architectures, namely a simple feed-forward network (FF) and two forms of recurrent neural networks (denoted Simple-Recurrent-Neural-Network (\gls{srnn}) and Iterative-Recurrent-Neural-Network (\gls{irnn}); see Methods) all of which incorporate the frequency time series of various Web search terms as exogenous variables, and provide uncertainty estimates by deploying BNN layers and inference techniques. The forecast targets are US national ILI proportions as published by the CDC. Evaluation is performed for the four flu seasons from 2015/16 to 2018/19 (both inclusive). For the overall best-performing NN model --- IRNN, we also confirm that the incorporation of exogenous data significantly improves performance. The best-performing networks for each forecasting horizon \gls{gamma}, SRNN when $\gamma=7$ and IRNN otherwise, are then compared with Dante~\cite{osthus2021multiscale}, a state-of-the-art conventional ILI forecasting model. Our experiments show that the proposed NN architectures incorporating Web search activity can significantly reduce forecasting error and provide significantly earlier insights about emerging ILI trends.

Our proposed NN architectures can efficiently and effectively incorporate Web search activity data. 
We use the daily frequency of a variety of search terms (keywords or phrases) related to ILI. These include symptoms, remedies, general advice seeking, and other relevant categories (Table~\ref{tab:sup_search_queries}). However, the different latencies of health reporting and Web search activity can introduce a level of confusion with respect to model configuration and evaluation. Generally in forecasting, we have a set of observed data points (samples) up to and including time (day) \gls{t0}, and aim to predict a future value at time $t > t_0$. Due to the different data reporting latencies, we can obtain historical ILI proportions up to time $t_0$ and exogenous data up to $t_0 + \delta$, where \gls{delta} is typically 14 days. When we refer to the number of days ahead to be forecast i.e. the forecast horizon denoted by $\gamma$, we need to specify from what time. For that purpose, we can either use $t_0$ (the time point of the last available ILI proportion) or $t_0 + \delta$ (the time point of the most recent exogenous information). Here, we adopt the convention from prior literature and use $t_0$. As such, a seven days ahead forecast i.e. for day $t = t_0 + 7$, with a latency of $\delta = 14$ days may actually use exogenous data that is available after the forecast horizon (days $t_0 + 8$, $\dots$, $t_0 + \delta$). This is a curious situation, but we note that it is accepted practice within the ILI forecasting community (often referred to as hindcasting), and hence we have chosen to include these results. Obviously, for forecast horizons greater or equal to $\delta$, no ``future'' exogenous data is available, which makes the outcomes of these experiments more relevant in practical terms.

Section~\ref{sec:chap1background} reviews literature and provides a background to uncertainty quantification, neural networks --- including how they can estimate uncertainty, and non-mechanistic ILI forecasting models. Next, we present our own methods before providing a comparative analysis with Dante. Finally, in Section~\ref{sec:IRNNs} we build on this work and introduce the IRNN$_s$ architecture, which improves the uncertainty estimation of the best-performing IRNN. 

\section{Background and Related Work}
\label{sec:chap1background}
Here we introduce uncertainty quantification and provide examples. We provide an overview of neural networks and show how they can be modified to estimate different kinds of uncertainty. We discuss non-mechanistic disease forecasting models, leaving a discussion of mechanistic models for Chapter~\ref{chapterlabel2}.

\subsection{Uncertainty Quantification}
Forecasts without an estimate of uncertainty are of little use to decision-makers. 
For decision-makers, a forecast spike in influenza cases with high confidence is very different to a forecast spike with low confidence.
The need to estimate uncertainty was clear during the Covid-19 pandemic~\cite{nixon2022real, Ioannidis2022, gibson2020real}, where uncertainty was integral to the interpretation of forecasts by policymakers. Additionally, estimation of uncertainty is a requirement for the ILI forecasting competition run by the CDC~\cite{centers2019flusight}. When considering the uncertainty around an estimate there are typically two types of uncertainty which should be considered~\cite{tagasovska2019single}: data, and model uncertainty.
    
\subsubsection{Data Uncertainty}
Data uncertainty, often referred to as aleatoric uncertainty, it is caused by randomness in the data. This arises due to measurement or sampling errors and appears as observational noise. This is an uncertainty that we learn about the data and cannot reduce with modelling --- we cannot exactly estimate values which are generated by a stochastic process, instead, we estimate the underlying distribution generated by that process. 

Data uncertainty in ILI data arises from the methodologies employed in its collection. In the United States, the ILI proportion is measured by a network of sentinel doctors who report the proportion of their patients exhibiting ILI symptoms. The reliability of these measurements, however, is subject to variation across states due to differences in patient volumes. 
The variability, and therefore uncertainty, in ILI measurements inversely correlates with the number of patient visits and population size of the measured region~\cite{osthus2021multiscale}. As the number of patient visits decreases, the data uncertainty increases correspondingly. The national average ILI proportion, denoted the weighted ILI proportion (wILI), is calculated by taking a population-weighted average of state-level ILI proportions. This averaging reduces the data uncertainty but does not remove it entirely. 
Although ILI data is supplied as a single value, without a measure of its uncertainty, it is important to model the underlying uncertainty. In some instances, modellers specify the uncertainty of ILI measurements based on the ILI proportion~\cite{shaman2013real,shaman2012forecasting}. In other cases, the modellers measure uncertainty in the variability of ILI proportions from past flu seasons and use it to estimate the data uncertainty in forecasts~\cite{osthus2019dynamic, osthus2021multiscale,osthus2022fast}.  Our machine learning methods do not require an ``uncertainty ground truth'', meaning that they do not require a predefined measure of the uncertainty associated with each data point. Instead, the uncertainty estimation is implicit in the loss function and training.

Data uncertainty can be further divided into homoscedastic and heteroscedastic uncertainty~\cite{gal2016uncertainty}. Homoscedastic uncertainty is constant for different inputs, while heteroscedastic uncertainty is variable and dependent on the inputs to the model. Data uncertainty in ILI forecasting is heteroscedastic, varying through the flu season. During the summer the ILI proportion is very low and nearly constant, thus the data uncertainty is very low. In the winter when flu is circulating the data uncertainty is much larger. 
    
\subsubsection{Model Uncertainty} 
Model uncertainty, often referred to as epistemic uncertainty, is uncertainty about how well models fit the data that they are trained and perform inference on. Deep learning models are universal approximators and assuming they are sufficiently large they can fit any data. Therefore, model uncertainty in deep learning ignores the structure of the model and is expressed as uncertainty in the model's parameters. This is caused by the model having insufficient data to generalise correctly for all possible data points~\cite{kendall2017uncertainties}. 
The model parameters are specified by a distribution instead of individual numbers. The distribution expresses the range of possible parameter values weighted by how well they fit the training data~\cite{o2004dicing}. A model trained on a large and diverse dataset is expected to be more confident than the same model trained on a subset of that dataset. Furthermore, models should be less confident on out-of-sample data, for example, a model trained on values \gls{x}$\in({0,1})$ should be more confident when tested on $x\in({0,1})$ than $x=100$. 

Model uncertainty is introduced into ILI forecasting due to variations in the flu season from year to year. Forecasts should be confident when the test season resembles a season included in the training set. 
Model uncertainty increases during unusual flu seasons which differ from the training data. For example, during the Covid-19 pandemic flu and Covid-19 co-circulated which was a novel scenario.

\subsubsection{Uncertainty Quantification Example}
\label{sec:uncertainty_linear_regression}
We demonstrate a simple linear regression problem with added uncertainty, similar to the example from \cite{bishop2006pattern}.  The structure of our model is chosen to mirror that of a neural network with a single unit. Most neural networks use non-linear functions referred to as activation functions to enable them to learn more complex patterns in data. We do not include an activation function in this example. An introduction into neural networks is provided in Section \ref{sec:ff_nn_intro}. The model is highly simplified enabling us to find the analytic solutions to the below examples, a neural network would require a numerical solution such as gradient descent. 

Data is generated based on the equation \gls{y}$=ax+b+\epsilon$, where $a=3$ and $b=5$, and \gls{epsilon}$\sim \mathcal{N}\left(0,\sigma^2 \right)$ is data noise, where \gls{sigma}$=1$. We create a model to fit this data on $a$ and $b$, we then modify it to estimate model uncertainty, data uncertainty, and a combination of both model and data uncertainty.

\paragraph{No Uncertainty\\}
The training data has $N=10$ datapoints from $y=ax+b+\epsilon$, where each value of $x$ has only one sample associated with it which includes unknown additive noise. The model is deterministic and makes predictions \gls{hat_y}$ = f($\gls{bmPhi}$, x)=ax+b$, where the parameters $\bm{\Phi} = [a,b]$.  We can find the analytic solution to model parameters using least-squares~\cite{rao1971further, van1996matrix}:
\begin{equation}
    \bm{\Phi}^\star = \left(\mathbf{A}^T\mathbf{A}\right)^{-1}\mathbf{A}^T\mathbf{y},
    \label{eq:least_squares}
\end{equation}
where \gls{A} is a design matrix and $^\star$ denotes a solution to an optimisation problem.
\begin{equation}
    \mathbf{A} = \left( 
    \begin{matrix}
        x_1 & 1 \\
        x_2 & 1 \\
        \vdots & \vdots \\
        x_N & 1
    \end{matrix}
    \right).
\end{equation}
Each column contains a basis function, and each row a training example. The first column is a linear basis function (the gradient), the second column is the intercept.

\begin{figure}[!ht]
    \centering
    \includegraphics[width=0.8\linewidth]{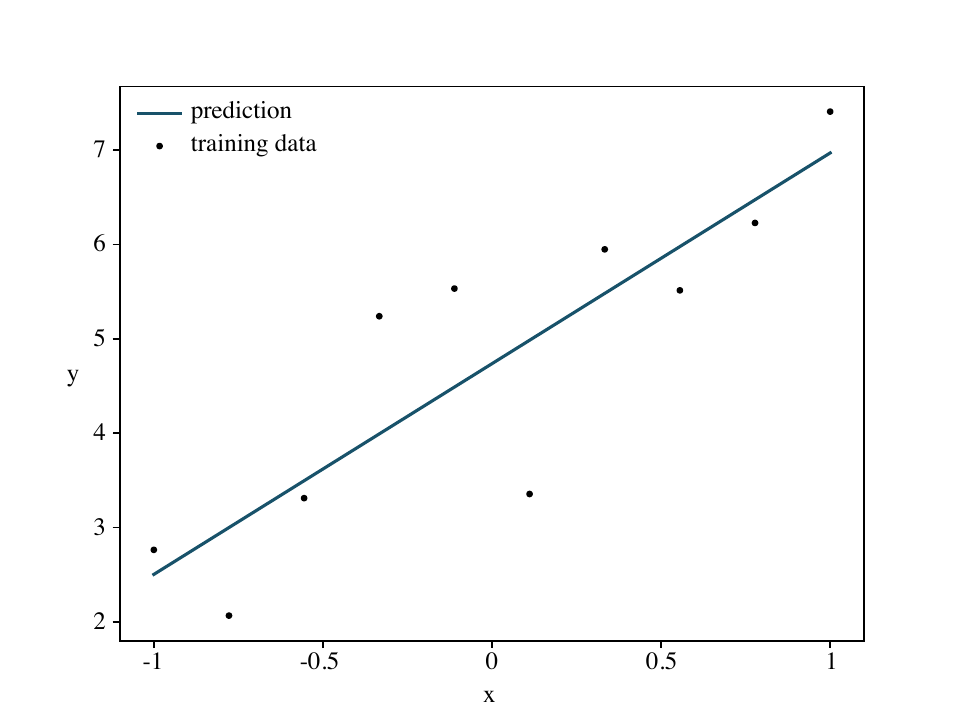}
    \caption[{Example linear model estimates}]{\textbf{Example linear model estimates}\newline
    Linear model estimate trained on 10 data points sampled from $y=3x+5+\mathcal{N}(0, 1)$. The trained model fit $y=2.23x+4.74$. The prediction is close to the data model and has no uncertainty estimate. }
    \label{fig:linear_model_example}
\end{figure}

We fit the model to the samples from the training data, resulting in $\bm{\Phi}^\star = [2.23, 4.74]$. These parameters fit the training data well, but there is no uncertainty estimate in how well they will generalise to new data (i.e., there is no model uncertainty). There is no estimate of the noise in the data that they have been trained on (i.e., there is no data uncertainty).  It would be possible to estimate the data uncertainty by extending the linear regression example, however doing so would make it no longer applicable to neural networks, where an analytic solution would not be available. Figure~\ref{fig:linear_model_example} shows the predictions of the linear model and the training samples. 

\begin{figure}[!ht]
    \centering
    \includegraphics[width=0.8\linewidth]{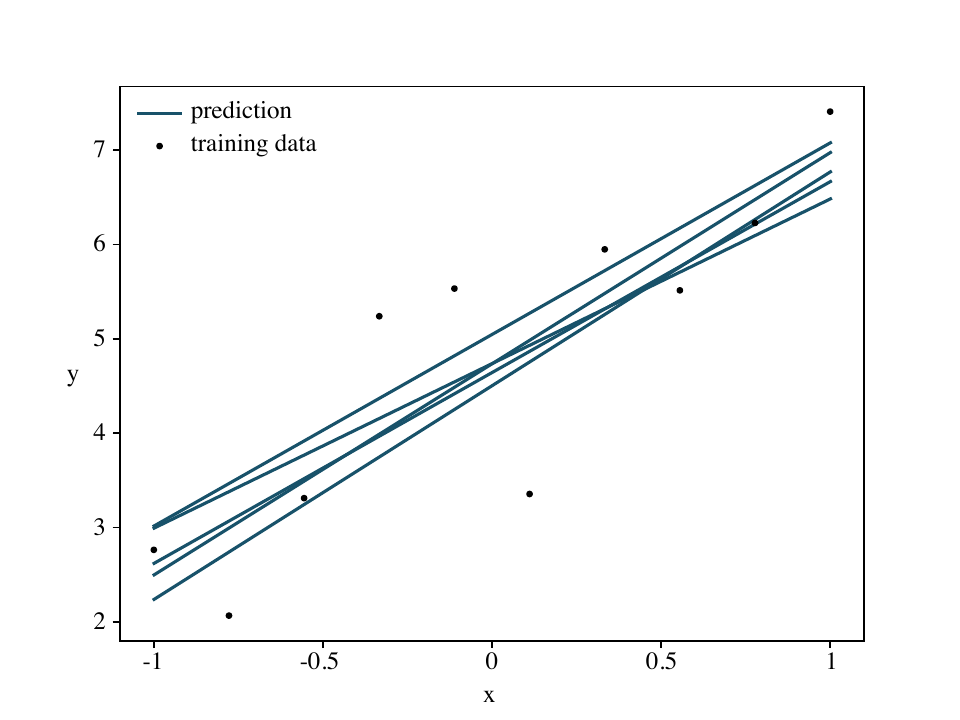}
    \caption[{Example Bayesian linear model estimates}]{\textbf{Example Bayesian linear model estimates}\newline
    Bayesian linear model estimates trained on 10 data points sampled from $y=3x+5+\mathcal{N}(0, 1)$. The posterior distribution over the parameters if given by $p(\bm{\Phi}|\mathbf{x},\mathbf{y}) = \mathcal{N}\left([4.74, 2.23],\begin{bmatrix}
0.50 & 0 \\
0 & 0.32
\end{bmatrix}\right)$, five samples from the model are shown. The prediction has the same mean as the deterministic linear model. The model uncertainty can be measured by integrating out the posterior. If the model is evaluated outside the training range then the predictions will have a higher variance.}
    \label{fig:model_uncert_example}
\end{figure}

\paragraph{Model Uncertainty\\}
To estimate model uncertainty the parameters $\bm{\Phi}$ are changed from single values to a distribution. This is referred to as a Bayesian model. We specify a prior distribution over $\bm{\Phi}$, during training, we observe data and find the trained distribution, referred to as the posterior. 
The prior is a normal distribution governed by a precision parameter \gls{zeta}$ = 1/\sigma^2$,  (the inverse of variance), and a mean of $0$:
\begin{equation}
\label{Eq:prior_analytic}
    p(\bm{\Phi}|\zeta) = \mathcal{N}(\bm{\Phi}|\mathbf{0},\begin{bmatrix}
1/\zeta & 0 \\
0 & 1/\zeta
\end{bmatrix}).
\end{equation}
Using the conjugate prior, we derive the posterior distribution of the form:
\begin{equation}
    \label{eq:posterior_example}
    p(\bm{\Phi}|\mathbf{x},\mathbf{y}) = \mathcal{N}(\bm{\Phi}|\mathbf{m}_N,\mathbf{S}_N),
\end{equation}
where $\mathbf{m}_N$ is the mean and $\mathbf{S}_N$ is the covariance, calculated by~\cite{bishop2006pattern}:
\begin{equation}
    \mathbf{m}_N=\iota\mathbf{S}_N\mathbf{A}^T\mathbf{y},
\end{equation}
\begin{equation}
    \mathbf{S}^{-1}_N=\zeta\mathbf{I}+\iota\mathbf{A}^T\mathbf{A},
\end{equation}
where \gls{iota} is a precision hyperparameter for the data uncertainty. For simplicity, we set $\zeta$ and $\iota$ to $1.0$ for the model uncertainty example.
Fitting the model to the same data as before yields $p(\bm{\Phi}|\mathbf{x},\mathbf{y}) = \mathcal{N}\left([4.74, 2.23],\begin{bmatrix}
0.50^2 & 0 \\
0 & 0.32^2
\end{bmatrix}\right)$. The mean is unchanged from the model trained without uncertainty. The uncertainty is determined by the covariance of the posterior. We numerically integrate the predictive distribution by sampling \gls{K} times from the posterior distribution $p(\bm{\Phi}|\mathbf{x},\mathbf{y}, \zeta, \iota)$, and use each sample to produce an output $\mathbf{\hat{y}^\prime} = f(\bm{\Phi}^\prime, \mathbf{x})$ where $\bm{\Phi}^\prime$ denotes a sample from $\bm{\Phi}$. The model uncertainty is the variance over the $K$ predictions: 
\begin{equation}
\label{eq:model_variance}
    \hat{\mathbf{\sigma}}^2_\text{model} \approx
    \frac{1}{K} \sum^K_{\kappa=1} \hat{\mathbf{y}}^{\prime 2}_\kappa
    - \left(\frac{1}{K} \sum^K_{\kappa=1} \hat{\mathbf{y}}^\prime _\kappa\right)^2.
\end{equation}
Increasing $K$ improves the approximation of the integral at the cost of computational time. Figure~\ref{fig:model_uncert_example} shows how the predictions vary when estimates are made using five samples from the posterior over $\bm{\Phi}$.  
For $x\in(-1,1)$ with $N=10$, the standard deviation due to model uncertainty is $\mathbf{\sigma}_\text{model}=0.44$. Outside the training range at $x=10$ the model uncertainty is greater at $\mathbf{\sigma}_\text{model}=4.85$. 
If the size of the dataset is increased from $N=10$ to $N=100$ then the model uncertainty for the predictions $x\in(-1,1)$ shrinks from $\mathbf{\sigma}_\text{model}=0.44$ to $\mathbf{\sigma}_\text{model}=0.14$. Thus the model uncertainty reduces in higher data regimes and increases in out-of-sample test data. 

\paragraph{Data Uncertainty\\}
Data uncertainty is incorporated by adding a noise component to the model:
\begin{equation}
    y=f(\bm{\Phi}, x)+\mathcal{N}\left(0,\sigma^2_\text{data} \right),
\end{equation}
where $\sigma^2_\text{data}$ is the data uncertainty. The parameters $\bm{\Phi}=[a,b]$ are set using Eq.~\ref{eq:least_squares} i.e., ignoring uncertainty and treating the model the same as the deterministic model. The data uncertainty is estimated as the variance between the estimates without uncertainty and the training data.
\begin{equation}
    \bm{\hat{\sigma}}^2_\text{data} = \frac{\sum\left(\hat{a} \mathbf{x} + \hat{b} - \mathbf{y}\right)^2}{N}.
\end{equation}

\paragraph{Combined Uncertainty\\}
Model and data uncertainty are combined into a single model. Here we use Eq.~\ref{eq:posterior_example} to compute $ p(\bm{\Phi}|\mathbf{x},\mathbf{y}) $. We can then sample from $ p(\bm{\Phi}|\mathbf{x},\mathbf{y}) $ $K$ times and compute the estimated data variance $ \bm{\hat{\sigma}}^2_\text{data} $ each time. The model uncertainty is the variance of the means of the $K$ predictions, while the data uncertainty is the mean of the $K$ data variances. The combined variance is the sum of the model and data uncertainty.
\begin{equation}
    \mathbf{\sigma}^2_\text{total} = \mathbf{\sigma}^2_\text{model}+\mathbf{\sigma}^2_\text{data} 
    % = \text{var}_{\bm{\Phi}} [\mathbb{E}(\hat{y}|x,\bm{\Phi})] + [\mathbb{E}_{\bm{\Phi}}\text{var}(\hat{y}|x, \bm{\Phi})].
\end{equation}

As the number of data points $N$ increases the model uncertainty should reduce and data uncertainty should converge to the true data uncertainty.

\begin{figure}[!ht]
    \centering
    \includegraphics[width=0.8\linewidth]{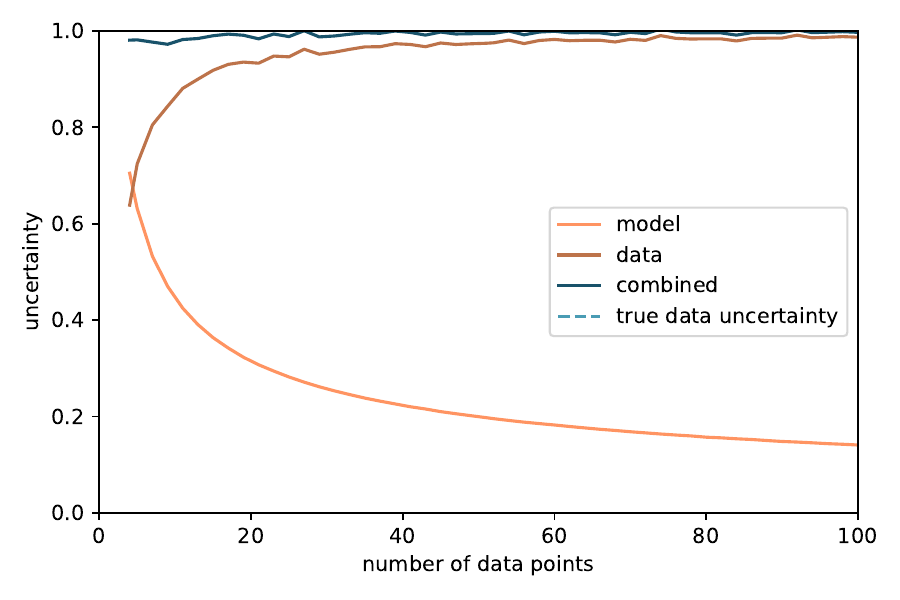}
    \caption[{Combined uncertainty with changing training set size}]{\textbf{Combined uncertainty with changing training set size} \newline
    Uncertainty averaged for $x\in[-1,1]$. As more data is used the model uncertainty reduces and the data uncertainty becomes more accurate. In almost all cases the combined uncertainty is the best estimate.}
    \label{fig:combined_example}
\end{figure}

Figure \ref{fig:combined_example} shows how the different uncertainties vary with the size of the dataset $N$ on the synthetic example provided. The uncertainty inherent in the data is known, and shown by the horizontal line. When the size of the dataset $N$ is small, the estimated data uncertainty (orange) is small, and the model uncertainty is large.  As the $N$ increases the model uncertainty reduces and the estimated data uncertainty approaches the uncertainty of the dataset. The total uncertainty is the closest to the true uncertainty in the test data regardless of the size of the training set.

\paragraph{Analytic Computation of Combined Uncertainty\\ }
We can compute the data uncertainty using the parameter $\iota$ and computing the predictive distribution. This method is not applicable to neural networks where no closed-form solution is available. For the linear-regression example, we can calculate the predictive distribution by integrating the parameter distribution:
\begin{equation}
p\left(\hat{y}|\mathbf{y}, \zeta, \iota \right) = \int p\left(\hat{y} | \mathbf{x}, \bm{\Phi}, \iota\right)p\left( \bm{\Phi}| \mathbf{y}, \zeta, \iota \right) d\bm{\Phi}.
\end{equation}
The left-hand side of the integral is calculated by $p\left(\hat{y}|\mathbf{x}, \bm{\Phi}, \iota\right) = \mathcal{N}\left(\hat{y}|f(\mathbf{x},\mathbf{y}), \iota^{-1} \right)$, where $f(\mathbf{x},\mathbf{y})$ is a deterministic function such that the model output is $\hat{y} = f(\mathbf{x}, \mathbf{y})+\epsilon$, and $\epsilon$ is additive noise.
The right-hand side of the integral is computed by Eq.~\ref{eq:posterior_example}. This gives the closed-form solution:
\begin{equation}
p\left(\hat{y}|x, \mathbf{D}, \zeta, \iota \right) = \mathcal{N}\left(\hat{y}|\mathbf{m}_N^T \phi(x), \hat{\sigma}^2_N(x)  \right), 
\end{equation}
where $\phi(x)$ is a basis function associated with an input $x$, \gls{D} is the dataset containing inputs and targets, and the variance is:
\begin{equation}
\hat{\sigma}^2_N(x) = \frac{1}{\iota}+ \phi(x)^T \mathbf{S}_N \phi(x). 
\end{equation}
Here, the first term represents the variance due to data uncertainty, and the model uncertainty is given by the second term.

In the previous examples, we assumed that $\iota$ was known to enable us to compute the posterior distribution. However, we must now estimate $\iota$ from the data. By fixing the prior precision $\zeta$, we can find an optimal value for $\iota$ using a numerical optimiser\footnote{We used the $\texttt{scipy.optimise.minimise}$ function in Python 3, using the default $\texttt{BFGS}$ settings} and maximising the likelihood $p(\mathbf{D}|\iota)$; or in this case minimising the negative-log-likelihood (\gls{nll}), computed by:
\begin{equation}
        \text{NLL}(y, \hat{y}, \hat{\sigma}^2_N) = 
        \frac{1}{N}\sum_{n=1}^{N}\left(\frac{1}{2 \hat{\sigma}_n^2}{\left(y_n-\hat{y}|\mathbf{m}_N^T \phi(x_n)\right)}^2 +\frac{1}{2}\log{\left(2 \pi \hat{\sigma}^2_N(x_n)\right)} \right) \, .
\end{equation}

\begin{figure}[!ht]
    \centering
    \includegraphics[width=0.8\linewidth]{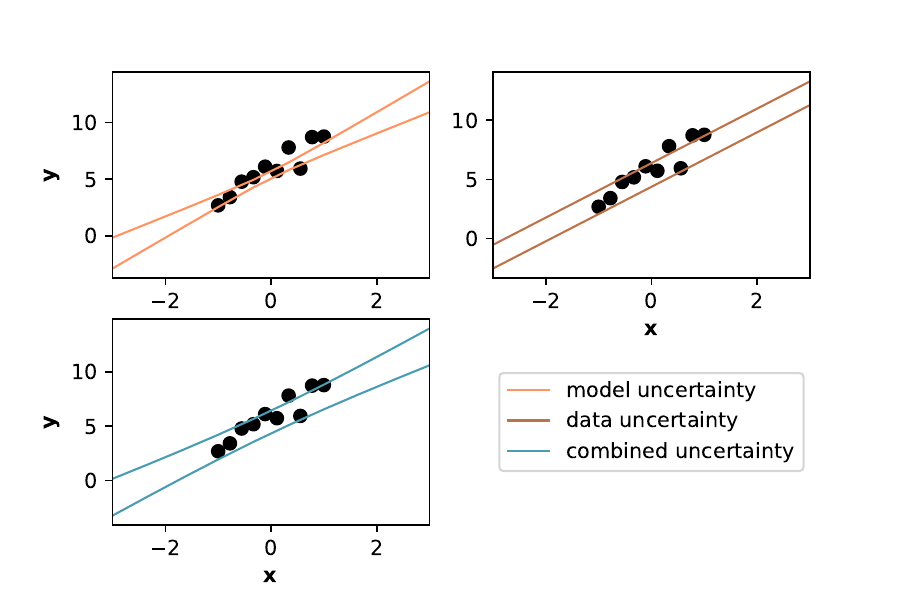}
    \caption[{Closed-form uncertainty estimates for $N=10$ training examples}]{\textbf{Closed-form uncertainty estimates for $N=10$ training examples} \newline
    Uncertainty estimates  for $N=10$ training examples, uncertainty intervals shown at one standard deviation from the mean. The model uncertainty is small for $x\in[-1,1]$ and grows for $x<-1$ and $x>1$ (out-of-sample). Data uncertainty is constant for all $x$. Combined uncertainty is the sum of the variances of model and data uncertainty. }
    \label{fig:analytic_model_uncertainty}
\end{figure}

 Figure \ref{fig:analytic_model_uncertainty} shows uncertainty estimates for $N=10$. The model uncertainty is low in the region where the model is trained, and grows on the out-of-sample predictions. Data uncertainty is constant across the estimates, and combined uncertainty is the sum of the two variances. 

\begin{figure}[!ht]
    \centering
    \includegraphics[width=0.8\linewidth]{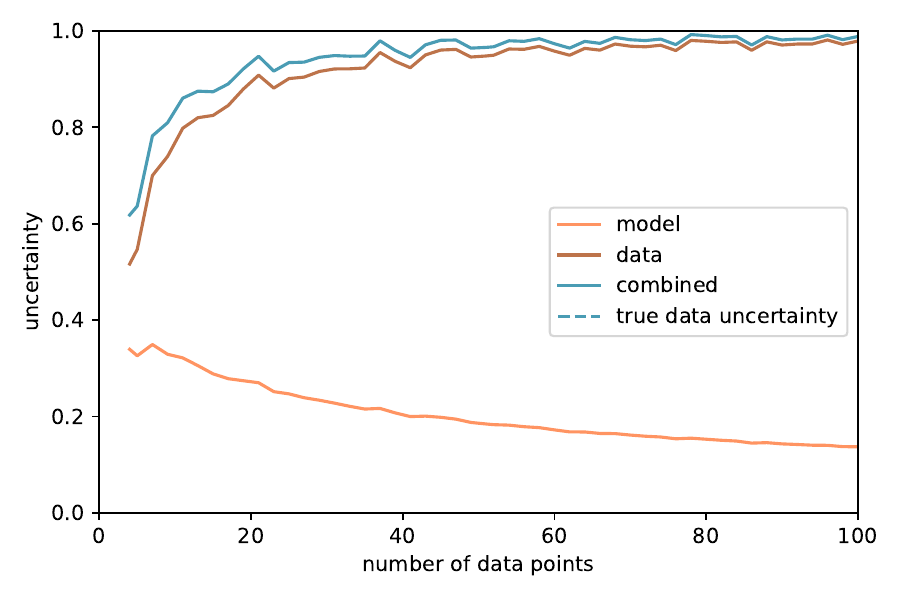}
    \caption[{Closed-form combined uncertainty with changing training set size}]{\textbf{Closed-form combined uncertainty with changing training set size} \newline
    Uncertainty averaged for $x\in[-1,1]$. As more data is used the model uncertainty reduces and the data uncertainty becomes more accurate. In almost all cases the combined uncertainty is the best estimate.}
    \label{fig:combined_example_analytic}
\end{figure}

 Figure \ref{fig:combined_example_analytic} shows the model, data, and combined uncertainties while varying $N$. Similarly to the example using sampling, when $N$ is small, the estimated data uncertainty (orange) is small, and the model uncertainty is large.  As the $N$ increases the model uncertainty reduces and the estimated data uncertainty approaches the uncertainty of the dataset. The total uncertainty is the closest to the true uncertainty regardless of the size of the training set. 

For very low $N$, the analytic solution has less model uncertainty than the solution using approximation by sampling. This is caused by the approximation using a preset $\iota$. However, both examples illustrate that as the size of the dataset increases, the estimated uncertainty becomes more accurate and that for low $N$ the model uncertainty will have a greater influence on the total uncertainty than for large $N$. We also observe that the model uncertainty is greater in out-of-sample estimates using both the approximation and closed-form solutions. In Section~\ref{sec:nn_uncertainy_example} we provide a similar uncertainty estimation example in the context of neural networks. 

\subsection{Neural Networks}
\label{sec:ff_nn_intro}
Here we provide a basic overview of neural networks and their function
\cite{goodfellow2016deep}.
A neural network is a mathematical model inspired by the structure of a biological brain. Neural networks consist of units or ``perceptrons'' that model biological neurons. Neural networks consist of an input layer, one or more hidden layers, and an output layer. 

\subsubsection{Feed-Forward Neural Network}
\ref{sec:ff_nn_intro}
The most basic neural network layer is a ``fully-connected-dense'' layer, which contains \gls{I} units and observes an input vector $\mathbf{x} = [x_1, x_2, ..., x_J]$. The output of the layer is a vector $\mathbf{y}=[y_1, y_2, ..., y_I]$ ($I$ units),  the output of the $i^\text{th}$ unit  is calculated by
\begin{equation}
    y_i = g\left(\sum_{j=1}^{J} w_{ij}x_j + b_i\right),
\end{equation}
where $y_i$ is the $i^\text{th}$ output, $x_j$ is the $j^\text{th}$ input, and \gls{g} is a non-linear activation function. The weight $w_{ij}$ is associated with the $j^\text{th}$ input and the $i^\text{th}$ unit, and $b_i$ is the $i^\text{th}$ bias. Thus, each unit effectively computes a multiple linear regression which is then transformed by a non-linear activation function.
In practice, the output of the entire layer is calculated by 
\begin{equation}
    \label{eq:ff_equation}
    \mathbf{y} = g\left(\mathbf{W}^T\mathbf{x} + \mathbf{b}\right), 
\end{equation}
where \gls{W} is a weight matrix and \gls{b} is a vector of biases. For example, a two-layer FF neural network for regression would typically have an activation function such as Rectified-Linear-Unit (ReLU  $g(x) = \max{(0,x)}$) on the first layer and no activation function on the output layer, having no activation function enables the model to output any value positive or negative. Without a non-linear activation function, the NN would behave like a multiple linear regression model since adding layers would be equivalent to summing linear functions. The model would have weights associated with layers $1$ and $2$: $\mathbf{W}_1$ and $\mathbf{W}_2$ and biases $\mathbf{b}_1$ and $\mathbf{b}_2$. Assuming an input $\mathbf{x}$, The output of the first layer is  $\mathbf{y}_1 = g\left(\mathbf{W}^T_1\mathbf{x} + \mathbf{b}_1\right)$, and the output of the network is $\mathbf{\hat{y}} = \left(\mathbf{W}^T_2\mathbf{y}_1 + \mathbf{b}_2\right)$. For simplicity, we denote the output of a neural network $\mathbf{\hat{y}}$ based on inputs $\mathbf{x}$, using parameters (including all weight matrices and bias vectors) $\bm{\Phi}$ as $\bm{\hat{y}} = f^{\bm{\Phi}}(\mathbf{x})$.

During training, the network weights are iteratively adjusted to minimise the difference between the network output and the training data. The most common way to train neural networks is gradient descent combined with back-propagation. Back-propagation computes gradients~\cite{rumelhart1986learning} of a loss function \gls{loss} with respect to the network parameters. These gradients are in turn used to update the network parameters. The gradients are calculated on subsets of the dataset called ``batches'' or ``minibatches''. On each batch the loss is calculated and the weights are updated. One full cycle through the dataset is referred to as an epoch, and training is typically performed over several epochs.

Neural networks are scalable and can easily model non-linear relationships between inputs and the target variable, this makes them well suited to time series forecasting\cite{zhang2003time}. NNs can use a wide range of inputs and exogenous variables and are able to make forecasts for multiple variables simultaneously. 
The simplest type of neural network for time series forecasting is a feed-forward Neural Network (FF), which is made up of fully-connected-dense layers. FF models simultaneously observe all inputs to produce all outputs. 

\subsubsection{Recurrent Forward Neural Network}
\label{sec:rnn_intro}
\begin{figure}[!ht]
    \centering
    \includegraphics[width=\linewidth]{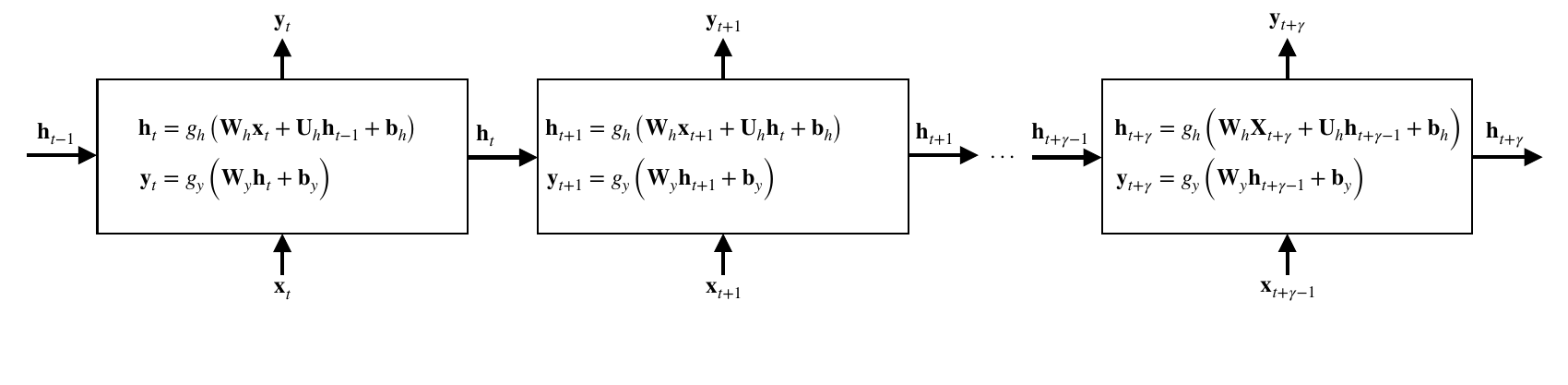}
    \caption[{Jordan Recurrent Neural Network Diagram}]{\textbf{Jordan Recurrent Neural Network Diagram} \newline
    Diagram of the Jordan RNN architecture. The RNN observes inputs $\mathbf{x}_t$ sequentially over time and updates the hidden state $\mathbf{h}_t$ and produces an output $\mathbf{y}_t$ each timestep. The hidden state is used to inform the models own predictions in future timesteps. The activation functions are $g_h$ and $g_y$; $\mathbf{W}_h$, $\mathbf{W}_u$ and $\mathbf{W}_y$ are weight matrices; and $\mathbf{b}_h$ and $\mathbf{b}_y$ are bias vectors.}
    \label{fig:Jordan_RNN_Diagram}
\end{figure}

Recurrent Neural Networks (RNNs) are more common for time series. While FF models observe all inputs at once, RNNs sequentially observe inputs from different timesteps, and retain important information in their hidden states. This allows RNNs to better handle sequential data - such as time series - and enables them to model sequences of any length~\cite{alom2019state}. Different versions of the RNN have been proposed, in the most basic architecture, referred to as the Jordan architecture~\cite{elman1990finding}, the RNN layer updates its hidden state \gls{ht} and output $\mathbf{y}_t$ based on an input vector $\mathbf{x}_t$ at time \gls{t}, according to the following:

\begin{align*}
    \mathbf{h}_t &= g_h\left(\mathbf{W}_h\mathbf{x}_t + \mathbf{W}_u\mathbf{h}_{t-1}+\mathbf{b}_h\right) \\
    \mathbf{y}_t &= g_y \left( \mathbf{W}_y\mathbf{h}_t + \mathbf{b}_y \right) .
\end{align*}
Where $g_h$ and $g_y$ are activation functions; $\mathbf{W}_h$, $\mathbf{W}_u$ and $\mathbf{W}_y$ are weight matrices; and $\mathbf{b}_h$ and $\mathbf{b}_y$ are bias vectors. The update is performed iteratively over time, re-using the same weights and biases but updating the hidden state and observing a new input.  A diagram of the basic Jordan RNN is provided in Figure~\ref{fig:Jordan_RNN_Diagram}.

A common problem with RNNs is the vanishing/exploding gradient problem~\cite{bengio1994learning} where gradients shrink to zero or explode to infinity as they are propagated through an RNN during training. This causes the parameters to go to zero or infinity, resulting in poor performance. This problem was overcome by the Long-Short-Term-Memory~\cite{hochreiter1997long} (\gls{lstm}) and Gated-Recurrent-Unit~\cite{cho2014properties} (\gls{gru}) networks. These use gating mechanisms to control the flow of information through the network over time. The gates determine whether information should be passed to the output and what information should be retained in the network's hidden state. This makes them more stable in training and allows them to better capture dependencies in longer sequences. LSTMs and GRUs decouple the memory from the output cell and only do additive updates to the memory state. Unlike traditional RNNs, they do not continually re-apply the recurrent weight matrix, which causes the gradients explode or vanish \cite{pascanu2013difficulty}.

LSTMs have been applied to various time series forecasting problems and have often outperformed traditional methods \cite{gers2001lstm}. An LSTM uses a hidden state $h_t$ and a memory cell  \gls{Ct} to pass information between timesteps. An LSTM unit is updated using the following equations:
\begin{align}
    \label{eq:LSTM_update}
    f_t &= g_\sigma(W_f \cdot [h_{t-1}, x_t] + b_f) && \text{Forget gate} \\
    i_t &= g_\sigma(W_i \cdot [h_{t-1}, x_t] + b_i) && \text{Input gate} \\
    \tilde{C}_t &= \tanh(W_C \cdot [h_{t-1}, x_t] + b_C) && \text{candidate memory-cell} \\
    C_t &= f_t \odot C_{t-1} + i_t \odot \tilde{C}_t && \text{Memory cell} \\
    o_t &= g_\sigma(W_o [h_{t-1}, x_t] + b_o) && \text{Output gate} \\
    h_t &= o_t \odot \tanh(C_t) && \text{Hidden state, also the output .}
\end{align}
Where $g_\sigma$, is the sigmoid activation function, $\tanh$ is the hyperbolic tangent activation function, and $\odot$ denotes element-wise multiplication. LSTMs have four separate sets of weights $W_f$, $W_i$, $W_o$,  $W_C$ and biases $b_f$, $b_i$, $b_o$, $b_C$. The hidden state and memory cell are fed back into the LSTM at each timestep. 

\begin{figure}[!ht]
    \centering
    \includegraphics[width=1.0\linewidth]{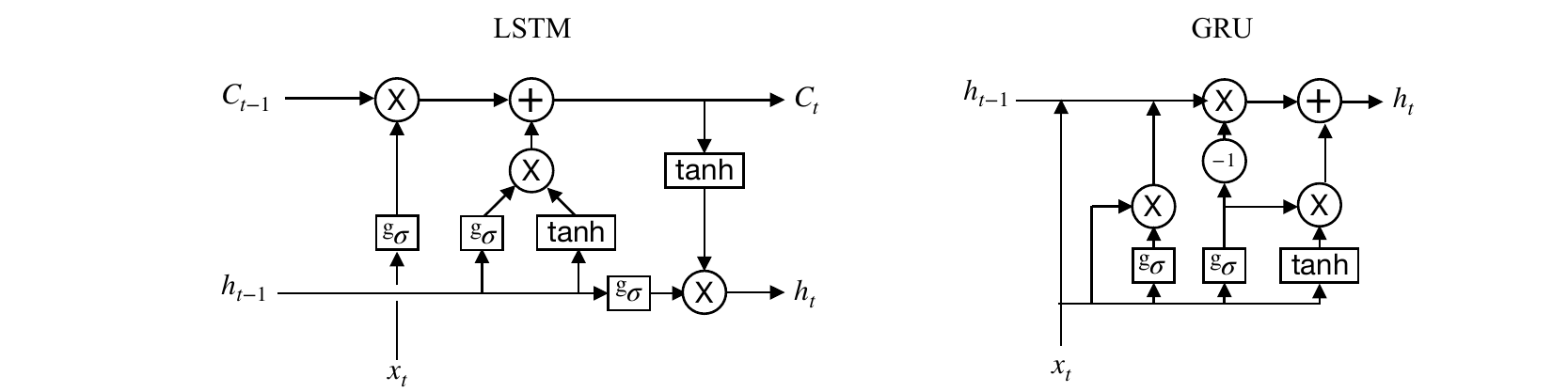}
    \caption[{LSTM and GRU Architectures Diagram}]{\textbf{LSTM and GRU Architectures Diagram} \newline
    Diagrams of the LSTM and GRU architectures, showing the updates for the RNN cells. $g_\sigma$ denotes a sigmoid activation and $\text{tanh}$ denotes a hyperbolic tangent activation function, $\text{X}$ denotes an element-wise multiplication. The equations to compute the updates for the LSTM and GRU are provided in Eq.~\ref{eq:LSTM_update} and Eq.~\ref{eq:GRU_update}, respectively.}
    \label{fig:architecture_diagram}
\end{figure}

A GRU performs similarly to an LSTM, with the advantage of requiring less memory due to having no separate memory cell and hidden state. A GRU unit is updated with the following equations:
\begin{align}
\label{eq:GRU_update}
    z_t &= g_\sigma(W_z \cdot [h_{t-1}, x_t] + b_z) && \text{Update gate} \\
    r_t &= g_\sigma(W_r \cdot [h_{t-1}, x_t] + b_r) && \text{Reset gate} \\
    \tilde{h}_t &= \tanh(W \cdot [r_t \odot h_{t-1}, x_t] + b) && \text{Candidate activation} \\
    h_t &= (1 - z_t) \odot h_{t-1} + z_t \odot \tilde{h}_t && \text{Hidden state, also the output}.
\end{align}
Where $W_z$, $W_r$ and $W$ are weights and $b_z$, $b_r$ and $b$ are biases. The hidden state is fed back into the GRU at each timestep. A diagram of the FF and RNN architectures, as well as the RNN, LSTM and GRU cells, is provided in Figure~\ref{fig:architecture_diagram}.

\subsection{Uncertainty in Neural Networks}
\label{sec:nn_uncertainty}
The application of neural networks to disease forecasting has been limited. Only simple formulations of NNs have been used prior to this work, and those do not include uncertainty estimation. In other fields of machine learning neural networks are the established state-of-the-art. 

A key reason why neural networks are not more common in disease forecasting is the difficulty in producing uncertainty estimates with them. This section discusses several methods for uncertainty estimation with neural networks.

Bayesian neural networks offer a promising avenue for estimating model uncertainty. They provide a probabilistic framework that allows us to quantify uncertainty in predictions. However, efficiently implementing Bayesian inference in neural networks, especially in larger models is challenging. The Bayesian linear regression in Section~\ref{sec:uncertainty_linear_regression}, used a closed-form solution which was made possible by the model's simplicity and linearity. Bayesian inference in neural networks involves complex, high-dimensional parameter spaces and the nonlinearities and large number of parameters make analytical solutions intractable. This necessitates the use of approximate methods such as variational inference and dropout. 

We provide an example using variational inference and dropout as Bayesian approximations. We discuss a simple way of estimating data uncertainty by modifying the output of the network to a distribution, along with alternative uncertainty estimation methods of Quantile Regression~\cite{taylor2000quantile} and Conformal Prediction~\cite{shafer2008tutorial} .

\subsubsection{Bayesian Neural Networks}
Bayesian statistics offer a principled method to quantify model uncertainty in neural networks. A prior distribution is placed over the network parameters, this is updated during training to find the posterior. Bayes theorem provides a mechanism to update prior beliefs as new data becomes available: 
\begin{equation}
    \label{eq:bayes_rule}
            p(\bm{\Phi}|\mathbf{D})=
            \frac{ p(\mathbf{D}|\bm{\Phi}) p(\bm{\Phi})} 
            {p(\mathbf{D})}\,,
    \end{equation}
where $\bm{\Phi}$ are model parameters, and $\mathbf{D}$ are observations i.e. training data containing inputs $\mathbf{x}$ and targets $\mathbf{y}$. During inference, the weight distribution is sampled $\bm{\Phi}^\prime\sim q(\bm{\Phi})$, and the sampled weights are used to make a prediction $\hat{y}^\prime = f^{\bm{\Phi}^\prime}(x)$. By taking $K$ samples from $q(\bm{\Phi})$ we can approximate the predictive distribution  $\mathcal{N}({\mathbf{y}},{\mathbf{\sigma}}^2_\text{model})$, which for simplicity we assume to be Gaussian, the model uncertainty is ${\mathbf{\sigma}}^2_\text{model}$ is approximated by Eq.~\ref{eq:model_variance}, and $\hat{\mathbf{y}}$ is the mean of the $K$ predictions
\begin{equation}
    \label{eq:model_mean}
    \hat{\mathbf{y}} \approx
    \frac{1}{K} \sum^K_{\kappa=1} \hat{\mathbf{y}}^{\prime}_\kappa
   .
\end{equation}
A higher value of $K$ will result in a better approximation of the predictive distribution, but this comes at the cost of computational time.

In theory, the posterior distribution in a Bayesian neural network can be calculated using Bayes rule. However, it is usually impossible to obtain an exact estimate of the posterior due to the denominator of Bayes rule, calculated by $p(\mathbf{D})=\int p(\bm{\Phi},\mathbf{D})d\bm{\Phi}$. For neural networks, this is unavailable in closed form and requires exponential time to compute~\cite{blei2017variational}. 
There are several approximate inference techniques which are tractable and provide an alternative to exact inference.

\paragraph{Variational Inference\\} 
Variational inference replaces Bayes Rule with an optimisation task. First, the form of the posterior is constrained to a family of distributions over the latent variables $q(\bm{\Phi}) \in $ \gls{Q_family}. 
The complexity of $\mathcal{Q}$ defines the difficulty of the optimisation. In the simplest case, $\mathcal{Q}$ is set to a multivariate Gaussian with an identity covariance matrix. The goal of variational inference is to minimise the Kullback-Leibler  (\gls{kl}) divergence of the potential posteriors within $\mathcal{Q}$ to the true posterior. 
\begin{equation}
    \label{eq:KL}
    q^\star(\bm{\Phi}) = \underset{q(\bm{\Phi}) \in \mathcal{Q}}{\text{argmin}} D_\text{KL}(q(\bm{\Phi})||p(\bm{\Phi}|\mathbf{D})),
\end{equation}
where $q^\star(\bm{\Phi})$ is the optimum solution. The KL divergence, also known as ``relative entropy''~\cite{hershey2007approximating}  is a statistical measure of the difference between two probability distributions, it quantifies how much extra information, in bits, is needed to approximate the distribution $q$ using the distribution $p$. If $\mathcal{Q}$ is chosen to be more complex, the resulting optimisation space is bigger and correspondingly more difficult, for example by using a Gaussian distribution with full covariance. However, Eq.~\ref{eq:KL} is not tractable because it requires computing the evidence $\log{p(\mathbf{D})}$. This is due to the calculation of the KL divergence~\cite{blei2017variational}, which is:
\begin{equation}
D_\text{KL}(q(\bm{\Phi})||p(\bm{\Phi}|\mathbf{D})) = \mathbb{E}\left[\log{q(\bm{\Phi}})\right]-\mathbb{E}\left[\log{p(\bm{\Phi}|\mathbf{D}})\right],
\end{equation}
where the expectations are taken with respect to $q(\bm{\Phi})$. The conditional $\mathbb{E}\left[\log{p(\bm{\Phi}|\mathbf{D}})\right]$ is expanded, resulting in:
\begin{equation}
D_\text{KL}(q(\bm{\Phi})||p(\bm{\Phi}|\mathbf{D})) = \mathbb{E}\left[\log{q(\bm{\Phi}})\right]-\mathbb{E}\left[\log{p(\bm{\Phi},\mathbf{D}})\right]+\log{p(\mathbf{D})}.
\end{equation}
Which introduces the dependence on $\log{p(\mathbf{D})}$. While we cannot compute this, we can instead optimise the evidence-lower-bound (\gls{elbo})~\cite{blei2017variational}:
\begin{equation}
\label{eq:elbo}
    \text{ELBO}(q)= \mathbb{E}\left[\log\left(p(\mathbf{D}|\bm{\Phi})\right)\right]- D_{\text{KL}}\left[q(\bm{\Phi})||p(\bm{\Phi})\right],
\end{equation}
which is equivalent to Eq.~\ref{eq:KL} up to an added constant $\log{p(\mathbf{D})}$, which itself is independent of $q(\bm{\Phi})$. 

The first component is the expected likelihood, encouraging the posterior to fit the training data. The second term is the KL divergence between the posterior and prior distributions. The KL divergence term behaves similarly to a regulariser, encouraging the model to choose a simple $q(\bm{\Phi})$. Without the KL divergence term, the trained model would be deterministic and the posterior would shrink to single values. Training models with variational inference can be challenging, partly due to the introduction of additional hyperparameters and partly due to training being unstable. 

\paragraph{Dropout\\}
Dropout~\cite{srivastava2014dropout} has been proposed as an approximation to a Bayesian neural network~\cite{gal2016dropout, gal2016theoretically, kendall2015Bayesian, gal2016uncertainty} which is trivial to implement. Dropout is a regularisation technique for reducing overfitting in neural networks. Throughout training, dropout works by randomly ``dropping out'' (setting to zero) a proportion of layer activations during each training step. The random de-activation ensures that the network is not reliant on any single activation and instead must learn general rules. It prevents complex adaptations to the training data where a neuron becomes fine-tuned to only work in the presence of specific activations from other neurons. Having learnt general rules, the network should perform better on unseen data. Dropout is analogous to changing a single neural network into an ensemble of networks, where each network corresponds to a different subset of active neurons. Each training iteration involves a slightly different architecture because different sets of neurons are active or inactive. The ensemble over networks can be seen as a Bayesian neural network where the dropout probability describes the weight distribution. 

When using dropout for regularisation, the parameters are dropped out during training but not during testing and inference i.e,. during testing and inference the dropout proportion is set to $0$. This uses the full predictive power of the neural network and makes it deterministic. In contrast, when dropout is used for estimating uncertainty, the parameters are dropped out during both training and inference i.e,. during testing and inference the dropout proportion is set the same as during training. . This introduces stochasticity into the predictions. To estimate uncertainty, Monte Carlo sampling is employed, where multiple predictions are made for a single input by sampling from the dropout distribution multiple times. These multiple predictions form a distribution of possible outcomes, allowing for the estimation of model uncertainty. 

An advantage of dropout is that the network can be kept largely the same, and can use common loss functions (mean squared error, cross-entropy, etc.), it is common to also use L2 regularisation
\begin{equation}
    \label{eq:DropoutLoss}
    \mathcal{L}_\text{dropout}=\frac{1}{N}\sum_{i=1}^{N}\mathcal{L}(y_i,\hat{y}_i)+\lambda\sum_{i=1}^{L}\bm{\Phi}_i^2,
\end{equation}
where $\mathcal{L}(y_i,\hat{y}_i)$ is a loss function measuring the accuracy of the predictions,  $\sum_{i=1}^{L}\bm{\Phi}_i^2$ is the L2 norm of the model parameters, and \gls{lambda} is a weighting for the two components of the loss function. However, dropout as a method to estimate uncertainty has attracted criticism as it does not have typical properties of a Bayesian model~\cite{osband2016risk, hron2017variational}. For example, neural networks using dropout to estimate uncertainty do not become more confident as they are shown more data, a key aspect of model uncertainty.

\subsubsection{Data uncertainty}
Uncertainty is comprised of both model uncertainty, discussed above, and data uncertainty, which is inherent in the data and not dependent on the modelling process. To estimate data uncertainty, the output layer of a neural network is modified from making a single estimate of the target to estimating the parameters of the distribution from which the target is sampled from~\cite{bishop1994mixture}
\begin{equation}
    \label{eq:data_uncertainty}
    f^{\bm{\Phi}}(\mathbf{x}) = \mathcal{N}\left(\hat{y}, \hat{\sigma}^2\right) ,
\end{equation}
where:
\begin{align}
    \hat{y} &= \hat{y}_1 \\
    \hat{\sigma} &= \texttt{softplus}(\log(e - 1) + \hat{y}_2),
\end{align}
and $\hat{y}_1$ and $\hat{y}_2$ are the outputs of the final layer of the neural network. The $ \texttt{softplus}$ activation applied to $\hat{y}_2$ ensures that the standard deviation is always positive
\begin{equation}
    \label{eq:softplus}
        \texttt{softplus}(x) = \ln({1+e^{x}}) \, ,
\end{equation}
$log(e - 1)$ shifts the standard deviation to $1$ when $\sigma=0$. We found empirically that shifting the standard deviation improves stability and reduces the training time. The network is trained using gradient descent with the negative log-likelihood (NLL) as the loss function:
\begin{equation}
    \label{eq:NLL}
        \text{NLL}(y, \hat{y}, \hat{\sigma}) = 
        \frac{1}{N}\sum_{n=1}^{N}\left(\frac{1}{2 \hat{\sigma}_n^2}{\left(y_n-\hat{y}_n\right)}^2 +\frac{1}{2}\log{\left(2 \pi \hat{\sigma}_n^2\right)} \right) \, ,
\end{equation}
where $\hat{y}$ is the predicted mean, \gls{hat_sigma} is the predicted standard deviation and $y$ is the ground truth. 
The first component of Eq.~\ref{eq:NLL} contains a residual term equivalent to the mean squared error (\gls{mse}) and an uncertainty normalisation term. The second component prevents the model from predicting an infinitely large uncertainty. Minimising the NLL allows us to train an NN despite not having ground truth estimates of the data uncertainty. Although other methods of estimating data uncertainty are available, this method is easy to combine with model uncertainty to concurrently estimate both uncertainties.

\subsubsection{Combining data and model uncertainty}
\label{sec:nn_uncertainy_example}
Data and model uncertainty can be estimated simultaneously by a neural network. For this, we turn our data uncertainty NN from Eq.~\ref{eq:data_uncertainty} into a Bayesian NN by placing a distribution over its weights.
During inference, we sample parameters from the approximate posterior $\bm{\Phi}^\prime\sim q(\bm{\Phi})$ and make an estimate
\begin{equation}
    [\hat{\mathbf{y}}^\prime, \hat{\mathbf{\sigma}}^{\prime 2}] = f^{\bm{\Phi}^\prime}(\mathbf{x}), 
\end{equation}
where $f$ is the neural network parameterised by $\bm{\Phi}^\prime$.  We repeat this $K$ times, drawing $K$ samples from $q(\bm{\Phi})$ to approximate the predictive distribution. The predictive uncertainty is given by\cite{kendall2017uncertainties}:
\begin{equation}
\label{eq:combine_uncertainties}
    \hat{\mathbf{\sigma}}^2 \approx
    {\frac{1}{K} \sum^K_{\kappa=1} \mathbf{\hat{y}}^{\prime 2}_\kappa
    - \left(\frac{1}{K} \sum^K_{\kappa=1} \mathbf{\hat{y}}^\prime _\kappa\right)^2
    + \frac{1}{K} \sum^K_{\kappa=1} \mathbf{\hat{\mathbf{\sigma}}}^{\prime 2}_\kappa} \, .
\end{equation}
In Eq.~\ref{eq:combine_uncertainties}, the first two terms are the variance of the means i.e. the model uncertainty from Eq.~\ref{eq:model_variance}. The third term is the mean of the data variances i.e. the data uncertainty. The predictive mean $\hat{y}$ is the mean of the $K$ forecasts i.e.
\begin{equation}
\label{eq:combine_uncertainties_mean}
    \hat{\mathbf{y}} = \frac{1}{K}\sum_{k=1}^K\hat{\mathbf{y}}^\prime_k \, .
\end{equation}

\subsubsection{Neural Network Uncertainty Example}
\label{sec:nn_uncertainty_example}
Here we provide an example of a Bayesian neural network which we train with both variational inference and dropout as Bayesian approximations. We create a synthetic dataset with $N=50$ training examples for $x\in(-1, 1)$:
\begin{equation}
    \mathbf{y} = 0.882 + \mathbf{x} * 0.2 + 0.489\mathbf{x}^2 + \frac{\sin(4\mathbf{x})(\mathbf{x}+0.5)^2}{2} + \mathcal{N}(0, (0.05)^2).
\end{equation}
The last term $\mathcal{N}(0, (0.05)^2)$ adds noise to the data. We use a neural network with three layers with $20$ hidden units, one output, and two inputs:$x$ and $x^2$. Here we cannot use a linear model as in Section~\ref{sec:uncertainty_linear_regression}, as the dataset contains non-linearities which a linear model could not capture. The previous example used a closed-form solution which was made possible by the model's simplicity and linearity, we cannot apply the same method to the high-dimensional parameter space and non-linearities in a neural network. Instead, we use the approximations described below.

\paragraph{Variational Inference Approximation\\}
For the model using variational inference, we define a prior distribution over the network parameters $\bm{\Phi}$ as an isotropic Gaussian $p(\mathbf{\bm{\Phi}}) = \mathcal{N}(0,1)$. The posterior $q_{\bm{\theta}}(\bm{\Phi})=\mathcal{N}(\mathbf{\mu}_q, \mathbf{\sigma}_q^2)$ is a Gaussian of the same form parameterised by \gls{theta} - which contains the mean \gls{mu} and standard deviation $\sigma$ of the parameters. Half of the values $\bm{\theta}$ are associated with the means $\mathbf{\mu}_q$, denoted $\theta_{\mathbf{\mu}_q}$, and the other half are associated with $\mathbf{\sigma}_q$, denoted $\theta_{\mathbf{\sigma}_q}$. The mean and standard deviation of $q_{\bm{\theta}}(\bm{\Phi})$ are as follows:
\begin{equation}
    \mathbf{\mu}_q = \theta_{\mathbf{\mu}_q},
\end{equation}
\begin{equation}
    \mathbf{\sigma}_q = \texttt{softplus}\left(log(e - 1) + \theta_{\mathbf{\sigma}_q}\right).
\end{equation}
Similarly to in Eq.\ref{eq:data_uncertainty},  the softplus ensures that the standard deviation is always positive.

During training, the dataset is split into $N_\text{batches}$ equally sized subsets (batches). Each gradient is averaged over all elements in one of these batches. Graves~\cite{graves2011practical} proposed minimising the batch-loss:
\begin{equation}
\label{eq:elbo_1m}
    \text{ELBO}(q)= \mathbb{E}\left[\log\left(p(\mathbf{D}_b|\bm{\Phi})\right)\right]- \frac{1}{N_\text{batches}}D_{\text{KL}}\left[q(\bm{\Phi})||p(\bm{\Phi})\right],
\end{equation}

where $\mathbf{D}_b$ is training data for one batch. In this example, we have a small dataset with $50$ examples and use the full dataset for each batch ($N_\text{batches}=1$). 
In one training step, $K=16$ predictions are made, $\hat{y}^\prime = f^{\bm{\Phi}^\prime}(x)$, each time resampling parameters $\bm{\Phi}^\prime\sim q(\bm{\Phi})$. The predictive distribution is calculated using the mean (Eq.~\ref{eq:model_mean} and variance (Eq.~\ref{eq:model_variance}), which is then used to compute the ELBO (Eq.~\ref{eq:elbo_1m}) and update the weights. 

\begin{figure}[!ht]
    \centering
    \includegraphics[width=1.0\linewidth]{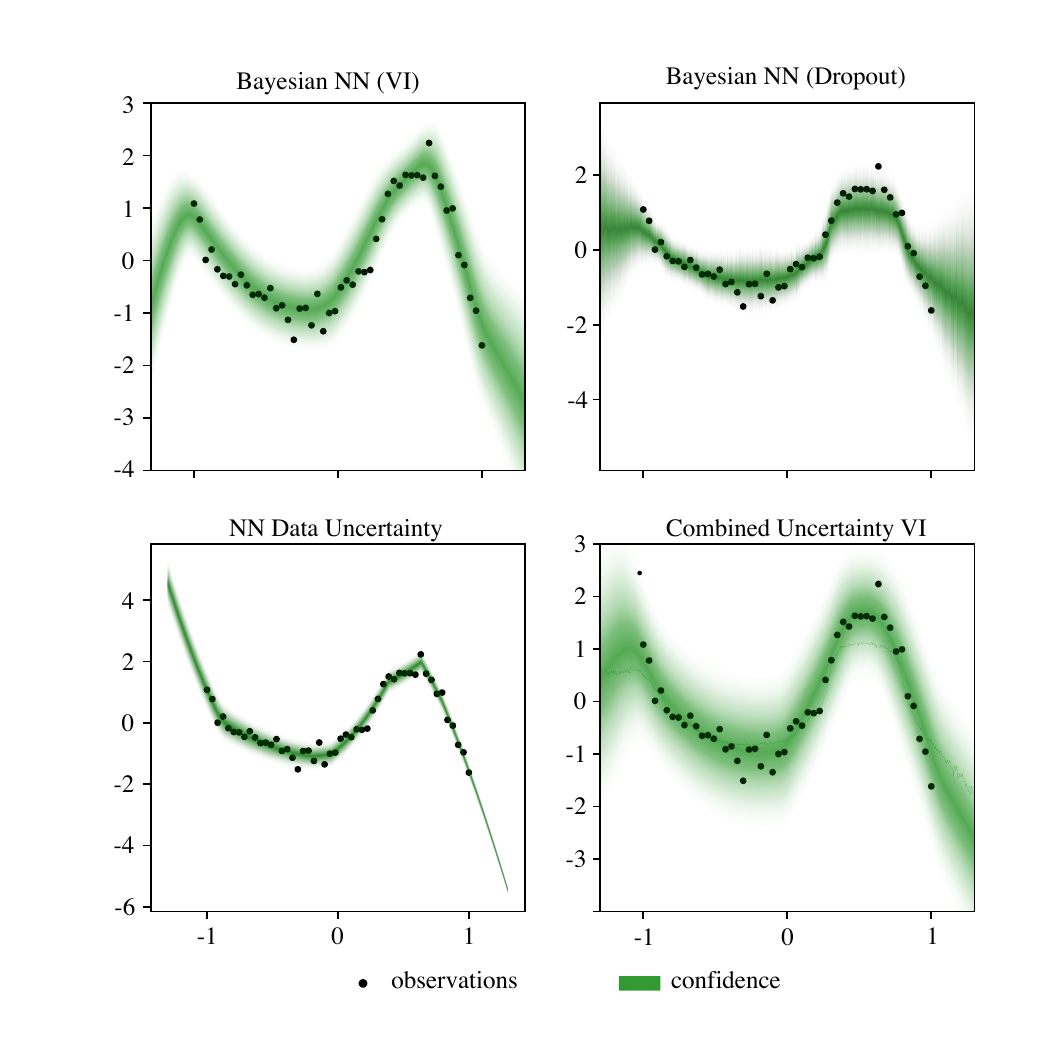}
    \caption[{Comparison of uncertainty estimates in neural networks on synthetic data}]{\textbf{Comparison of uncertainty estimates in neural networks on synthetic data} \newline
    The models in the top row use Bayesian approximations to estimate model uncertainty --- VI denotes variational inference. The bottom left model estimates data uncertainty only. The bottom right model uses both model and data uncertainty by combining the Bayesian neural network trained with VI and the NN Data Uncertainty. The opacity of the confidence interval corresponds to the confidence of the model.}
    \label{fig:VI_Dropout}
\end{figure}

The model converges slowly, requiring $1000$ epochs to train using an Adam optimiser~\cite{kingma2014adam} with a learning rate of $0.001$. Slow convergence is a common problem with variational methods~\cite{graves2011practical, blundell2015weight}. More recently, techniques borrowed from other areas of deep learning such as batch normalisation and learning rate scheduling have been shown to make variational inference faster and more practical~\cite{ioffe2015batch, loshchilov2016sgdr, goyal2017accurate, osawa2019practical}. For this example, we keep the most basic formulation possible and simply train on a large number of epochs. All results are shown in Figure~\ref{fig:VI_Dropout}.

\paragraph{Dropout Example\\}
We use the same network architecture, modifying it to use dropout on each layer. We use dropout with a probability of $0.1$ after the first two layers and use the mean squared error (MSE) and L2 regularisation, with $\lambda = 1e-4$:
\begin{equation}
    \label{eq:DropoutLoss2}
    \mathcal{L}(\hat{y}, y, \bm{\Phi})=\frac{1}{N}\sum_{i=1}^{N}(y_i-\hat{y}_i)^2+\lambda\sum_{i=1}^{L}\bm{\Phi}_i^2,
\end{equation}
We train the model using the same learning rate, optimiser, and number of epochs as the variational inference example. Training time for dropout is significantly faster than variational inference because there are fewer parameters and only one sample per epoch. The uncertainty in predictions is made at inference time using the same method as for the variational inference example, however here we sample from the dropout distributions rather than the distribution over the parameters.. 

\paragraph{Data Uncertainty Example\\}
We again use the same underlying architecture, however, we do not use a Bayesian approximation to estimate model uncertainty. The output layer is modified to have two units which estimate the mean and standard deviation of a normal distribution for the data uncertainty, using Eq.~\ref{eq:data_uncertainty}.
The model is trained using negative-log-likelihood (Eq~\ref{eq:NLL}) as a loss function, keeping the same number of epochs, optimiser, and learning rate as the previous models. 

\paragraph{Combined Uncertainty Example\\}
Finally, we combine the Bayesian neural network with variational inference and the data uncertainty model. 
We use the same underlying architecture, using the Bayesian approximation over the weights which we have already described. The output layer is modified to have two units, which estimate the data uncertainty using Eq.~\ref{eq:data_uncertainty}. For each training step we use $K=16$ samples of the network parameters $\bm{\Phi}^\prime\sim q(\bm{\Phi})$, and compute corresponding outputs $\mathcal{N}\left(\hat{y}^\prime, \hat{\sigma}^{2 \prime}\right) = f^{\bm{\Phi}^\prime}(x)$. The combined uncertainty is calculated using Eq~\ref{eq:combine_uncertainties}, whereas the mean is calculated using Eq.~\ref{eq:model_mean}. We again use the ELBO (Eq.~\ref{eq:elbo_1m}) with $N_\text{batches}=1$. We keep the same $1000$ epochs, Adam optimiser, and $0.001$ learning rate as the other models. 

Figure~\ref{fig:VI_Dropout} presents the results of the four models. The models are trained for $x\in(-1, 1)$ and tested for $x\in(-1.25, 1.25)$. The region outside the training set is out of sample and should exhibit more model uncertainty than examples within the training data range. 
On the in-sample estimates, the model trained with variational inference performs better and fits the training data more closely when compared with the dropout model. This is most noticeable at the peak around $x=0.6$. It is difficult to quantify out-of-sample performance as there is no correct amount of uncertainty. Both models are less confident on the out-of-sample test points, but this is more noticeable on the dropout model than the variational inference model.

The model using only data uncertainty is by far the most accurate on the training set, however, the model's confidence does not change on the out-of-sample predictions, which would be problematic in a real-world situation. The combined uncertainty example performs similarly to the variational inference model containing only model uncertainty; both examples have high model uncertainty which is due to the relatively low data regime and the regularisation caused by the prior. 

\subsubsection{Quantile Regression}
Quantile regression~\cite{koenker2001quantile} is an extension of linear regression which estimates the specific percentiles (quantiles) of a dataset rather than just a mean. This gives an estimation of data uncertainty at various levels. In a typical linear regression model, the output  is calculated by:
\begin{equation}
    \hat{y}=a_1x_{1} + ... + a_px_{p} + b,
\end{equation}
where $\mathbf{a}$ and $b$ are parameters and $\mathbf{x}$ is a vector of inputs of length $p$. Linear models can be trained by minimising the mean squared error (MSE):
\begin{equation}
    \text{MSE}(\hat{y}, y) = \frac{1}{N}\sum^N_{i=1}\left(y_i-\hat{y}_i\right)^2,
\end{equation}
where $N$ is the number of examples in a dataset. In quantile regression, the goal is to find the \emph{median} and quantiles rather than the mean. An estimate is calculated by:
\begin{equation}
    Q_\tau(y)=b_0(\tau)+a_1(\tau)x_{1} + ... + a_\rho(\tau)x_{p},
\end{equation}
where $Q_\tau$ is an estimate for quantile $\tau$. Instead of calculating the MSE the mean absolute deviation (\gls{mad}) is calculated:
\begin{equation}
    \text{MAD} = \frac{1}{N}\sum^N_{i=1}\rho_\tau\left(y_i-Q_\tau(y_i)\right)^2
\end{equation}
where $\rho_\tau$ is an asymmetric function associated with quantile $\tau$ which is calculated by:
\begin{equation}
    \rho_\tau(u)=\tau~\text{max}(\mu, 0)+(1-\tau)~\text{max}(-\mu, 0).
\end{equation}
This loss function is weighted depending on if the error is positive or negative, and is computed separately for different quantiles, i.e. values of $\tau$. For example, when calculating the $10^{\text{th}}$ percentile $\tau$ is $0.1$. The gradient of the loss will be $0.1$ for positive errors and $-0.9$ for negative errors. For $\tau=0.1$ the outputs should be above the true value $10\%$of the time, and below the true value  $90\%$ of the time. 

Quantile regression has been applied to many different models, including feed-forward neural networks~\cite{taylor2000quantile}, recurrent neural networks~\cite{gasthaus2019probabilistic} and sequence-to-sequence models~\cite{wen2017multi}(models which forecast at multiple horizons simultaneously). These papers highlight that an advantage of quantile regression is the model can learn any shape of output distribution. However, quantile regression introduces additional computational complexity to estimating data uncertainty and generally requires large amounts of data to train a model to estimate multiple quantiles. As disease forecasting is a low data regime and estimating model uncertainty already introduces significant computation overhead, we instead favour the more simple data uncertainty estimation method of modifying the output to a distribution and learning its parameters. 

\subsubsection{Conformal Prediction}
Conformal Prediction is another way of measuring confidence however it is difficult to compare with the methods outlined above as it does not make comparable uncertainty estimates. Conformal prediction~\cite{shafer2008tutorial} generates prediction intervals based on a model's accuracy on a training and validation set. Prediction intervals are not confidence intervals in the traditional sense but are instead empirical measures of the model's ability to make predictions that align with observations seen during training. In conformal prediction, the training data is divided into two parts: the training set and the calibration set. A model, such as a neural network, is trained on the training set using traditional techniques such as minimising the MSE via gradient descent. 
Finally, predictions are made on the calibration set using the trained model and the predictions are used to calculate the nonconformity score e.g., the MSE for regression tasks. The nonconformity score gauges how atypical a given example is considering both the training and calibration data. 
This score guides the construction of prediction regions for new data points based on their similarity to examples in the calibration set. 

Conformal prediction rests on the assumption of data exchangeability, meaning that the underlying distribution of the data remains consistent over time and across instances. Given exchangeability, conformal prediction offers guaranteed coverage: a $95\%$ prediction interval, for instance, will encompass the true value $95\%$ of the time. Conformal prediction is adaptable and does not rely on rigid assumptions about data distribution or model specifics, but it does demand data exchangeability. Time series data, particularly in epidemic forecasting, is non-exchangeable as statistical properties shift over time and each epidemic season presents unique challenges. Due to the issues with exchangeability, we do not attempt to apply conformal prediction to epidemic modelling.

\subsection{ILI Forecasting Models}
\label{sec:nonmechILImodels}
Here we discuss existing ILI forecasting models. The focus is on non-mechanistic models, leaving a discussion of mechanistic models for Chapter \ref{chapterlabel2}. Note that non-mechanistic models are typically more accurate for ILI forecasting~\cite{reich2019collaborative}.

\subsubsection{Auto-Regressive-Integrated-Moving-Average}
The Box-Jenkins Auto-Regressive-Integrated-Moving-Average (\gls{arima}) model~\cite{box2015time} is a time series forecasting technique which uses autoregressive (AR) and moving average (MA) models. 

ARIMA models consist of three parts, Autoregressive (AR), Integrated (I) and Moving Average (MA). 
The autoregressive (AR) aspect predicts future values based on its own past values, defined mathematically as:
\begin{equation}
    y_t = c + \sum^p_{n=1}\alpha_ny_{t-n}+\epsilon_t
\end{equation}
where $\alpha_n$ is a coefficient associated with the time series at a lag $n$, $c$ is the intercept, and $\epsilon_t$ is an error term. Note that we use standard notation when describing ARIMA models, but in other sections of the thesis, the notation is used differently.

The integrated (I) component makes the time series stationary. Stationarity ensures that the properties of the time series remain consistent over time~\cite{kwiatkowski1992testing}. A non-stationary time series can be transformed to a stationary one using differencing using one or more times. This procedure constructs a new series, $y^\prime$, derived from the differences between consecutive points of the original series $y$:
\begin{equation}
    y_t^\prime = y_t-y_{t-1} .
\end{equation}
Other techniques, such as logarithmic transformations, can be used in conjunction with differencing. For instance, in~\cite{soebiyanto2010modeling}, the authors enhanced the accuracy of ARIMA by differencing the logarithm of the ILI proportion, thereby providing more linear variances for influenza cases. The Augmented Dickey Fuller test~\cite{cheung1995lag}, a statistical test based on a t-statistic, is used to determine the differencing order needed to ensure stationarity.

The Moving Average (MA) models focus solely on lagged forecast errors:
\begin{equation}
    y_t = c + \sum^q_{n=1}\theta_n\epsilon_{t-n}+\epsilon_t
\end{equation}

Where $\epsilon_t$ is the error terms at time $t$ and $\theta_t$ is an associated weight. The ARIMA model is the sum of these components:
\begin{equation}
    y_t = c + \sum^p_{n=1}\alpha_ny_{t-n} + \sum^q_{n=1}\theta_n\epsilon_{t-n}+\epsilon_t.
\end{equation}

Incorporating external variables such as environmental factors can improve ARIMA models. 
In~\cite{soebiyanto2010modeling} the authors incorporated humidity, which is known to influence influenza transmissibility~\cite{shaman2010absolute, shaman2013real}. In warmer climates such as Arizona and Hong Kong, ILI primarily spreads through physical contact. Weather plays a crucial role by affecting human behaviour, thereby indirectly dictating disease transmission patterns. The inclusion of humidity weather data into the model significantly improved forecast accuracy. In another study conducted in Wuhan\cite{he2018epidemiology}, researchers used ARIMA to predict the positive rate of influenza tests among hospitalised children up to a month in advance. 
Notably, the authors tested their model over a six-month timeframe after the peak of the flu season, bringing into question how well this model will generalise across different flu seasons.

The Seasonal-Auto-Regressive-Integrated-Moving-Average (\gls{sarima}) model extends ARIMA by incorporating a Seasonal (S) component; it has been used by the ReichLab group as a baseline~\cite{ray2017infectious} and in an ensemble approach for the FluSight competition~\cite{ray2018prediction}. 
SARIMA works by extending the ARIMA model with components that are back-shifted by the seasonal period, $s$: 
\begin{equation}
    y_t = c + \sum^p_{n=1}\alpha_ny_{t-n} + \sum^q_{n=1}\theta_n\epsilon_{t-n}+      
                   \sum^P_{n=1}\phi_ny_{t-sn} + \sum^Q_{n=1}\eta_n\epsilon_{t-sn} +
                   \epsilon_t,
\end{equation}
where $\phi_t$ and $\eta_t$ are the coefficients associated with the lagged autoregressive and lagged moving averages, respectively. Further extensions to the model included non-linear basis functions which improved model flexibility and allowed improved accuracy, as well kernel density methods to provide uncertainty estimates. Delphi group's basis regression~\cite{brooks2018nonmechanistic} uses similar basis functions and kernel methods for the same task.

Both the ReichLab and Delphi groups extended ARIMA models using non-linear basis functions, enhancing model versatility. However, this also reduced their capability to handle high-dimensional data. To simplify model fitting, Delphi compresses their input data, which limits how Web search data can be used. Neural networks are scaleable while being able to estimate uncertainty and thus present a better method for forecasting with large input spaces. The inflexibility of the Reichlab and Delphi models meant in the Flusight competition that they were beaten by Dante, which we use as a baseline. 

\subsubsection{Dante}
\label{sec:Dante}
Dante~\cite{osthus2021multiscale} is an influenza forecasting model that learns spatial, temporal, and data structure at a state, regional, and National level. Dante uses random walk models~\cite{ibe2013elements} conditioned on ILI data at different spatial levels. State-level forecasts are aggregated based on census data to create forecasts for larger geographic areas. The disease propagation within regions as well as the interaction between regions are modelled explicitly by separate models.

Dante uses separate models for data and model uncertainty, denoted the data and process models. We describe Dante using the original notation, which is used differently in other parts of the thesis. 
The data model estimates the distribution of the observed ILI proportion $y_{\text{rst}}$ in state $r$, season $s$ during week $t$ with a Beta distribution. The variance of the distribution is given by:
\begin{equation}
    \sigma^2(y_{\text{rst}}|\theta_{\text{rst}}, \lambda_r)=\frac{\theta_{\text{rst}}(1-\theta_{\text{rst}})}{1+\lambda_r},
\end{equation}
where $\theta_{\text{rst}}$ is the unobservable true ILI proportion i.e. the ILI proportion if there were no data uncertainty. A state-specific parameter $\lambda_r>0$ captures the amount of noise in the measurements for each state. Different states have different measurement errors for their ILI proportions, hence each state has its own value $\lambda$. A small state like Hawaii will have noisier ILI measurements than a large state like California due to the smaller population, and hence lower number of outpatient visits each week. The authors note a negative logarithmic correlation between the average number of outpatients per week and the week-to-week ILI volatility. Volatility quantifies the noise in ILI proportions at a state, regional, and national level, from week to week. The relationship between $\lambda_r$ and the outpatient count is unknown, so it is learnt from the data.

The process model uses four components to model the ILI proportion:
\begin{equation}
    y_{\text{rst}} = 
    \mu^{\text{all}}_t + 
    \mu^{\text{state}}_{rt} + 
    \mu^{\text{season}}_{st} +
    \mu^{\text{interaction}}_{\text{rst}}, 
\end{equation}
where $\mu^{\text{all}}_t$ and $\mu^{\text{state}}_{rt}$ are season-independent noise terms which are modelled by random walks. There are two season-dependent terms $\mu^{\text{season}}_{st}$ and $\mu^{\text{interaction}}_{\text{rst}}$ which model the state on its own and interactions between states, respectively. 

Finally, there is an aggregation model which linearly combines state-level estimates based on population to either a state or national level. The use of state-level noise parameters allows the model to work well on states which may have significantly more or less noise in their ILI reporting than others. Dante is fit to data using Markov-Chain-Monte-Carlo (\gls{mcmc}) sampling to learn posterior distributions for the process model and fit the state-level $\lambda$ parameters. 

Dante won the Flusight competition~\cite{osthus2021multiscale}, and as such we consider it to be state-of-the-art. We use Dante as a baseline which we compare our models with. We found that training Dante can be difficult --- forecasts are made individually, one observation at a time, and the model is retrained between forecasts. The authors suggested that Dante could be improved by using web search queries, however, due to the complexities of the existing model it is not clear how this could be achieved without significant computational overhead. A more recent model ``Inferno''  by the same authors~\cite{osthus2022fast} instead focused on speeding up Dante by removing the interaction between states. This speeds up training the model at the cost of a slight reduction in accuracy.

\subsection{Web Search Data for Disease Modelling}
Search query frequency data reports the frequency of internet searches for given terms. Query data can be used to give information about a population. For example, if the frequency of searches for ``flu symptoms'' suddenly increases, then there is a good chance that many people think that they might have the flu. This is not foolproof --- search query frequencies can change erroneously due to unassociated causes such as news or social media etc.. Additionally, search queries can also correlate with the target time series despite being completely unrelated - a case in point being "Christmas". There are several models using Web search data to inform disease modelling.

\subsubsection{Google Flu Trends}
Google Flu Trends (\gls{gft})~\cite{ginsberg2009detecting} was a web service which used Google search query frequencies to nowcast (estimate the current amount) the ILI proportion at a city level in the United States. GFT used a linear model to predict the ILI proportion based on the frequency of searches for set term. We use the original notation to describe the model:
\begin{equation}
    \text{logit}(ILI) = c+a~\text{logit}(Q) + \epsilon.
\end{equation}
Where $Q$ is an aggregated set of queries, $c$ is the intercept, $a$ is a multiplicative coefficient, $\epsilon$ is an error term and $\text{logit}(x)$ is $\log{\frac{x}{1-x}}$. The authors used the $45$ search queries which correlated closely with influenza proportions and were on a related topic. For example, ``high school basketball'' was not included despite it having a good correlation with the ILI proportion. These queries were then averaged into a single variable. 

The model had good performance when nowcasting ILI and GFT became a Google service. However, the service attracted criticism after overestimating the ILI proportion in $2013$ (four years after publication), culminating with the service being abandoned in $2015$. There were several causes attributed to this failure~\cite{santillana2014can}. Combining multiple queries into a single feature ignored variability between different queries and made the model susceptible to sudden changes in an individual query's usage. Ignoring queries based on the author's opinion introduced bias into the model. The model used the same queries each year, selecting queries once and using them for many years ignoring changes in user-search behaviour over time. The query-selection method also ignored how closely a query semantically correlated to what they were trying to estimate. Finally, the model itself was simplistic and could not account for non-linear relationships between variables. 

\subsubsection{Advanced Models Using Web Search Data}
There have been several works improving on GFT. An Elasticnet was used in~\cite{lampos2015advances} to evaluate the utility of Web search queries for ILI nowcasting after the failure of GFT. An Elasticnet is a linear regression model with two regularisation terms to penalise large weights and encourage the model to base estimates of the minimum number of features. The objective function for an Elasticnet is:
\begin{equation}
    \label{eq:elasticnet}
    \text{argmin}{\mathbf{w}, \beta}\left(
    \sum^T_{t=1}(\mathbf{w}^{\mathbf{T}}\mathbf{x}_t+b-y_t)^2 
    +\lambda_1\sum^I_{j=1}|w_j|
    +\lambda_2\sum^I_{j=1}w_j^2\right),
\end{equation}
where $\mathbf{w}$ is a vector of $I$ weights, $b$ is the intercept, $\mathbf{x}_t$ is a vector of query frequencies at time $t$, $y_t$ is the ILI proportion at time $t$, and $\lambda_1$ and $\lambda_2$ are hyperparameters which determine the degree of regularisation. The first regularisation term (determined by $\lambda_1$) is an $L1$ lasso regularisation term. The second (determined by $\lambda_2$) is an $L2$ ridge regularisation term. The model's trained weights can be inspected to determine which inputs are useful and which are not. This allows the modeller to make informed decisions about which queries to use, without introducing user bias. The Elasticnet can use multiple inputs rather than a single aggregated input thereby increasing flexibility and making the model more robust to changes in search behaviour. 

In the same paper by Lampos et al., a Gaussian Process (\gls{gp}) regression model improved on the Elasticnet. As GP models do not work well with high dimensional inputs, queries were clustered together to create a low dimensional input. Finally, the GP and Elasticnet models were combined with an ARMAX, using the same notation as we used for ARIMA, this is defined by:
\begin{equation}
    y_t = c + \sum^p_{n=1}\alpha_ny_{t-n} + \sum^q_{n=1}\theta_n\epsilon_{t-n}+               \sum^r_{n=1}\beta_nx_{n_t} + \epsilon,
\end{equation}
where $\mathbf{x}_{t}$ are exogenous inputs associated with time $t$, in this case, outputs from the GP and Elasticnet models. The ARMAX uses the lagged ILI proportion to improve estimates. The combination of ARMAX and GP regression gave the best set of results and significantly outperformed GFT in both accuracy and stability. This work showed that Web search query data can improve estimates compared to purely autoregressive models. However, the work focused on nowcasting and did not attempt to forecast ILI based on Web search query data.

Further work by Lampos et al.~\cite{lampos2017enhancing} refined the query selection method using word embeddings learnt from Twitter data to determine the semantic similarity between a search query and a concept related to influenza. A ``similarity score'' was proposed:
\begin{equation}
    S(\mathcal{Q},\mathcal{C})=\frac{\sum_{i=1}^{k}{\cos(\mathbf{e}_\mathcal{Q},\mathbf{e}_{P_i})}}{\sum_{j=1}^{z}{\cos(\mathbf{e}_\mathcal{Q},\mathbf{e}_{N_j})+\gamma}}
\end{equation}
where gls{Q} is the query being evaluated, $\mathbf{e}_\mathcal{Q}$ is an embedding of a query, $\gamma=0.001$ is a constant to stop divide-by-zero errors, and $\mathcal{C}$ is a ``concept'' that contains both positive embedding $\mathbf{e}_\mathcal{P}$ and negative embedding $\mathbf{e}_\mathcal{N}$ examples of a query. For a concept relating to influenza, positive examples included: ``flu'', ``flu fever'', ``flu symptoms'' and ``flu treatment'', and negative examples included: ``ebola'' and ``reflux''. Cosine similarities were transformed by $(x+1)/2$ to avoid negative sub-scores. The word embeddings were constructed using a continuous bag of words (CBOW) from Twitter data. This query selection method was compared to a `hybrid' method which additionally used a correlation score --- the bivariate correlation between the ILI proportion and the time series of search frequencies. Queries were filtered according to their two scores. The similarity score removed queries such as `skiing' which may correlate well to the ILI proportion but were unrelated. The hybrid method provided the best nowcasting results and improved on the previous work. We build on this work and use a similar method for query selection. 

Other work attempted to use Twitter data to improve ILI modelling~\cite{culotta2010towards, aramaki2011twitter, paul2014twitter}. However, Twitter data is harder to interpret than query data, and query data is better for nowcasting the ILI proportion~\cite{wagner2018added}. In the rest of this chapter, we combine search queries and neural networks and focus on forecasting ILI in the US. 

\section{Methods}
\label{sec:nn_methods}
We first describe the data sets used, then introduce the neural network architectures we have deployed, and finally detail how training and validation were performed.

\subsection{Datasets and Web Search Query Selection}
\label{sec:data}
\subsubsection{Influenza-like illness (ILI) proportions} 
CDC defines ILI as fever (temperature of $37.8^{\circ}\text{C}$ or greater) and a cough and/or sore throat without a known cause besides influenza. ILI is monitored through several surveillance efforts including the Outpatient Influenza-like Illness Surveillance Network (ILINet) which collects weekly state-level ILI proportions from over $2{,}000$ healthcare providers from all states. The state-level ILI proportions are weighted by population size to report the wILI at different geographic levels~\cite{cdc_flu}. Our models use weekly wILI proportions for the flu seasons $2004/05$ to $2018/19$ inclusive.\footnote{Data obtained from \href{https://gis.cdc.gov/grasp/fluview/fluportaldashboard.html}{gis.cdc.gov/grasp/fluview/fluportaldashboard.html}} Note that this data is not final i.e. it can be revised by the CDC. To ensure reproducibility of our results, a copy of all the ILI data used can be found in our GitHub repository.\footnote{ \href{https://github.com/M-Morris-95/Forecasting-Influenza-Using-Neural-Networks-with-Uncertainty}{github.com/M-Morris-95/Forecasting-Influenza-Using-Neural-Networks-with-Uncertainty}} A week in the CDC data represents a seven-day period that starts on a Sunday and ends on a Saturday. We assume the weekly ILI proportion is representative of Wednesday (middle day) and use cubic interpolation\footnote{As implemented in \texttt{interpolate.interp1d} from Python's \texttt{SciPy} library.} to generate daily ILI proportions. This not only increases the number of samples (seven-fold) but also provides an aligned time series with the daily temporal resolution of the Web search activity data. The deployment of a cubic as opposed to a linear interpolation to generate daily ILI proportions resulted in slightly better forecasting accuracy on the test sets. We hypothesise that this is because of the increased level of smoothness (see~\ref{fig:sup_daily_and_weekly_ILI_rate}), but we have not fully assessed this data manipulation choice. For training the Dante forecasting model, we have also obtained regional wILI proportions for the 53 US states/locations and the 10 US Health and Human Services regions. These were downloaded from the CDC for the same period above and are also available on our GitHub repository.

\subsubsection{Search query frequency time series} 
Search query frequencies for the US are obtained from the Google Health Trends API, as for similar studies~\cite{Lampos2017WWW,lampos2021covid}. A frequency represents the fraction of searches for a certain term or set of terms divided by the total amount of searches (for any term) for a day and a certain location. We initially downloaded the daily search frequencies of a predetermined pool of $20{,}856$ unique US health-related search queries for the period from March 2004 to May 2019 inclusive, for the US. Query frequencies are smoothed using a seven-day moving average, and min-max normalisation is applied to each query's time series during training (i.e. without using any future data). For a given test season, for each query, $\mathcal{Q}$, we compute the bivariate correlation $r_\mathcal{Q}$ with the ILI proportion over the five seasons preceding the test season. We also compute a semantic similarity score $s_\mathcal{Q}$ that measures each query's similarity to a predefined flu concept as described in Lampos et al. (2017)\cite{Lampos2017WWW}. Both scores are then normalised between 0 and 1 and a composite score $U_q=r_\mathcal{Q}^2+s_\mathcal{Q}^2$ for each query is calculated. Only the \gls{m} queries with the highest $U_q$ are used, where $m$ is a hyperparameter (see ``Hyperparameter optimisation''). 

\begin{table*}[!ht]
    \centering
    \small
    \begin{tabular}{lll}
    \toprule
    Topic                   &   Proportion (\%)  & Examples \\
    \midrule
    Symptoms                & 	37          &	cold flu symptoms, flu sore throat, flu nausea \\
    General/Strains         &	19          &	flu, b flu, flu strain                         \\
    Medicine                & 	18          &	flu tamiflu, flu medicine, flu treatment       \\
    Influenza in Children   &	11          &	flu infants, flu toddler                       \\
    Incubation/Spread       &	8           &	flu incubation period, cold contagious         \\
    Testing/Advice	        &   7           &	flu swab, flu help                             \\
    \bottomrule
    \end{tabular}
    \caption[{Manually curated topics for web search queries}]{\textbf{Manually curated topics for web search queries}\newline
    A demonstration of manually curated topics based on the Web search queries used in forecasting models trained for and tested on the 2015/16 flu season. Please note that we do not use query topics in our forecasting models.}
    \label{tab:sup_search_queries}
\end{table*}

\newpage
\subsection{Neural network architectures}
The three NN architectures we have deployed are described next. Each NN outputs two values, namely an ILI proportion forecast estimate ($\hat{y}$) and an associated data uncertainty ($\hat{\sigma}$). Each architecture also has an additional Bayesian layer where the weights are specified by an associated probability distribution $q(\bm{\Phi})$. The predictive distribution is approximated by sampling network parameters $\bm{\Phi}^\prime\sim q(\bm{\Phi})$ and computing the output each sample. The outputs are used to estimate model uncertainty. 

\subsubsection{Feed-Forward Neural Network (FF)} 
\begin{figure}[!ht]
\centering
\includegraphics{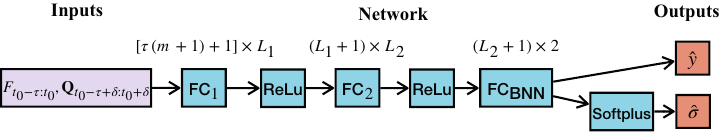}
\caption[{Diagram of FF architecture}]{\textbf{Diagram of FF architecture} \newline
Diagram of the feed-forward (FF) NN architecture with dimensions of parameter matrices shown. $L_1$ and $L_2$ denote the number of units in fully connected layers $\text{FC}_1$ and $\text{FC}_2$, respectively. $\text{FC}_{\text{BNN}}$ denotes a fully connected layer with a distribution over its weights.}
\label{fig:sup_ff_diagram}
\end{figure}

The FF model has two hidden feed-forward neural layers with a ReLU ($\max(0, x)$) activation function, and a Bayesian feed-forward layer (Figure~\ref{fig:sup_ff_diagram}). 
Feed-forward layers are described in Section~\ref{sec:ff_nn_intro} - the layer outputs are $\mathbf{y} = g\left(\mathbf{W}^T\mathbf{x} + \mathbf{b}\right)$  where $\mathbf{x}$ are inputs, $g$ is an activation function, $\mathbf{W}$ are weights and $\mathbf{b}$ are biases.
the NN estimates model and data uncertainty in the output layer using the combined uncertainty mechanism in Section~\ref{sec:nn_uncertainty_example}.
The input to the network is a window of \gls{tau}$ + 1$ days of ILI proportions and $m$ search query frequencies. There is an ILI proportion collection delay of $\delta$ days, in that at day $t_0$ we know (CDC has published) the ILI proportion of day $t_0 -\delta$. The delay is assumed to be $\delta=14$ days throughout our experiments. Thus, at day $t_0$, the input to the network consists of a window of \gls{tau} ILI proportions, \gls{F_input}$_{t_0-\tau}$ to $F_{t_0}$, and search query frequencies, \gls{Q_input}$_{t_0 + \delta - \tau}$ through $Q_{t_0+\delta}$. Because there is no mechanism to use temporal structure in feed-forward neural networks, we ignore the temporal structure of the data and use an $(m+1) \times (\tau + 1)$ vector as the input to the neural network. The output of the network is an estimate of the ILI proportion and corresponding data uncertainty $\gamma$ days ahead.
\begin{figure}[!ht]
    \centering
    \includegraphics{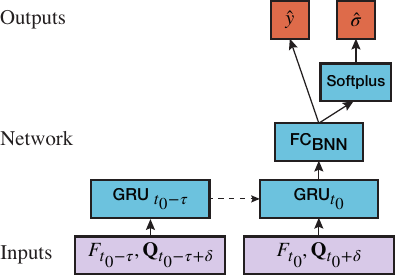}
    \caption[{Diagram of SRNN architecture}]{\textbf{Diagram of SRNN architecture} \newline
    Diagram of the Simple RNN (SRNN) architecture. GRU (Gated Recurrent Unit) is a recurrent layer, and FC denotes a fully connected dense layer. $\text{FC}_{\text{BNN}}$ denotes a fully connected layer, which uses a distribution over its weights to computer uncertainty. Note that the query frequencies ($Q$) and the ILI proportions ($F$) are temporally misaligned by $\delta$ days. 
    }
    \label{fig:sup_srnn_diagram}
\end{figure}
\subsubsection{Simple Recurrent Neural Network (SRNN)} This is a recurrent neural network which observes a time series of ILI proportions and search frequencies (\ref{fig:sup_srnn_diagram}). The input to the network is the same as for FF, but without flattening into a vector. The inputs are an $(m+1) \times (\tau + 1)$ matrix, which is fed into a Gated Recurrent Unit (GRU) layer (Section~\ref{sec:rnn_intro}) one day at a time i.e., $m+1$ dimensional vectors are iteratively passed into the GRU $\tau+1$ times. The last output of the GRU is passed to a feed-forward layer with a distribution over its weights, which calculates uncertainty in the same way as the FF model.

\subsubsection{Iterative Recurrent Neural Network (IRNN)} 
\begin{figure*}[t]
\centering
\includegraphics[width=0.99\linewidth]{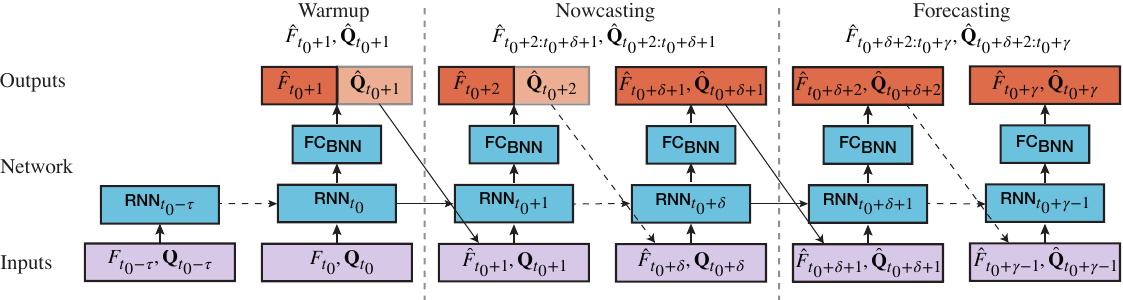}
\caption[{IRNN architecture Diagram}]{\textbf{IRNN architecture Diagram} \newline Diagram of the~IRNN architecture where for the recurrent layers (RNN) we have used a Gated Recurrent Unit. An ILI proportion, $F \in [0,1]$, and $m$ search query frequencies, $\mathbf{Q}$ $\in \mathbf{R}_{\ge 0}^m$, beginning from time point (day) $t_0 -\tau$ are fed into the network a day at a time. $\tau$ denotes the window size of past observations that we consider ($\tau + 1 = 56$ days). The reporting delay of the ILI proportions means that when ILI proportions are available up to day $t_0$, search query frequencies are available up to day $t_0 + \delta$, where $\delta = 14$ days in our experiments. Dashed arrow lines denote that the model is called for multiple timesteps (where a timestep is a day). For days $t_0-\tau$ to $t_0$, IRNN enters a warm-up phase where it sets the hidden states in the RNN layer without making any predictions. For days $t_0$ to $t_0+\delta$, we can observe search query frequencies, but we cannot observe ILI proportions. At this stage, IRNN performs nowcasting with respect to input $\mathbf{Q}$. During nowcasting the estimated ILI proportion $\hat{F}_t$ is combined with the true search frequencies $\mathbf{Q}_t$ and used as the input for the next timestep. The query search frequency estimates which are not used (as they are known to us) are shown by a faded box. For days $t_0+\delta+1$ to $t_0+\gamma$, where $\gamma$ denotes the forecasting horizon, IRNN conducts pure forecasting as neither search query frequencies nor ILI proportions are known for that period. Forecasted values for both of them are used as inputs for subsequent timesteps. The full sequence of both predicted ILI proportions and search query frequencies is used in the training loss.}
\label{fig:rnn_fdbk_full_diagram}
\end{figure*}

This is a recurrent neural network which makes forecasts of the ILI proportion and search frequencies one day at a time. It bases forecasts on its own previous forecasts. IRNN comprises a recurrent GRU layer and a feed-forward Bayesian layer as shown in Figure~\ref{fig:rnn_fdbk_full_diagram}. We have also described how model training works with pseudocode in the Supporting Information (\ref{fig:sup_algorithm}). Given its special structure, IRNN does not incorporate future (for a period of seven days after the target forecast) search query frequencies when $\gamma = 7$. Hence, for both $\gamma = $ $7$ and $14$, the only minor difference may be due to the more recent past ILI proportion inputs. As a result, the difference in performance between $\gamma = $ $7$ and $14$ is expected to be minor given that search query frequencies are always the more recent information source (as opposed to past ILI proportions). This is also empirically confirmed by our experiments (see Table~\ref{tab:nn_comparison}). A caveat of the current formulation of IRNN is that the model is agnostic of the actual forecast horizon and hence its uncertainty might be underestimated for larger forecasting horizons.

\subsection{Experiments}
We first introduce the training setup for a BNN, and the variations which are used for the different architectures, then we discuss hyperparameter optimisation, and finally how the evaluation is performed in our experiments.

\subsubsection{Training} 
When training the FF and SRNN models, each training step takes an $\left[(m+1) \times (\tau+1)\right]$-dimensional input (where $m$ denotes the number of search queries and $\tau+1$ denotes the window of days, from $t_0$ and back, for which query frequencies and ILI proportions are used) and produces a forecast estimate $\hat{y}_{t_0+\gamma}$ containing both a mean and standard deviation for the ILI proportion for time (day) $t_0+\gamma$. The parameters $\bm{\Phi}$ are updated by minimising Eq.~\ref{eq:elbo}. During each training step, one sample is taken from $q_{\bm{\theta}}(\bm{\Phi})$ and used to compute the ELBO, where $\bm{\theta}$ describes the mean and standard deviation of parameter distribution (Section~\ref{sec:nn_uncertainty_example}). We use back-propagation to compute gradients and update the parameters in both the Bayesian and non-Bayesian layers. The model is retrained for each time horizon $\gamma$, where $\gamma = 7$, $14$, $21$ or $28$ days, and for each test period.

The output of the IRNN is a sequence of ILI proportions and search frequencies. Although we have search data from $t_0$ to $t_0+\delta$, we use the full sequence of estimated query frequencies when back-propagating the ELBO (Eq.~\ref{eq:elbo}) through time. When evaluating the model's performance we are only concerned with the model's ILI proportion forecasts. The Bayesian layer is called once for each iterative prediction.

\subsubsection{Hyperparameter optimisation} We use Bayesian hyperparameter optimisation\cite{shahriari2015taking} with five-fold cross-validation where each fold is 365 days covering a full flu season (see \ref{tab:sup_val_periods}, \ref{fig:sup_validation_periods}, and \ref{fig:sup_k_fold_diagram}). We tune the hyperparameters once before the first test period and keep the same hyperparameters for all subsequent test seasons. For the FF and SRNN, the hyperparameters are re-tuned for each of the four forecast horizons. For the IRNN the hyperparameters are tuned once, considering all four forecasting horizons (the average NLL is computed across them). The hyperparameters are the following: the size of the hidden NN layers $\in [25, 125]$, the number of queries $m$ $\in [20, 150]$, the weighting of the KL divergence term in the ELBO loss $KL_w \in [0.0001, 1.0]$, the scaling factor of the output's standard deviation $s \in [1.0, 100]$, the prior standard deviation $\sigma_p \in [0.0001, 0.1]$, the number of epochs $\in [10,100]$, and the learning rate $\in [0.0001, 0.01]$ for training the NNs. After the hyperparameters are tuned we re-train the model using the full training set for the number of epochs chosen. The derived model is then used for forecasting on the test set. Note that hyperparameters are not re-tuned for comparison with Dante (when Web search activity data that are more recent than the last observed ILI proportion are removed), which may have disadvantaged our NN models.

\subsubsection{Inference} When making an estimate with a BNN based on inputs $\mathbf{X}$, and with training data $\mathbf{D}$, the goal is to compute an output $\mathbf{\hat{y}}$ for the entire distribution over $\bm{\Phi}$:
\begin{equation}
\label{eq:BNN_output}
    p(\mathbf{\hat{y}}|\mathbf{X}, \mathbf{D}) = 
    \int_{\bm{\Phi}} 
    p(\mathbf{\hat{y}}^\prime|\mathbf{X},\bm{\Phi}^\prime) 
    p(\bm{\Phi}^\prime|\mathbf{D}) 
    d\bm{\Phi}^\prime \,.
\end{equation}
In practice, $ p(\mathbf{\hat{y}}|\mathbf{X}, \mathbf{D})$ is estimated using Monte-Carlo sampling from $p(\bm{\Phi}|\mathbf{D})$\cite{jospin2020hands}. At prediction time, the posterior distribution over the weights is sampled $K$ times, each giving an output $\mathcal{N}(\hat{y}^\prime, \sigma^\prime)$. The $K$ estimates are combined using Eq.~\ref{eq:combine_uncertainties} which makes an estimate for the combined model and data uncertainty. $K$ is chosen by sampling until the final estimate of $\hat{y}$ stabilises. Initially, we sample $10$ times and produce an estimate using Eq.~\ref{eq:combine_uncertainties}. We then run the model a further $10$ times and produce a new estimate using the $20$ samples. We repeat this process until increasing $K$ by $10$ does not change the estimated mean by more than $0.1\%$. Despite averaging over $K$ instances of the model, we observed some instability in training the models. 

To resolve this each model was trained $10$ times with different initialisation seeds i.e. the seed controlling the initial parameter values of the NN. The mean of the $10$ estimates of forecasts and associated variances are our final forecast and variance. We considered alternate methods of combining estimates, such as Eq.~\ref{eq:combine_uncertainties} and averaging the probability density functions. Ultimately, we found that averaging the means and variances gave the best final forecasts. Thus, the total number of samples for making a forecast is equal to $\sum_{i=1}^{10} K_c \times 10$, where $i \in \{1,\dots,10\}$ denotes a different seed, and $K_c \ge 20$ is the number of samples required for this seed to converge.

To estimate with the SRNN and FF models, the inputs are passed through the model's layers up to the Bayesian layer. The weights in the Bayesian layer are then sampled $K$ times, and the estimates from the $K$ samples are combined with Eq.~\ref{eq:combine_uncertainties} as discussed in the previous paragraph. Estimating with IRNN has three distinct phases: warm-up, nowcasting, and forecasting (\ref{fig:sup_algorithm}). During the warm-up phase, the model observes ILI proportions and $m$ search queries from  $t_0-\tau$ to $t_0$. This sets the hidden states of the GRU layer based on all ILI proportions and search frequencies from the same days. The output of the GRU is fed into the Bayesian layer (denoted by FC$_\text{BNN}$ in Figure~\ref{fig:rnn_fdbk_full_diagram}), which estimates the input for the next timestep. The Bayesian layer estimates model and data uncertainty and has $2 \times (m+1)$ units. The first half of the units estimate the means of the query frequencies and ILI proportion; the second half of the units estimate the corresponding standard deviations. The estimated ILI proportion is a distribution which cannot be directly interpreted by a NN layer. Therefore, a sample from this distribution is combined with the true search query frequencies and fed back into the GRU layer. This is repeated from $t_0$ to $t_0+\delta$ (nowcasting phase). After time $t_0+\delta$, no more search query frequencies are available. The estimated search query frequencies and ILI proportions from each timestep are fed back into the model to make subsequent forecasts. The process of making daily estimates can be repeated indefinitely, so $\gamma$, the forecasting horizon, could increase arbitrarily. 

\subsubsection{Evaluation} We evaluate the performance for forecasting horizons $\gamma = 7$, $14$, $21$ and $28$ days ahead. We choose weekly test dates starting from week $45$ and lasting for $25$ weeks. We use the $2015/16$, $2016/17$, $2017/18$ and $2018/19$ flu seasons to evaluate our model. We did not consider running experiments on data from $2019/20$ or $2020/21$ as the ILI proportion has significantly declined, and ILI proportion estimates from the CDC became less reliable due to the COVID-19 pandemic. We train models for the period $05/06/2004$ until the Wednesday of the 33rd week of the year in which the test flu season starts (around mid-August). We test the models on the period from the Sunday of week 44 until the Saturday of week 23 in the following year. Exact training and test periods are provided in the Supporting Information (\ref{tab:sup_train_periods} and \ref{fig:sup_test_periods}). To compare our NN models to Dante, we evaluate the model scores on the same test weeks as specified in Reich et al. (2019)\cite{reich2019collaborative}. When comparing the best-performing NNs to Dante, the training set included all seasons except the test season i.e. it also included data after the test season (models NN and \gls{NNb} in Table~\ref{tab:dante_comparison}). We did not re-tune hyperparameters to account for training on future seasons. As discussed later, we do not consider training on data after the test period to be appropriate, but it allows the most direct comparison to the training setup used by Dante. We also report the performance of our best-performing NNs when trained using only data prior to the test season (model \gls{NNa} in Table~\ref{tab:dante_comparison}).

\section{Results}
\label{sec:nn_results}
We first provide a comparative performance analysis of the NN based models. Then, we compare it with the established state-of-the-art in ILI forecasting. Details about the models, training, and evaluation can be found in the Methods section.

\subsection{Forecasting performance of NNs} 
We investigate the performance of three Bayesian NN architectures, a feed-forward network (FF), a simple recurrent NN optimised for a single forecast horizon (SRNN), and an iterative RNN which feeds back daily forecasts to itself up to and including the horizon window (IRNN). We forecast the national-level weighted ILI proportion (wILI) in the US over four flu seasons, namely 2015/16 to 2018/19 from late October until June (exact dates are provided in \ref{tab:sup_train_periods} and corresponding ILI proportions are displayed in \ref{fig:sup_test_periods}). We evaluate our models for four forecast horizons $\gamma=$ 7, 14, 21, and 28 days ahead of the last available ILI proportion. The input to all NNs is both past ILI proportions and a time series of Web search query frequencies. In addition to that, for a more complete comparison, we also report performance results for the best-performing NN, IRNN, after excluding Web search activity data. We deploy six metrics to compare estimated forecasts to reported ILI proportions (ground truth). Mean absolute error (MAE) and bivariate correlation ($r$) compare forecasts without considering the associated uncertainty. Negative log likelihood (NLL), continuous ranked probability score (\gls{crps}), and Skill weight the error by its corresponding uncertainty. For NLL, CRPS, and MAE a lower score is better, while for $r$ and Skill higher scores are better. When average metrics are calculated across several seasons or forecast horizons, the arithmetic mean is used for all metrics besides Skill, for which the geometric mean is used~\cite{reich2019collaborative}.

\afterpage{
\begin{sidewaystable}
    \centering
    \footnotesize
    % \tiny
    \setlength{\tabcolsep}{0.5pt}
    \begin{tabular}{p{0.7cm}p{1.2cm}p{1.2cm}p{1.2cm}p{1.1cm}p{1.2cm}p{1.2cm}p{1.1cm}p{1.2cm}p{1.2cm}p{1.1cm}p{1.2cm}p{1.2cm}p{1.1cm}p{1.2cm}p{1.2cm}p{1.1cm}}

    & &\multicolumn{3}{c}{2015/16} & \multicolumn{3}{c}{2016/17} & \multicolumn{3}{c}{2017/18} & \multicolumn{3}{c}{2018/19} & \multicolumn{3}{c}{Avg(2015-19)} \\ 
   \cmidrule(lr){3-5}\cmidrule(lr){6-8}\cmidrule(lr){9-11}\cmidrule(lr){12-14}\cmidrule(lr){15-17}
   $\gamma$ & & IRNN & SRNN & FF & IRNN & SRNN & FF & IRNN & SRN & FF & IRNN & SRNN & FF & IRNN & SRNN & FF\\
   \midrule
   7    &    NLL       &            0.20 &       \textbf{-0.38} &           -0.19 &            0.38 &       \textbf{-0.29} &           -0.06 &               0.65 &       \textbf{0.07} &            0.52 &            0.32 &       \textbf{-0.49} &            0.12 &            0.39 &       \textbf{-0.27} &            0.10 \\
        &   CRPS       &            0.18 &        \textbf{0.09} &            0.10 &            0.24 &        \textbf{0.12} &            0.14 &               0.31 &       \textbf{0.17} &            0.33 &            0.21 &        \textbf{0.08} &            0.19 &            0.23 &        \textbf{0.12} &            0.19 \\
        &  Skill       &            0.73 &        \textbf{0.95} &            0.89 &            0.59 &        \textbf{0.85} &            0.78 &               0.49 &       \textbf{0.77} &            0.54 &            0.62 &        \textbf{0.95} &            0.70 &            0.60 &        \textbf{0.87} &            0.72 \\
        &    MAE       &            0.26 &        \textbf{0.13} &            0.13 &            0.33 &        \textbf{0.18} &            0.20 &               0.37 &       \textbf{0.25} &            0.49 &            0.28 &        \textbf{0.11} &            0.27 &            0.31 &        \textbf{0.17} &            0.27 \\
        &   $r$        &            0.85 &        \textbf{0.97} &            0.97 &            0.92 &        \textbf{0.99} &            0.99 &               0.98 &       \textbf{0.99} &            0.98 &            0.93 &        \textbf{0.99} &            0.97 &            0.92 &        \textbf{0.98} &            0.98 \\
   \midrule
   14   &    NLL       &       \textbf{0.18} &             0.35 &            0.49 &       \textbf{0.36} &             0.53 &            0.46 &          \textbf{0.64} &            3.28 &            0.94 &            0.30 &        \textbf{0.27} &             0.64 &       \textbf{0.37} &             1.11 &            0.63 \\
        &   CRPS       &       \textbf{0.18} &             0.21 &            0.20 &            0.24 &             0.27 &       \textbf{0.19} &          \textbf{0.31} &            0.72 &            0.39 &   \textbf{0.21} &                 0.21 &            0.27 &       \textbf{0.24} &             0.35 &            0.26 \\
        &  Skill       &       \textbf{0.73} &             0.69 &            0.61 &       \textbf{0.59} &             0.54 &            0.59 &          \textbf{0.49} &            0.11 &            0.39 &            0.63 &        \textbf{0.67} &            0.52 &       \textbf{0.60} &             0.40 &            0.52 \\
        &    MAE       &                0.26 &             0.29 &   \textbf{0.25} &            0.33 &             0.36 &       \textbf{0.22} &          \textbf{0.38} &            0.88 &            0.50 &   \textbf{0.28} &                 0.29 &            0.40 &       \textbf{0.31} &             0.45 &            0.34 \\
        &   $r$        &                0.85 &             0.85 &   \textbf{0.91} &            0.92 &             0.92 &       \textbf{0.97} &          \textbf{0.98} &            0.88 &            0.94 &            0.93 &        \textbf{0.94} &            0.91 &            0.92 &             0.90 &       \textbf{0.93} \\
   \midrule
   21   &    NLL       &       \textbf{0.32} &             0.85 &            0.85 &       \textbf{0.63} &             0.80 &            0.88 &          \textbf{0.78} &            2.08 &            1.50 &       \textbf{0.47} &             0.88 &            1.06 &       \textbf{0.55} &             1.15 &            1.07 \\
        &   CRPS       &       \textbf{0.23} &             0.33 &            0.28 &            0.31 &        \textbf{0.29} &            0.30 &          \textbf{0.42} &            0.70 &            0.61 &       \textbf{0.27} &             0.33 &            0.38 &       \textbf{0.30} &             0.41 &            0.39 \\
        &  Skill       &       \textbf{0.65} &             0.51 &            0.45 &       \textbf{0.49} &             0.46 &            0.42 &          \textbf{0.43} &            0.15 &            0.24 &       \textbf{0.56} &             0.45 &            0.37 &       \textbf{0.52} &             0.36 &            0.36 \\
        &    MAE       &       \textbf{0.32} &             0.45 &            0.36 &            0.42 &             0.39 &       \textbf{0.35} &          \textbf{0.56} &            0.90 &            0.79 &       \textbf{0.37} &             0.46 &            0.53 &       \textbf{0.42} &             0.55 &            0.51 \\
        &   $r$        &       \textbf{0.81} &             0.68 &            0.77 &            0.85 &             0.89 &       \textbf{0.93} &          \textbf{0.94} &            0.89 &            0.86 &       \textbf{0.88} &             0.85 &            0.84 &       \textbf{0.87} &             0.83 &            0.85 \\
   \midrule
   28   &    NLL       &       \textbf{0.53} &             1.07 &            1.07 &       \textbf{0.73} &             1.19 &            1.09 &          \textbf{1.32} &            3.82 &            1.80 &       \textbf{0.54} &             0.98 &            1.27 &       \textbf{0.78} &             1.76 &            1.31 \\
        &   CRPS       &       \textbf{0.31} &             0.40 &            0.35 &       \textbf{0.35} &             0.35 &            0.38 &          \textbf{0.57} &            0.88 &            0.80 &       \textbf{0.30} &             0.36 &            0.46 &       \textbf{0.38} &             0.50 &            0.50 \\
        &  Skill       &       \textbf{0.54} &             0.41 &            0.36 &       \textbf{0.45} &             0.34 &            0.35 &          \textbf{0.27} &            0.06 &            0.18 &       \textbf{0.53} &             0.40 &            0.30 &       \textbf{0.43} &             0.24 &            0.29 \\
        &    MAE       &            0.45 &             0.56 &       \textbf{0.40} &            0.49 &             0.44 &       \textbf{0.41} &          \textbf{0.73} &            1.05 &            1.07 &       \textbf{0.43} &             0.49 &            0.63 &       \textbf{0.53} &             0.64 &            0.63 \\
        &   $r$        &       \textbf{0.80} &             0.55 &            0.60 &            0.81 &             0.89 &       \textbf{0.92} &          \textbf{0.91} &            0.84 &            0.76 &            0.85 &        \textbf{0.86} &            0.77 &       \textbf{0.84} &             0.78 &            0.76 \\
   \bottomrule
   \end{tabular}
    \caption[{Regression performance metrics for three Bayesian NN models}]{\textbf{Regression performance metrics for three Bayesian NN models} \newline Performance metrics for shown for four forecast horizons ($\gamma =$ 7, 14, 21, and 28 days ahead). Negative log likelihood (NLL), continuous ranked probability score (CRPS) and Skill compare the accuracy weighted by the uncertainty of forecasts. MAE is the mean absolute error, and $r$ is the bivariate correlation between forecasts and reported ILI proportions. The best results for each metric and forecast horizon are shown in bold. The last three columns are performances averaged over the four test flu seasons (from 2015/16 to 2018/19).}
    \label{tab:nn_comparison}
\end{sidewaystable}
}

\begin{figure}[t]
\centering
\includegraphics[width=0.99\linewidth]{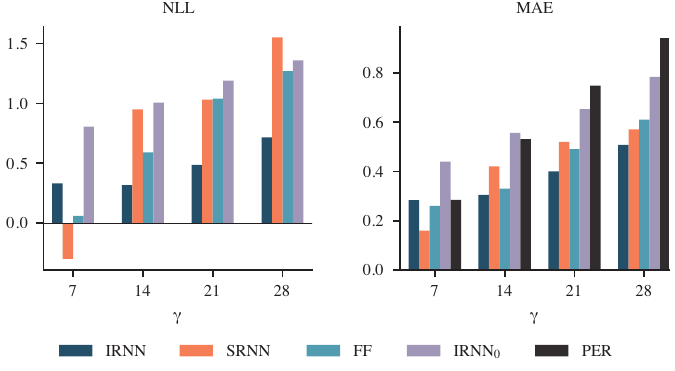}
\caption[{Negative log-likelihood (NLL) and mean absolute error (MAE) for each NN model averaged over all four test flu seasons (2015/16 to 2018/19)}] {\textbf{Negative log-likelihood (NLL) and mean absolute error (MAE) for each NN model averaged over all four test flu seasons (2015/16 to 2018/19)} \newline Scores for different forecast horizons ($\gamma$) are shown. Lower values are better. We also provide a comparison with IRNN trained without using any Web search activity data (\gls{IRNN0}), and a simple persistence model (PER). Note that NLL cannot be determined for PER as it does not provide an associated uncertainty. \ref{fig:sup_avg_metrics} shows the results for all metrics.}
\label{fig:avg_metrics}
\end{figure}

\begin{figure*}[!ht]
\centering
\includegraphics[width=0.99\linewidth]{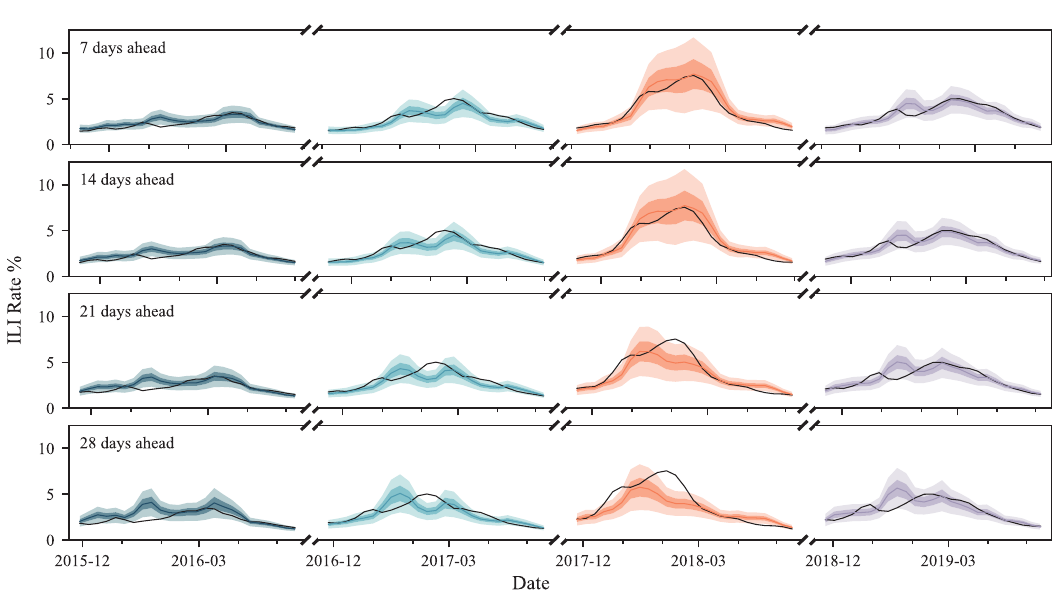}
\caption[{IRNN Forecasts}]{\textbf{IRNN Forecasts} \newline IRNN forecasts for all four test seasons (2015/16 to 2018/19) and forecasting horizons ($\gamma =$ 7, 14, 21, and 28). Confidence intervals (uncertainty estimates) are shown at $50\%$ and $90\%$ levels and are visually distinguished by darker and lighter colour overlays respectively. The influenza-like illness (ILI) proportion (ground truth) is shown by the black line.}
\label{fig:rnn_fdbk_full_forecasts}
\end{figure*}

Table~\ref{tab:nn_comparison} enumerates the performance metrics for the three NNs in each flu season and forecast horizon. The IRNN performs best for all forecast horizons, except for $\gamma = 7$ days ahead where SRNN is the best-performing model. As we detail in Methods, this is not unexpected given the model design. IRNN, contrary to SRNN and FF, does not use future query frequencies (from the seven days following the target forecast date) for the hindcasting task ($\gamma = 7$). Interestingly, we also observe that the performance of IRNN does not change for $\gamma = 7$ and $14$, something that can probably be explained by a model behaviour that gives significantly more importance to the more recent inputs (search query frequencies are ahead of the past ILI proportions by $\delta = 14$ days). IRNN, the most advanced NN that we propose, compared to the next best NN architecture reduces error by 14.87\% in terms of MAE, 20\% in terms of CRPS, and improves Skill by 32.48\%, when averaged across all test seasons and forecasting horizons $\gamma =$ 14, 21, and 28 days. IRNN yields further improvements in the rest of the metrics, although these have a more limited interpretability. The fact that IRNN improves more between MAE and CRPS (by 4.15 percentage points) means that it is also a better model for the uncertainty bounds compared to FF and~SRNN. 

Figure~\ref{fig:avg_metrics} provides an alternative visual of the forecasting performance metrics of the different NN models when averaged over the four flu seasons  (NLL and MAE are depicted, the rest of the metrics are displayed in~\ref{fig:sup_avg_metrics}). In addition to the three NNs, we also provide performance metrics for an IRNN variant that does not use any search query frequency data (denoted by IRNN$_0$), along with a simple persistence model (denoted by PER; see S1 Appendix for a definition). IRNN consistently performs better than IRNN$_0$, which confirms our hypothesis that Web search activity information provides a significant performance improvement. On the other hand, IRNN$_0$ displays competitive performance when compared to SRNN or FF which highlights that IRNN is a more suitable model for handling search query frequency time series. In the Supporting Results, we have also provided an additional baseline comparison with an Elasticnet~\cite{zou2005regularization} model that, in line with our previous work~\cite{Lampos2015GFT}, provides inferior performance (\ref{tab:sup_dante_comp_full} and \ref{fig:sup_elasticnet_forecasts}). A fair comparison with Gaussian Processes models~\cite{Rasmussen2006}, which we have also deployed in the past~\cite{Lampos2017WWW, Zou2018www}, was not practically tractable given the high dimensionality of the task and the relatively large amount of training samples. Finally, the persistence model baseline is always inferior to at least one of the NN models.

\begin{figure*}[!ht]
    \centering
    \includegraphics[width=0.98\linewidth]{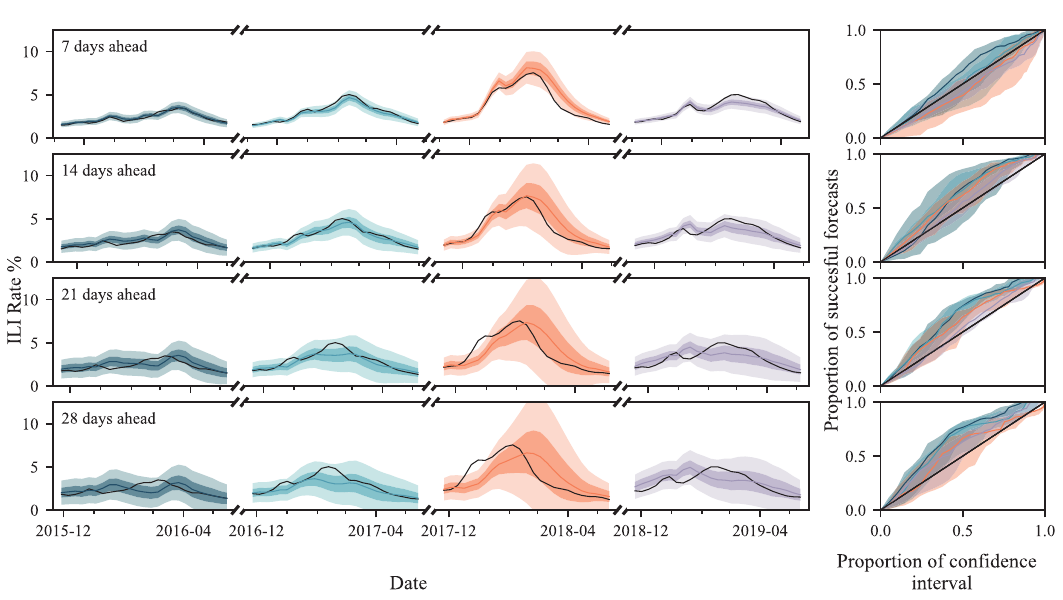}
    \caption[{FF Forecasts}]{\textbf{FF Forecasts} \newline
    FF forecasts for all four test seasons (2015/16 to 2018/19) and forecasting horizons ($\gamma =$ 7, 14, 21, and 28). Confidence intervals (uncertainty estimates) are shown at $50\%$ and $90\%$ levels and are visually distinguished by darker and lighter colour overlays respectively. The influenza-like illness (ILI) proportion (ground truth) is shown by the black line. The flu seasons are shown in different colours, corresponding with the calibration plots on the right. The calibration lines show how frequently the ground truth falls within a confidence interval (\gls{ci}) of the same level. To be more precise, a point $(x,y)$ denotes that the proportion $y \in [0, 1]$ of the forecasts when combined with a CI at the $x \times 100\%$ level includes the ground truth (successful forecasts). The optimal calibration is shown by the diagonal black line. Points above or below the diagonal indicate an over- or under-estimation of uncertainty, and hence an under- or over-confident model, respectively. The shadows show the upper and lower quartile of the calibration curves when the models are trained multiple times with different initialisation seeds.}
    \label{fig:sup_ff_forecasts}
\end{figure*}

\begin{figure*}[!ht]
    \centering
    \includegraphics[width=0.98\linewidth]{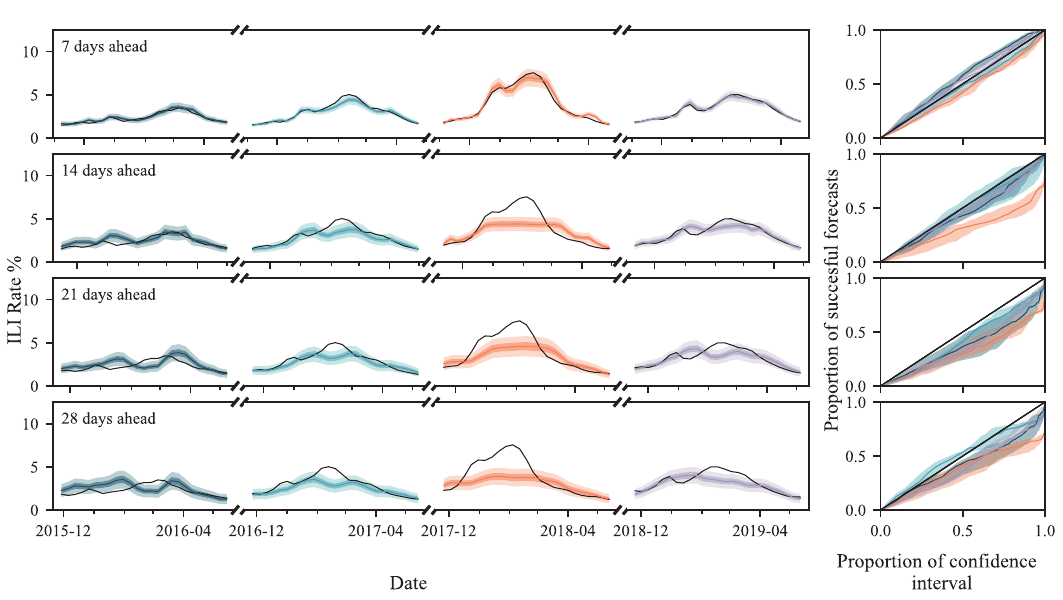}
    \caption[{SRNN Forecasts}]{\textbf{SRNN Forecasts} \newline
    SRNN forecasts for all four test seasons (2015/16 to 2018/19) and forecasting horizons ($\gamma =$ 7, 14, 21, and 28). Confidence intervals (uncertainty estimates) are shown at $50\%$ and $90\%$ levels and are visually distinguished by darker and lighter colour overlays respectively. The influenza-like illness (ILI) proportion (ground truth) is shown by the black line. The flu seasons are shown in different colours, corresponding with the calibration plots on the right. The calibration lines show how frequently the ground truth falls within a confidence interval (CI) of the same level. To be more precise, a point $(x,y)$ denotes that the proportion $y \in [0, 1]$ of the forecasts when combined with a CI at the $x \times 100\%$ level includes the ground truth (successful forecasts). The optimal calibration is shown by the diagonal black line. Points above or below the diagonal indicate an over- or under-estimation of uncertainty, and hence an under- or over-confident model, respectively. The shadows show the upper and lower quartile of the calibration curves when the models are trained multiple times with different initialisation seeds.}
    \label{fig:sup_SRNN_forecasts}
\end{figure*}

Forecasts from IRNN in every season and forecast horizon are shown in Figure~\ref{fig:rnn_fdbk_full_forecasts}, whereas forecasts from the FF and SRNN architectures are shown in the Supporting Information (\ref{fig:sup_ff_forecasts} and \ref{fig:sup_SRNN_forecasts}, respectively). The expected decline in accuracy as the forecast horizon increases is visually evident for all models. Interestingly, forecasts from the FF NN closely follow the estimates of a persistence model (i.e. shifted ground truth) and also have quite pronounced uncertainty bounds for $\gamma = $ 21 and 28. SRNN provides smoother but generally flatter forecasts that, in principle, may capture the underlying ILI trend. However, they quite often underestimate the exact ILI proportion and are over-confident (visualised by tight uncertainty bounds). The IRNN makes more independent forecasts that do not necessarily follow previous trends in recently observed ILI proportions. Uncertainty bounds increase slightly with $\gamma$, albeit we note that this model does not directly differentiate between forecasting horizons. Overall, forecasts from IRNN have a better correspondence to the ILI proportion range and provide an early flu onset warning (in at least three of the four test seasons).

\begin{figure*}[t]
\centering
\includegraphics[width=0.99\linewidth]{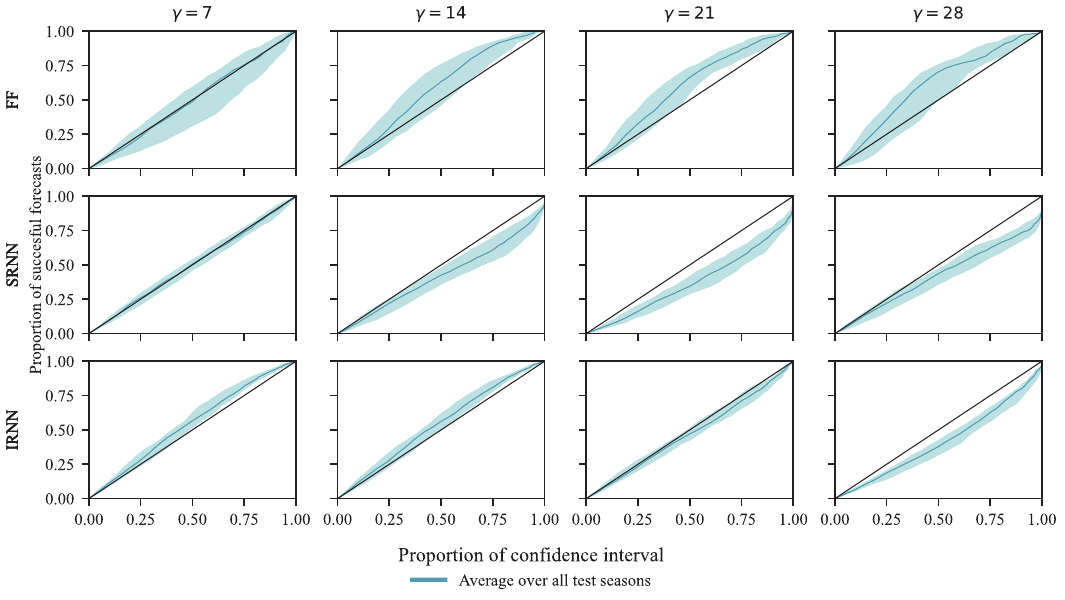}
\caption[{Averaged calibration plots for all test periods}]{\textbf{Averaged calibration plots for all test periods} \newline
Calibration plots for the forecasts made by the three NN models (FF, SRNN, and IRNN) averaged over the four test periods (2015/16 to 2018/19) and shown for the four forecasting horizons ($\gamma$). The lines show how frequently the ground truth falls within a confidence interval (CI) of the same level. To be more precise, a point $(x,y)$ denotes that the proportion $y \in [0, 1]$ of the forecasts when combined with a CI at the $x \times 100\%$ level includes the ground truth (successful forecasts). The optimal calibration is shown by the diagonal black line. Points above or below the diagonal indicate an over- or under-estimation of uncertainty, and hence an under- or over-confident model, respectively. The shadows show the upper and lower quartile of the calibration curves when the models are trained multiple times with different initialisation seeds. The plot broken out into separate test periods is shown in the Supporting Information (\ref{fig:sup_calibration_plot}).}
\label{fig:calibration_plot}
\end{figure*}

Figure~\ref{fig:calibration_plot} shows the calibration of the confidence intervals (CI) for each of the NNs. The $x$-axis represents the expected frequency that the ground truth data will be present in a specified region of confidence, while the $y$-axis represents the empirical frequency as measured from the test results. Remember that each forecast has an associated uncertainty represented by a Gaussian distribution. For a specified probability, $p_\rho$, we can determine the confidence region around each forecast such that we expect the ground truth to fall within these regions with probability $p_\rho$. $p_\rho$ can be computed by $ p_\rho = \texttt{cdf}(n) - \texttt{cdf}(-n)$, where $n$ is the number of standard deviations away from the mean, and $\texttt{cdf}$ denotes the cumulative distribution function. For a given probability (on the $x$-axis), we compute the empirical probability for each of the four test seasons. The diagonal line ($y=x$) represents perfect calibration i.e. the expected and empirical probabilities are the same. Points above the diagonal indicate that the uncertainty estimates are too large. Conversely, the points below indicate that the uncertainty estimates are too low. The shadow around the calibration curve shows the variation due to different initialisation seeds over $10$ NN training runs (see Methods for further details). Uncertainties produced by the IRNN are closer to the diagonal (i.e. better estimates of uncertainty) for horizon windows greater than seven. Overall, we see that FF is an under-confident model, SRNN is an over-confident model, and IRNN is generally more balanced, but the error in confidence increases for the largest forecast horizon ($\gamma = 28$).

\begin{table*}[t]
    \centering
    \small
    \setlength{\tabcolsep}{2pt}

    \begin{tabular}{p{1.1cm} p{1.1cm} p{0.7cm} p{0.7cm} p{0.7cm} p{0.7cm} p{0.7cm} p{0.7cm} p{0.7cm} p{0.7cm} p{0.7cm} p{0.7cm} | p{0.87cm} p{0.7cm}}
        Horizon & Metric  &\multicolumn{2}{c}{2015/16} & \multicolumn{2}{c}{2016/17} & \multicolumn{2}{c}{2017/18} & \multicolumn{2}{c}{2018/19} & \multicolumn{4}{c}{Avg (2015-19)} \\ 
        \cmidrule(lr){3-4}\cmidrule(lr){5-6}\cmidrule(lr){7-8}\cmidrule(lr){9-10}\cmidrule(lr){11-14}
        $\gamma$  & & Dte & NN & Dte & NN & Dte & NN & Dte & NN & Dte & NN & NN$_a$ & NN$_b$\\
        \midrule
        7   & Skill         & 0.67      & \textbf{0.75} & \textbf{0.63} & 0.53      & 0.45      & \textbf{0.53} & \textbf{0.62} & 0.61      & 0.59      &\textbf{0.60}  & 0.85  & 0.88\\
            & MAE           & \textbf{0.22} & 0.26      & \textbf{0.19} & 0.35      & 0.39      & \textbf{0.38} & \textbf{0.21} & 0.28      & \textbf{0.25} & 0.32      & 0.18  & 0.17\\
            & $r$           & \textbf{0.88} & 0.81      & \textbf{0.96} & 0.91      & 0.97      & \textbf{0.98} & \textbf{0.97} & 0.90      & \textbf{0.94} & 0.90      & 0.98  & 0.98\\
        \midrule
        14  & Skill         & 0.54      & \textbf{0.74} & \textbf{0.54} & 0.53      & 0.29      & \textbf{0.53} & 0.52      & \textbf{0.61} & 0.46      & \textbf{0.59} & 0.55  & 0.59\\
            & MAE           & 0.38      & \textbf{0.28} & \textbf{0.32} & 0.35      & 0.64      & \textbf{0.39} & 0.33      & \textbf{0.28} & 0.42      & \textbf{0.33} & 0.35  & 0.34\\
            & $r$           & 0.64      & \textbf{0.79} & 0.91      & \textbf{0.91} & 0.90      & \textbf{0.98} & \textbf{0.92} & 0.90      & 0.84      & \textbf{0.89} & 0.89  & 0.89\\
        \midrule
        21  & Skill         & 0.44      & \textbf{0.64} & \textbf{0.48} & 0.43      & 0.21      & \textbf{0.30} & 0.46      & \textbf{0.52} & 0.38      & \textbf{0.45} & 0.47  & 0.48\\
            & MAE           & 0.48      & \textbf{0.37} & \textbf{0.38} & 0.45      & 0.86      & \textbf{0.62} & \textbf{0.40} & 0.44      & 0.53      & \textbf{0.47} & 0.48  & 0.46\\
            & $r$           & 0.36      & \textbf{0.67} & \textbf{0.87} & 0.83      & 0.82      & \textbf{0.94} & \textbf{0.89} & 0.82      & 0.73      & \textbf{0.81} & 0.81  & 0.81\\
        \midrule
        28  & Skill         & 0.37      & \textbf{0.53} & \textbf{0.46} & 0.38      & \textbf{0.17} & 0.14      & 0.42      & \textbf{0.45} & 0.33      & \textbf{0.33} & 0.37  & 0.40\\
            & MAE           & 0.54      & \textbf{0.47} & \textbf{0.39} & 0.50      & 1.06      & \textbf{0.85} & \textbf{0.45} & 0.58      & 0.61      & \textbf{0.60} & 0.61  & 0.58\\
            & $r$           & 0.23      & \textbf{0.63} & \textbf{0.88} & 0.79      & 0.76      & \textbf{0.92} & \textbf{0.86} & 0.79      & 0.68      & \textbf{0.78} & 0.78  & 0.79\\
        \bottomrule
        \end{tabular}

    \caption[{Performance metrics for best NN compared with Dante}]{\textbf{Performance metrics for best NN compared with Dante} \newline
    Forecasting performance metrics for the best-performing neural network (SRNN for $\gamma=7$, IRNN for $\gamma \ge 14$) compared with Dante. The NNs are trained using search query frequencies generated only up to the last available  (Dte)ILI proportion (the 2-week advantage of using Web search data is removed). We use leave-one flu season-out to train models, similarly to Dante. The best results for this comparison are shown in bold. The very last column (NN$_b$) presents the average performance results of NNs where the temporal advantage of Web search activity information is maintained (see also \ref{fig:sup_IRNN_LOE_forecasts} that depicts IRNN's forecasts when leave-one flu season-out is applied). The penultimate column (NN$_a$) holds results for the same experiment as NN$_b$ with the addition of disabling leave-one flu season-out training.}
    \label{tab:dante_comparison}
\end{table*}

\subsection{Comparison with state-of-the-art} 
We compare our best model for each forecasting horizon i.e. SRNN for $\gamma=7$ and IRNN for $\gamma \ge 14$, to a state-of-the-art ILI proportion forecasting model, known as `Dante'\cite{osthus2021multiscale}. In its original implementation, Dante produces a binned forecast and does not permit comparison based on CRPS or NLL (see S1 Appendix). Therefore, for this analysis, we restrict the performance metrics to Skill, MAE, bivariate, and correlation.

To be consistent with prior published literature and conduct a fair comparison, we adopt exactly the same training setup as proposed in the original paper that proposed Dante~\cite{osthus2021multiscale}. However, we would like to make the reader aware of various caveats in this comparison. First, Dante's national US ILI proportion forecasts are based on ILI proportions from 63 subnational US geographical regions (50 US states, 10 Health and Human Services regions, the District of Columbia, Puerto Rico, and Guam) as well as ILI proportions at the national level. The NNs use only national US ILI proportions, augmented with a US national aggregate of Web search activity data. The latter is more recent i.e. search query frequencies are available until $t_0 + \delta$ which is after the last observed ILI proportion ($t_0$). To remove this temporal advantage, we do not use Web search activity data generated after $t_0$ when training models for comparison with Dante. Secondly, Dante is trained using a leave-one flu season-out methodology, training on all other flu seasons (past and future) but the test one. Thus, for example, for the test season 2016/17, Dante will use historical data prior to 2016 and after 2016/17. We do not consider this appropriate as, in practice, a deployed system has no knowledge of future seasons. However, for comparison purposes, we train our models using leave-one flu season-out as well. We note that we were not able to successfully train Dante when restricting training data to exclude future seasons; Dante's performance was too poor to be considered for comparison. We emphasise that training on dates after the test season is only done when compared to Dante. Another caveat is that Dante exploits regional ILI data to produce a national forecast -- this can sometimes provide an earlier warning as outbreaks will first be recorded sub-nationally. Our models are not built this way, and cannot leverage this information. The final remark is that Dante performs retraining prior to conducting a forecast. Although that is possible for the NN models as well, running complete experiments (across many seasons, different NN architectures, and different initialisation seeds) with retraining every time prior to making a forecast would have taken a considerable amount of time. Hence, NNs make forecasts for an entire flu season without retraining.

\begin{figure*}[!ht]
    \centering
    \includegraphics[width=0.98\linewidth]{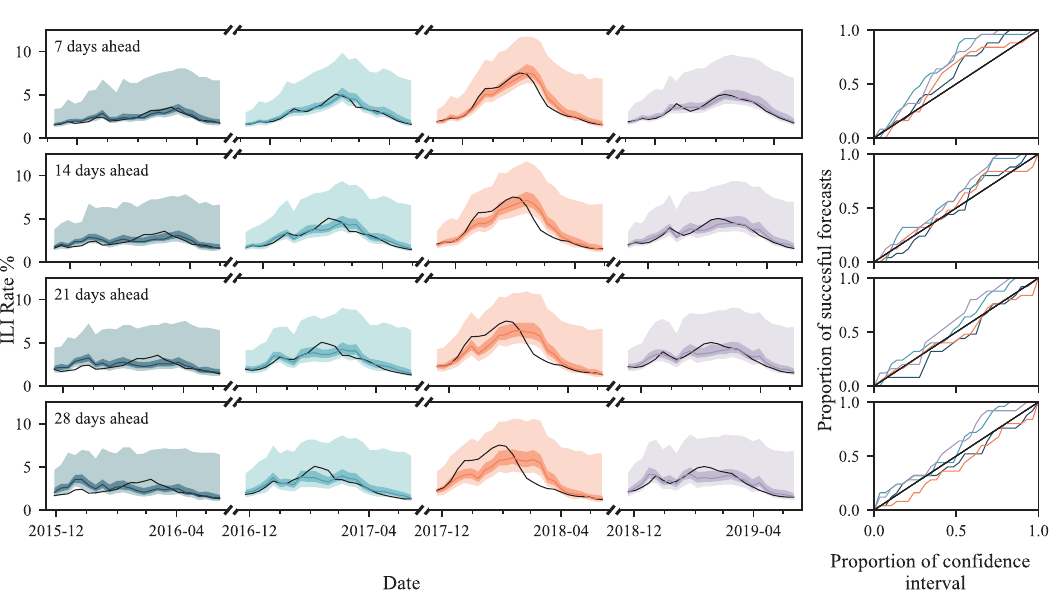}
    \caption[{Dante Forecasts}]{\textbf{Dante Forecasts} \newline
    Dante forecasts for all four test seasons (2015/16 to 2018/19) and forecasting horizons ($\gamma =$ 7, 14, 21, and 28). Confidence intervals (uncertainty estimates) are shown at $50\%$ and $90\%$ levels and are visually distinguished by darker and lighter colour overlays, respectively. The influenza-like illness (ILI) proportion (ground truth) is shown by the black line. The flu seasons are shown in different colours which correspond to the calibration plots on the right. The calibration lines show how frequently the ground truth falls within a confidence interval (CI) of the same level. To be more precise, a point $(x,y)$ denotes that the proportion $y \in [0, 1]$ of the forecasts when combined with a CI at the $x \times 100\%$ level includes the ground truth (successful forecasts). The optimal calibration is shown by the diagonal black line. Points above or below the diagonal indicate an over- or under-estimation of uncertainty, and hence an under- or over-confident model, respectively.}
    \label{fig:sup_dante_forecasts}
\end{figure*}

Table~\ref{tab:dante_comparison} shows the metrics for the best NN for each forecast horizon $\gamma$, trained with leave-one flu season-out and with search data from $t \le t_0$, and results for Dante taken on identical forecast dates. When averaged over all forecasting tasks, the NNs have $11.93\%$ higher Skill, $4.97\%$ lower MAE, and $5.96\%$ higher correlation than Dante. Dante has a better-calibrated uncertainty compared to IRNN, but this can be interpreted by its significantly larger uncertainty estimates that sometimes are over 2 times greater than the ones produced by IRNN (\ref{fig:sup_dante_forecasts}). In general, a better-calibrated uncertainty is less important when forecast error metrics indicate overall inferior performance. The last column (NN$_b$) of Table~\ref{tab:dante_comparison} provides an expanded comparison (full results are shown in \ref{tab:sup_dante_comp_full}) whereby we have enabled training with Web search activity data that maintain their actual latency ($t_0 + \delta$). As expected, the performance benefits increase, obtaining $33.52\%$ higher Skill, $14.37\%$ lower MAE, and $8.78\%$ higher correlation compared to Dante. Disabling leave-one flu season-out training on just our models also results in a better performance compared to Dante (which maintains its knowledge of future flu seasons) (see column NN$_a$ of Table~\ref{tab:dante_comparison}).

\section{Discussion}
\label{sec:nn_discussion}
We have demonstrated the ability of neural networks to forecast ILI proportions by incorporating exogenous Web search activity data while providing uncertainty estimates. IRNN exhibits superior performance (averaged over all test years) for forecast horizons greater than seven days, whereas SRNN is superior for the $\gamma = 7$ days ahead forecast horizon, a prediction task also referred to as hindcasting. As discussed extensively (see Methods and Results), this is expected because when $\gamma = 7$ days, SRNN is using all the available Web search activity data, which extends seven days beyond the target forecasting horizon. We have also demonstrated that the proposed forecasting framework can provide very competitive performance that is better than the established state-of-the-art in ILI proportion forecasting.

Our experiments highlight the importance of including Web search activity for forecasting ILI proportions with or without their expected temporal advantage. This is consistent with previous literature whereby the added value of online user-generated data streams (e.g., Web search, but also social media) has been evaluated~\cite{nsoesie2013forecasting, paul2014twitter, Lampos2015GFT}. However, our experiments present the most comprehensive analysis to date, assessing performance over four consecutive flu seasons, and utilising an open-ended, non-manually curated set of search queries. In addition, we have cross-examined accuracy with several different error metrics, including CRPS and NLL, that can incorporate the validity of uncertainty estimates. We have seen that adding Web search information not only improves accuracy but also provides better estimates of confidence (Figure S1).

By examining ILI seasons in our training and test sets, we can deduce that the 2015/16 test season is the least similar season to previously seen ones (mean bivariate correlation of 0.74), whereas the 2018/19 is the most similar (mean bivariate correlation of 0.81). With that in mind, we observe that in comparison to Dante the NNs that utilise Web search activity perform better when the flu season has a more novel trajectory (Table~\ref{tab:dante_comparison}). As Dante is utilising ILI proportions only (including subnational ones), it is expected to be a more focused model on previously seen ILI proportion trajectories. In contrast, the search query frequency time series provides an opportunity to capture more complex underlying patterns, and hence seem to be a more informative source during novel flu seasons.

From an epidemiological perspective, accurate forecast estimates might not always be the sole determinant of model superiority. Although our model performance analysis is comprehensive, and contrary to most of the related literature, providing a clean depiction of seasonal forecasts, it does focus on the accuracy of a forecast and its associated uncertainty. Table~\ref{tab:comp_dante_discuss} attempts to partially address that by offering a few additional comparative insights following aspects of a similar analysis for ILI proportion nowcasting models in England~\cite{Wagner2018}. Focusing on the most challenging forecasting horizons ($\gamma = 21$ and $28$ days), we compute the delay in forecasting the peak of the flu season as well as the difference in magnitude between the predicted and the estimated peak ILI proportion. We see that Dante is making either very invalid early estimates (e.g., 70 days before the actual peak) or otherwise lags by 1 or 2 weeks (i.e. no early warning), whereas the NN models tend to always provide reasonable early warnings of the peak. While there is no definitive winner in estimating the ILI proportion peak magnitude, by examining forecasts when the ILI proportion was relatively high (above the seasonal mean plus one standard deviation), we observed that Dante's estimates were significantly worse in terms of MAE and relative MAE (symmetric mean absolute percentage of error). A similar analysis across NN variants is provided in \ref{tab:sup_comp_nn_discuss} highlighting the expected superiority of~IRNN.

Existing disease forecasting frameworks are difficult to scale, and incorporating additional features or more training data can result in excessive computational costs. This results in a trade-off between model flexibility and the number of exogenous variables a model can handle effectively~\cite{osthus2022fast, brooks2018nonmechanistic, ray2017infectious}. An advantage of neural networks is that they are easy to scale; increasing the amount of training instances often results in better overall performance~\cite{alom2019state}. Overfitting issues, which become more apparent when working with relatively small data sets, are alleviated to an extent by the deployment of a Bayesian layer which averages over parameter values instead of making single point estimates~\cite{hernandez2015probabilistic}. A lingering disadvantage, however, is that there is no current consensus on estimating uncertainty with NNs in a principled manner. Our methodological approach, presented in the following section, has attempted to address that by considering two modes of uncertainty (epistemic and aleatoric). In addition, given the relatively restricted amount of samples of training neural networks, our experimental approach provides novel insights for model derivation, training, and hyperparameter validation for similar time series forecasting tasks.

\begin{table*}[t]
    \centering
    \small
    \setlength{\tabcolsep}{6pt}
    \begin{tabular}{l ccc}
    \toprule
    Horizon
            & \multicolumn{3}{c}{$\gamma = 21$} \\
    \toprule
    Metric
            & Dante     & NN        & NN$_b$ \\
    \midrule
    $\delta$-p (days)          &
            \textbf{-70}, 14, 14, 14 & -49, 14, -14, -35     & -49, 14, -28, -35  \\
    Avg. $\delta$-$y_{\text{p}}$ &
            0.99        & 0.70      & 0.59      \\
    MAE-p               &
            0.84        & 0.75      & 0.76      \\
    SMAPE-p (\%)        &
            20.19       & 15.57     & 15.69     \\
    \bottomrule
    \\
    \\
    \toprule
    Horizon & \multicolumn{3}{c}{$\gamma = 28$} \\
    \toprule
    Metric
            & Dante     & NN        & NN$_b$ \\
    \midrule
    $\delta$-p (days)          &
            \textbf{-70}, 14, 7, 14  & -42, -21, -21, -28    & -42, -21, -21, -28 \\
    Avg. $\delta$-$y_{\text{p}}$ &
            0.67        & 1.01      & 0.87      \\
    MAE-p               &
            1.09        & 0.89      & 0.88      \\
    SMAPE-p (\%)        &
            \textbf{26.24}       & 17.72     & 17.57     \\
    \bottomrule
    \end{tabular}

    \caption[{Meta-analysis of ILI forecasts}]{\textbf{Meta-analysis of ILI forecasts} \newline Meta-analysis of ILI proportion forecasts around the peak of a flu season for Dante, NN (the best NN variant when the temporal advantage of Web search activity data is removed), and NN$_b$ (same as NN but after reinstating the temporal advantage of Web search activity data). $\delta$-p denotes the temporal difference (in days) in forecasting the peak of the flu seasons 2015/16, 2016/17, 2017/18, and 2018/19, respectively. Negative/positive values indicate an earlier / later forecast; averaging $\delta$-p across the four test flu seasons would remove this information and that is why we enumerate all four values. Avg. $\delta$-$y_{\text{p}}$ measures the average magnitude difference in the estimate of the peak of the flu season between a forecasting model and the CDC. MAE-p is the MAE when the ILI proportion is above the seasonal mean plus one standard deviation. SMAPE-p (\%) is the symmetric mean absolute percentage of error for the same time periods. Outcomes that yield an unfavourable interpretation for the underlying forecasting model are provided in bold. Detailed outcomes for all NNs are shown in \ref{tab:sup_comp_nn_discuss}.}
    \label{tab:comp_dante_discuss}
\end{table*}
% correlation of ILI proportions in the 4 test seasons including future ones
% -- 15/16: 0.74 (0.25) <== most unique to others!
% -- 16/17: 0.80 (0.27) <== easiest
% -- 17/18: 0.78 (0.24)
% -- 18/19: 0.81 (0.26) <== easiest

It is equally important to acknowledge the limitations of our methodological approach, and more broadly, of this research task as a whole. We note that the retrospective analysis provided in this thesis cannot be the only determinant for model deployment within established syndromic surveillance systems. This would also require real-time assessments during ongoing influenza seasons in collaboration with public health organisations. Furthermore, an ILI consultation proportion is not always representative of the true influenza proportion in a population. It is a proxy indicator, and as such it might be biased~\cite{yang2015inference, baltrusaitis2018evaluation}. Therefore, any model that is trained and evaluated based on these rates is inherently limited by this property. An additional factor that could arguably yield misleading inferences is the co-existence of COVID-19 and influenza, given their similar symptom profiles. Although this is outside the remit of this thesis, early results from our ILI models for England during the 2022/23 flu season have showcased that ILI proportions can be accurately estimated during COVID-19 outbreaks~\cite{UKHSA2023}. From a methodological perspective, we note that our approach to estimating uncertainty can be improved --- IRNN, the best-performing NN, is currently not explicitly aware of the actual forecasting horizon ($\gamma$) when conducting a prediction (see Methods). Addressing this appropriately will most likely result in better-calibrated uncertainty estimates. From an empirical evaluation perspective, our experiments have been conducted on the US at a national level. Hence, although we expect that these results will generalise sub- and internationally, we have no evidence of this, apart from the fact that past research on similar types of models has shown promise in various different US subregions or countries~\cite{Lampos2017WWW, clemente2019improved, Zou2018www, Zou2019, ning2019accurate}. Finally, the application presented in this thesis relies on the existence of Web search activity data. Access to this data is not assured as it both depends on sufficient Internet usage rates and on the willingness of private corporations to provide this information for research and epidemiological modelling. Nonetheless, the presented forecasting models do provide a general machine learning approach applicable to different input (e.g., social media activity, body sensors) and output streams of information (e.g., different disease indicators).

\section{IRNN Uncertainty Propagation Analysis and Refinement}
\label{sec:IRNNs}
The Iterative Recurrent Neural Network (IRNN) developed in Sections~\ref{sec:nn_methods} to \ref{sec:nn_discussion} produces good quality forecasts, outperforming Dante in terms of Skill and MAE. However, during testing, it was noted that the forecast uncertainty behaves unexpectedly and does not significantly increase with the forecast horizon. In this section, we scrutinise the forecasting process and propose modifications to improve the uncertainty estimation for longer forecast horizons. 

The IRNN estimates data uncertainty by outputting the mean and standard deviation of a Normal distribution i.e., $\hat{\mu}$ and $\hat{\sigma}$. Model uncertainty is estimated by specifying a distribution $Q(\bm{\Phi})$ over the weights ($\bm{\Phi}$) in the dense layer. Monte-Carlo sampling of the posterior distribution is used to approximate the combined uncertainty using Eq~\ref{eq:combine_uncertainties_mean} and \ref{eq:combine_uncertainties} for the mean and variance, respectively. 

Ignoring the ability of the IRNN to observe Web search data produced after the ILI proportions, the process of producing a forecast for each of the $K$ Monte-Carlo samples, for $\gamma$ days ahead is as follows.
\begin{enumerate}
    \item Initialise RNN layer hidden states. 
    \item Sequentially feed inputs from $t_0-\tau$ to $t_0$ into the RNN layer to set the RNN hidden state.
    \item Feed the output from the RNN layer into the Bayesian dense layer. Sample from the weight distribution to estimate the mean and standard deviation of the inputs for the subsequent timestep.
    \item Feed the mean of the estimated inputs back into the RNN layer to update the hidden state and produce the next prediction.
    \item Steps $3$ and $4$ are repeated up until $t_0 + \gamma$.
\end{enumerate}
This process is repeated $K$ times, each using different samples from the weight distributions to produce a range of predictions that are combined to estimate model uncertainty. The prediction means are fed back into the RNN since RNNs are ill-equipped to interpret distributions. We empirically found that the mean produced better results than sampling from the output distribution, in between timesteps. Next, we provide a simplified example based on the IRNN, evaluating the sampling of the model to acquire the best uncertainty estimates.  

\subsection{IRNN Uncertainty Propagation Example}
The IRNN iteratively estimates values for subsequent timesteps based on its own estimates; this can be summarised as $\hat{x}_{t+1} = f^{p(\bm{\Phi})}(x_{t})$ where $p(\bm{\Phi})$ is a distribution over the parameters. For simplicity let $f^{p(\bm{\Phi})}(x) = x+p(\bm{\Phi}_a)+\mathcal{N}(0, p(\bm{\Phi}_\sigma))$ where $p(\bm{\Phi}_a)$ is the change each timestep and $p(\bm{\Phi}_\sigma)$ is the data uncertainty. The model iteratively produces estimates of $\mathbf{x}$ up to $x_{t_0+\gamma}$: 
\begin{align}
    \hat{x}_{t_0+1} &= x_{t_0} + p(\bm{\Phi}_a) + \mathcal{N}(0, p(\bm{\Phi}_\sigma)) \\
    \hat{x}_{t_0+2} &= \hat{x}_{{t_0}+1} + p(\bm{\Phi}_a) + \mathcal{N}(0, p(\bm{\Phi}_\sigma)) \\
                  &~~\vdots \\
    \hat{x}_{t_0+\gamma} &= \hat{x}_{t_0+\gamma-1} + p(\bm{\Phi}_a)+ \mathcal{N}(0, p(\bm{\Phi}_\sigma));
\end{align}

\subsubsection{Model Uncertainty Only\\}
To isolate the model uncertainty, we remove the data uncertainty $\mathcal{N}(0, p(\bm{\Phi}_\sigma)$ from the example model. This also allows the assessment of how the different sampling regimes affect the uncertainty propagation. The parameters can be sampled in three ways: using the mean of $p(\bm{\Phi})$, sampling $p(\bm{\Phi})$ once for every new prediction (i.e. every timestep), or sampling from $p(\bm{\Phi})$ only at the start before making any predictions. We discuss these in turn. 

Let $p(\bm{\Phi}_a) = \mathcal{N}(1.0, 0.1)$ be a Normal distribution with a mean of $1.0$ and a standard deviation of $0.1$. By taking the mean and iteratively making predictions with the model, then if ${x}_{t_0} = 0$, $[\hat{x}_{t_0+1}, \hat{x}_{t_0+2}, ...,\hat{x}_{t_0+\gamma}] = [1,2, ...,\gamma]$, there is no uncertainty in the predictions. 

To evaluate resampling once for every timestep, the model is run for $\gamma=100$, with $1000$ samples from $p(\bm{\Phi}_a)$ each timestep. Uncertainty is computed as the standard deviation of the predictions at each timestep. The uncertainty increases according to $\hat{\sigma} = \sqrt{\gamma a_\sigma}$, where $a_\sigma=0.1$ is the standard deviation of $p(\bm{\Phi}_a)$. Thus, the uncertainty increases with the square root of the forecast horizon. 

Finally, $p(\bm{\Phi}_a)$ is sampled $1000$ times before making any predictions and is not resampled between timesteps. Here the uncertainty increases linearly with the forecast horizon according to $\hat{\sigma} = \gamma a_\sigma$. Sampling once is akin to instantiating an ensemble of models and running each individually for all forecast horizons. The individual trajectories are smooth and diverge from one another. Contrastingly, sampling every timestep yields noisy trajectories. There is no ground truth for model uncertainty,  however, sampling only at $t_0$ produces models that are instantiated just once rather than at every timestep and produce intuitively more reasonable results: model uncertainty diverges uniformly with time.

\subsubsection{Data Uncertainty\\}
Data uncertainty is independent of the modelling process, therefore long-term and short-term forecasts can have indistinguishable data uncertainty. The underlying distribution of the ILI proportion on a set day is independent of the forecast horizon. For example, when making a forecast for the ILI proportion on the $30^\text{th}$ of September, the data uncertainty should be identical to a forecast made on the $1^\text{st}$ or  $29^\text{th}$ of September. The extra uncertainty for long-term forecasts comes from the modelling process, not the data.

Adding the data uncertainty term back in, the estimate is now given by
\begin{equation}
    f^{p(\bm{\Phi})}(x) = x+p(\bm{\Phi}_a)+\mathcal{N}(0, p(\bm{\Phi}_\sigma)),
\end{equation}
keeping sampling from $p(\bm{\Phi})$ to only at $t_0$, but outputting a distribution instead of a single value. We feed back values from this distribution each timestep to make subsequent forecasts. 
Sampling from the data uncertainty distribution mirrors sampling from the parameter distribution i.e., the uncertainty only increases if it is sampled. Given that data uncertainty should be independent of time, we chose to feed back the mean of each prediction into the model, recombining it with the model uncertainty post-predictions by using Eq~\ref{eq:combine_uncertainties}.

We modify the IRNN to improve estimate uncertainty by sampling the parameter distribution once before making predictions and changing the RNN layer to use weight distributions.

\subsection{Modifications to the IRNN}
\label{sec:IRNN_Change}
The IRNN estimates model uncertainty in its dense layer, while the RNN layer is deterministic and provides a low dimensional representation of the data which the dense layer utilises to forecast. Changing to a fully Bayesian neural network where all the parameters are defined by distributions increases the model's capacity to express uncertainty, at the cost of computational complexity.

The training setup for the IRNN uses $K=1$ samples for each training step. This works well for the existing version of the model which samples each timestep, but not when sampling only once at the start of making predictions. We found that using only one sample for the entire prediction resulted in the data uncertainty shrinking to zero during training.
We can instead use $K>1$ samples and calculate the combined uncertainty during training. The higher the value of $K$ the more accurate the approximation of the combined uncertainty, but with a trade-off of increased training time. We found empirically that $K=3$ was the minimum value which improves the model performance and still trains at a reasonable speed. 

All our models use a \texttt{tensorflow-probability} implementation of a dense layer: the modeller specifies the form of the prior and posterior distributions while the \texttt{tensorflow} backend handles the sampling. However, at the time of writing, there is no built-in implementation of Bayesian RNN layers and it is impossible to change the sampling method in the \texttt{tensorflow-probability} Bayesian dense layer. To allow different sampling options, we made custom versions of a GRU and dense layer. The backend of \texttt{tensorflow} utilises the ``reparametrisation trick''~\cite{kingma2013auto} to allow sampling to occur during training. It is impossible to calculate the gradients of a random process, required for training, so the trick moves the randomness outside the model by adding it as an additional input. We implement this method: we sample from additional variable $\epsilon_\text{repr} \sim \mathcal{N}(0,1)$ externally to the training loop. Subsequent reparameterisation by  $\theta^\prime = \mu_\theta + \epsilon_\text{repr} \odot \sigma_\theta$, where $\odot$ represents element-wise multiplication, makes the model deterministic for the sake of back-propagation.

Thus we have a modified IRNN to use only Bayesian layers, train using the combined uncertainty, and sample only at $t_0$ rather than every timestep. We denote the modified IRNN as IRNN$_s$, where the $_s$ refers to the change in the sampling methodology. 

\subsection{IRNNs Results}
We compare the IRNN$_s$ with the IRNN from~\cite{morris2023neural}. We maintain the same metrics and training and evaluation periods. We re-run the hyperparameter tuning for the IRNN$_s$ using the same methodologies employed for the IRNN. 

Table~\ref{tab:sampling_comparison_dante} enumerates the performance metrics for the IRNN$_s$, the IRNN, and Dante for each flu season and forecast horizon. The IRNN$_s$ has superior Skill, MAE, and bivariate correlation to the IRNN for $\gamma=7$ and $\gamma=14$. For $\gamma=21$ IRNN$_s$ performs similarly to the IRNN, with marginally lower Skill. For ~$\gamma=28$ days ahead the gap between the two NNs is greatest and the IRNN is slightly better. 

\begin{figure}[!ht]
    \centering
    \includegraphics[width=0.98\linewidth]{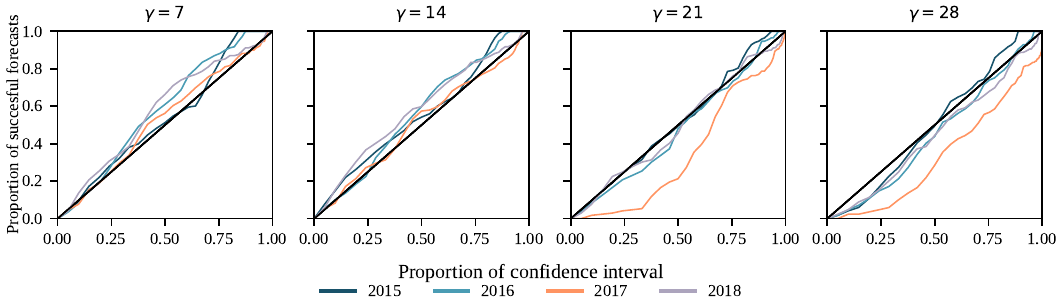}
    \caption[{calibration plots for all test periods for IRNN$_s$}]{\textbf{Calibration plots for all test periods for IRNN$_s$} \newline
    Calibration plots for the forecasts made by the IRNN$_s$ model for (2015/16 to 2018/19) and shown for the four forecasting horizons ($\gamma$). The lines show how frequently the ground truth falls within a confidence interval (CI) of the same level. To be more precise, a point $(x,y)$ denotes that the proportion $y \in [0, 1]$ of the forecasts when combined with a CI at the $x \times 100\%$ level includes the ground truth (successful forecasts). The optimal calibration is shown by the diagonal black line. Points above or below the diagonal indicate an over- or under-estimation of uncertainty, and hence an under- or over-confident model, respectively. The area between each calibration curve and the optimal calibration is given in Table~\ref{tab:sampling_comparison_dante}.
    }
    \label{fig:sampling_calibration_plot}
\end{figure}

We introduce \CalibMetric~which is the area between the calibration curve (Figure~\ref{fig:sampling_calibration_plot}) and the ideal calibration. A lower \CalibMetric~indicates a better-calibrated uncertainty, meaning the confidence interval size more closely aligns with the proportion of successful forecasts. Note that the \CalibMetric~ metric is unaffected by the overall accuracy of the model. As such, a model with poor accuracy but good confidence intervals can score well, whereas Skill combines these into one metric --- weighting the accuracy of the forecast with the quality of the confidence interval. The IRNN$_s$ has the best uncertainty calibration of the three models. 

\begin{figure}[!ht]
    \centering
    \includegraphics[width=0.98\linewidth]{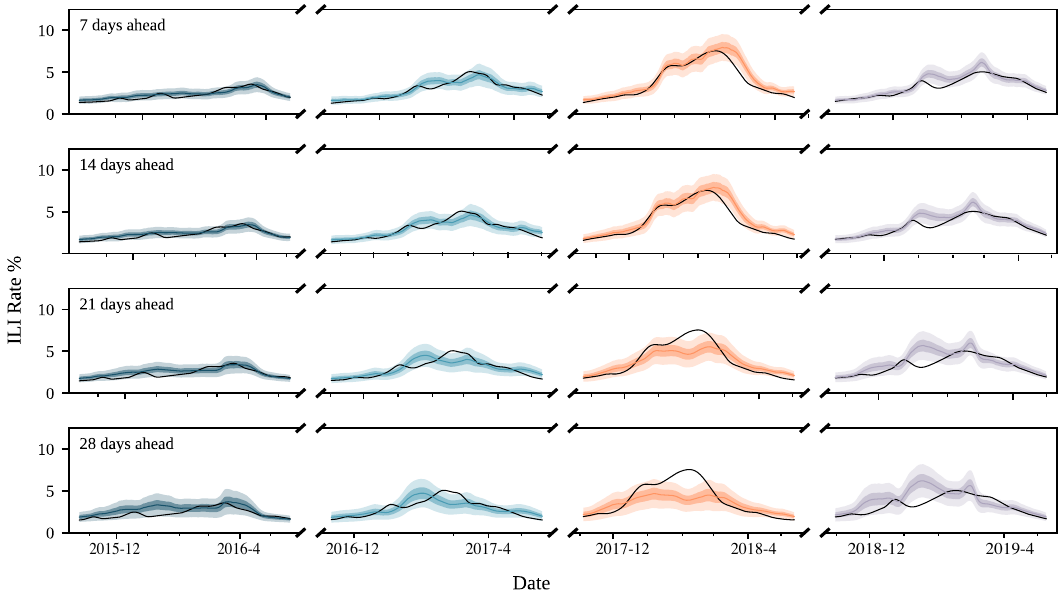}
    \caption[{IRNN$_s$ Forecasts}]{\textbf{IRNN$_s$ Forecasts} \newline IRNN$_s$ forecasts for all four test seasons (2015/16 to 2018/19) and forecasting horizons ($\gamma =$ 7, 14, 21, and 28). Confidence intervals (uncertainty estimates) are shown at $50\%$ and $90\%$ levels and are visually distinguished by darker and lighter colour overlays, respectively. The influenza-like illness (ILI) proportion (ground truth) is shown by the black line.}
\label{fig:IRNNs_forecasts}
\end{figure}

\begin{figure}[!ht]
    \centering
    \includegraphics[width=0.98\linewidth]{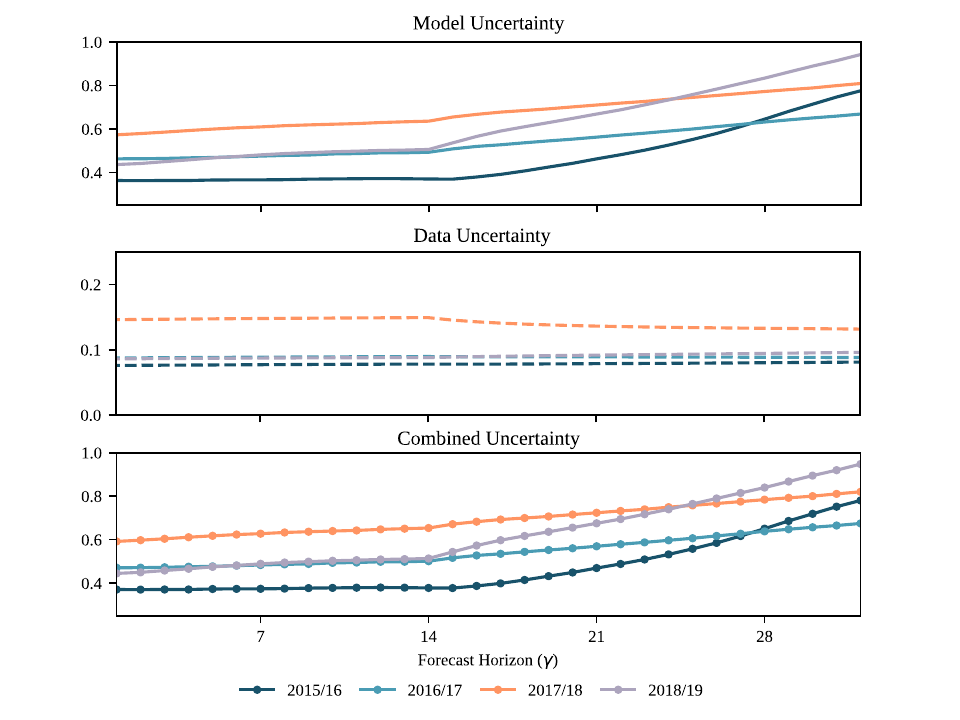}

    \caption[{Averaged uncertainty estimates for varying forecast horizon in IRNN$_s$}]{\textbf{Averaged uncertainty estimates for varying forecast horizon in IRNN$_s$}\newline
    Uncertainty shown is one standard deviation --- computed by finding the square root of the average variance for each day in the forecast season. In each season the uncertainty stays constant up to the most
     recent search data ($\gamma=14$). 
    }
    \label{fig:sampling_uncertainty_plot}
\end{figure}

Forecasts from IRNN$_s$ in every season and forecast horizon are shown in Figure~\ref{fig:IRNNs_forecasts}. Similarly to the IRNN, the decline in accuracy for longer horizons is obvious from the forecast plots. As with the IRNN, the IRNN$_s$ makes independent forecasts that do not necessarily follow the previously observed ILI proportions. Unlike with the IRNN, it is clear that the uncertainty bounds increase with $\gamma$. 

Figure~\ref{fig:sampling_uncertainty_plot} shows how the various uncertainties (model, data, combined) change as $\gamma$  increases. The model uncertainty is larger than the data uncertainty --- varying from $\approx 0.4$ to $\approx 0.95$, while the data uncertainty varies from $\approx 0.075$ to $\approx 0.15$. The model uncertainty is near constant for horizons with Web search data, then for $\gamma>14$ the model uncertainty increases linearly. Data uncertainty is independent of the forecast horizon, observable in $2015/16, 2016/17$ and $2018/19$. 

Figure~\ref{fig:sampling_calibration_plot} shows the calibration of the confidence intervals (CI) for the IRNN$_s$. This figure corresponds to Figure~\ref{fig:calibration_plot} for the IRNN. We compute the empirical probability for each of the four test seasons. The diagonal line ($y=x$) represents perfect calibration i.e. the expected and empirical probabilities are the same. Points above the diagonal indicate that the uncertainty estimates are too large and points below it indicate that the uncertainty estimates are too low. 

\begin{sidewaystable}[!ht]
    \centering
    \small
    \setlength{\tabcolsep}{2pt}
    \begin{tabular}{p{1.1cm}p{1.1cm}p{1.2cm}p{1.2cm}p{1.2cm}p{1.2cm}p{1.2cm}p{1.2cm}p{1.2cm}p{1.2cm}p{1.2cm}p{1.2cm}p{1.2cm}p{1.2cm}p{1.2cm}p{1.2cm}p{1.2cm}}
Horizon   & Metric  & \multicolumn{3}{c}{2015/16}       & \multicolumn{3}{c}{2016/17}       & \multicolumn{3}{c}{2017/18}       & \multicolumn{3}{c}{2018/19}       &\multicolumn{3}{c}{Avg (2015-19)}  \\ 
\cmidrule(lr){3-5}\cmidrule(lr){6-8}\cmidrule(lr){9-11}\cmidrule(lr){12-14}\cmidrule(lr){15-17}
$\gamma$  &         & Dante     & IRNN      & IRNN$_s$  & Dante     & IRNN      & IRNN$_s$  & Dante     & IRNN      & IRNN$_s$  & Dante     & IRNN      & IRNN$_s$  & Dante     & IRNN      & IRNN$_s$  \\
\midrule
7         & Skill   & 0.67      & 0.74      & \textbf{0.76} & 0.63      & 0.60      & \textbf{0.64} & 0.45      & \textbf{0.51} & 0.49      & \textbf{0.62} & 0.61      & 0.60      & 0.59      & 0.61      & \textbf{0.62} \\
          & MAE     & \textbf{0.22} & 0.25      & 0.24      & \textbf{0.19} & 0.32      & 0.29      & 0.39      & \textbf{0.36} & 0.47      & \textbf{0.21} & 0.28      & 0.34      & \textbf{0.25} & 0.30      & 0.33      \\
          & $r$     & 0.88      & 0.86      & \textbf{0.92} & \textbf{0.96} & 0.93      & 0.94      & 0.97      & \textbf{0.98} & 0.97      & \textbf{0.97} & 0.93      & 0.93      & \textbf{0.94} & 0.93      & 0.94      \\
     & \CalibMetric & 0.09      & 0.10      & \textbf{0.04} & 0.18      & \textbf{0.05} & 0.08      & 0.09      & 0.06      & \textbf{0.03} & 0.16      & 0.14      & \textbf{0.08} & 0.13      & 0.09      & \textbf{0.06} \\
\midrule
14        & Skill   & 0.54      & 0.69      & \textbf{0.75} & 0.54      & 0.53      & \textbf{0.61} & 0.29      & 0.43      & \textbf{0.47} & 0.52      & 0.58      & \textbf{0.59} & 0.46      & 0.55      & \textbf{0.60} \\
          & MAE     & 0.38      & 0.28      & \textbf{0.24} & \textbf{0.32} & 0.38      & \textbf{0.32} & 0.64      & \textbf{0.44} & 0.50      & 0.33      & \textbf{0.31} & 0.35      & 0.42      & \textbf{0.35} & \textbf{0.35} \\
          & $r$     & 0.64      & 0.79      & \textbf{0.91} & 0.91      & 0.90      & \textbf{0.92} & 0.90      & \textbf{0.97} & \textbf{0.97} & \textbf{0.92} & 0.90      & \textbf{0.92} & 0.84      & 0.89      & \textbf{0.93} \\
     & \CalibMetric & 0.05      & \textbf{0.03} & 0.05      & 0.10      & \textbf{0.04} & 0.06      & 0.04      & 0.07      & \textbf{0.02} & 0.10      & 0.14      & \textbf{0.08} & 0.07      & 0.07      & \textbf{0.05} \\
\midrule
21        & Skill   & 0.44      & 0.60      & \textbf{0.65} & 0.48      & 0.43      & \textbf{0.49} & 0.21      & \textbf{0.37} & 0.33      & 0.46      & \textbf{0.51} & 0.41      & 0.38      & \textbf{0.47} & 0.46      \\
          & MAE     & 0.48      & 0.35      & \textbf{0.32} & \textbf{0.38} & 0.50      & 0.46      & 0.86      & \textbf{0.64} & 0.77      & \textbf{0.40} & 0.42      & 0.59      & 0.53      & \textbf{0.48} & 0.54      \\
          & $r$     & 0.36      & 0.71      & \textbf{0.80} & \textbf{0.87} & 0.80      & 0.82      & 0.82      & \textbf{0.92} & \textbf{0.94} & \textbf{0.89} & 0.82      & 0.74      & 0.73      & 0.81      & \textbf{0.83} \\
     & \CalibMetric & 0.06      & 0.03      & \textbf{0.02} & 0.07      & 0.10      & \textbf{0.03} & \textbf{0.03} & 0.06      & 0.15      & 0.10      & 0.08      & \textbf{0.02} & 0.07      & 0.07      & \textbf{0.06} \\
\midrule    
28        & Skill   & 0.37      & 0.49      & \textbf{0.51} & \textbf{0.46} & 0.39      & 0.43      & 0.17      & \textbf{0.21} & 0.20      & 0.42      & \textbf{0.48} & 0.32      & 0.33      & \textbf{0.37} & 0.34      \\
          & MAE     & 0.54      & 0.51      & \textbf{0.46} & \textbf{0.39} & 0.57      & 0.54      & 1.06      & \textbf{0.86} & 1.01      & \textbf{0.45} & 0.50      & 0.78      & \textbf{0.61} & \textbf{0.61} & 0.70      \\
          & $r$     & 0.23      & \textbf{0.69} & 0.68      & \textbf{0.88} & 0.74      & 0.76      & 0.76      & \textbf{0.90} & 0.88      & \textbf{0.86} & 0.79      & 0.63      & 0.68      & \textbf{0.78} & 0.74      \\
     & \CalibMetric & 0.05      & \textbf{0.03} & 0.05      & 0.08      & 0.09      & \textbf{0.05} & \textbf{0.08} & 0.12      & 0.17      & 0.10      & \textbf{0.05} & \textbf{0.05} & \textbf{0.08} & \textbf{0.08} & \textbf{0.08} \\
\bottomrule
\end{tabular}

    \caption[{IRNN$_s$ comparison}]{\textbf{IRNN$_s$ comparison} \newline 
    Forecasting performance metrics for the IRNN$_s$ compared with Dante and the IRNN. The best results for this comparison are shown in bold. Both NNs maintain the temporal advantage of Web search queries and do not use leave-one-season-out. \CalibMetric~ is the area between the calibration curve (see Figure~\ref{fig:sampling_calibration_plot}) and the optimum calibration. }
    \label{tab:sampling_comparison_dante}
\end{sidewaystable}

\subsection{IRNNs Discussion}
The IRNN$_s$ outperforms Dante in terms of Skill for every forecast horizon. The average Skill for all horizons and seasons is $0.49$ for IRNN and IRNN$_s$, whereas Dante's average skill is $0.43$ --- approximately $13\%$ lower.  The IRNN$_s$ has lower Skill for $21$ and $28$ days ahead. We can attribute this drop off in Skill to the $2018/19$ season where it deteriorates dramatically between $\gamma=14$ and $\gamma=21$ days ahead. 
Ignoring this season from all model averages results in the IRNN$_s$ being $4\%$ and $2.5\%$ better in terms of skill for $\gamma=21$ and $28$ days ahead, respectively, compared with the IRNN (which is superior to Dante). Despite the poor Skill and MAE in $2018/19$, IRNN$_s$ produces significantly better-calibrated uncertainty estimates than the other two models as evidenced by \CalibMetric. Of note, the $2016/17$ and $2018/19$ seasons have similar epidemic trajectories, and the model uncertainty for the two is almost identical up to $\gamma=14$. For $\gamma>14$, the uncertainty increases more sharply for $2018/19$ indicating that the search queries in $2018/19$ increase the model uncertainty, highlighting that the season has unusual search trends thus making it more challenging to forecast.

The performance of the IRNN$_s$ model, in terms of mean absolute error (MAE), consistently lags behind that of the standard IRNN model across all forecast horizons, except for the $\gamma=14$ days ahead horizon. However, despite its inferior accuracy in predicting the means of the forecast, the IRNN$_s$ exhibits comparable or superior skill. This suggests that the modifications made to the model have notably improved its ability to estimate uncertainty.

The standard IRNN model outperforms the IRNN$_s$ by approximately $8\%$ in terms of MAE when averaged over all seasons and horizons despite both models sharing equivalent architectures. This discrepancy in MAE highlights the importance of refining hyperparameters and optimising the training setup for the IRNN$_s$. Further improving its predictive accuracy would lead to a greater improvement in skill. 

In terms of \CalibMetric,~the IRNN$_s$ is unparalleled when averaged over all seasons and horizons at $0.06$, compared with $0.07$ for IRNN, and $0.09$ for Dante. This also includes the $2017/18$ season where the \CalibMetric~ is unexpectedly poor for $21$ and $28$ days ahead. For $2017/18$ the data uncertainty is initially much higher than the other seasons, then drops off when $\gamma>14$. Figure~\ref{fig:sampling_calibration_plot} shows that the uncertainty is well calibrated for $\gamma\le14$, implying the greater initial uncertainty is appropriate for the difficulty of the season. 
For $\gamma>14$ the data uncertainty decreases since the model forecasts the query data beyond $\gamma=14$ and it will tend to produce forecasts of queries (and ILI proportions) in line with its training experience. Therefore, inputs for later timesteps are likely to be more in line with the training data, and thus may exhibit lower data uncertainty. As discussed, model uncertainty increases with time, which should offset this. However, for $2017/18$ the model uncertainty does not increase enough to result in well-calibrated uncertainty for longer horizons. We previously found this season is the most difficult to forecast of the four seasons\cite{morris2023neural}, and all models perform worst here. Removing the $2017/18$ season from the average \CalibMetric, then the IRNN$_s$ improves to $0.05$, the IRNN stays the same at $0.07$, and Dante's \CalibMetric score increases to $0.10$. 

Despite the IRNN$_s$ model exhibiting a higher MAE compared to the standard IRNN across most forecasting tasks, its improved skill highlights its utility in practical forecasting scenarios. The IRNN$_s$'s better ability to estimate uncertainty, as evidenced by its comparable or superior skill despite its slightly worse accuracy in predicting mean values, shows it is a more valuable tool for decision-making under uncertainty.

In essence, a model's effectiveness in capturing and quantifying uncertainty is critical in many real-world applications. Thus, the IRNN$_s$'s capability to provide more accurate and reliable uncertainty estimates outweighs its minor deficiency in mean forecast accuracy. Consequently, the modifications introduced to the IRNN$_s$ architecture represent a significant advancement, aligning it as the preferred choice for forecasting tasks where precise uncertainty estimation is critical for decision-making.

The presented methodology improves the uncertainty estimation of the IRNN without significantly changing the underlying architecture. Further tuning and refinement of the training process would result in further improved performance on the ILI forecasting task. The architecture remains applicable to different data sources, diseases, and potentially even different forecasting problems altogether.
\newcommand{\FA}{$\text{ODE}_{\text{B}}$}
\newcommand{\FAQS}{$\text{ODE}_{\text{B}~\text{Q}}$}
\newcommand{\SIRSIMPLE}{$\text{SIR}_\text{B}$}
\newcommand{\SIRNN}{$\text{SIR}_{\text{Adv}}$}
\newcommand{\SIRFA}{$\text{SIR}_{\text{Adv}~\text{U}}$}
\newcommand{\SIRQS}{$\text{SIR}_{\text{Adv}~\text{Q}}$}
\newcommand{\SEIRNN}{$\text{SEIR}_\text{Adv}$}
\newcommand{\SEIRFA}{$\text{SEIR}_{\text{Adv}~\text{U}}$}

\chapter{Physics Informed Neural Ordinary Differential Equations for Disease Forecasting}
\label{chapterlabel2}
In this chapter, we combine mechanistic models with neural networks to incorporate the complimentary benefits of both. We evaluate the performance of eight such models for ILI forecasting in the United States.

\section{Introduction}
Mechanistic and non-mechanistic models are suited to different applications and forecast targets.  
% Forecasts of infectious diseases have different requirements depending on their use case. For example, during the Covid-19 pandemic, the peak number of hospital beds required was a more meaningful forecasting target than the case count, and the relationship between the two varied over time \cite{nyberg2022comparative}. When forecasting for scenario planning, the capability to vary known parameters such as mobility is important in order to see the effect on disease transmission~\cite{ferguson2006strategies, ferguson2020impact, chang2021mobility}. For other interventions such as vaccination trials, estimating the case counts over time in different areas is more useful. 
Non-mechanistic forecasting models, such as neural networks, are well suited to dealing with noisy data with non-linear relationships between variables. Consequently, in ILI forecasting, they produce good-quality forecasts which tend to be more accurate than mechanistic models~\cite{reich2019collaborative}. However, mechanistic models have several advantages over non-mechanistic models. Their physical constraints mean that they automatically incorporate expected patterns and behaviours
% e.g., mechanistic models consistently produce epidemic curves that rise and eventually decline to zero.
which reduces the need for acquired knowledge through training and so reduces the training set size when compared with NNs~\cite{karpatne2017physics}. By modelling physical properties which we can measure and interpret, mechanistic models enable a better understanding of transmission dynamics which allows easier scenario planning.

A disadvantage of mechanistic models is that they rely on (often restrictive) assumptions to enable them to model the real world. Simple models have more restrictive assumptions and are consequently less flexible. They also have a reduced number of parameters, which makes them easy to train, at the expense of the ability to model complex phenomena. A more complex model is the inverse - they relax their assumptions, going so far as to model each member of a population individually. However, this increases the number of modelled parameters which makes training harder~\cite{burnham2004multimodel}. Increasing the number of parameters can increase the model's flexibility, enabling it to fit more detailed nuances of the real world, but it also poses risks. Overly complex models with too many parameters can overfit to the training data, and so fail to generalise to unseen data.

Mechanistic models are key for understanding the processes behind disease transmission, however, in their basic forms, many disease models lack the flexibility to be able to fit complex disease trajectories which limits their accuracy for forecasting~\cite{reich2019collaborative}. More advanced mechanistic disease models require more inputs, such as mobility data~\cite{chang2021mobility}. A generalised way of increasing the flexibility of mechanistic models would be beneficial for both forecasting and understanding the driving factors behind disease transmission.
We can combine a mechanistic model with a non-mechanistic component to account for the discrepancies arising from an imprecise mechanistic model. Neural ordinary differential equations ~\cite{chen2018neural} (described in Section~\ref{sec:background}) provide a method of doing this which brings together neural networks and ordinary differential equations (ODEs) in a unified framework. These in turn can be combined with mechanistic ODE models to create universal differential equations (UDEs) i.e., an ODE which is defined in full or in part by a universal approximator, (something which can approximate any function) and can therefore fit to any trajectory. Universal differential equations benefit from the physical constraints of mechanistic models and the modelling ability of neural networks.

The next section provides a background on mechanistic models and discusses how they can be combined with neural ODEs to create epidemic models which can fit more complex epidemic curves than a purely mechanistic model. We then evaluate several models on synthetic data, showing how traditional ODEs, neural ODEs, and UDEs are used for estimation. We also provide a parameter sensitivity analysis of an SIR model, which we use to inform our design of forecasting models. We then apply UDEs for forecasting influenza in the US over four flu seasons and four forecast horizons. We provide a comparative analysis with the IRNN from Chapter~\ref{chapterlabel1}. Although the ODE models are unable to provide the same accuracy, they have several useful properties, and we discuss avenues for future work to improve their forecast performance.

\section{Background and Related Work}
\label{sec:background}
In this section, we first provide an overview of ordinary differential equations (ODEs). A basic understanding of ODEs is required for an understanding of existing mechanistic models for epidemic modelling. We then provide an overview of neural ODEs and examples of how they can be combined with mechanistic models and neural networks to create forecasts. 

\subsection{Ordinary Differential Equations}

\begin{figure}
    \centering 
    \includegraphics[width=\textwidth]{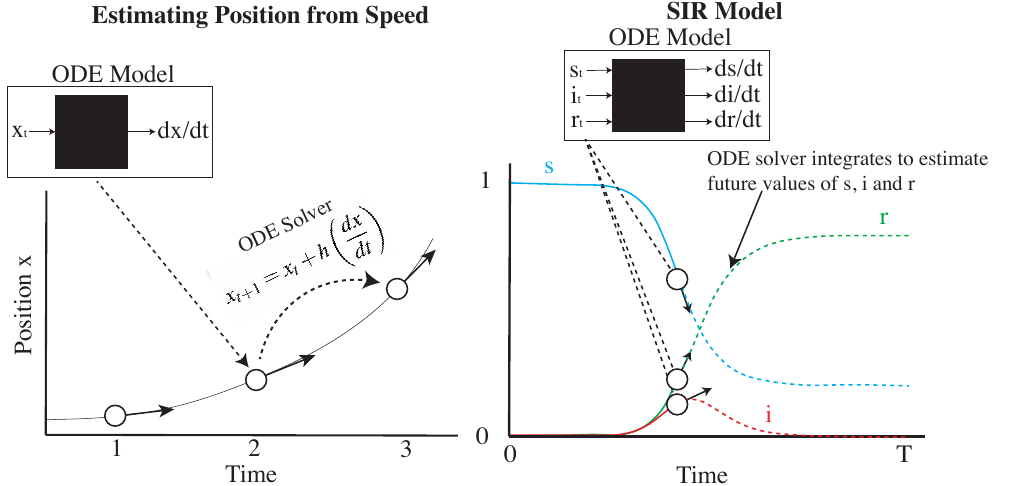}
    \caption[{Overview of ODE models for time series}]{\textbf{Overview of ODE models for time series} \newline 
    The figure on the left shows the one-dimensional problem of estimating position $x$ based on the speed $dx/dt$. An ODE model observes the position $x$ at time $t$ and estimates the speed $dx/dt$. An ODE solver (namely Euler's method) integrates the speed to give the position. 
    On the right-hand side, an SIR model gives the susceptible $s$, infected $i$ and recovered $r$ fractions. Here the ODE model observes the fractions in each compartment and gives the gradient of each trajectory. This works similarly to the position/speed example, but the ODE has three components instead of one.}
    \label{fig:basic_ODE_diagram}
\end{figure}

An ordinary differential equation contains the derivative of an unknown function. The ODE can be integrated (either analytically or using an ODE solver) to find the function itself.  The most basic ODE solver is Euler's method:
\begin{equation}
x_{t+h} = x_t + h \left(\frac{dx}{dt}\right),
\end{equation}
where $x_{t+h}$ is the estimated value of a time series $x$ at time $t+h$, $\frac{dx}{dt}$ is the ODE i.e., the rate of change of $x$ at time $t$, and \gls{time_step} is the step size. By re-evaluating $dx/dt$ after each timestep the original trajectory can be constructed from the gradients at each timestep. Reducing the timestep $h$ improves the accuracy of the ODE solver at the cost of more function evaluations. Euler's method is the most simple ODE solver, more complex methods such as Runge-Kutte are more accurate for the same step size (See Appendix \ref{sec:appendixRK4} for details). Other ODE solvers can adapt the step size to reduce the number of function evaluations while still producing accurate estimates. A diagram showing an overview of how ODEs are used to estimate a function is provided in Figure~\ref{fig:basic_ODE_diagram}. Here we show an imagined one-dimensional time series. The speed of an object ($dx/dt$) is known and is integrated to give its position for three timesteps in the future using an ODE solver. We also show the most basic epidemiological model --- the SIR model, which estimates the susceptible, infectious and recovered fractions (\gls{s_i_r}) of a population at time $t$. Their derivatives are described by the ODE model which is used in combination with an ODE solver to give an estimate of their trajectories. More detail on the SIR model and its derivatives is provided in the next section. 

In the provided examples, the initial values of the time series (initial conditions) are known. This does not always reflect reality as we cannot accurately measure people's interactions with a disease at a population level and thus do not know the initial conditions for epidemic models. In real-world modelling situations, the initial conditions must be estimated. Neural networks provide a method of estimating the initial conditions based on observations of the target time series. This is discussed further in Section~\ref{sec:VAE}.

A further challenge of ODE models is that in complex real-world systems, we do not always know the true form of the ODE. This is often the case in disease modelling, where complex interactions between individuals are not captured by simplified population-level equations. To minimise the error caused by poorly specified equations, we can include a discrepancy model~\cite{osthus2019dynamic},  a non-mechanistic model which minimises the error between the ODE solver's output and the target time series. 
However, we still may not know the parameters of our underlying ODE. In this case, it is common to learn the ODE parameters in real-time~\cite{osthus2019dynamic, shaman2009absolute, shaman2012forecasting}. 
It is possible to jointly apply each of these solutions using neural ODEs~\cite{chen2018neural}, a framework where a neural network acts as an ODE and an ODE solver is used to estimate the original function. This is discussed in Section~\ref{sec:NODES}.

\subsection{Mechanistic Models}

Mechanistic models for epidemic modelling tend to either model individual behaviour with agent-based models, or homogeneous mixing, where groups exhibiting identical behaviour are modelled. Agent-based models simulate individual behaviour of entities called agents, which represent individuals in the real world. Agents interact with one another according to pre-defined rules based on real-world behaviour. These are the most flexible mechanistic epidemic models, but due to the large number of parameters and the impossibility of differentiating through them, fitting their parameters to data is difficult. This limitation means agent-based models are seldom used for influenza forecasting, though they have been used to for scenario planning for hypothetical situations in an epidemic~\cite{ferguson2006strategies, ferguson2020impact}. 

Conversely, compartmental models separate a population of individuals or hosts according to their disease status~\cite{keeling2005networks}, referred to as compartmental models. Each compartment represents a different stage of the disease, such as susceptible, infected, and recovered (\gls{sir}). Disease characteristics, such as transmission rate and recovery time, determine how the population moves between compartments. Compartmental models tend to use ODEs to model transitions between compartments, however, these can also be modelled using agents

\subsubsection{Susceptible, Infected, Recovered Model}
Susceptible, infected, recovered (SIR) models~\cite{kermack1927contribution} are the simplest family of compartmental models. They are defined by the following set of ODEs which describe how the population moves from being susceptible to a disease, catching it and being infected, and then recovering (or dying):
\begin{align}
    \frac{dS}{dt} &= -\beta SI \\
    \frac{dI}{dt} &= \beta SI - \omega I \\
    \frac{dR}{dt} &= \omega I ~.
    \label{eq:SIR}
\end{align}
Here, \gls{beta} represents the rate at which infected individuals transmit the infection to susceptible individuals per unit of time, and $\omega$ is the probability of an infected individual recovering per unit of time. \gls{S_I_R} are the absolute number of susceptible, infected and recovered individuals, respectively. $S+I+R=N_\text{pop}$, where \gls{N_pop} is the total population.  We use $s=S/N_\text{pop}$, $i=I/N_\text{pop}$ and $r=R/N_\text{pop}$ as the susceptible, infected and recovered fraction of the population. Modelling the infected fraction is a design choice which we make which simplifies the model design by keeping the parameters in similar ranges to other compartmental models. Another consideration is that ILI is recorded as a percentage of doctor visits so there is no relevant population size. 

SIR models allow the measurement of key properties of a disease. The effective reproductive number \gls{R_e}$ = \frac{s\beta}{\omega}$ is the average number of secondary infections which an infected individual will produce before recovering. 
\gls{R_0}$=\frac{\beta}{\omega}$~\cite{weiss2013sir} is the number of secondary infections the average infectious person would produce in a fully susceptible population. When $R_e>1$ the gradient $dI/dt$ will increase exponentially. Throughout an epidemic, $s$ will decrease, in turn decreasing $R_e$. When $R_e<1$ the disease will die out as carriers recover faster than they cause new infections. 
Disease spread can be managed by reducing the rate of individual infection $\beta$, which is a product of mobility and the infectiousness of a disease. Mobility can be changed through interventions such as lockdowns or school closures which reduce the number of interactions between. The infectiousness can change naturally through mutations in the disease or environmental factors like weather~\cite{shaman2009absolute}, it can also be changed by vaccination~\cite{anderson1992infectious}. SIR models are used to give information to public health workers to help calculate the percentage of a population which needs to be vaccinated in order to prevent an epidemic by keeping $R_e<1$~\cite{bjornstad2020modeling}. 

While SIR models serve as vital tools for informing public health decisions, the fundamental assumptions they rely on may not fully capture the complexities of real-world epidemics, particularly in their simplest form. Here we discuss some of the assumptions which are made by the basic SIR model, as we note, each assumption are removed in more sophisticated versions used for public health decision-making. However, doing so invariably introduces complexity, and each assumption requires its own modification. We will later introduce a universal differential equation model which can remove all assumptions in a unified solution. 

\begin{itemize}[topsep=0pt, partopsep=0pt, itemsep=0pt,parsep=0pt]
\item \textbf{Homogeneous mixing}: Basic SIR models assume uniform mixing, with every individual having an equal chance of encountering an infected individual. However, models used for public health decisions often incorporate heterogeneous mixing patterns, acknowledging that interactions between individuals vary across populations and contexts.
\item \textbf{Fixed Population}: The basic model assumes a constant population size with no births deaths or immigration. However, more advanced models may incorporate demographic factors such as births, deaths, and migration to better reflect real-world dynamics.
\item \textbf{No latency}: SIR models assume that individuals immediately become infectious upon infection. However, more sophisticated models may include latency periods, recognising that there can be a delay between infection and the onset of infectiousness.
\item \textbf{Perfect immunity}: Recovered individuals are immune to the disease for as long as the model is run. Advanced models may account for waning immunity over time, allowing for more realistic representations of disease dynamics, particularly for diseases with temporary immunity.
\item \textbf{Constant rates}: The disease mechanics are constant throughout the epidemic. While basic models assume fixed disease parameters throughout an epidemic, more complex models may allow for time-varying parameters to capture changes in transmission dynamics over time.
\item \textbf{Disease-only deaths}: The natural death rate is not considered. In more comprehensive models, natural death rates may be incorporated alongside disease-related mortality.
\item \textbf{Deterministic}: Basic SIR models are deterministic and do not account for stochastic fluctuations or uncertainty. However, stochastic versions of the model are commonly employed in public health contexts to better capture the inherent randomness in disease transmission.
\end{itemize}

While some of these assumptions are still simplified representations of reality, extensions and modifications to the basic SIR framework continue to improve the accuracy of models used in public health decision-making. 

\subsubsection{Extensions to SIR Model}
Here we discuss several extensions to the basic SIR model. When describing the different models we use the original notation from each paper, in some instances this differs from the notation we use elsewhere in this thesis.

Shaman et al.~\cite{shaman2013real, yang2014comparison}, employ a humidity-forced SIRS model. A SIRS model is similar to an SIR model but allows individuals to return from recovered to susceptible after a set time. This is the case with illnesses that may be caught more than once, such as influenza. Their SIRS model can only model a single circulating disease, making it unable to estimate the signal for ILI ---- a signal which is affected by multiple circulating diseases with overlapping symptoms such as RSV and Covid-19. The SIRS model is defined as:
\begin{align}
    \label{EQ:SIRS_S}
    \frac{dS}{dt} &= \frac{R}{L} - \frac{\beta I S}{N_\text{pop}} - \alpha \\
    \frac{dI}{dt} &= \frac{\beta I S}{N_\text{pop}} - \frac{I}{D} + \alpha ,
\end{align}
where $L$ is the average duration of immunity, $D$ is the infectious period, and $\alpha$ is the rate of travel-related import of influenza. The contact rate is calculated for each timestep, $\beta(t) = R_0(t)/D$, where $R_0(t)$ is the time-varying reproductive number.  This is affected by the absolute humidity: increasing when the humidity decreases and vice versa. It is calculated at time $t$ by:
\begin{equation}
    R_0(t) = R_{0_\text{min}} + \left(R_{0_\text{max}} - R_{0_\text{min}}\right) e^{-aq(t)}
\end{equation}
where $R_{0_\text{max}}$ and $R_{0_\text{min}}$ are maximum and minimum values of $R_0$, $a=180$ is a constant determined by lab work \cite{shaman2009absolute}, and $q(t)$ is the time-varying specific humidity. This method relaxes the assumption that the parameters are constant, instead making them dependent on humidity. Consequently, the Shaman SIRS model requires humidity data throughout the forecasting period (ahead of $t_0$); therefore to produce forecasts, forecasts of the specific humidity are also required. 

The initial conditions are set for week $40$ and are determined by running the model from $1973$ to $2012$ and evaluating the distribution over model states in the final year. 
In our own experiments, we found this method to be unreliable and required manual tuning to get good estimates from the model. 

Instead of creating a proxy for influenza, Osthus et al.~\cite{osthus2019dynamic} estimate the logit function of the ``true but unobservable proportion of influenza-like illness'' in week $t$ and flu season $j$ as the sum of three components,
\begin{equation}
    \text{logit}(\pi_{j,t}) = \text{logit}(I_{j,t}) + \mu_t + \delta_{j,t},
\end{equation}
where $\text{logit}(I_{j,t})$ is the infected population from an SIR model in flu season $j$ at week $t$, $\mu_t$ is a discrepancy component common to all flu seasons, and $\delta_{j,t}$ is a discrepancy component specific to each flu season. The discrepancy models terms are defined by non-mechanistic models, more specifically, reverse-random-walks~\cite{osthus2019dynamic} \footnote{A random walk is a stochastic model where each step is determined by a random process, often involving sampling from a specific distribution. A reverse random walk, on the other hand, involves retracing the steps of a random walk in reverse order. It is particularly useful for understanding how a certain state in a process was reached or for analyzing the system from a time-reversed perspective. The choice of model is a design decision taken by the authors.}. The SIR component uses a parameter distribution to estimate model uncertainty, while the discrepancy components minimise the error between the SIR output and the measured ILI proportion. The discrepancy components use random walk models which are trained jointly with the SIR model. 

The estimate of the observed ILI proportion $y_{j,t} \sim \mathcal{N}(\pi_{j,t}, \sigma)$ is a Normal distribution with mean and standard deviation given by:
\begin{equation}
    \mu(y_{j,t}) = \pi_{j,t}
\end{equation}
\begin{equation}
    \sigma(y_{j,t}) = \left(\frac{\pi_{j,t}(1-\pi_{j,t})}{1+\lambda}\right)^{0.5} ~~,
\end{equation}
where  $\pi_{j,t}$ is the true but unobservable ILI proportion, and $\lambda$ relates to the data uncertainty in the measured ILI proportion. The authors attribute the data uncertainty to sampling variability, ILI diagnosis errors, and reporting variability. 

Empirical Bayes is used to set the prior distribution over the initial conditions and parameters; models are fit to previous seasons, and the trained parameters inform the prior in the current season. The initial susceptible population $s_0$ is set to $0.9$ for all flu seasons as there is insufficient information to estimate it from the data. As we discuss later, we found that the models are sensitive to the initial susceptible population, so find this solution of fixing $s_0$ unsatisfactory. 

This model is difficult to generalise as it was developed specifically in the context of ILI forecasting in the United States. The model is retrained from scratch each time a new ILI proportion is observed and requires a large amount of data to set the initial conditions, making it inapplicable to novel diseases. 

In Chang et al.~\cite{chang2021mobility} the authors use a population-based Susceptible Exposed Infectious Removed (\gls{seir}) model to estimate the spread of Covid-19. The SEIR model is an extension of the standard SIR model, containing an additional compartment for members of the population who have been exposed to a disease but are not yet infectious. The authors use multiple SEIR models in a metapopulation model. These separate a population into discrete ``patches'' which interact according to predefined rules. Specifically, the model separates the population into census block groups (CGBs) where each CBG has its own SEIR model that interacts with one another, and the inter-patch interactions model the spread of Covid-19 between CBGs. 
Mobility data influences the infectiousness of the disease in each CGB. 

The estimates from the metapopulation model align well with measured Covid-19 case data in the UK. However, the model was only used in hindsight and does not apply to forecasting as future mobility is unknown. This method introduces significant computational complexity and was not used for forecasting; instead modelling how Covid-19 affected people based on their socio-economic status. 

Other metapopulation models have estimated the diffusion of diseases in a population~\cite{balcan2009seasonal, tizzoni2012real}. 
Like agent-based models, metapopulation models can show the effects of interventions designed to reduce the spread of a disease~\cite{balcan2009seasonal}. They are seldom used for forecasting; in Tizzoni et al.~\cite{tizzoni2012real} a mechanistic model is used to predict the peak of an epidemic. However, the authors note their dependency on good quality and continually updated data from many sources. The models are unsuitable for real-world forecasting because this data is not available in real time. 

A recurring issue with mechanistic models is the disease prevalence they are modelling is difficult or impossible to measure. 
If instead, they are modelling an available proxy, such as ILI, the issue becomes that the relationship between the proxy and the case count is unknown. The method proposed by Osthus et al.~\cite{osthus2019dynamic} goes some way to address this: modelling the discrepancy between the estimated and observed ILI proportions. 
However, their model is specific to the task of ILI forecasting, in that their method of estimating the initial conditions does not apply to other diseases where less data is available, or where $s_0$ is unknown. This means that the model does not have applicability beyond ILI forecasting.

Next, we discuss neural ODEs, an alternative method of discrepancy modelling which are more flexible and generalisable than random walks. We then introduce variational autoencoders which provide an architecture which can combine neural ODEs with methods to estimate initial conditions for any point in the flu season.

\subsection{Neural ODEs}
\label{sec:NODES}
\begin{figure}
    \centering 
    \includegraphics[width=0.471\textwidth]{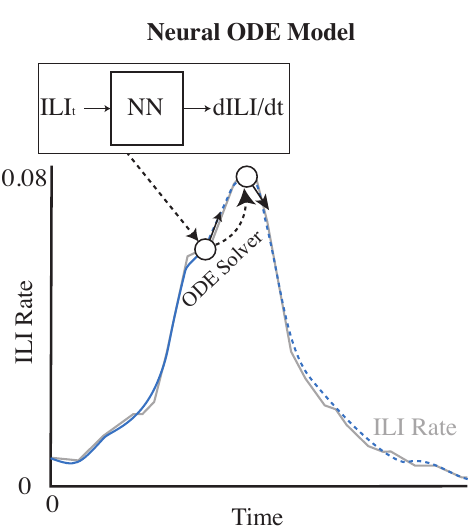}
    \caption[{Overview of a Neural Ordinary Differential Equation model for time series}]{\textbf{Overview of a Neural Ordinary Differential Equation model for time series} \newline 
    The figure shows how an ODE function can be replaced by a neural network, and used with an ODE solver to create a neural-ODE. }
    \label{fig:neural_ode_example}
\end{figure}
Neural ODEs (\gls{n_ode}s)~\cite{chen2018neural} bring together neural networks and ordinary differential equations (ODEs) in a unified framework. Neural ODEs frame an ODE as a neural network, which estimates the gradient of a function over time. An ODE solver integrates the gradient to give an estimate of the original function.
Mathematically, a neural ODE takes the form:
\begin{equation}
\frac{d\mathbf{x}_t}{dt} = f\left(\mathbf{x}, t, \bm{\Phi}\right),
\end{equation}
where $\mathbf{x}_t$ are inputs, $t$ is the current time, and $f(.)$ is a function parametrised by $\bm{\Phi}$ --- the learnable parameters of the network. The model produces estimates for $t\in\{0...T\}$ by observing a values at $\mathbf{x}_0$ and integrating them forwards to $\mathbf{x}_T$. In contrast, a traditional NN takes an observation $\mathbf{x_t}$ and applies discrete transformations to it by 
\begin{equation}
    \mathbf{x}_{t+1} = f\left(\mathbf{x}_t, \bm{\Phi}\right).
\end{equation}
Traditional neural networks operate in discrete, predefined steps, but neural ODEs evolve over time with dynamically changing step sizes. This offers a more fluid understanding of data transformations. An overview of how a neural ODE can be used for time series modelling is provided in Figure~\ref{fig:neural_ode_example}.

The ability to continuously transform functions has opened up several use cases for neural ODEs, including continuous depth neural networks \cite{poli2020hypersolvers}, normalising flows\cite{yang2019pointflow}, which are common in image and text generation, and time series modelling\cite{rubanova2019latent, de2019gru, NEURIPS2020_4a5876b4}. The continuous nature of neural ODEs allows them to capture temporal dependencies in time-series-data which is beneficial when data is missing or irregularly sampled~\cite{rubanova2019latent}. Neural ODEs are scalable~\cite{poli2020hypersolvers} and easy to combine with other ODE models~\cite{rackauckas2020universal}. Consequently, the powerful modelling capabilities of neural networks can be combined with the physical constraints of mechanistic models; these are referred to as universal differential equations. 

\subsubsection{Universal Differential Equations}
\begin{figure}
    \centering 
    \includegraphics[width=\textwidth]{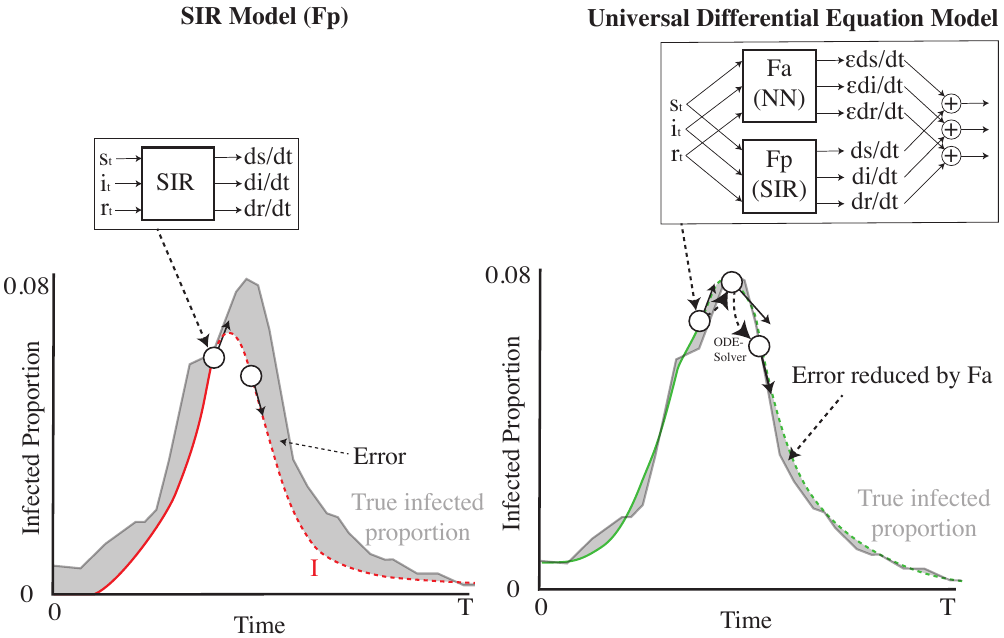}
    \caption[{Overview of a Universal Differential Equation (UDE) model for time series}]{\textbf{Overview of a Universal Differential Equation (UDE) model for time series} \newline 
    The figure on the left shows an SIR model's estimate for the infected proportion in an ILI modelling task. This example assumes that the infected proportion is known, however as we discuss in Section~\ref{sec:VAE}, the true infected proportion is unknown and must be estimated from the ILI proportion. The figure highlights that the best fit of a simple model (the SIR model) may not be accurate for a complex time series - there is a significant error highlighted in grey. The SIR model produces estimates for $ds/dt$, $di/dt$ and $dr/dt$ using the SIR equations (Eq.~\ref{eq:SIR}).
    The figure on the right shows how a universal differential equation can be constructed from a physical model \gls{Fp}, namely an SIR model and a neural network augmentation model \gls{Fa}. Fa reduces the error in $ds/dt$, $di/dt$ and $dr/dt$ resulting in a model which can fit to more complex time series. The UDE's output is $f^{\beta, \omega}(\mathbf{x})+f^{\bm{\Phi}}(\mathbf{x})$.}
    \label{fig:UDE_Example}
\end{figure}

\label{sec:UDEs}
Universal differential equations (UDEs) are differential equations which are defined in full or part by a universal approximator. Neural networks are a common universal approximator which works well in high dimensions~\cite{rackauckas2020universal}. We can embed a mechanistic ODE such as an SIR model with a neural ODE to create a universal differential equation. The neural ODEs represent the unknown or complex dynamics of the system and enable it to fit more complex data than the original mechanistic model allows. A UDE can be constructed from an SIR model, where the states $[s, i, r] = \mathbf{x}$:
\begin{equation}
    \frac{ds}{dt} = -\beta si + f^{\bm{\Phi}}(\mathbf{x})_1
\end{equation}
\begin{equation}
    \frac{di}{dt} = \beta si - \omega i  + f^{\bm{\Phi}}(\mathbf{x})_2
\end{equation}
\begin{equation}
    \frac{dr}{dt} = \omega i + f^{\bm{\Phi}}(\mathbf{x})_3 ~,
\end{equation}
where $f^{\bm{\Phi}}(\mathbf{x})$ is a neural network with parameters $\bm{\Phi}$ and conditioned on the states of model $\mathbf{x}$. The subscript $_1,_2,_3$ refers to the three outputs of the NN. The physical component of the equations is commonly denoted $F_p$, and the augmentation model (the neural network) is $F_a$. These models are flexible and applicable to a variety of situations~\cite{rackauckas2020universal}. 

UDEs, and ODEs in general, require accurate initial conditions to produce good forecasts. In epidemic modelling, it is impossible to measure the initial conditions directly from the population. This necessitates developing a robust method of estimating the initial conditions. We investigate variational autoencoders (\gls{vae}s), which provide an end-to-end method of estimating initial conditions and forecasting, using UDEs.

\subsubsection{Variational Autoencoders}
\label{sec:VAE}
The measured ILI proportion is different to the true infected proportion which is modelled by a compartmental model.  To produce a forecast using a compartmental model it is therefore important to estimate the initial conditions (proportions) in each compartment e.g., $s_0$, $i_0$ and  $r_0$. The initial conditions are then integrated forwards using an ODE such as an SIR model. The outputs of the SIR model are time series of the proportions in each compartment. However, we need to convert these back into the measured ILI proportion to compare it with the ground truth. We can use neural networks to estimate the initial conditions from measurements and to estimate the measurements based on the outputs of the ODE. Variational autoencoders\cite{kingma2013auto} (VAEs) provide a good framework for doing this.  

VAE are a class of generative neural networks which consist of an encoder and a decoder. The encoder observes inputs and encodes them into latent representations (in our case initial conditions of the ODE). The decoder reconstructs observations based on the latent representation. 
The latent representation compresses the data, forcing the encoder to learn the underlying distribution of the data which is then reconstructed back into the original data by the decoder. 
The latent representation is a distribution which is sampled at reconstruction time; different samples will produce different outputs, thus expressing uncertainty. 

In \cite{yildiz2019ode2vae, portwood2019turbulence, xie2019neural} forecasting models inspired by variational autoencoders (VAE) use neural ODEs to produce continuous-time forecasts. These architectures consist of three parts: the encoder, latent-ODE model, and decoder. The encoder observes a time series and produces the latent representation for a single timestep. The latent representation is integrated forward in time using the latent-ODE model to create a latent trajectory, which is reconstructed into the target time series by the decoder. Here, the latent time series is the output of a compartmental model containing estimates of the true proportion of infection.
Typically, the latent-ODE model is a neural ODE, but we show that UDEs can be used instead. An advantage of using a VAE framework is that the data does not have to be at the same scale as the latent representation. For example, we can observe the ILI proportion, a measure of how many people are visiting the doctor with ILI symptoms, and estimate the true infected proportion in the population which may be very different. 

\begin{figure}
    \centering
    \includegraphics[width=\textwidth]{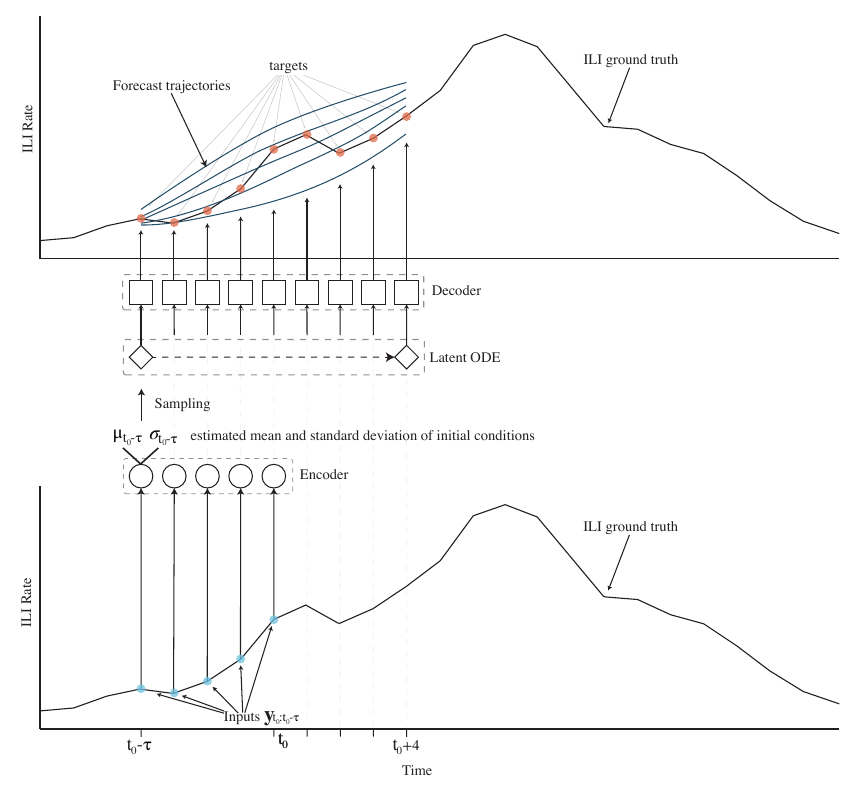}
    \caption[{Variational-autoencoder (VAE) diagram for forecasting with neural ODEs}]{\textbf{Variational-autoencoder (VAE) diagram for forecasting with neural ODEs} \newline 
    The encoder at the bottom of the diagram observes inputs backwards in time $\mathbf{y}_{t_0:t_0-\tau}$ where $\tau$ is the window, and $t_0$ is the time from which the forecasts are made. The encoder creates a distribution of the initial conditions for the ODE model $q(z_{t_0-\tau}|\mathbf{y}_{t_0:t_0-\tau})$. Initial conditions are sampled from the distribution and the latent ODE model integrates the initial conditions forwards in time from $t_0-\tau$ to $t_0+\gamma$. The decoder observes the states from the latent ODE and converts each trajectory back into the same domain as the inputs $\mathbf{\hat{y}}^\prime_{t_0-\tau:t_0+\gamma}$, where $^\prime$ denotes a sample from the latent initial conditions. The mean and standard deviation of the $K$ samples are used to construct a distribution predictive distribution. 
    }
    \label{fig:VAE_Diagram}
\end{figure}

Figure~\ref{fig:VAE_Diagram} shows the operation of a time series VAE using a neural ODE to integrate the latent variables forward in time. The encoder observes a window of the time series $\mathbf{y}_{t_0:t_0-\tau}$ which is used to estimate the distribution over the latent initial conditions \gls{z}$_{t_0-\tau}=\mathcal{N}\left(\mu_{t_0-\tau}, \sigma_{t_0-\tau}\right)$.  Recurrent neural networks (RNN) are a common encoder architecture for time series VAEs, having found use in representation learning, classification, and forecasting~\cite{srivastava2015unsupervised, lotter2016deep, li2018disentangled, locatello2019challenging}.  
The distribution over the initial conditions is sampled $K$ times, and each of the samples is integrated forward in time, using an ODE solver from $t_0-\tau$ to $t_0+\gamma$. Each latent trajectory $\mathbf{z}^\prime_{t_0-\tau:t_0+\gamma}$ is decoded back into the domain of inputs, giving forecasts $\hat{\mathbf{y}}^\prime_{t_0-\tau:t_0+\gamma}$. The $K$ samples from the initial conditions result in $K$ forecasts which are used to construct a Normal distribution which is used for training and to create confidence intervals. 
The VAE is trained by maximising the evidence-lower-bound (ELBO, Eq~\ref{eq:elbo}), the same loss function as the IRNN in Chapter \ref{chapterlabel2}
\begin{equation}
\label{eq:ELBOchapter2}
        \text{ELBO}(q)= \mathbb{E}\left[\log\left(p(\mathbf{x}|\mathbf{z})\right)\right]- D_{\text{KL}}\left[q(\mathbf{z})||p(\mathbf{z})\right] .
\end{equation}

The ELBO is calculated over the full trajectory from $t_0-\tau$ to $t_0+\gamma$.
The loss maximises the probability of observing the data given the parameters with a KL divergence regularisation term. The KL divergence is calculated between the distribution over latent initial conditions $q_{\bm{\Phi}}(z_{t_0-\tau}| \mathbf{y}_{t_0:t_0-\tau})$ and a prior $p(z_{t_0-\tau})$ that is specified at the start of training. 

\subsubsection{VAE Uncertainty Estimation}
\label{sec:VAE_uncertainty}
VAEs primarily focus on modelling data uncertainty but also have implications for model uncertainty. VAEs explicitly estimate the inherent variability in the data and express it in their latent representations. 
In Bayesian neural networks, uncertainty is typically estimated by placing a distribution over parameters as discussed in Chapter~\ref{chapterlabel1}. Whilst the VAE framework does not explicitly address model uncertainty, the latent distribution can be seen as expressing uncertainty in the latent variables, given the observed data. VAEs typically optimise parameters to a point estimate, so do not fully capture the uncertainty in model parameters. To explicitly estimate model uncertainty in a VAE, the weights can be represented by distributions referred to as a Bayesian VAE~\cite{daxberger2019bayesian}, where the encoder and decoder are themselves Bayesian neural networks. We choose not to develop Bayesian VAEs at this time because it introduces additional modelling complexities, and, as we discuss later, the uncertainty estimation is not the performance bottleneck in our models.

Next, we experiment using synthetic data to evaluate a simple ODE problem. We then apply neural ODEs, compartmental models, and UDEs to synthetic epidemiological data, we also investigate the parameter sensitivity of an SIR model. Finally, we built a variational autoencoder based model and then applied eight variations of a VAE model to forecasting ILI in the US. These include traditional compartmental models, neural ODEs, and UDEs. Variational autoencoders have been combined with neural ODEs~\cite{yildiz2019ode2vae}, however to our knowledge they have not been combined with UDEs. We provide a comparative analysis of ODE models with the IRNN from Chapter~\ref{chapterlabel1}.

\section{Synthetic Data Experiments}
Here we provide examples of various ODEs. We begin with a basic description of how an ODE is integrated to give a prediction, and then how its parameters can be estimated from measured values via back-propagation. We then evaluate the sensitivity of SIR models on synthetic data, show that a neural ODE can approximate an SIR model, and demonstrate that a UDE with a simple physical model can approximate a more complex physical model. Finally, we apply UDEs to real-world ILI data for England and show that a UDE can be accurate whilst maintaining a simple physical component.

\subsection{ODE Example}
We show how an ODE can be integrated with Euler's method --- the simplest ODE solver.
Euler's method is used to estimate the position $x_2$ of an object at time $t=2$ given an ODE which describes its velocity (the derivative of position) $\frac{dx}{dt} = 3$ and position at $x_0=0$. We can approximate the integral of the ODE by taking small steps along the gradient of the function, using Euler's method:
\begin{equation}
x_{t+1} = x_t + h \left(\frac{dx}{dt}\right),
\end{equation}
where $x_{t+1}$ is the estimated position at time $t+1$, $\left(\frac{dx}{dt}\right)$ is the velocity at time $t$, and $h$ is the step size. Given the initial position $x_0=0$, for a step size $h=1$ we can integrate from $t_0$ to $t_2$:
\begin{align*}
x_1 &= x_0 + h \left(\frac{dx}{dt}\right) = 3 \\
x_2 &= x_1 + h \left(\frac{dx}{dt}\right) = 6 ~ .
\end{align*}
Therefore, the position $x_2$ at time $t=2$ is $6$.

Next, we show how the parameters of an ODE can be found if points on its trajectory are known. Using the same example, if the positions $x_0 = 0$ and $x_2 = 6$ are known, we can specify an equation $\frac{dx}{dt} = f^{\phi}(x,t)$, where $f^{\phi}(x,t)$ is an unknown function with parameters $\phi$. The function can take any form, but for simplicity, we use $f^{\phi}(x,t) = \phi$. To find $\phi$ using back-propagation, the following steps are taken:
\begin{enumerate}[topsep=0pt, partopsep=0pt, itemsep=0pt,parsep=0pt]
    \item Initialise $\phi$ with a random value.
    \item Use an ODE solver to find the estimated position $\hat{x}_2$ given $x_0$, $\phi$ and $h$.
    \item Calculate the loss $\mathcal{L}(\phi) = (\hat{x_2} - x_2)^2$ i.e. as the squared difference between the predicted and actual positions at $t=2$.
    \item Compute the gradient of the loss with respect to $\phi$: $\frac{d\mathcal{L}(\phi)}{d\phi}$ using auto-differentiation through the ODE solver. (Auto-differentiation is a standard computational tool for computing gradients, and is used almost universally in the machine-learning community.)
    \item Update $\phi$ using gradient descent, with $\alpha$ as the learning rate:
    \begin{equation*}
    \phi_{\text{new}} = \phi_{\text{old}} - \alpha \frac{d\mathcal{L}(\phi)}{d\phi}.
    \end{equation*}
    \item Repeat steps $2$-$5$ until the loss reduces to an acceptable value. 
\end{enumerate}
This method iteratively updates the value of $\phi$ to minimise the difference between the predicted and measured positions, thus estimating the unknown parameter $\phi$.

\subsection{Neural ODE Example}
We show that a neural ODE can reproduce the output of an SIR model. 
We first specify initial conditions: $[s_0, i_0, r_0]$ and use a neural network in place of the ODE equations. 
The inputs to the NN are the time $(t)$ and the states of the ODE at time $t$, $[s_t, i_t, r_t]$; the outputs are $[\frac{ds}{dt}, \frac{di}{dt}, \frac{dr}{dt}]$. An ODE solver integrates the neural ODE forwards to preset times $[t_0, t_1, t_2, ..., t_n]$, where $t_n$ is the final timestep (or forecast horizon). We compute the mean squared error loss between the target time series and predicted values, and train the weights using back-propagation. Back-propagating through an ODE solver is straightforward but can introduce a high memory cost and numerical error.  The ``adjoint sensitivity method''~\cite{chen2018neural} is an alternative to back-propagation in neural ODEs --- it avoids the numerical problems, has linear complexity, and has low memory cost. In our experiments, we found that the adjoint sensitivity method works well in simple neural ODEs, such as in this example, but not in UDEs. For this reason, we use standard back-propagation in all our experiments. 

\begin{figure}[ht]
    \centering
    \includegraphics[width=0.8\textwidth]{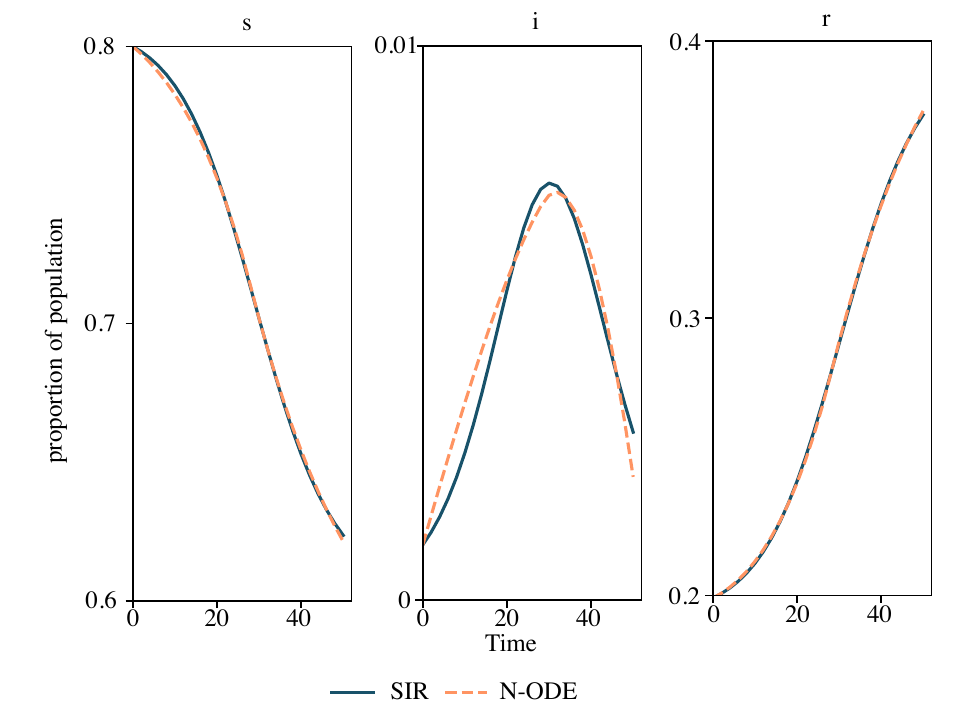}
    \caption[{N-ODE replicating trajectory of an SIR model}]{\textbf{N-ODE replicating trajectory of an SIR model} \newline
    The trajectory of an N-ODE trying to replicate the results an SIR model where $\beta = 2.0$, $\omega = 1.4$ and initial conditions $[s_0, i_0, r_0] = [0.8, 0.001, 0.199]$. The model is integrated for $t \in (0,26)$.}
    \label{fig:SIR_NODE_Comparison}
\end{figure}

Figure~\ref{fig:SIR_NODE_Comparison} shows that a neural ODE can approximate the trajectories of an SIR model. We create a target time series from the SIR model with $\beta = 2.0$ and $\omega = 1.4$. The initial conditions are $[s_0, i_0, r_0] = [0.8, 0.001, 0.199]$).  The neural ODE is a $3$ layer neural network with a hidden layer size of $32$ , and an eLu activation function\cite{clevert2015fast}:
\begin{align}
    \label{eq:elu}
    f(x) = x~\text{if}&~x > 0 \\
    f(x) = e^{x}-1~\text{is}&~x\le{0}.
\end{align}
 The neural ODE is trained for $1000$ epochs by minimising the mean squared error with an Adam optimiser~\cite{kingma2014adam}  (learning rate = $1\times10^{-3}$). The number of epochs is chosen arbitrarily as we are only concerned with the network's modelling capacity and overfitting is not a concern. 
 
 Figure~\ref{fig:SIR_NODE_Comparison} shows that the N-ODE is able to capture the dynamics of the system very closely, and fits the curve for susceptible and recovered almost exactly. For the infected population, there is a small deviation. The deviation in the infected population is caused by the loss function being on a different scale for infected compared to susceptible and recovered. The magnitude of $s$ and $r$ are much greater, so they are prioritised by the optimiser.

\subsection{Universal Differential Equation Example}
\begin{figure}[ht]
    \centering
    \includegraphics[width=0.8\textwidth]{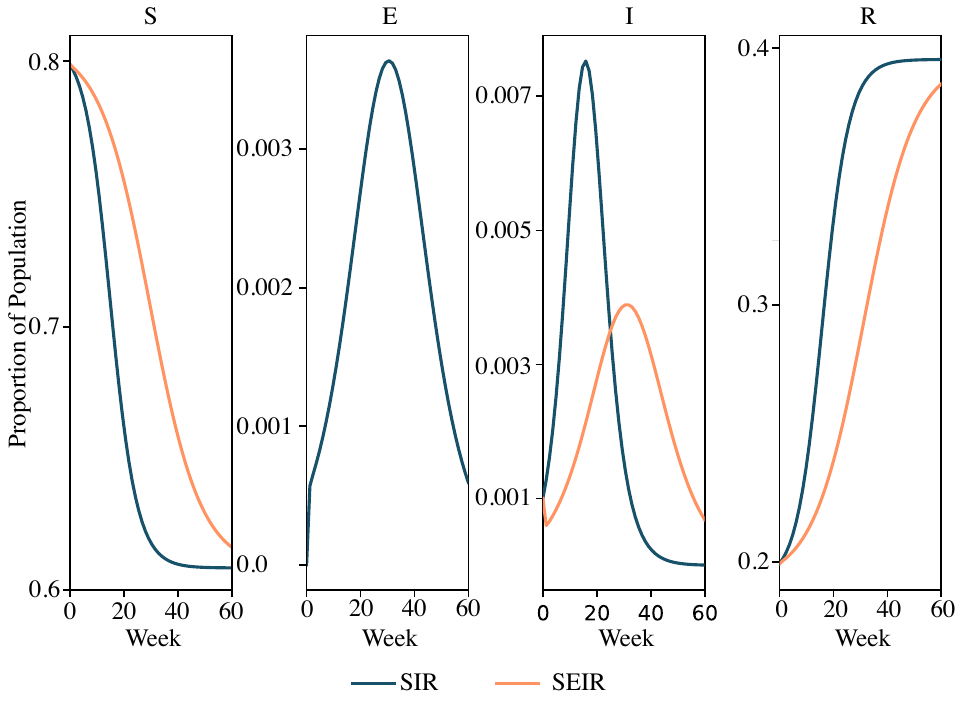}
    \caption[{Prediction of a SEIR and SIR models}]{\textbf{Prediction of a SEIR and SIR models} \newline Both models use $\beta = 2.0$, $\omega = 1.4$, the SEIR model has $\rho = 1.5$. The initial conditions for the SIR model are $[s_0, i_0, r_0] = [0.8, 0.001, 0.199]$, and for the SEIR model are $[s_0, e_0, i_0, r_0] = [0.8, 0.001, 0, 0.199]$. Both models are integrated for $t \in (0,60)$}
    \label{fig:SIR_SEIR}
\end{figure}

We show that a UDE with a simple physical component (SIR) can recreate the results of a more complex compartmental model, namely, a susceptible exposed infectious removed (SEIR) model, defined by the following:
\begin{align}
    \frac{ds}{dt} &= -\beta si \\
    \frac{de}{dt} &= \beta si - \rho e \\
    \frac{di}{dt} &= \rho e - \omega i \\
    \frac{dr}{dt} &= \omega i ~,
\end{align}
where \gls{rho} is the rate of movement from the exposed population to the infected population. The SEIR model parameters are $\beta = 2.0$,  $\omega = 1.4$, $\rho=1.5$. This model is compared to an SIR model using $\beta = 2.0$ and $\omega = 1.4$. The initial conditions for the SIR model are, $[s_0, i_0, r_0] = [0.8, 0.001, 0.199]$, and for the SEIR model are $[s_0, e_0, i_0, r_0] = [0.8, 0.001, 0, 0.199]$. 
Figure~\ref{fig:SIR_SEIR} shows that these two models produce very different epidemic trajectories; the SEIR model has a less severe season but is drawn out over a longer time. Self evidently, but importantly, it is impossible to make the SIR model produce the same infected trajectory as the SEIR model.

\begin{figure}[ht]
    \centering
    \includegraphics[width=0.8\textwidth]{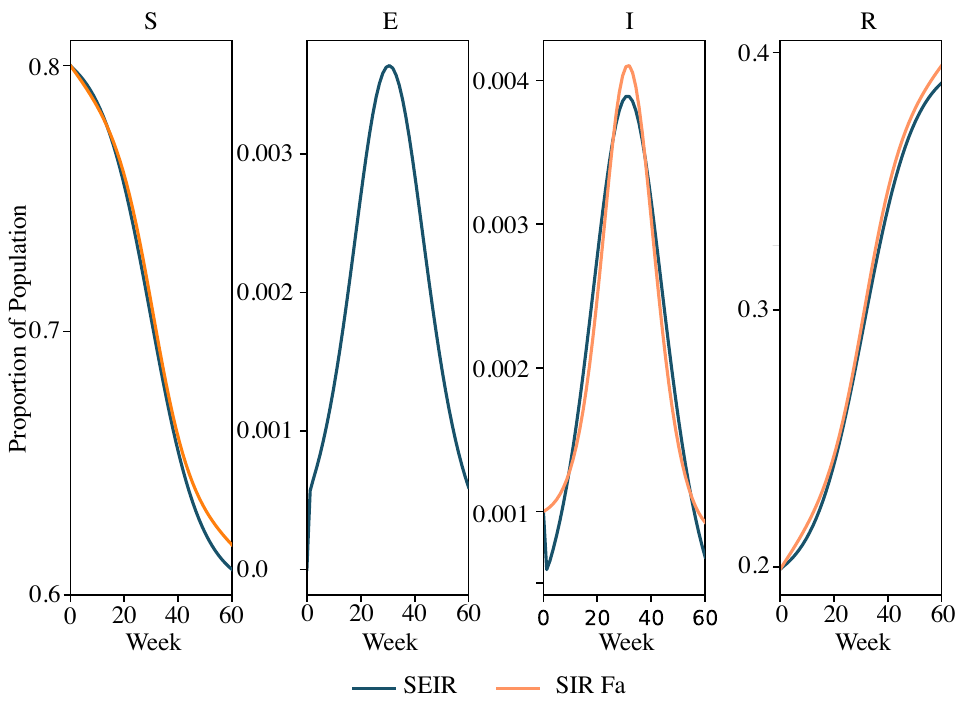}
    \caption[{Prediction of SEIR and SIR-UDE models}]{\textbf{Prediction of SEIR and SIR-UDE models} \newline Both models use $\beta = 2.0$, $\omega = 1.4$, and the SEIR model has $\rho = 1.5$. The initial conditions for the SIR model are $[s_0, i_0, r_0] = [0.8, 0.001, 0.199]$, and for the SEIR model are $[s_0, e_0, i_0, r_0] = [0.8, 0.001, 0, 0.199]$. The SIR model is augmented with a neural ODE which increases the complexity of the physical model. Both models are integrated for $t \in (0,60)$.}
    \label{fig:SEIR_SIR_NODE}
\end{figure}

Next, we convert the SIR model into a UDE, keeping the same $\beta$ and $\omega$  parameters as before. Our model is defined as $f^{\beta, \omega}(\mathbf{x})+f^{\bm{\Phi}}(\mathbf{x})$, where  $f^{\beta, \omega}(\mathbf{x})$  is the SIR model with fixed $\beta$ and $\omega$.  We define the augmentation model $f^{\bm{\Phi}}(\mathbf{x})$ as a Feed Forward neural network with three layers, also referred to as $F_a$. The hidden layers have $20$ units and use an eLu activation function. The NN minimises the error in the output of the ODE. As this is a regression task, $F_a$ does not require an activation function in the final layer. The model uses \gls{rk4} as an ODE solver (Runge-Kutte 4$^\text{th}$ order - see appendix \ref{sec:appendixRK4} for details). We train by minimising the mean squared error between the infected compartment in the SEIR and UDE models from week $0$ to week $60$. We use an Adam optimiser with a learning rate of $0.001$, a batch size of $60$, and $1000$ epochs. 
Figure~\ref{fig:SEIR_SIR_NODE} shows that the UDE can significantly reduce the difference between the SIR and the SEIR models. Thus, modifying an ODE into a UDE with a neural network can improve the model's flexibility and allow it to model data that would otherwise be impossible without a more advanced model.

Although the UDE was trained to minimise the infected compartment, the susceptible and recovered components were similar to the same components in the SEIR model. This is because we force the model to maintain the assumption that the total population size does not change. Adding a neural network to the outputs of a compartmental model will usually cause the sum of the compartments (the population) to vary over time. However, we can prevent this with a simple modification to the output.

\section{Maintaining Physical Assumptions in Neural ODEs}
\label{sec:maintaining_physical_assumptions_odes}
Each term a compartmental model's equations is made up of expressions for groups leaving or joining a compartment. For example, in an SIR model, $\beta si$ is the fraction in the susceptible group who become infected and join the infected group. This term is added to the infected fraction, and subtracted from the susceptible fraction. Similarly, $\omega i$ is the number of people in the infected group who recover from the disease and join the recovered group; this expression is added to the recovered population and subtracted from the infected population. Adding these expressions to one compartment and subtracting them from another ensures that $\frac{ds}{dt} + \frac{di}{dt}+ \frac{dr}{dt} = 0$, i.e. the population is constant without deaths or births. If we add a neural network to the output, i.e., we change our SIR model $f^{\beta, \omega}(\mathbf{x})$ to a UDE $f^{\beta, \omega}(\mathbf{x})+f^{\bm{\Phi}}(\mathbf{x})$, then the neural network will change the population if its outputs do not sum to zero.

We can ensure that the output of an NN sums to zero by modelling the movement between compartments rather than modelling the change to each compartment. 
For example, in an SIR model, we can estimate the error in $\beta s i$ and $\omega i$ rather than the error in $\frac{ds}{dt}$, $\frac{di}{dt}$ and $ \frac{dr}{dt}$. 
We can do this by setting $F_a$ to have two units in its penultimate layer $\mathbf{y}_{l-1} = [y_{l-1,1}, y_{l-1,2}]$, where $l$ is the number of layers in the neural network $F_a$. 
The final layer calculates the output as $f^{\bm{\Phi}}(\mathbf{x}) = [-y_{l-1,1}, y_{l-1,1}-y_{l-1,2}, y_{l-1,2}]$, by fixing the weights and biases. 
From Section~\ref{sec:ff_nn_intro}, the output of a feed-forward neural network layer is $\mathbf{W}^T\mathbf{x} + \mathbf{b}$, we can fix $\mathbf{W}$ and $\mathbf{b}$
\begin{equation*}
\mathbf{W} = \begin{bmatrix}
-1 & 0 \\
1 & -1 \\
0 & 1 \\
\end{bmatrix}, \quad
\mathbf{x} = \begin{bmatrix}
\mathbf{y}_{l-1, 1} \\
\mathbf{y}_{l-1, 2} \\
\end{bmatrix}, \quad
\mathbf{b} = \begin{bmatrix}
0 \\
0 \\
\end{bmatrix}.
\end{equation*}
This ensures that $Fa$ models the change from one compartment to another, and does not change the population. More generally, for a model with three compartments, we can use the weight matrix:
\begin{equation*}
    \mathbf{W} = \begin{bmatrix}
    1  & 1 & 0 \\
    -1  & 0  & 1 \\
    0  & -1  & -1
    \end{bmatrix} ,
    \label{eq:rescale_output}
\end{equation*}
where the penultimate layer should have $3$ units. This models every possible movement from one compartment to another, and is equivalent to a SIRS model, as it allows individuals to move from $r$ to $s$ (somebody losing their immunity) or $s$ to $r$ (somebody gaining immunity without infection e.g., through vaccination).  

We can generalise this method to any number of compartments \gls{n}, the penultimate layer requires $\text{Tri}(n-1)$ units, where $\text{Tri}$ denotes a triangle number, calculated by $\text{Tri}(n) = \sum^n_{k=1} k$. The weights in the output layer will be a $n \times \text{Tri}(n-1)$ matrix. The weights will follow the same pattern of each output being added to one compartment and subtracted from another, such that every output models the movement between compartments. For example, a $5$-compartment model would use the following weight matrix in the output layer:
\begin{equation*}
    \begin{bmatrix}
        1  &  1 &  1 &  1 &  0 &  0 &  0 &  0 &  0 &  0 \\
        -1 &  0 &  0 &  0 &  1 &  1 &  1 &  0 &  0 &  0 \\
        0  & -1 &  0 &  0 & -1 &  0 &  0 &  1 &  1 &  0 \\
        0  &  0 & -1 &  0 &  0 & -1 &  0 & -1 &  0 &  1 \\
        0  &  0 &  0 & -1 &  0 &  0 & -1 &  0 & -1 & -1 
    \end{bmatrix}
    ~.
\end{equation*}

\section{Fitting Compartmental Models to Real World Data}
We fit variations of an SIR model to ILI data for England. The RCGP define the ILI rate as the number of infections per 100,000. We choose to model the fraction in each compartment, using an SIR model with a population size of $1$, therefore $i_t = I/100,000$, where $I$ is the ILI rate from the RCGP. Assuming that the population starts the flu season entirely susceptible (this is rarely true as some people will be vaccinated or may have resistance from something else), then the susceptible fraction $s$ at time $t_n$ is  $s_{t_n} = \int_0^{t_n} 1-i_t \,dt$ and the recovered fraction at time $t$ is $r_t = 1-s_t-i_t$. 

\begin{figure}[!ht]
    \centering
    \includegraphics[width=0.8\textwidth]{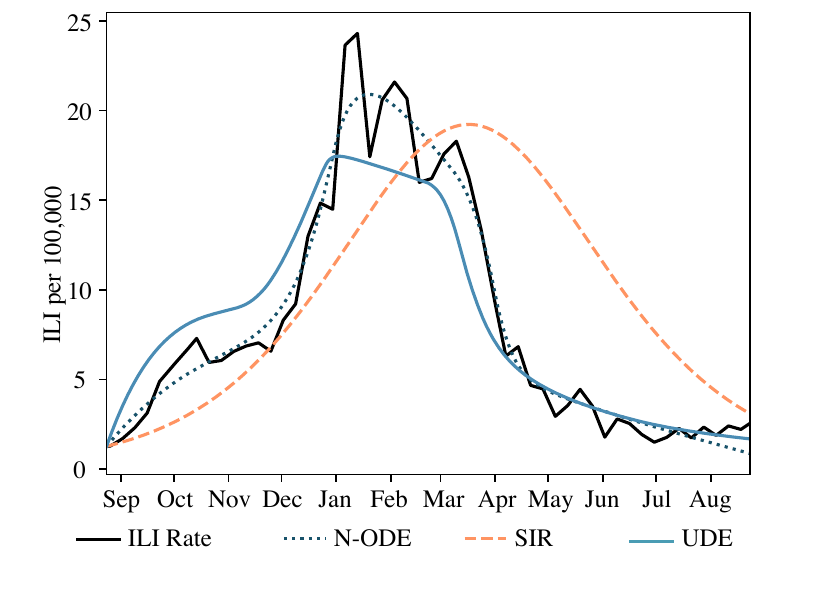}
    \caption[{Predictions of SIR, N-ODE, and UDE models for English ILI rates (2014/15)}]{\textbf{Predictions of SIR, N-ODE, and UDE models for English ILI rates (2014/15)} \newline 
    The three models are trained to minimise the mean squared error in hindsight i.e. no forecasting. The SIR model is too inflexible to fit the measured ILI, and misses the peak of the season. The N-ODE model fits the flu season closely but has no physical component. The UDE model uses an SIR model + N-ODE to augment the output. This obtains benefits of both models but is not as accurate as the N-ODE model because the N-ODE model is able to overfit the data more easily.}
    \label{fig:SIR_NN_UDE}
\end{figure}

We fit $\beta$ and $\omega$ in the SIR model to ILI data from England for the $2014/15$ season using mean squared error as a loss function. Figure~\ref{fig:SIR_NN_UDE} shows that the SIR model is accurate at the start of the season but misses the peak of the season and is unable to model the sudden drop in the ILI rate after the peak. This is caused by the model being too inflexible to fit a real epidemic curve, and not knowing the initial conditions. 

Next, we train an SIR-based UDE model to fit the same data.
Neural networks are typically initialised based on the assumption that the inputs will vary from $0$ to $1.0$ but in compartmental models, we expect that the compartments will be over different scales. Depending on the disease, the susceptible and recovered compartments may vary by up to $1$, but typically the infected fraction is much smaller. In the experiments with an SIR model, $s$ varies from $1.0$ to $0.97$, $i$ varies from $0$ to $0.00025$, and $r$ varies from $0$ to $0.035$. 
As there is no way to rescale the data inside the ODE solver, we instead scale the data inside $F_a$ using a Feed Forward layer with preset fixed weights. Minmax rescales an input $x$ by applying $\frac{x - \text{min}(x)}{\text{max}(x) - \text{min}(x)}$ to the data. A Feed Forward layer applies the transformation $g\left(\mathbf{W}^T\mathbf{x} + \mathbf{b}\right)$, therefore we can replicate a minmax function by setting $\mathbf{W}={1}/\left(\text{max}(X)-\text{min}(\mathbf{x})\right)$, and $b = -{\text{min}(\mathbf{x})}/\left(\text{max}(\mathbf{x})-\text{min}(\mathbf{x})\right)$ where $\mathbf{x}$ are the inputs, in this case $[s,i,r]$. Rescaling in the network ensures that the subsequent layers see inputs which vary from $0.0$ to $1.0$. The rescaling function for the input reduces training time and improves stability. We train the model using the same setup as before, including using the setup from Eq~\ref{eq:rescale_output}.

Figure~\ref{fig:SIR_NN_UDE} compares the trajectories of the SIR-based UDE $f^{\beta, \omega}(\mathbf{x})+f^{\bm{\Phi}}(\mathbf{x}) = F_p+F_a$, an N-ODE trained with no physical component $f^{\bm{\Phi}}(\mathbf{x})$, and an SIR model $f^{\beta, \omega}(\mathbf{x})$. The SIR model is the least able to fit the real-world ILI data. The N-ODE is the most accurate, this is due to its flexibility. However, the results from the N-ODE are no more useful than the predictions of any other neural network, since both are black boxes. The UDE model is slightly less accurate than the N-ODE, but is a significant improvement on the SIR model alone. 

However, there remains the question of how much $F_a$ is contributing to the overall prediction. Ideally, the mechanistic model would contribute as much as possible with $F_a$ contributing a much smaller amount to minimise prediction error. 

\begin{figure}[ht]
    \centering
    \includegraphics[width=0.8\textwidth]{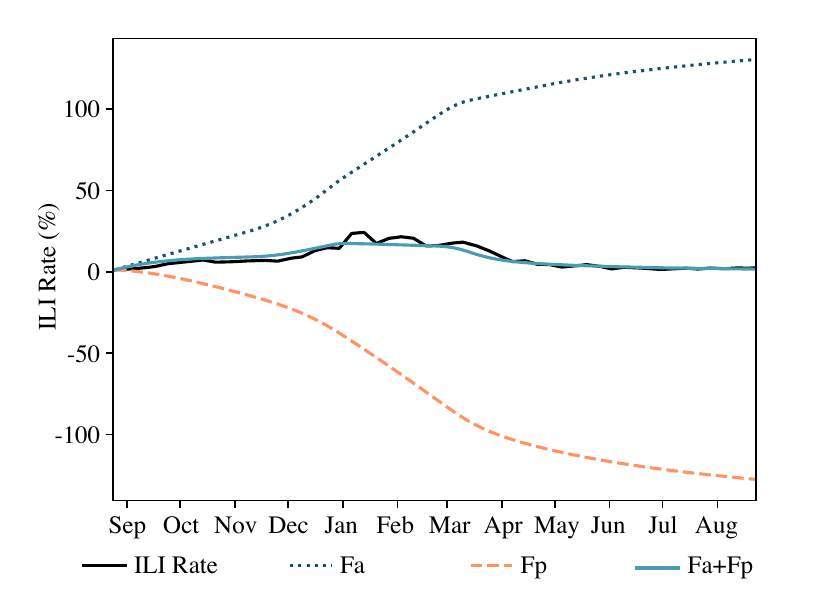}
    \caption[{Individual components of the UDE model}]{\textbf{Individual components of the UDE model} \newline Predictions of the $F_a$ and $F_p$ components of the UDE model. The combined prediction ($F_a + F_p$) is accurate, but the physical model ($F_p$) does not make meaningful predictions. Instead, the augmentation model ($F_a$) corrects the spurious SIR model. Experiments were done for English ILI data in the $2014/15$ season.}
    \label{fig:SIR_NN_no_reg}
\end{figure}

We evaluate the individual components ($F_a$ and $F_p$) of the SIR-based UDE to see what effects they have on the prediction, this is shown in Figure~\ref{fig:SIR_NN_no_reg}. While the combination of the two models works well,  the output of $F_p$ predicts a negative ILI rate as it learned negative values for $\beta$. $F_a$ is sufficiently flexible that it can account for any error, and even though $F_p$ is wrong, $F_a+F_p$ is accurate. The erroneous output from $F_p$ removes the benefit of using a UDE over a neural ODE.

\begin{figure}[!ht]
    \centering
    \includegraphics[width=0.8\textwidth]{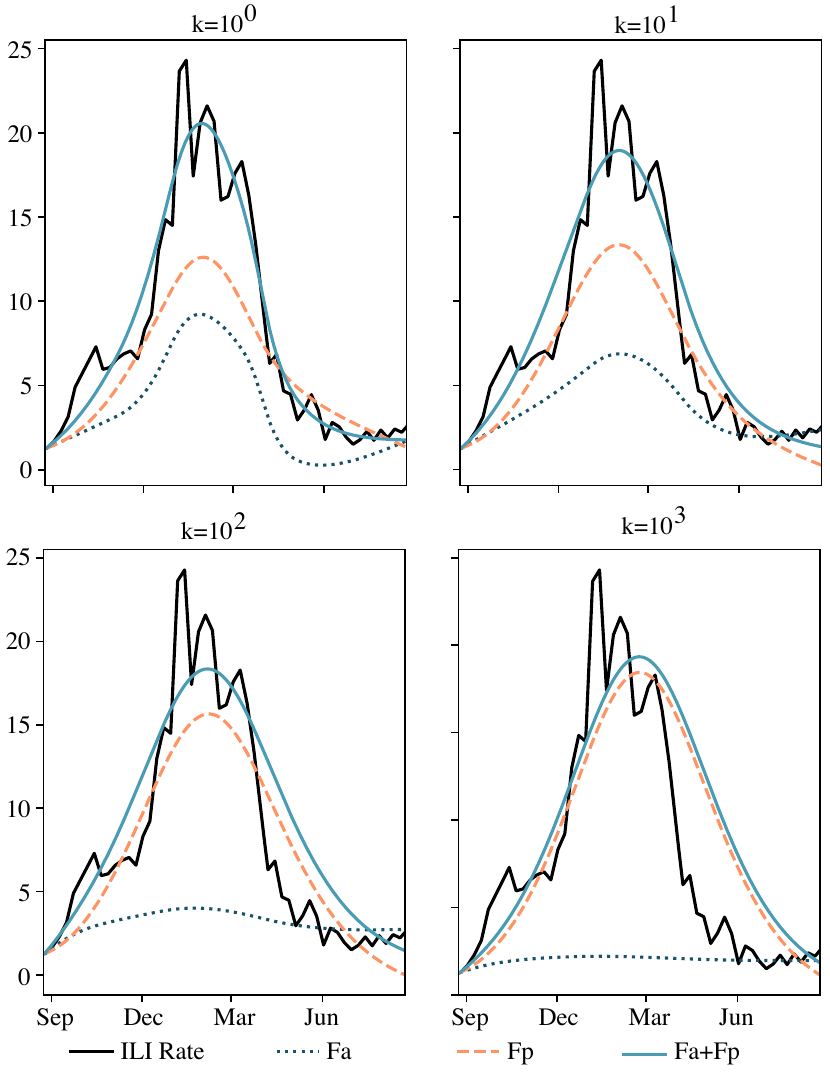}
    \caption[{Changing regularisation amount for the UDE model}]{\textbf{Changing regularisation amount for the UDE model} \newline Predictions for the physical model ($F_p$) and augmentation model ($F_a$) components of the UDE model  ($F_a + F_p$). The regularisation weighted by $\kappa$ reduces the accuracy of the model but forces it to use the physical model ($F_p$). ILI values are shown for England in the $2014/15$ season.}
    \label{fig:SIR_NN_norm}
\end{figure}

To ensure that the model learns meaningful parameters for the physical component, we can regularise $F_a$ \cite{rackauckas2020universal}. The loss function is modified to:
\begin{equation}
    \mathcal{L}(\mathbf{y}, \mathbf{\hat{y}}, \mathbf{F_a}) = \frac{1}{T}\sum_{t=1}^T(\mathbf{y} - \mathbf{\mathbf{\hat{y}}})^2 + \kappa || \mathbf{F_a} ||
    \label{eq:loss_norm}
\end{equation}
where \gls{kappa} is a weighting for the norm of the output of the NN, $\mathbf{F_a}$ is the trajectory of augmentation model outputs, $\mathbf{y}$ is the ground truth, and $\mathbf{\hat{y}}$ is the output. The larger the value of $\kappa$ the more emphasis the model will place on the physical component. However, this comes at the cost of flexibility, thus there is a trade-off and  $\kappa$ should defined by the modeller.
Figure~\ref{fig:SIR_NN_norm} shows the prediction of the model using different values of $\kappa$. Increasing $\kappa$ correspondingly decreases the effect of $\mathbf{F_a}$. The most accurate model has the lowest value of $\kappa$, but the SIR model's contribution is furthest from the ILI rate. 

\section{Forecasting Methods}
\label{sec:ODEmethods}
We propose and evaluate the performance of eight models for forecasting ILI in the United States. We compare our best-performing ODE models (\SEIRNN and \SIRFA --- described next) with the IRNN. We compare models with and without Web search activity data. All of our ODE models use a VAE framework with an encoder, latent ODE model and decoder. The encoder observes inputs such as ILI proportions and estimates latent initial conditions for the ODE model e.g., $s_0, i_0, r_0$. The ODE model integrates the initial conditions forwards in time, producing a latent trajectory, for example susceptible, infected and recovered fractions. The decoder estimates ILI proportions from the latent trajectories.
 
\subsection{Encoder Architectures}
\begin{figure}
    \centering
    \includegraphics[width=0.95\textwidth]{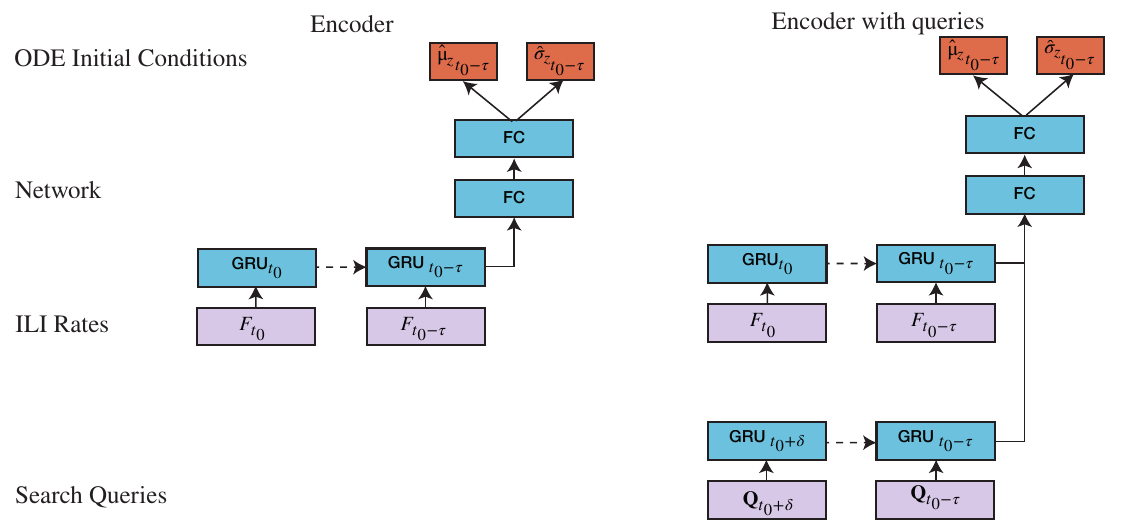}
    \caption[{Encoders for estimate initial conditions}]{\textbf{Encoders for estimate initial conditions} \newline 
    The encoders observe ILI proportions $\mathbf{y}_{t_0:t_0-\tau}$, backwards in time from $t_0$ to $t_0-\tau$ where $\tau$ is the window size, and $t_0$ is the time from which the forecasts are made. The ILI proportions are fed through a GRU layer, for the encoder with queries there is an additional GRU which observes query data $\mathbf{Q}_{t_0+\delta:t_0-\tau}$ for $t_0+\delta$ to $t_0-\tau$. The outputs from the GRUs are concatenated and fed into a fully connected dense layer. The dense layer feeds into the output layer which outputs means and standard deviations for the latent conditions $z_{t_0-\tau}$ at $t_0-\tau$.
    }
    \label{fig:Encoder_Diagram}
\end{figure}

We use two kinds of encoder, the first uses only ILI proportions, and the second uses ILI proportions and web search queries. A diagram of the encoder architectures is provided in Figure~\ref{fig:Encoder_Diagram}. In both models, a gated-recurrent-unit (GRU) sequentially observes ILI data $\mathbf{y}_{t_0:t_0-\tau}$ backwards in time from $t_0$ to $t_{0}-\tau$.  The output of the GRU at the $t_0-\tau$ is fed into a dense layer with a hyperbolic tangent (tanh) activation function which was chosen to be consistent with existing literature~\cite{chen2018neural}. The dense layer feeds into the output layer which has two outputs, one estimating the means and the other the standard deviations of the latent initial conditions. The key difference for the encoder using queries is a second GRU layer which observes $m$ web search frequencies $\mathbf{Q}_{t_0+\delta:t_0-\tau}$ from time $t_0+\delta$ to $t_{0}-\tau$. In the encoder with queries, the outputs of the two GRUs are concatenated into a single vector which is fed into the dense layer. The two models using web search data are \SIRQS and \FAQS.

We pre-train the encoder using only the KL divergence term of the ELBO:
\begin{equation}
    \mathcal{L}_{\text{KL}_z} = \text{D}_{\text{KL}}\left(q_{\bm{\Phi}_{enc}}(z_{t_0-\tau}| \mathbf{y}_{t_0:t_0-\tau})||p(z_{t_0-\tau})\right), 
\end{equation}
Or for the encoder using Web activity data:
\begin{equation}
    \mathcal{L}_{\text{KL}_z} = \text{D}_{\text{KL}}\left(q_{\bm{\Phi}_{enc}}(z_{t_0-\tau}| \mathbf{y}_{t_0:t_0-\tau}, \mathbf{Q}_{t_0+\delta:t_0-\tau})||p(z_{t_0-\tau})\right), 
\end{equation}
where $\bm{\Phi}_{enc}$ are the encoder parameters, $z_{t_0-\tau}$ are the latent initial conditions, and $p(z_{t_0-\tau})$ is a prior distribution over the latent variables which is specific to the ODE model. Pre-training using the KL divergence was found to be an easy way of forcing the encoder to produce reasonable outputs because it ensures that the model starts from a realistic condition.
This significantly reduces the training epochs and improves training stability.

\subsection{Latent ODE Architectures}
Here we describe five latent ODE architectures, we experiment with neural ODEs, mechanistic models and UDEs.

\paragraph{Neural ODE (\FA) \\}
We use a basic neural ODE as a baseline, denoted \FA. The neural network has three layers, the first two use eLu activations (Eq~\ref{eq:elu}) and $20$ units, and the output layer has no activation function and $8$ units to match the latent space. The network is defined as $f^{\bm{\Phi}}(\mathbf{x})$, where  $f^{\bm{\Phi}}$ denotes a neural network with parameters $\bm{\Phi}$.

The latent dimension is $8$, with a prior $p(z_{t_0-\tau}) = \mathcal{N}(0, 1)$.  The initial conditions distribution is sampled using the ``reparametrisation trick''~\cite{kingma2013auto} (discussed in Section~\ref{sec:IRNN_Change}) and then fed into the ODE model. The model is trained using the ELBO given by Eq~\ref{eq:ELBOchapter2}. We evaluate the neural ODE both with and without queries, denoted \gls{ODEB} and \gls{ODEBQ}, respectively.

\paragraph{Basic SIR model (\SIRSIMPLE)\\}
We evaluate an SIR model (\gls{SIRB}) as a baseline where the parameters $\beta$ and $\omega$ are learnt during training but then fixed i.e., they do not vary with time. \SIRSIMPLE~is the simplest physical model which we evaluate. By using a VAE architecture this model can produce different trajectories each season, however the model's flexibility is severely limited. 

The encoder approximates the latent initial conditions $s_{t_0-\tau}$ and $i_{t_0-\tau}$. The prior over the latent initial conditions is $p(z_{t_0-\tau}) = \mathcal{N}\left( {z_{\mu_{t_0-\tau}}}, I[0.1, 0.01] \right)$. The mean is the output of the encoder, and the standard deviation is $0.1$ for $s$ and $0.01$ for $i$. This is chosen because we found that regularising the standard deviations is important for performance, with the standard deviation of $i_t$ much smaller than $s_t$ resulting in the best performance. 
However, regularising the initial conditions to a preset distribution can negatively impact performance as the initial conditions vary significantly depending on when the forecast is made. For example, the susceptible population is much larger at the start of a flu season than at the end.

When sampling from $q(z_{t_0-\tau})$, we compute the absolute values of the samples to ensure non-negative fractions in each compartment. The recovered fraction is not needed to update the SIR equations but can be calculated by $r_t=1-s_t-i_t$.

\paragraph{Advanced SIR Model (\SIRNN)\\}
We use a more advanced SIR model, which uses the same SIR equations but $\beta$ and $\omega$ are estimated by a neural network as part of the ODE. We evaluate the model both with and without queries, denoted \gls{SIRadv} and \gls{SIRadvQ}, respectively.
The neural network output is $[\beta, \omega] = f^{\bm{\Phi}_\text{ode}}(x)$, where $\bm{\Phi}_\text{ode}$ are the network parameters. The inputs to the network are $\mathbf{x}_t = [s_t, i_t, z_{3t} ..., z_{8t}]$, the first $2$ are the susceptible and infected fractions, $z_{3t}$ to $z_{8t}$  are set by the encoder but are not updated by the ODE function. 
The SIR equations (Eq~\ref{eq:SIR}) compute $\frac{ds}{dt}$ and $\frac{di}{dt}$ the resulting vector comprises two values which are padded with zeros to align with the size of the latent variables, i.e., $[\frac{ds}{dt}, \frac{di}{dt}, 0, 0, 0, 0, 0, 0]$. 
The purpose of $z_{3t}$ to $z_{8t}$ is to provide additional information to the network which determines $\beta$ and $\omega$, for example, information about the virulence of the disease which may have been observed by the encoder, but there is otherwise no way of passing that information to the ODE. We ignore $\frac{dr}{dt}$ because the recovered fraction can be calculated after by $r_t=1-s_t-i_t$.

The network consists of three layers. The first two use the eLu activation function with $20$ units. The final layer has $8$ units an \texttt{abs} $(|x|)$ activation function to ensure $\beta$ and $\omega$ are positive. We regularise 
$f^{\bm{\Phi}_\text{ode}}$ to choose reasonable values for $\beta$ and $\omega$. 

Due to the samples from the initial conditions we have a time-varying distribution over $\beta$ and $\omega$, denoted  $q(\beta, \omega)$. We regularise the parameters to a prior  $p(\beta, \omega) = \mathcal{N}([0.8, 0.55], I[0.1, 0.1])$ using the KL divergence:
\begin{equation}
    \mathcal{L}_{\text{KL}_p} = \text{D}_{\text{KL}}(q(\beta, \omega) || p(\beta, \omega)),
\end{equation}
The prior is chosen based on the experiments discussed in Section \ref{sec:background}.

We regularise the latent trajectory to ensure that it stays between $0$ and $1$:
\begin{equation}
    \mathcal{L}_\text{reg}(z) = \sum_{t} \begin{cases} 
        |z_t| - 1 & \text{if } z_t > 1 \\
        |z_t| & \text{if } z_t < 0 \\
        0 & \text{otherwise} .
    \end{cases}
\end{equation}
This is only necessary when the initial conditions are poorly specified and $s_0+i_0 > 1$. We found that this addition to the loss function speeds up convergence and improves stability, but does not affect the model performance after training; for trained models $\mathcal{L}_\text{reg}(z)$ should always equal $0$. 

The full loss function for the physical model is therefore:
\begin{equation}
    \text{NLL}(\mathbf{y}, \mathbf{\hat{y}}, \mathbf{\hat{\mathbf{\sigma}}})+\mathcal{L}_{\text{KL}_z}+\mathcal{L}_{\text{KL}_p} + \mathcal{L}_\text{reg}(z)~,
    \label{eq:Fp_loss}
\end{equation}
this is the sum of the NLL and the three regularisation terms. For $s$ and $i$ the prior for the initial latent variables is the same as for \SIRSIMPLE, for the other values the prior is $\mathcal{N}(0,1)$. 

\paragraph{Advanced SEIR Model (\SEIRNN) \\}
We construct an SEIR model denoted \gls{SEIRadv}, which uses the same method as the \SIRNN model. The model uses a neural network to estimate the parameters of an SEIR model $[\beta, \omega, \rho] = f^{\bm{\Phi}_\text{ode}}(x)$. It uses an input vector $\mathbf{x}_t = [s_t, e_t, i_t, z_{4t} ..., z_{8t}]$ where the first 3 values correspond to $[s_t, e_t, i_t]$ and the next 5 are the variables set by the Encoder. The network is the same as in the \SIRNN but with $3$ units in the output. We again regularise the parameters $q_{\bm{\Phi}_\text{ode}}(\beta, \omega, \rho)$ to a prior distribution $p(\beta, \omega, \rho) = \mathcal{N}([2.0, 1.4, 0.2], I[0.1, 0.1, 0.2])$ using the KL divergence. We otherwise keep the same training setup as \SIRNN. We compute $r_t$ as $1-s_t-e_t-i_t$, as this enables us to sample from $z_t$ while maintaining the total population size as $1$.

\begin{figure}
    \centering
    \includegraphics[width=0.8\textwidth]{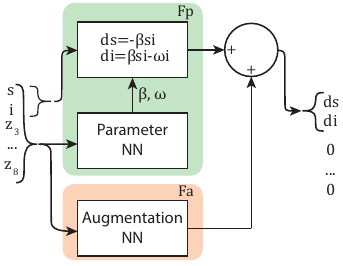} 
    \caption[{Diagram of \SIRFA ~architecture}]{\textbf{Diagram of \SIRFA ~architecture} \newline The inputs are $\mathbf{x}_t = [s, i, z_{3} ..., z_{8}]$ (we omit the subscript $_t$ denoting time.  
    The physical model $F_p$ in the green box contains the parameter NN, which estimates the parameters for $\beta$ and $\omega$ which are used by the SIR model. The augmentation NN $F_a$ minimises the error in the output of the model. The additional inputs $z_{3:8}$ provide information to the NNs and are outputs from the encoder which are not updated by the models.}
    \label{fig:UDE_Diagram}
\end{figure}

\paragraph{UDE Models (\SIRFA~and \SEIRFA) \\}
We use two UDE models \gls{SIRadvU} and \gls{SEIRadvU},  using \SIRNN~and \SEIRNN~as their physical models, respectively. These models use an additional neural network as an augmentation component to convert the models into UDEs. A diagram of the \SIRFA~model is provided in Figure \ref{fig:UDE_Diagram}, the physical model $F_p$ is \SIRNN, the the augmentation model $F_a$ reduces error in the output of the model. Both the parameter NN and the augmentation NN have three layers, the first two have 20 units and use eLu activation functions. The output layer of the parameter NN contains two units for the SIR model and three units for the SEIR model, both use an $\text{abs}$ activation function to ensure $\beta$, $\omega$ and $\rho$ are positive. The output of the augmentation NN uses a layer with preset weights following the method outlined in Section \ref{sec:maintaining_physical_assumptions_odes} to ensure that $\frac{ds}{dt}+\frac{di}{dt}+\frac{dr}{dt} = 0$ for \SIRFA~and $\frac{ds}{dt}+\frac{de}{dt}+\frac{di}{dt}+\frac{dr}{dt} = 0$ for \SEIRFA.

To train the model we use Eq~\ref{eq:Fp_loss} as a loss with but also regularise how much the model uses the augmentation NN by adding the norm of $F_a$ to the loss $\kappa || \mathbf{F_a} ||$, this prevents the model from relying on the augmentation component. The full loss function is
\begin{equation}
    \text{NLL}(\mathbf{y}, \mathbf{\hat{y}}, \mathbf{\hat{\mathbf{\sigma}}})+\mathcal{L}_{\text{KL}_z}+\mathcal{L}_{\text{KL}_p} + \mathcal{L}_\text{reg}(z) + \kappa || \mathbf{F_a} ||.
    \label{eq:FpFa_loss}
\end{equation}

\subsection{Decoder Architecture}
To decode the latent trajectories back into the same domain as $\mathbf{y}$ we use a neural network with a single layer. The decoder observes the latent trajectory and outputs forecasts. For the models which have mechanistic models in them, we only decode the latent variables corresponding with their respective compartmental models i.e., $[s, i, r]$ and $[s, e, I, r]$.

\section{Results}
\label{sec:oderesults}
We evaluate the eight models --- \FA, \FAQS, \SIRSIMPLE, \SIRNN, \SIRQS, \SIRFA, \SEIRNN, \SEIRFA --- by forecasting ILI at a national level in the US. We use the same test periods from Chapter~\ref{chapterlabel1} for the flu seasons in $2015/16$, $2016/17$, $2017/18$ and $2018/19$. We use weekly ILI proportions as inputs with a window size of 5 weeks, a forecast horizon of $\gamma=7$ to $\gamma=28$ days. Models which use search queries use daily query data from $t_0+14$ to $t_0-35$ days. We train the models for $2000$ epochs with a batch size of $16$. The learning rate starts at $0.001$ and decays by being multiplied by $0.999$ at the end of each epoch. We do not allow the learning rate to drop below $0.0001$.

We first compare the different neural-ODE architectures. Then we compare with the IRNN and IRNN$_0$ --- the IRNN which does not use Web search activity data.

\subsection{Forecasting performance of Neural ODEs}
We investigate the performance of the eight neural ODEs using four metrics --- Mean absolute error (MAE) and bivariate correlation ($r$) compare forecasts without considering the associated uncertainty. Negative log likelihood (NLL) and Skill weight the error by its corresponding uncertainty. When average metrics are calculated across several seasons or forecast horizons the arithmetic mean is used for all metrics besides Skill, where the geometric mean is used~\cite{reich2019collaborative}.

\begin{figure}
    \centering
    \includegraphics[width=1.0\textwidth]{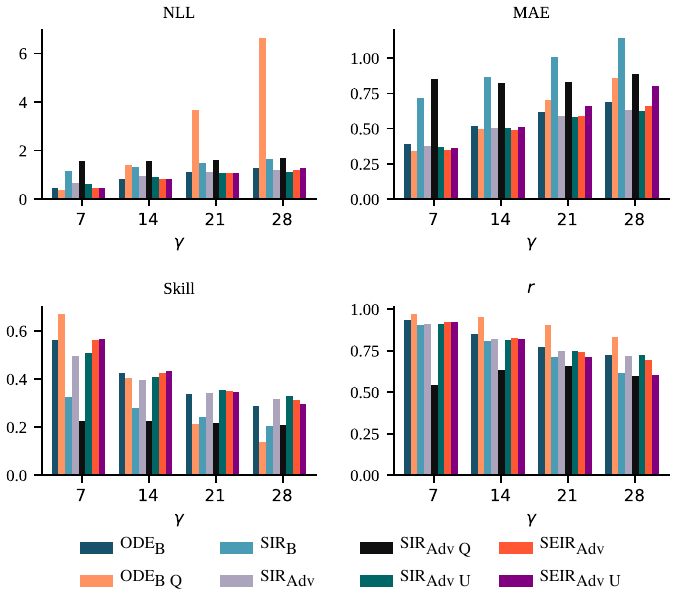} 
    \caption[{Metrics for each neural ODE model averaged over all four test flu seasons (2015/16 to 2018/19)}]{\textbf{Metrics for each neural ODE model averaged over all four test flu seasons (2015/16 to 2018/19)} 
    \newline 
    Scores for different forecast horizons ($\gamma$) are shown. Lower values for negative-log-likelihood (NLL) and mean absolute error (MAE) are better, and higher values for Skill and bivariate correlation $r$ are better.}
    \label{fig:avg_metrics_nodes}
\end{figure}

We enumerate the performance of the eight models in all performance metrics and flu seasons in Appendix Tables~\ref{tab:neuralodecomparison1517}, \ref{tab:neuralodecomparison1719} and \ref{tab:neuralodecomparison1519}. We provide a visual of the forecasting performance metrics of the different models when averaged over the four flu seasons in Figure~\ref{fig:avg_metrics_nodes}. 

\begin{figure*}[!h]
    \centering
    \includegraphics[width=0.98\linewidth]{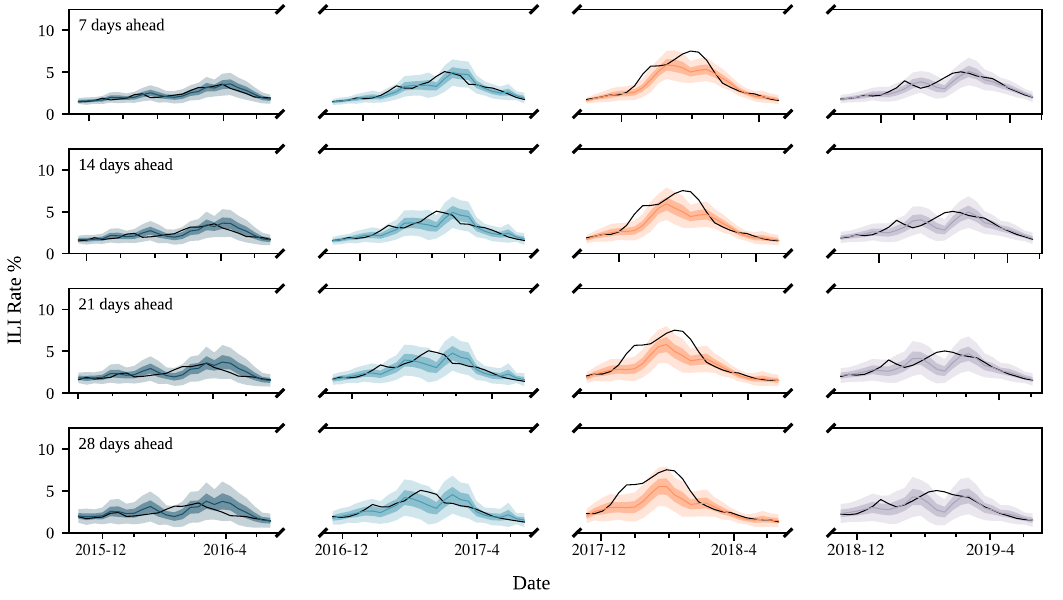}
    \caption[{\SIRFA~forecasts}]{\textbf{\SIRFA~forecasts}\newline
    \SIRFA~forecasts for all 4 test seasons ($2015/16$ to $2018/19$) and forecasting horizons ($\gamma = 7$, $14$, $21$, and $28$). Confidence intervals are shown at $50\%$ and $90\%$ levels and are visually distinguished by darker and lighter colour overlays respectively. The influenza-like illness (ILI)proportion (ground truth) is shown by the black line.
    }
    \label{fig:SIRFA_forecasts}
\end{figure*}

\begin{figure*}[!h]
    \centering
    \includegraphics[width=0.98\linewidth]{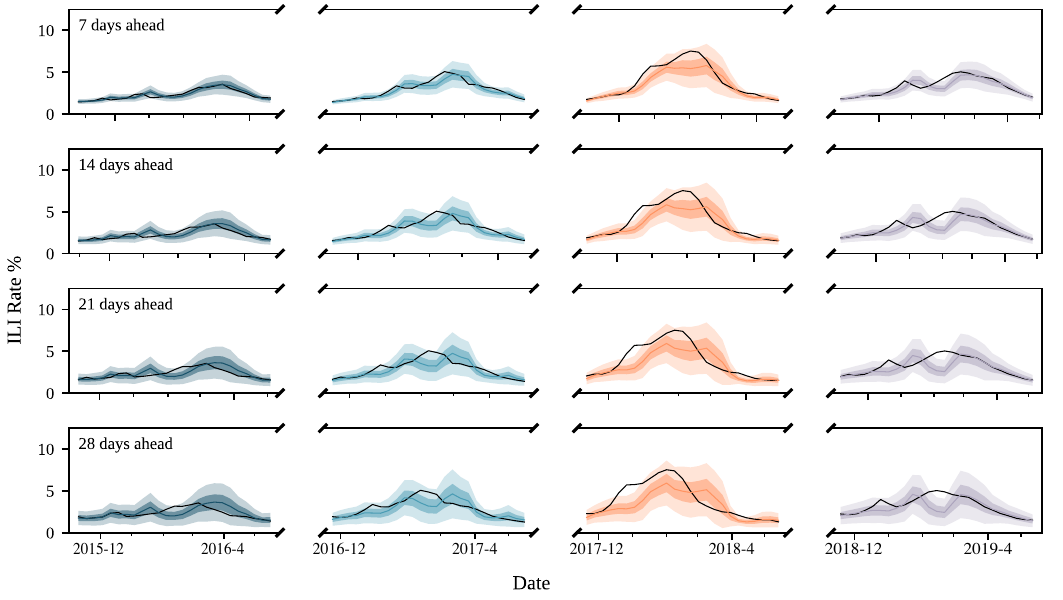}
    \caption[{\SEIRNN forecasts}]{\textbf{\SEIRNN forecasts}\newline
    \SEIRNN~forecasts for all 4 test seasons (2015/16 to 2018/19) and forecasting horizons ($\gamma = 7$, $14$, $21$, and $28$). Confidence intervals are shown at $50\%$ and $90\%$ levels and are visually distinguished by darker and lighter colour overlays respectively. The influenza-like illness (ILI) proportion (ground truth) is shown by the black line.
    }
    \label{fig:SEIRNN_forecasts}
\end{figure*}

\begin{figure*}[!h]
    \centering
    \includegraphics[width=0.98\linewidth]{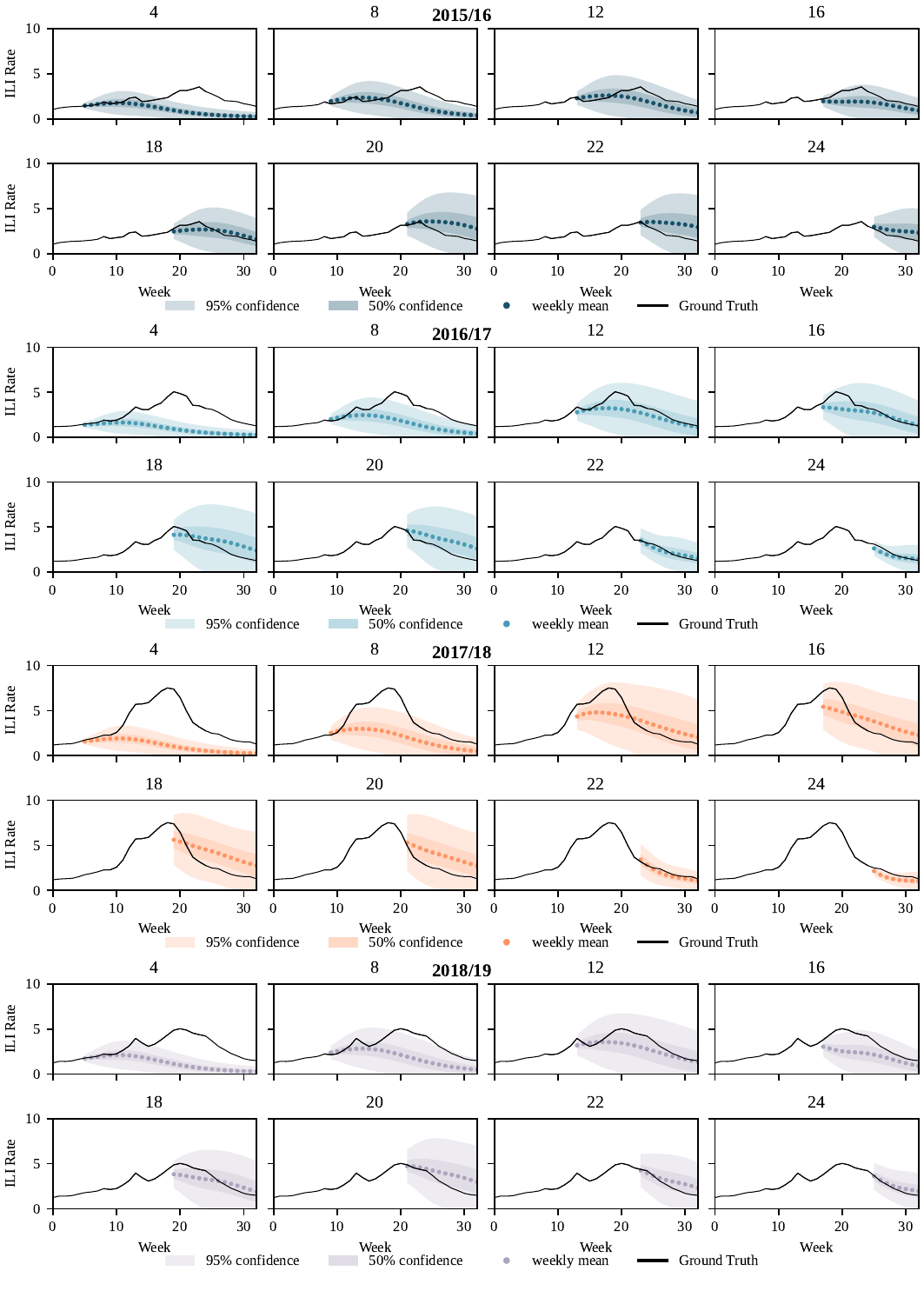}
    \caption[{\SEIRNN~trajectories}]{\textbf{\SEIRNN~trajectories}\newline
    Forecast trajectories for the \SEIRNN~from a given epidemic week (indexed from week $40$ in the year) to the end of the season. Each subplot shows the model's forecast from the given epidemic week (starting at week $40$ in the year). Trajectories are shown for the mean, $50\%$ and $90\%$ confidence intervals.
    }
    \label{fig:SEIRNN_trajectories}
\end{figure*}

\begin{figure*}[!h]
    \centering
    \includegraphics[width=0.98\linewidth]{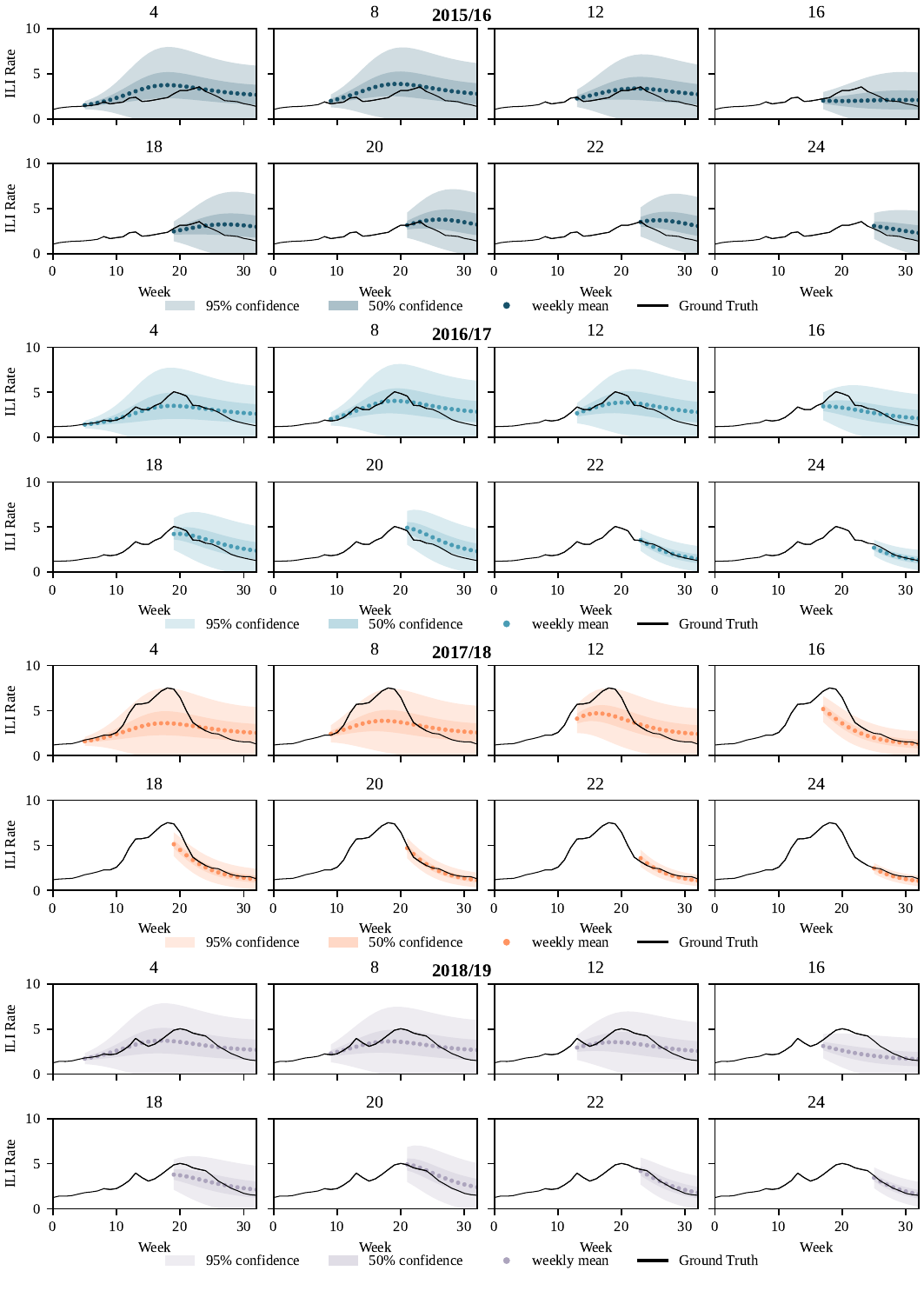}
    \caption[{\SIRFA trajectories}]{\textbf{\SIRFA trajectories}\newline
    Forecast trajectories for the \SIRFA~from a given epidemic week (indexed from week $40$ in the year) to the end of the season. Each subplot shows the model's forecast from the given epidemic week (starting at week $40$ in the year). Trajectories are shown for the mean, $50\%$ and $90\%$ confidence intervals.
    }
    \label{fig:SIRFA_trajectories}
\end{figure*}
To evaluate the forecasts in more detail we provide Figures~\ref{fig:SEIRNN_trajectories} and \ref{fig:SIRFA_trajectories}. These plots show how a forecast will develop from fixed points in the flu season. Corresponding figures for the other models provided in the appendix Figures~\ref{fig:FA_trajectories}, \ref{fig:FAQS_trajectories}, \ref{fig:SIRSIMPLE_trajectories}, \ref{fig:SIRNN_trajectories}, \ref{fig:SIRQS_trajectories}, and \ref{fig:SEIRFA_trajectories}. Each figure shows how the models forecast to the end of each flu season from 8 points throughout the season. 

% \clearpage
\begin{sidewaystable}
    \centering
    \small
    \setlength{\tabcolsep}{2pt}
    \begin{tabular}{p{0.8cm}p{0.8cm}p{0.8cm}p{0.8cm}p{0.8cm}p{0.8cm}p{0.8cm}p{0.8cm}p{0.8cm}p{0.8cm}p{0.8cm}p{0.8cm}p{0.8cm}p{0.8cm}p{0.8cm}p{0.8cm}p{0.8cm}p{0.8cm}p{0.8cm}p{0.8cm}p{0.8cm}p{0.8cm}}
& Metric  &\multicolumn{4}{c}{2015/16} & \multicolumn{4}{c}{2016/17} & \multicolumn{4}{c}{2017/18} & \multicolumn{4}{c}{2018/19} & \multicolumn{4}{c}{Avg (2015-19)} \\

\cmidrule(lr){3-6}\cmidrule(lr){7-10}\cmidrule(lr){11-14}\cmidrule(lr){15-18}\cmidrule(lr){19-22}
$\gamma$   &     &  \rotatebox{-45}{\IRNNQs} &  \rotatebox{-45}{\IRNN} &  \rotatebox{-45}{\SEIRNN} &  \rotatebox{-45}{\SIRFA} &  \rotatebox{-45}{\IRNNQs} &  \rotatebox{-45}{\IRNN} &  \rotatebox{-45}{\SEIRNN} &  \rotatebox{-45}{\SIRFA} &  \rotatebox{-45}{\IRNNQs} &  \rotatebox{-45}{\IRNN} &  \rotatebox{-45}{\SEIRNN} &  \rotatebox{-45}{\SIRFA} &  \rotatebox{-45}{\IRNNQs} &  \rotatebox{-45}{\IRNN} &  \rotatebox{-45}{\SEIRNN} &  \rotatebox{-45}{\SIRFA} &  \rotatebox{-45}{\IRNNQs} &  \rotatebox{-45}{\IRNN} &  \rotatebox{-45}{\SEIRNN} &  \rotatebox{-45}{\SIRFA} \\
\midrule

7       &  Skill &     0.56 &            0.74 &  \textbf{0.76} &       0.68 &            0.47 &            0.60 &  \textbf{0.61} &            0.56 &     0.34 &  \textbf{0.51} &        0.42 &            0.35 &     0.42 &  \textbf{0.61} &        0.52 &            0.51 &     0.44 &  \textbf{0.61} &        0.56 &        0.51 \\
     &    MAE &     0.26 &            0.25 &  \textbf{0.20} &       0.22 &            0.38 &            0.32 &  \textbf{0.31} &  \textbf{0.31} &     0.73 &  \textbf{0.36} &        0.58 &            0.60 &     0.39 &  \textbf{0.28} &        0.32 &            0.34 &     0.44 &  \textbf{0.30} &        0.35 &        0.37 \\
     &   $r$ &     0.86 &            0.86 &  \textbf{0.92} &       0.89 &            0.90 &  \textbf{0.93} &            0.92 &            0.91 &     0.92 &  \textbf{0.98} &        0.96 &            0.94 &     0.86 &  \textbf{0.93} &        0.89 &            0.90 &     0.89 &  \textbf{0.93} &        0.92 &        0.91 \\
\midrule
14      &  Skill &     0.50 &  \textbf{0.69} &            0.60 &       0.56 &            0.42 &  \textbf{0.53} &            0.47 &            0.46 &     0.26 &  \textbf{0.43} &        0.31 &            0.26 &     0.36 &  \textbf{0.58} &        0.38 &            0.41 &     0.37 &  \textbf{0.55} &        0.43 &        0.41 \\
     &    MAE &     0.35 &  \textbf{0.28} &            0.30 &       0.35 &            0.45 &  \textbf{0.38} &            0.45 &            0.44 &     0.94 &  \textbf{0.44} &        0.76 &            0.77 &     0.49 &  \textbf{0.31} &        0.45 &            0.46 &     0.56 &  \textbf{0.35} &        0.49 &        0.51 \\
     &   $r$ &     0.78 &            0.79 &  \textbf{0.82} &       0.77 &            0.83 &  \textbf{0.90} &            0.83 &            0.82 &     0.86 &  \textbf{0.97} &        0.90 &            0.90 &     0.76 &  \textbf{0.90} &        0.76 &            0.78 &     0.81 &  \textbf{0.89} &        0.83 &        0.82 \\
\midrule
21      &  Skill &     0.43 &  \textbf{0.60} &            0.49 &       0.47 &            0.38 &  \textbf{0.43} &            0.40 &            0.41 &     0.20 &  \textbf{0.37} &        0.24 &            0.23 &     0.32 &  \textbf{0.51} &        0.31 &            0.35 &     0.32 &  \textbf{0.47} &        0.35 &        0.35 \\
     &    MAE &     0.45 &  \textbf{0.35} &            0.37 &       0.41 &  \textbf{0.50} &  \textbf{0.50} &            0.54 &            0.51 &     1.09 &  \textbf{0.64} &        0.92 &            0.91 &     0.56 &  \textbf{0.42} &        0.52 &            0.50 &     0.65 &  \textbf{0.48} &        0.59 &        0.58 \\
     &   $r$ &     0.66 &  \textbf{0.71} &            0.70 &       0.64 &            0.78 &  \textbf{0.80} &            0.76 &            0.76 &     0.80 &  \textbf{0.92} &        0.85 &            0.89 &     0.69 &  \textbf{0.82} &        0.66 &            0.71 &     0.74 &  \textbf{0.81} &        0.74 &        0.75 \\
\midrule
28      &  Skill &     0.35 &  \textbf{0.49} &            0.42 &       0.40 &            0.35 &  \textbf{0.39} &            0.37 &  \textbf{0.39} &     0.16 &            0.21 &        0.21 &  \textbf{0.23} &     0.30 &  \textbf{0.48} &        0.29 &            0.33 &     0.28 &  \textbf{0.37} &        0.31 &        0.33 \\
     &    MAE &     0.68 &            0.51 &  \textbf{0.43} &       0.51 &            0.56 &            0.57 &            0.56 &  \textbf{0.51} &     1.25 &  \textbf{0.86} &        1.09 &            0.99 &     0.65 &            0.50 &        0.55 &  \textbf{0.48} &     0.78 &  \textbf{0.61} &        0.66 &        0.62 \\
     &   $r$ &     0.54 &  \textbf{0.69} &            0.56 &       0.50 &  \textbf{0.78} &            0.74 &            0.77 &  \textbf{0.78} &     0.78 &            0.90 &        0.80 &  \textbf{0.91} &     0.66 &  \textbf{0.79} &        0.65 &            0.71 &     0.69 &  \textbf{0.78} &        0.70 &        0.73 \\
\bottomrule
\end{tabular}

    \caption[{Neural ODE comparison with IRNN}]{\textbf{Neural ODE comparison with IRNN} \newline Performance metrics for the two best performing N-ODE models \SEIRNN~and \SIRFA~for four forecast horizons ($\gamma = 7$, $14$, $21$, and $28$ days ahead) compared with the IRNN and IRNN$_0$ --- the IRNN but without using web search queries. Skill compares the accuracy weighted by the uncertainty of forecasts. MAE is the mean absolute error, and $r$ is the bivariate correlation between forecasts and reported ILI proportions. The best results for each metric and forecast horizon are shown in bold. The last three columns are performances averaged over the four test flu seasons (from 2015/16 to 2018/19).}
    \label{tab:NODE_comparison2}
\end{sidewaystable}

\begin{figure*}[!h]
    \centering
    \includegraphics[width=0.98\linewidth]{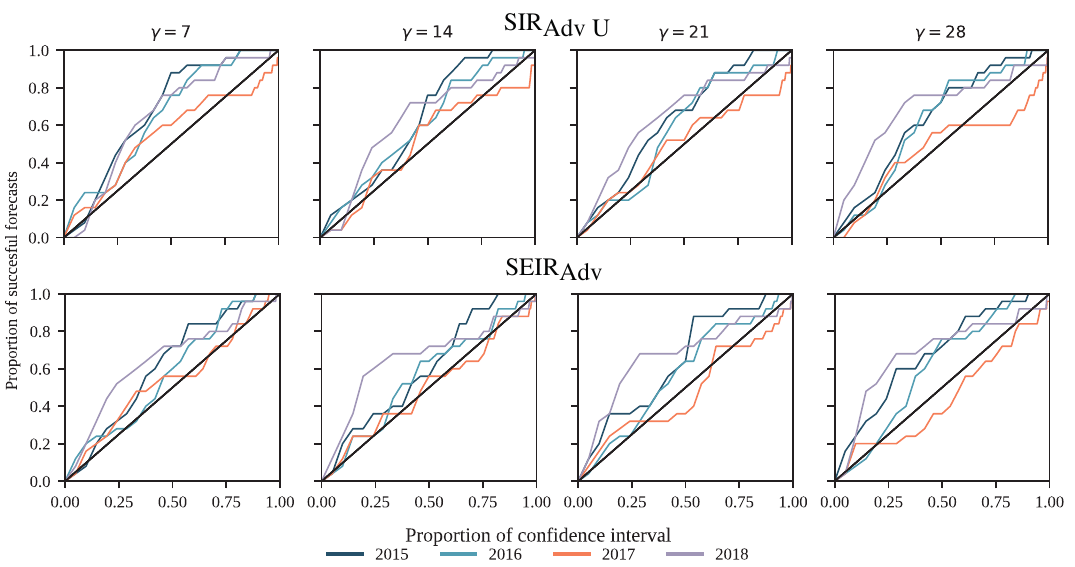}
    \caption[{Calibration plots for Neural ODEs}]{\textbf{Calibration plots for Neural ODEs}\newline
    Calibration is shown for \SIRFA~and \SEIRNN~for each of the four test periods (2015/16 to 2018/19) and forecasting horizons ($\gamma$). The lines show how frequently the ground truth falls within a confidence interval (CI) of the same level. To be more precise, a point $(x,y)$ denotes that the proportion $y \in [0, 1]$ of the forecasts when combined with a CI at the $x \times 100\%$ level includes the ground truth (successful forecasts). The optimal calibration is shown by the diagonal black line. Points above or below the diagonal indicate an over- or under-estimation of uncertainty, and hence an under- or over-confident model, respectively.
    }
    \label{fig:main_ode_calibration}
\end{figure*}

We focus our evaluation on the two best-performing models: \SIRFA~and \SEIRNN, although metrics and figures for every model are provided in the Appendix. Forecasts from \SEIRNN~and \SIRFA~ for every season and forecast horizon are shown in Figures~\ref{fig:SIRFA_forecasts} and \ref{fig:SEIRNN_forecasts} respectively. Forecasts for the other models are provided in the appendix Figures~\ref{fig:FA_forecasts}, \ref{fig:FAQS_forecasts}, \ref{fig:SIRSIMPLE_forecasts}, \ref{fig:SIRNN_forecasts}, \ref{fig:SIRQS_forecasts}, and \ref{fig:SEIRFA_forecasts}. There is a visible decline in the forecasting performance as the forecast horizon increases, and a corresponding increase in the size of the confidence intervals. We expect from the previous experiments that the \SEIRFA~would be able to produce more complex forecast trajectories. In practice, we do not see this in the \SEIRFA forecasts. 

Calibration plots from \SEIRNN~and \SIRFA~ for every season and forecast horizon are shown in Figures~\ref{fig:main_ode_calibration}. Calibration plots for the other models are provided in the appendix Figures~\ref{fig:sup_ode_calibration}. The lines show how frequently the ground truth falls within a confidence interval (CI) of the same level. The optimal calibration is shown by the diagonal black line. Points above or below the diagonal indicate an over- or under-estimation of uncertainty, and hence an under- or over-confident model, respectively.

\subsection{Comparison with IRNN}
Table~\ref{tab:NODE_comparison2} enumerates the performance of the best performing \SEIRNN~and \SIRFA~ in comparison to the IRNN. We also include IRNN$_0$, which is the same IRNN architecture but trained without web-search data. This has the same ILI input and targets as the neural ODE models.

\section{Discussion and Conclusions}
\label{sec:odediscussion}
We have demonstrated that the ODE forecasting models are able to produce reasonable forecasts for up to four weeks ahead. The VAE architecture successfully observes ILI proportions, estimates latent initial conditions of an ODE model, integrates them forwards using an ODE, and then decodes the latent trajectories back into the same domain as the ILI proportion. The \FA~model performs well and is outperforms the basic \SIRSIMPLE~ and \SIRNN~models. Thus demonstrating the benefits of using more flexible models for forecasting the complex ILI signal. 

The compartmental models introduce physical constraints to the modelling. The trajectory plots (Figure~\ref{fig:SIRFA_trajectories} and \ref{fig:SEIRNN_trajectories} show that the models produce forecasts with one peak following a smooth trajectory, which gradually decays to zero at the end of the season. At the peak of the season, the models are far less confident than at the start and end of the season. When the models forecast from before the peak of the season the forecasts have one peak with more uncertainty around the peak. The \SIRFA~ makes forecasts which tend to tail off towards the end of the season. This trend is not as defined as for the \SEIRNN~model, highlighting that the augmentation component of the model can significantly change the forecasts and cause them to stop following the trends of physical models. 

The SIR models generally produce better forecasts when their constraints are relaxed. This is evident from the NLL averaged over all seasons and horizons; the \SIRSIMPLE~scores $1.40$, \SIRNN~scores $0.98$, and \SIRFA~scores $0.93$. It follows that introducing a more complex physical model would improve the forecasts further. The \SEIRNN improves on the most complex SIR model with an NLL of $0.90$, however, the augmentation component in \SEIRFA~ does not improve the forecasting performance. This indicates that the performance of the more complex models has asymptoted. 

Table \ref{tab:node_average_metrics} enumerates the average metrics for \FA, \SIRNN, \SIRFA, \SEIRNN, \SEIRFA. The five models perform similarly --- the standard deviation for the Skill, NLL, MAE and $r$ are all under $0.03$. The similar performance of the latent ODE models suggests that the limiting factor to forecast performance comes from another area of the VAE architecture.
\begin{table}[h!]
    \centering
    \begin{tabular}{c|ccccc|cc}
     & \textbf{\FA} & \textbf{\SIRNN} & \textbf{\SIRFA} & \textbf{\SEIRNN} & \textbf{\SEIRFA}  &\textbf{IRNN}& \textbf{IRNN$_0$}\\
    \hline
    Skill & 0.39 & 0.38 & \textbf{0.40}& \textbf{0.40}& \textbf{0.40} &0.49& 0.35\\
    NLL   & 0.92 & 0.98 & 0.93 & \textbf{0.90}& 0.92  & 0.71& 1.05\\
    MAE   & 0.55 & 0.53 & \textbf{0.52}& \textbf{0.52}& 0.58  & 0.44& 0.61\\
    $r$  & \textbf{0.82}& 0.80 & 0.80 & 0.80 & 0.77  & 0.85& 0.78\\
    \end{tabular}
    \caption{Metrics averaged for all seasons and forecast horizons for similarly performing neural ODEs. IRNN and IRNN$_0$ i.e., the best performing neural network from Chapter~\ref{chapterlabel1}, both with and without queries are shown. The best neural ODE results are shown in bold.}
    \label{tab:node_average_metrics}
\end{table}

% The decoder was developed specifically to force the model to rely on the physical components of the ODE models. It is a very simple NN which does little more than rescaling the latent trajectory, increasing its flexibility by adding more layers or inputting data directly from the encoder may improve the accuracy but would also limit the impact of the physical constraints of the ODE.
The encoder sets the initial conditions and provides information on the observed trajectory to the ODE model. We can see that for the most part, the encoder works well, but there are situations where its performance is quite poor. However, as the encoder observes data backwards in time and estimates the latent conditions at $t_0-\tau$, a problem $\tau$ weeks ago can affect the performance at $t_0$. 

This occasionally causes poor short-term forecasts as seen in Figures \ref{fig:SEIRNN_trajectories} and \ref{fig:SIRFA_trajectories} where the performance in weeks $12$ and $16$ is significantly lower. This is around the time of Christmas when there is a seasonal rise and then a dip in the ILI proportion\cite{osthus2019dynamic}, instead of predicting the rise and fall around Christmas, the models extrapolate the first rise, and when they observe the dip they treat it as the peak of the season. This is caused by poor initial conditions and is therefore an issue with the encoder. However, modifications to the encoder estimating the initial conditions at $t_0$ did not yield good results, so improvements must be found somewhere else.

An obvious improvement for the encoder is to use Web search activity data. However, attempts to incorporate them were largely unsuccessful. The \FAQS~overfits and fails to generalise to unseen data. The average NLL for \FAQS~was $3.02$ whereas the best models all have NLLs below $1.0$. Similarly, the \SIRQS~is the worst performing physical model which we evaluate, with an average NLL of $1.62$. The \FAQS~model works best seven days ahead for all metrics, but its performance drops off for longer forecast horizons. We experimented with changing the latent dimension, tuning the number of layers, size of layers and prior for the encoder. However, we found that results after tuning tended to either be very similar to the results we have presented, or significantly worse, with very little in between. If the encoder was able to capture more information from the data then the latent dimension could potentially be increased, which would then give the ODE more information to estimate initial conditions.

Comparing models which do not use  Web search activity data, the ODEs outperform the IRNN for almost all horizons and metrics. The only scenario where the IRNN$_0$ is better is in terms of MAE and $r$ in $2016/17$ for $21$ days ahead. 
The average skill for the IRNN$_0$~is $0.35$, and the average skill for the \SEIRNN~and \SIRFA is $0.40$ and $0.39$, respectively. However, the IRNN is successful at incorporating Web search activity data to improve its forecasts, improving its Skill from $0.35$ to $0.49$. When comparing the ODE models to the IRNN with queries, the IRNN is easily the best model with the best metrics in almost all seasons and horizons. 

One of the anticipated advantages of mechanistic models is that they can model novel scenarios~\cite{baker2018mechanistic, maino2016mechanistic}. We observe this in the $2017/18$ season, where the ODEs match or outperform the IRNN for $\gamma=28$ days ahead. The \SIRFA~ has a Skill of $0.23$, while the IRNN and \SEIRNN~models both score $0.21$. The calibration plots in Figure\ref{fig:main_ode_calibration} show that the calibration is similar for all seasons, albeit generally underconfident. However, in $2017/18$ the calibration for both models is close to the optimum calibration for both the \SEIRNN~and the \SIRFA.  This is promising as we previously found that the IRNN and IRNN$_s$ had poor calibration in $2017/18$ due to the novelty of the season. The calibration plots for the other ODE models (Figure~\ref{fig:sup_ode_calibration}) show that all the ODEs follow a similar trend apart from those which use queries --- which have not been implemented successfully. 

The neural ODEs are effectively combining data and model uncertainty in a single mechanism. This is because the latent variables are both a model parameter and an output. Changing the latent ODE model to use being fully Bayesian would potentially improve the uncertainty estimates by allowing model uncertainty to exist at different points in the model. However, this modification goes beyond the scope of this thesis, as it would increase complexity both in implementation and training, as well as increase computational overhead by introducing additional points of sampling into the model. 

Our neural ODE models work as a proof of concept, but for them to work as well as neural networks further work is needed. The results show that in some cases the neural ODEs match the IRNN. The $2017/18$ season was by far the most challenging season for the IRNN and Dante, but the neural ODEs were able to outperform the best-performing IRNN for $28$ days ahead in terms of skill, with comparable bivariate correlation. This is despite the IRNN using web-search data. 

We found that training UDEs is a difficult balance of the different components of the loss function. Adding the regularisation to prevent the model from choosing erroneous physical model parameters significantly improved the reliability of training, but in some instances, it would be necessary to turn down the regularisation during training. We tried setting the parameter regularisation $0$ after $500$ training epochs, in some cases this improved performance, but in others would result in exploding/vanishing gradients and a correspondingly unstable model.
Developing the training procedure further could improve the models and enable easier hyper-parameter tuning. However, the main performance bottleneck is the encoder's inability to use Web search activity data to improve accuracy. Potential modifications include: changing the regularisation, more careful specification of the prior, or forecasting the web-search queries as well as the ILI proportion. Forecasting queries would increase the complexity of the model but would force the latent dimension to express more information about the queries, thus limiting its ability to overfit. 
Forecasting search queries with a VAE would provide a low-dimensional representation of the search queries which could inform design decisions for future architectures.

If the encoder performance bottleneck can be overcome then more complex physical models could be used. More advanced physical models could include richer compartmentalisation of the population or simultaneous modelling of multiple geographic regions~\cite{osthus2021multiscale,osthus2019dynamic, chang2021mobility} would both allow more accurate forecasts and closer targeting of public health interventions.

The neural ODE framework presented in this chapter provides a framework for forecasting which combines neural networks and physical models. The method is not without its faults, and we were unable to improve upon the neural networks presented in Chapter \ref{chapterlabel1}. However, the models behave according to the physical properties of the compartmental models and there are clear avenues for future research efforts to further improve forecasting performance.
\chapter{Conclusions}
\label{chapterlabel4}
In this work, we have shown how neural networks can be used to forecast infectious disease prevalence. Existing work using neural networks has been limited by a lack of understanding of uncertainty, and limited testing over multiple seasons and forecast horizons. 

In Chapter~\ref{chapterlabel1} we demonstrated the ability of neural networks to forecast ILI rates by incorporating exogenous Web search activity data while providing uncertainty estimates. The Iterative-Recurrent-Neural-Network (IRNN) exhibits superior performance (averaged over all test years) for forecast horizons greater than 7 days, whereas SRNN is superior for the $\gamma = 7$ days ahead forecast horizon. We also demonstrated that the proposed forecasting framework can provide very competitive performance that is better than the established state-of-the-art in ILI rate forecasting.

We found that including Web search activity data significantly improved forecast performance, with or without a temporal advantage --- caused by ILI rates being delayed due to collection whereas Web search activity data can be collected immediately. This is consistent with previous literature however, our experiments are the most comprehensive analysis to date, assessing performance over 4 consecutive flu seasons, and utilising an open-ended, non-manually curated set of search queries. 
We have also cross-examined accuracy with a number of different error metrics, including CRPS and NLL that can incorporate the validity of uncertainty estimates. We have seen that adding Web search information not only improves accuracy but also provides better estimates of confidence.

Existing disease forecasting frameworks are difficult to scale, and incorporating additional features or more training data can result in excessive computational cost. An advantage of neural networks is that they are easy to scale; increasing the amount of training instances often results in better overall performance~\cite{alom2019state}. Overfitting issues, which become more apparent when working with relatively small data sets, are alleviated to an extent by the deployment of a Bayesian neural network which averages over parameter values instead of making single point estimates~\cite{hernandez2015probabilistic}.

From a methodological perspective, we found that our approach in estimating uncertainty can be improved ---- IRNN, the best-performing NN, was not explicitly aware of the actual forecasting horizon which resulted in the uncertainty not increasing with the forecast horizon. 
With this in mind, we developed IRNN$_s$, which changed the IRNN to a fully Bayesian neural network, the sampling was changed from once per time step to once for all horizons, and we also trained the model using multiple samples, estimating the combined uncertainty during training. The IRNN$_s$ improved the calibration over all seasons by $17\%$ compared with the IRNN. The model maintained the same Skill --- a probabilistic forecasting metric, but with a worse mean-absolute-error. Further tuning may alleviate the issues with accuracy which should be the focus of future work, alongside evaluating the model on more recent flu seasons, especially post-Covid. 

Despite this, the presented methodology improves on the uncertainty estimation of the IRNN without significantly changing the underlying approach. The architectures are applicable to different data sources, diseases, and potentially even different forecasting problems altogether.

In Chapter~\ref{chapterlabel2} we developed a framework for combining neural networks with existing mechanistic models using variational auto-encoders. However, we were unable to successfully use the neural ODEs with web search data. Consequently, the neural ODEs were not competitive with the neural networks developed in Chapter \ref{chapterlabel1}. However, in the absence of Web search activity data the neural ODEs were competitive with the neural networks. We also found that neural ODEs were better able to forecast unusual seasons, notably 2017/18, and had well-calibrated uncertainty even for longer forecast horizons where the IRNN performed especially poorly. The neural ODEs also outperformed the IRNN without Web search activity data by $14\%$ in terms of Skill. 

We previously highlighted the importance of including Web search activity for forecasting ILI rates, by incorporating more information such into the inputs for the model's Encoder the accuracy could be improved. Future research efforts should be focused on this integration of search data and neural ODEs. If the encoder was more capable then this would open up the possibility of developing more complex physical models and richer representations of disease spread in a population. 

The potential implications of this work are significant. Infectious diseases present a significant burden on society, and forecasting models are a critical tool in being able to reduce their impact. As illustrated in Chapters \ref{chapterlabel1} and \ref{chapterlabel2}, the Bayesian Neural Networks and neural ODEs, with their respective strengths and limitations are promising avenues towards achieving more precise, calibrated, and timely epidemic forecasts.

\addcontentsline{toc}{chapter}{Appendices}

\appendix
\chapter{UCL Research Paper Declaration Form: referencing the doctoral candidate’s own published work(s)}
	
	\begin{enumerate}\itemsep0em
		\item \textbf{1.	For a research manuscript that has already been published} (if not yet published, please skip to section 2)\textbf{:}
		\begin{enumerate}\itemsep0em
			\item \textbf{What is the title of the manuscript?}
			    Neural network models for influenza forecasting with associated uncertainty using Web search activity trends
			\item \textbf{Please include a link to or doi for the work:}
			        \url{https://journals.plos.org/ploscompbiol/article?id=10.1371/journal.pcbi.1011392}
			\item \textbf{Where was the work published?}
			        PLOS Computational Biology
			\item \textbf{Who published the work?}
			        Public Library of Science
			\item \textbf{When was the work published?}
			        28/8/2023
			\item \textbf{List the manuscript's authors in the order they appear on the publication:}
			        Michael Morris, Peter Hayes, Ingemar J. Cox, Vasileios Lampos 
			\item \textbf{Was the work peer reviewed?}
			        Yes
			\item \textbf{Have you retained the copyright?}
			        No
			\item \textbf{Was an earlier form of the manuscript uploaded to a preprint server (e.g. medRxiv)? If ‘Yes’, please give a link or doi} 
	    	  Yes: \url{https://arxiv.org/abs/2105.12433}
		\end{enumerate}
		
		\item \textbf{For multi-authored work, please give a statement of contribution covering all authors} (if single-author, please skip to section 3)\textbf{:}
		\begin{itemize}
		    \item Michael Morris: Conceptualization, Data curation, Formal analysis, Investigation, Methodology, Software, Visualization, Writing – original draft, Writing – review \& editing
                \item Peter Hayes: Conceptualization, Methodology
                \item Ingemar J. Cox: Conceptualization, Formal analysis, Supervision, Writing – original draft, Writing – review \& editing
                \item Vasileios Lampos: Conceptualization, Formal analysis, Methodology, Supervision, Validation, Visualization, Writing – original draft, Writing – review \& editing
		\end{itemize}
		\item \textbf{In which chapter(s) of your thesis can this material be found?}
            Chapter~\ref{chapterlabel1}
	\end{enumerate}

		\textbf{e-Signatures confirming that the information above is accurate}
		(this form should be co-signed by the supervisor/ senior author unless this is not appropriate, e.g. if the paper was a single-author work)\textbf{:}\\

	    % \textbf{Candidate:} \includegraphics{Figures/MM_Signature.pdf}\\
		% \textbf{Date:} 21/11/2023
        % \includegraphics[width=0.9\linewidth]{Figures/IJC_Signature.png}

\chapter{Supplementary Information for Chapter \ref{chapterlabel1}}
\label{appendixlabel1}

\section{Supplementary Methods}

\paragraph{Multibin logarithm score and forecast Skill score} The CDC use forecast Skill as a metric to compare forecasting models. For a forecast estimate $\hat{y} \in [0,1]$, they define an `accuracy of practical significance' as being within $\pm0.5\%$ of the correct ILI rate $y \in [0,1]$. The sum of the probability assigned to this region defines the Skill which is given by:
\begin{equation}
    \text{Skill}\left(\hat{y},y\right) = \left(\sum_{i=-5}^5p(y+i\times0.01|\hat{y})\right) \, ,
\end{equation}
where $p(y+i|\hat{y})$ is the probability assigned to a bin of size $0.1$ around the true ILI rate $y$. To compute the Skill score for a normal distribution $\hat{\mathcal{N}} = \mathcal{N}(\hat{y}, \hat{\sigma})$, we first obtain the lower value of the correct ILI bin, i.e. $y_b = 0.1 \times \text{floor}(y \times 10)$, and then use the cumulative density function (\texttt{cdf}) of $\hat{\mathcal{N}}$ to compute:
\begin{equation}
  \text{Skill}(\texttt{cdf}, y_b) = (\texttt{cdf}(y_b+0.6) - \texttt{cdf}(y_b-0.5)) \, . 
\end{equation}

\begin{figure}[H]
    \centering
    \includegraphics[width=0.85\linewidth]{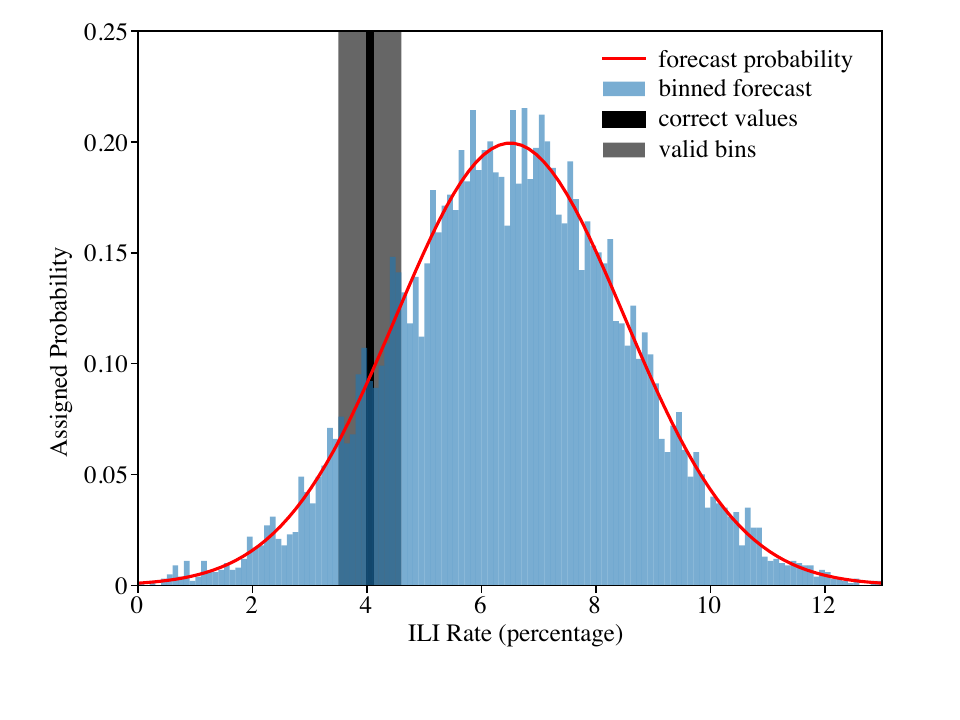}
    \caption[{Illustration of a binned probabilistic forecast}]{\textbf{Illustration of a binned probabilistic forecast} \newline Illustration of a binned probabilistic forecast showing the correct bin and area of practical significance.}
    \label{fig:sup_binned_forecast}
\end{figure}

\noindent An example of a forecast and the binned ILI rate is given in Figure~\ref{fig:sup_binned_forecast}. The true ILI rate $y$ is equal to $4.03\%$, so the correct bin is between $4.0\%$ and $4.1\%$. The area within $\pm0.5$ percentage points of the correct bin is considered an `accuracy of practical significance'~\cite{reich2019collaborative}. That is, from the bin between $3.5$ and $3.6$ up to $4.5$ and $4.6$. The probabilities assigned to these bins are: $0.0047$, $0.0046$, $0.0076$, $0.0080$, $0.01074$, $0.0051$, $0.0092$, $0.0033$, $0.0081$, and $0.0060$. The Skill score is simply the sum of these probabilities, i.e. $0.0673$. When calculating the Skill score for several forecasts the geometric average is taken. For example, if for weeks 1 to 5 of a flu season we have forecast Skills of $0.53$, $0.61$, $0.40$, $0.45$, the skill for the four week period is equal to
\begin{equation}
  \sqrt[4]{0.53 \times 0.61 \times 0.40 \times 0.45} = 0.49 \, .
\end{equation}
When computing the average Skill for multiple seasons or forecast horizons the same geometric average is used.

The Skill score is not a strictly proper metric~\cite{gneiting2007strictly}, i.e. it does not have a single unique ideal solution. This has been a source of criticism~\cite{bracher2019multibin}. For example, with mean absolute error (MAE) there is only one estimate which will result in a $0$ score. With forecast Skill, any forecast which places 100\% probability within the area of practical significance will achieve a Skill score of 1.

\paragraph{Continuous ranked probability score} Contrary to forecast Skill, both negative log likelihood (NLL) and continuous ranked probability score (CRPS) are strictly proper~\cite{gneiting2007strictly}. A criticism of NLL is that it over-penalises errors where the difference between the actual and forecasted value is much greater than the associated uncertainty~\cite{gneiting2007strictly}. CRPS is more forgiving. It is defined by 
\begin{equation}
        \text{CRPS}(\mathbf{y}, \mathbf{\hat{y}},\mathbf{\hat{\sigma}}) = \frac{1}{T} \sum^T_{t=1}\hat{\sigma}  \Bigg[ \frac{1}{\sqrt{\pi}}-2\varphi_t  \left(\frac{y_t-\hat{y}_t}{\hat{\sigma}_t} \right)
        - \frac{y_t-\hat{y}_t}{\hat{\sigma}_t} \left( 2\Psi_t \left(\frac{y_t-\hat{y}_t}{\hat{\sigma}_t} \right) -1\right)\Bigg] \, ,
\end{equation}
where \(\varphi_t\) and \(\Psi_t\) respectively denote the probability density function and the cumulative distribution function of a standard Gaussian variable $\mathcal{N}(\hat{y}_t,\hat{\sigma}_t)$. CRPS is a probabilistic metric that generalises to MAE when the standard deviation is \(0\).

\begin{figure}[t]
    \centering
    \includegraphics[width=0.85\linewidth]{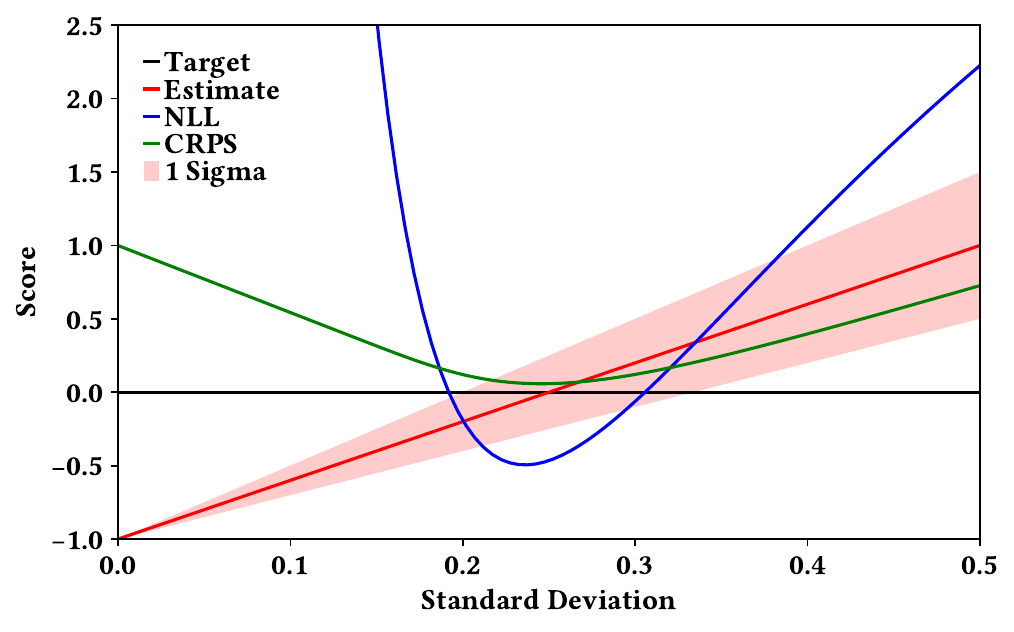}
    \caption[{NLL and CRPS variation}]{\textbf{NLL and CRPS variation}\newline NLL and CRPS variation with changing error and confidence. The red line shows a model's estimate with standard deviation shown in pink where the true value is $y=0$. As the accuracy and confidence change, the CRPS and NLL values have different trajectories.}
    \label{fig:sup_NLL_crps}
\end{figure}

Both CRPS and NLL favour a confident and accurate forecast. CRPS, however, is more forgiving when the confidence is high and the accuracy is poor. Figure~\ref{fig:sup_NLL_crps} illustrates this point. Here the blue and green curves depict the NLL and CRPS scores, respectively. Estimates are represented by the red diagonal line. The true value to be predicted is $y=0$. The first point on the diagonal line has zero standard deviation, but predicts $y=-1$, i.e. an erroneous value with perfect confidence (zero uncertainty). The CRPS penalises this with a score of $1$ (which in this case is equal to the MAE). In contrast, the NLL tends to infinity. As we move from left to right, the error in $y$ is initially decreasing while our uncertainty is increasing. As our estimate approaches the true value of $y=0$ which occurs when the standard deviation (x-axis) is $\approx 0.34$, both curves approach a minimum value. We would like to note that the minimum value of NLL is closer to when $y = 0$, but the minimum of CRPS happens prior to that (i.e. around point $0.3$ on the x-axis). As we continue to move from left to right, the error in $y$ increases along with the uncertainty. At the right-most side, we have $y=0.5$ with a standard deviation of $0.5$. Here, the CRPS score is approximately $0.2$, and the NLL $0.7$. Overall, the NLL metric much more strongly penalises errors that are outside of the uncertainty region. We note this, but do not favour one metric over the other, reporting both in our results.

\paragraph{Mean absolute error and bivariate correlation} We report the mean absolute error (MAE) and bivariate correlation ($r$). We measure MAE and $r$ between the means of the forecasted estimates ($\hat{\mathbf{y}} \in \mathbb{R}_{[0,1]}^T$) and the ground truth ILI rates ($\mathbf{y} \in \mathbb{R}_{[0,1]}^T$) during a flu season. The MAE evaluates how close the forecasted values are to the true values 
\begin{equation}
    \text{MAE} = \frac{1}{T}\sum_{t=1}^T{\lvert\hat{y}_t - y_t\rvert} \, .
\end{equation}
The bivariate correlation $r$ evaluates how similar the shape of the forecasted flu season is to the ground truth \begin{equation}
    r = \frac{\sum_{t=1}^T{(\hat{y}_t - \bar{\hat{y}}) (y_t - \bar{y}) }}{\sqrt{\sum_{t=1}^T{(\hat{y}_t - \bar{\hat{y}})^2}\sum_{t=1}^T{(y_t - \bar{y})^2}}} \, .
\end{equation}

\paragraph{Persistence model} A simple persistence model (PER) uses the last available ground truth value to make a forecast. For example, assume that the last observed ILI rate at time point $t_0$ is equal to $y_0$. In this case, at time point $t_0$ the $n$-day ahead forecast of a persistence model will always be equal to $y_0$, i.e. $\hat{y}_{t_0+n} = y_0$.

\section{Supplementary Figures and Tables}
\begin{figure}[!ht]
    \centering
    \includegraphics[width=0.85\linewidth]{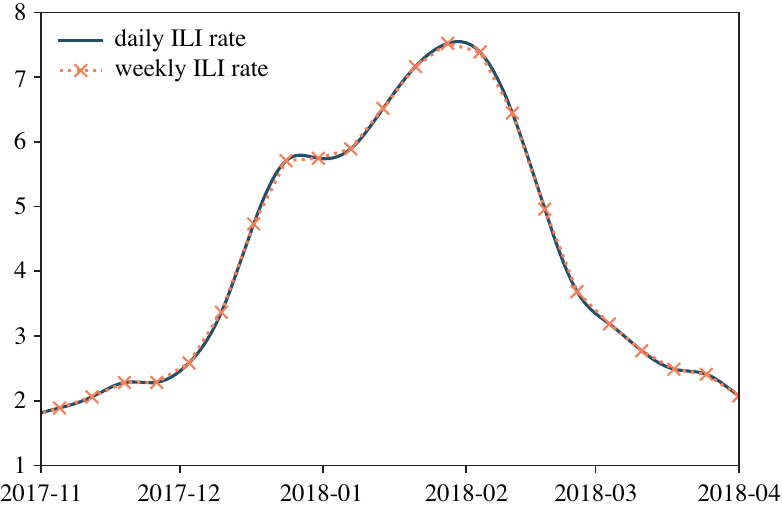}
    \caption[{Cubic interpolation example}]{\textbf{Cubic interpolation example} \newline
    Cubic interpolation of weekly ILI rates (as reported by the CDC for US) to produce pseudo-daily ones. Although cubic interpolation differs slightly from linear interpolation (straight line), it does not distort the weekly signal significantly and produces a more smoothed trend.}
    \label{fig:sup_daily_and_weekly_ILI_rate}
\end{figure}

\begin{figure}[!ht]
    \begin{algorithmic}
    \State {$x$} \Comment{inputs of shape size of batch $\times$ number of time-steps $\times$ number of search queries queries + 1}
    \State $\hat{y} \gets [~]$  \Comment{initialise prediction list}
    \State $\hat{x}, s  \gets \text{RNN}(x[:, :\tau, :])$   \Comment{warmup, data from $\tau$ to 0 is fed into the RNN to set the hidden states ($s$)}
    \State $\hat{x} \gets \text{FC}(\hat{x})$   \Comment{make the first forecast -- this contains the query frequencies and the ILI rate for $t_1$}
    \State $\hat{y}.\text{append}(\hat{x})$  \Comment{update list of forecasts}
    \For{$t$ in $1:\gamma-1$}
        \State $\hat{x} \gets \hat{x}.\text{sample}()$ \Comment{sample from last prediction to get an input which the RNN can use}
        
         \If{$t>\delta$}
            \State $\hat{x} \gets \text{concat}([x[:, \tau+t, :-1], \hat{x}[:, -1:]], 1)$ \Comment{concatenate ground truth query frequencies with forecasted ILI rate}
        \EndIf
        
        \State $\hat{x}, s \gets \text{RNN}(\hat{x}, s$)	
            \Comment{run the new inputs ($\hat{x}$) through the RNN layer}
        \State $\hat{x} \gets \text{FC}(\hat{x})$ 
        \State $\hat{y}\text{.append}(\hat{x})$  \Comment{update list of forecasts}
    \EndFor
    
    \State $\text{return}~\hat{y}$
    
    \end{algorithmic}
    \caption[{IRNN Pseudocode}]{\textbf{IRNN Pseudocode} \newline
    Pseudocode describing how the IRNN model makes one sequence of forecasts up to $\gamma$ days ahead.}
    \label{fig:sup_algorithm}
\end{figure}

\begin{table}[!ht]
    \centering
    \small
    \begin{tabular}{lcccc}
    \toprule
    \textbf Fold & \textbf Training start & \textbf Training end & \textbf Validation start & \textbf Validation end \\
    \midrule
    1 &  2004-03-24 &  2014-08-12 &  2014-08-13 &  2015-08-11 \\
    2 &  2004-03-24 &  2013-08-12 &  2013-08-13 &  2014-08-11 \\
    3 &  2004-03-24 &  2012-08-13 &  2012-08-13 &  2013-08-11 \\
    4 &  2004-03-24 &  2011-08-13 &  2011-08-14 &  2012-08-11 \\
    5 &  2004-03-24 &  2010-08-13 &  2010-08-14 &  2011-08-12 \\
    \bottomrule
    \end{tabular}
    \caption[{Train and validation intervals}]{\textbf{Train and validation intervals} \newline
    The training and validation date intervals of the 5 validation folds. These are used to validate and determine the hyperparameter values of the NNs in our experiments.}
    \label{tab:sup_val_periods}
\end{table}

\begin{figure}[!ht]
    \centering
    \includegraphics[width=0.95\linewidth]{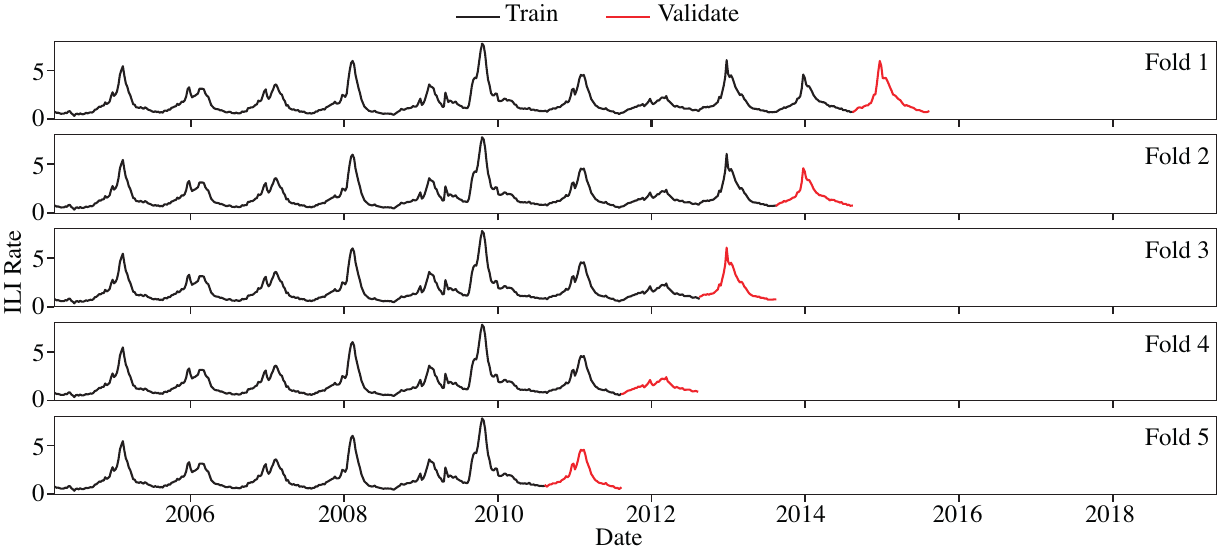}
    \caption[{Train and validation ILI rates}]{\textbf{Train and validation ILI rates} \newline
    National US ILI rates (as reported by the CDC) that we used for determining the hyperparameters of the NNs. We have denoted the 5 training and validation periods with black and red colours respectively.}
    \label{fig:sup_validation_periods}
\end{figure}

\begin{figure}[!ht]
    \centering
    \includegraphics[width=0.85\linewidth]{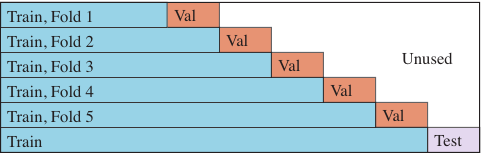}
    \caption[{Validation set diagram}]{\textbf{Validation set diagram} \newline
    Validation set diagram. Hyperparameters are validated using $5$-fold cross validation, where the validation periods are the last five available seasons before the test period. Error (NLL) is averaged over these validation periods. After hyperparameter optimisation, the full training set is used.}
    \label{fig:sup_k_fold_diagram}
\end{figure}

\begin{table}[!ht]
    \centering
    \small
    \begin{tabular}{lcccc}
    \toprule
    \textbf Flu season  & \textbf Training start   & \textbf Training end &  \textbf Testing start   & \textbf Testing end \\
    \midrule
    2015/16         & 2004-03-24        & 2015-08-12    & 2015-10-19        & 2016-05-14 \\
    2016/17         & 2004-03-24        & 2016-08-11    & 2016-10-17        & 2017-05-13 \\
    2017/18         & 2004-03-24        & 2017-08-10    & 2017-10-16        & 2018-05-12 \\
    2018/19         & 2004-03-24        & 2018-08-09    & 2018-10-15        & 2019-05-11 \\
    \bottomrule
    \end{tabular}
    \caption[{Train and test intervals}]{\textbf{Train and test intervals} \newline
    The training and testing date intervals (all inclusive) for the four flu seasons used to evaluate forecasting methods in our experiments. Dates given are the days from which forecasts are made.}
    \label{tab:sup_train_periods}
\end{table}

\begin{figure*}[!ht]
    \centering
    \includegraphics[width=0.95\linewidth]{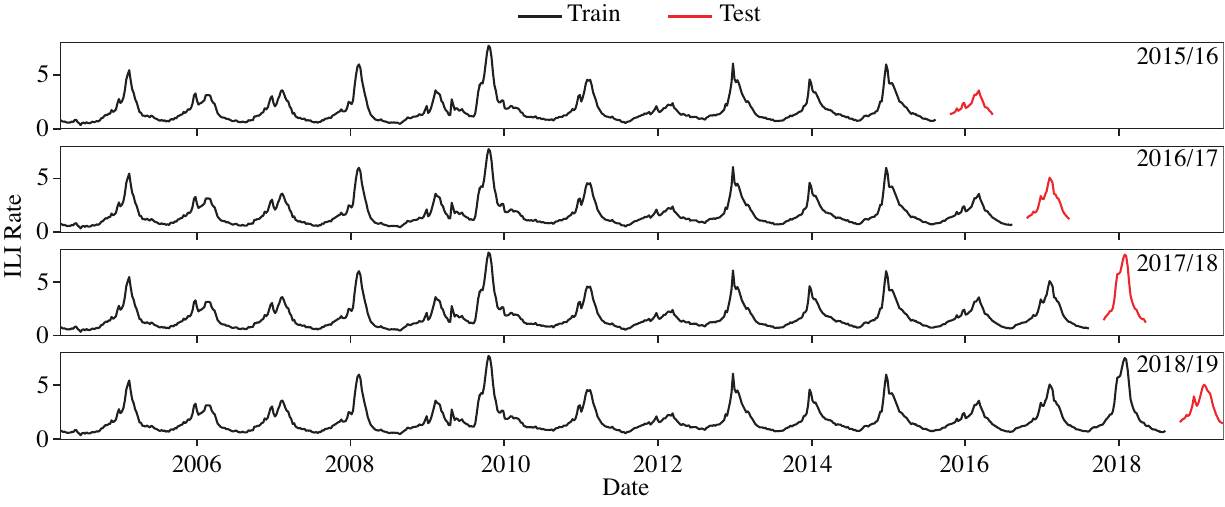}
    \caption[{ILI rates for training and testing}]{\textbf{ILI rates for training and testing} \newline
    National US ILI rates (as reported by the CDC) for the training (black) and testing (red) periods for each of the $4$ test folds that we used in our experiments.}
    \label{fig:sup_test_periods}
\end{figure*}

\begin{figure*}[!ht]
    \centering
    \includegraphics[width=0.95\linewidth]{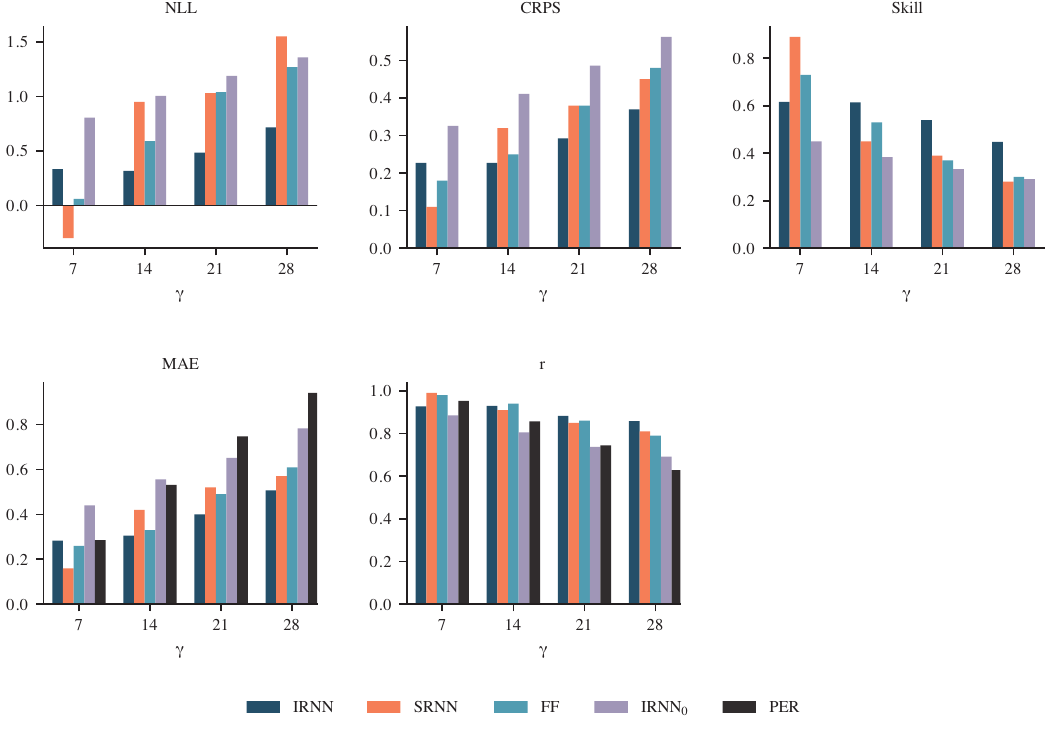}
    \caption[{Average metrics for all models}]{\textbf{Average metrics for all models} \newline
        Negative log-likelihood (NLL), continuous ranked probability score (CRPS), Skill, mean absolute error (MAE), and bivariate correlation for each NN model averaged over all four test flu seasons (2015/16 to 2018/19). Scores for different forecast horizons ($\gamma$) are shown. We also provide a comparison with IRNN trained without using any Web search activity data (IRNN$_0$) and a simple persistence model (PER) wherever applicable. This figure is a supplement to Figure 1 from the main manuscript.}
    \label{fig:sup_avg_metrics}
\end{figure*}

\begin{figure*}[!ht]
    \centering
    \includegraphics[width=0.98\linewidth]{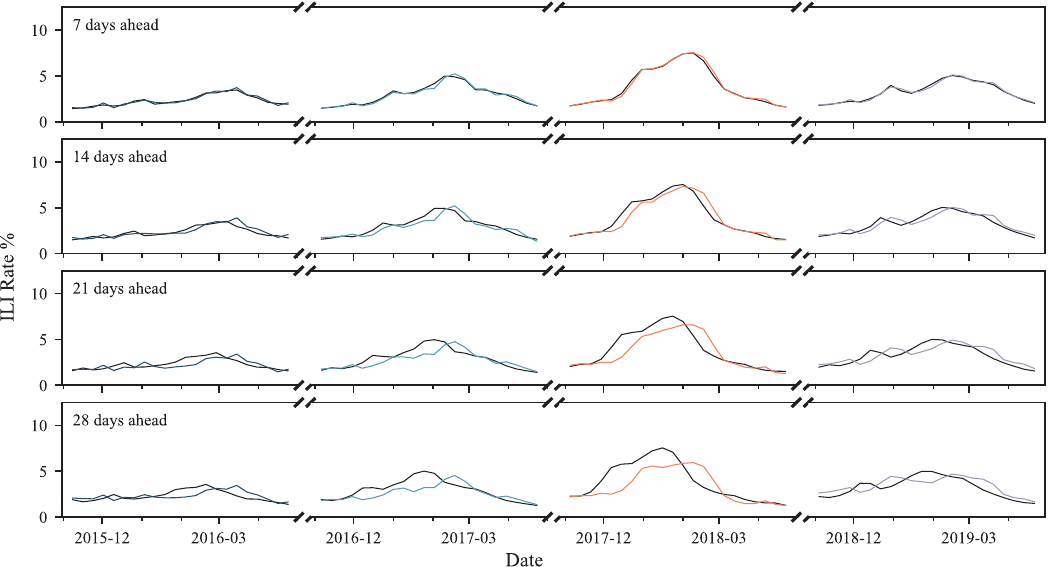}
    \caption[{Elastic-Net forecasts}]{\textbf{Elastic-Net forecasts} \newline
    Elastic-Net forecasts for all 4 test seasons (2015/16 to 2018/19) and forecasting horizons ($\gamma =$ 7, 14, 21, and 28). The influenza-like illness (ILI) rate (ground truth) is shown by the black line.}
    \label{fig:sup_elasticnet_forecasts}
\end{figure*}

\begin{sidewaystable}[!ht]
    \centering
    \footnotesize
    \setlength{\tabcolsep}{0.5pt}
    \begin{tabular}{p{1.1cm}p{0.7cm}p{0.7cm}p{0.7cm}p{0.7cm}p{0.7cm}p{0.7cm}p{0.7cm}p{0.7cm}p{0.7cm}p{0.7cm}p{0.7cm}p{0.7cm}p{0.7cm}p{0.7cm}p{0.7cm}p{0.7cm}p{0.7cm}p{0.7cm}p{0.7cm}p{0.7cm}p{0.7cm}p{0.7cm}p{0.7cm}p{0.7cm}p{0.7cm}p{0.7cm}}
\ Horizon & Metric  &\multicolumn{5}{c}{2015/16} & \multicolumn{5}{c}{2016/17} & \multicolumn{5}{c}{2017/18} & \multicolumn{5}{c}{2018/19} & \multicolumn{5}{c}{Avg (2015-19)} \\ 

\cmidrule(lr){3-7}\cmidrule(lr){8-12}\cmidrule(lr){13-17}\cmidrule(lr){18-22}\cmidrule(lr){23-27}
$\gamma$   &     &      Dte   &     Eln    &    NN$_a$ &     NN$_b$ &         NN &      Dte   &     Eln    &    NN$_a$ &    NN$_b$ &         NN &      Dte   &     Eln            &  NN$_a$ &     NN$_b$ &         NN &      Dte   &     Eln    &     NN$_a$ &     NN$_b$ &         NN &       Dte   &     Eln    &     NN$_a$ &    NN$_b$ &          NN \\
\midrule
7       &  Skill &       0.67 &           &      0.94 &  \textbf{0.99} &       0.75 &       0.63 &            &      0.82 &  \textbf{0.83} &       0.53 &      0.45 &            &       0.73 &  \textbf{0.77} &       0.53 &       0.62 &            &  \textbf{0.94} &  \textbf{0.94} &       0.61 &        0.59 &            &       0.85 &  \textbf{0.88} &       0.60 \\
        &    MAE &       0.22 &     0.13   &      0.13 &  \textbf{0.12} &       0.26 &  \textbf{0.19} &     0.20   &      0.20 &       0.20 &       0.35 &      0.39 &  \textbf{0.22} &       0.29 &       0.25 &       0.38 &       0.21 &     0.15   &  \textbf{0.11} &  \textbf{0.11} &       0.28 &        0.25 &     0.18   &       0.18 &  \textbf{0.17} &       0.32 \\
        &    $r$ &       0.88 &     0.96   &      0.96 &  \textbf{0.97} &       0.81 &       0.96 &     0.97   &      0.98 &  \textbf{0.99} &       0.91 &      0.97 &  \textbf{0.99} &       0.98 &  \textbf{0.99} &       0.98 &       0.97 &     0.98   &  \textbf{0.99} &  \textbf{0.99} &       0.90 &        0.94 &  \textbf{0.98} &  \textbf{0.98} &  \textbf{0.98} &       0.90 \\
\midrule
14      &  Skill &       0.54 &            &      0.69 &       0.72 &  \textbf{0.74} &       0.54 &            &      0.53 &  \textbf{0.57} &       0.53 &      0.29 &            &       0.43 &       0.50 &  \textbf{0.53} &       0.52 &            &       0.58 &       0.58 &  \textbf{0.61} &        0.46 &            &       0.55 &  \textbf{0.59} &  \textbf{0.59} \\
        &    MAE &       0.38 &  \textbf{0.26} &      0.28 &       0.28 &      0.28 &  \textbf{0.32} &     0.37   &      0.38 &       0.35 &       0.35 &      0.64 &     0.42   &       0.44 &       0.43 &  \textbf{0.39} &       0.33 &     0.37   &       0.31 &       0.31 &  \textbf{0.28} &        0.42 &     0.35   &       0.35 &       0.34 &  \textbf{0.33} \\
        &    $r$ &       0.64 &  \textbf{0.87} &      0.79 &       0.78 &      0.79 &  \textbf{0.91} &     0.90   &      0.90 &  \textbf{0.91} &  \textbf{0.91} &      0.90 &     0.95   &       0.97 &  \textbf{0.98} &  \textbf{0.98} &  \textbf{0.92} &     0.89   &       0.90 &       0.90 &       0.90 &        0.84 & \textbf{0.91}  &       0.89 &  \textbf{0.89} &      0.89 \\
\midrule
21      &  Skill &       0.44 &            &      0.60 &       0.62 &  \textbf{0.64} &       0.48 &            &      0.43 &  \textbf{0.49} &       0.43 &      0.21 &            &  \textbf{0.37} &       0.35 &       0.30 &       0.46 &            &       0.51 &       0.51 &  \textbf{0.52} &        0.38 &            &       0.47 &  \textbf{0.48} &        0.45 \\
        &    MAE &       0.48 &     0.41   & \textbf{0.35} &       0.38 &       0.37 &  \textbf{0.38} &     0.61   &      0.50 &       0.44 &       0.45 &      0.86 &     0.96   &       0.64 &  \textbf{0.62} &  \textbf{0.62} &  \textbf{0.40} &     0.52   &       0.42 &       0.42 &       0.44 &        0.53 &     0.62   &       0.48 &  \textbf{0.46} &        0.47 \\
        &    $r$ &       0.36 &     0.59   & \textbf{0.71} &       0.66 &       0.67 &  \textbf{0.87} &     0.71   &      0.80 &       0.85 &       0.83 &      0.82 &     0.78   &       0.92 &       0.93 &  \textbf{0.94} &  \textbf{0.89} &     0.77   &       0.82 &       0.82 &       0.82 &        0.73 &     0.71   &  \textbf{0.81} &  \textbf{0.81} &   \textbf{0.81} \\
\midrule
28      &  Skill &       0.37 &            &      0.49 &       0.50 &  \textbf{0.53} &       0.46 &            &      0.39 &  \textbf{0.48} &       0.38 &      0.17 &            &       0.21 &  \textbf{0.22} &        0.14 &       0.42 &            &  \textbf{0.48} &  \textbf{0.48} &       0.45 &        0.33 &            &       0.37 &  \textbf{0.40} &        0.33 \\
        &    MAE &       0.54 &  \textbf{0.41} &      0.51 &       0.50 &  \textbf{0.47} &  \textbf{0.39} &     0.81   &      0.57 &       0.51 &       0.50 &      1.06 &     1.35   &       0.86 &  \textbf{0.82} &        0.85 &  \textbf{0.45} &     0.63   &       0.50 &       0.50 &       0.58 &        0.61 &     0.80   &       0.61 &  \textbf{0.58} &        0.60 \\
        &    $r$ &       0.23 &     0.47   & \textbf{0.69} &       0.62 &       0.63 &  \textbf{0.88} &     0.56   &      0.74 &       0.81 &       0.79 &      0.76 &     0.61   &       0.90 &  \textbf{0.92} &   \textbf{0.92} &  \textbf{0.86} &     0.69   &       0.79 &       0.79 &       0.79 &        0.68 &     0.58   &       0.78 &  \textbf{0.79} &        0.78 \\
\bottomrule
\end{tabular}

    \caption[{Performance metrics for best neural network models compared with Dante}]{\textbf{Performance metrics for best neural network models compared with Dante} \newline
    Forecasting performance metrics for the best-performing neural network (SRNN for $\gamma=7$, IRNN for $\gamma \ge 14$) compared with Dante (Dte) and Elastic Net (Eln). The NNs are trained using search query frequencies generated only up to the last available ILI rate (the 2-week advantage of using Web search data is removed). We use leave-one flu season-out to train models, similarly to Dante. The best results for this comparison are shown in bold. NN$_b$ denotes results where the temporal advantage of Web search activity information is maintained (compared to NN). NN$_a$ holds results for the same experiment as NN$_b$ with the addition of disabling leave-one flu season-out training, i.e. training does not include data after the test year. Eln uses the same data sets (inputs, targets) as the NNs. Therefore, it is trained using look ahead and without leave-one flu season-out. Eln does not estimate uncertainty and hence, the Skill metric is not available (empty cell). This Table supplements Table~\ref{tab:dante_comparison} in the main manuscript.}
    \label{tab:sup_dante_comp_full}
% \end{sidewaystable}
\end{sidewaystable}

\begin{figure*}[t]
    \centering
    \includegraphics[width=0.99\linewidth]{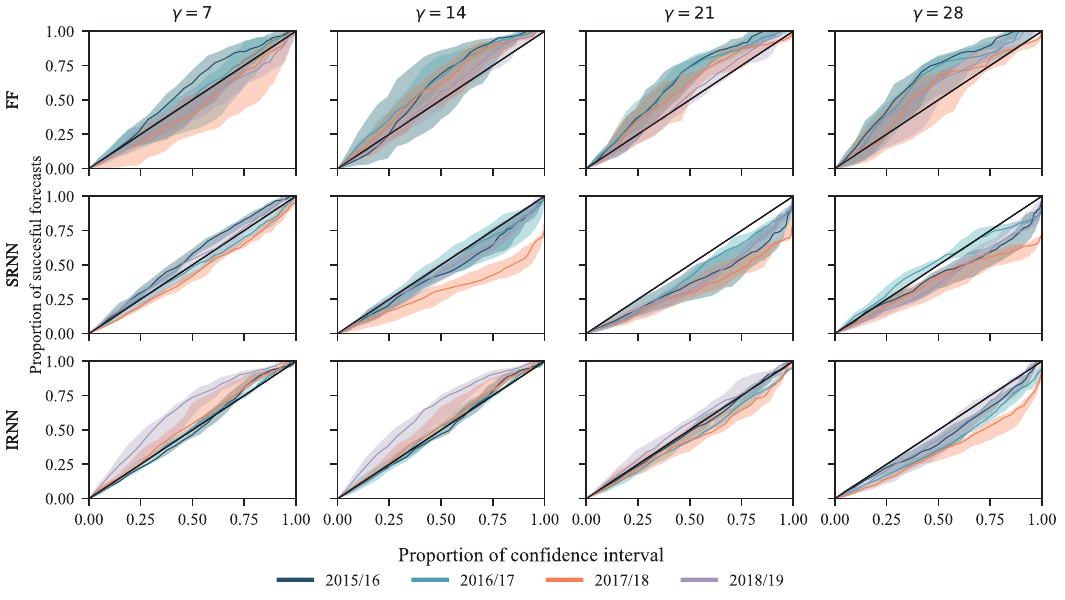}
    \caption[{Detailed calibration plots for neural network models}]{\textbf{Detailed calibration plots for neural network models} \newline
    Calibration plots for the forecasts made by the three NN models (FF, SRNN, and IRNN) for each of the four test periods (2015/16 to 2018/19) and forecasting horizons ($\gamma$). The lines show the how frequently the ground truth falls within a confidence interval (CI) of the same level. To be more precise, a point $(x,y)$ denotes that the proportion $y \in [0, 1]$ of the forecasts when combined with a CI at the $x \times 100\%$ level include the ground truth (successful forecasts). The optimal calibration is shown by the diagonal black line. Points above or below the diagonal indicate an over- or under-estimation of uncertainty, and hence an under- or over-confident model, respectively. The shadows show the upper and lower quartile of the calibration curves when the models are trained multiple times with different initialisation seeds.}
    \label{fig:sup_calibration_plot}
\end{figure*}

\begin{figure*}[!ht]
    \centering
    \includegraphics[width=0.98\linewidth]{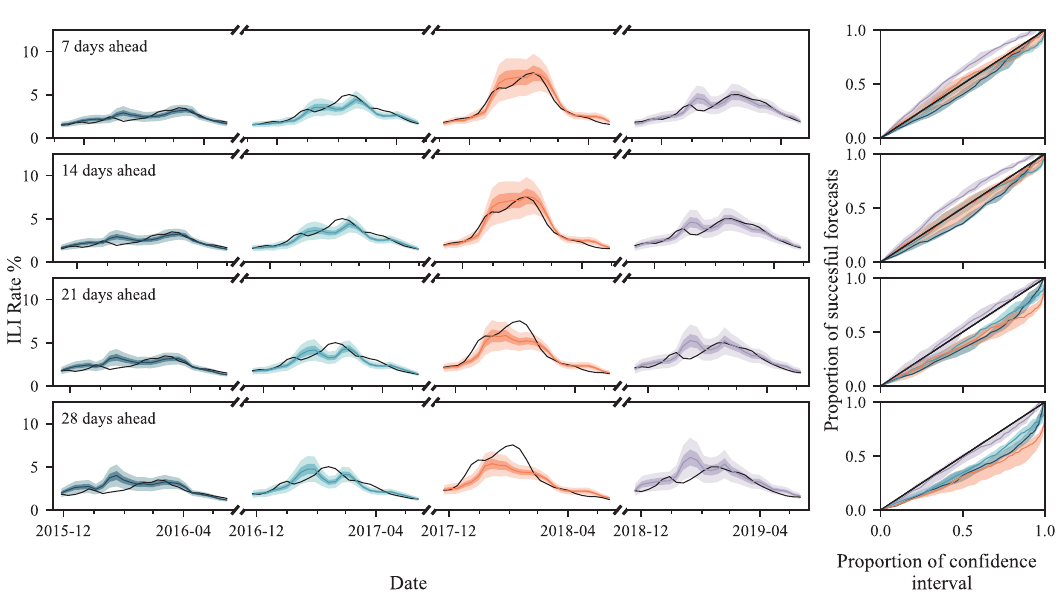}
    \caption[{IRNN Forecasts}]{\textbf{IRNN Forecasts}\newline
    IRNN forecasts with leave-one flu season-out and using all available Web search data for all 4 test seasons (2015/16 to 2018/19) and forecasting horizons ($\gamma =$ 7, 14, 21, and 28). Confidence intervals (uncertainty estimates) are shown at $50\%$ and $90\%$ levels, and are visually distinguished by darker and lighter colour overlays respectively. The influenza-like illness (ILI) rate (ground truth) is shown by the black line. The flu seasons are shown in different colours which correspond with the calibration plots on the right. The calibration lines show the how frequently the ground truth falls within a confidence interval (CI) of the same level. To be more precise, a point $(x,y)$ denotes that the proportion $y \in [0, 1]$ of the forecasts when combined with a CI at the $x \times 100\%$ level include the ground truth (successful forecasts). The optimal calibration is shown by the diagonal black line. Points above or below the diagonal indicate an over- or under-estimation of uncertainty, and hence an under- or over-confident model, respectively. The shadows show the upper and lower quartile of the calibration curves when the models are trained multiple times with different initialisation seeds.}
    \label{fig:sup_IRNN_LOE_forecasts}
\end{figure*}

\begin{table*}[!ht]
    \centering
    \small
    \setlength{\tabcolsep}{6pt}
    \begin{tabular}{l ccc}
        \toprule
        Horizon & \multicolumn{3}{c}{$\gamma = 21$} \\
        \toprule
        Metric                  & FF               & SRNN              & IRNN              \\
        \midrule
        $\delta$-p (days)       & 12, 15, 9, -33   & 10, 15, 5, -28    & -4, 14, -25, -28  \\
        Avg. $\delta$-$y_{\text{p}}$ & 0.50        & 1.31              & 0.52              \\
        MAE-p                   &  0.89            & 1.28              & 0.73               \\
        SMAPE-p (\%)            & 21.34            & 29.70             & 15.66      \\
        \bottomrule
        \\
        \\
        \toprule
        Horizon & \multicolumn{3}{c}{$\gamma = 28$} \\
        \toprule
        Metric                        & FF                & SRNN                  & IRNN     \\
        \midrule
        $\delta$-p (days)             & 9, -15, 15, -26   & -40, -14, -16, -23    & -47, -20, -21, -25 \\
        Avg. $\delta$-$y_{\text{p}}$  & 0.70              & 1.54                  & 0.76     \\
        MAE-p                         & 1.22              & 1.59                  & 0.93     \\
        SMAPE-p (\%)                  & 30.55             & 38.17                 & 19.83    \\
        
        \bottomrule
\end{tabular}
    \caption[{Meta-analysis metric for Bayesian neural networks}]{\textbf{Meta-analysis metric for Bayesian neural networks} \newline 
    Meta-analysis of ILI rate forecasts around the peak of a flu season for FF, SRNN, and IRNN. $\delta$-p denotes the temporal difference (in days) in forecasting the peak of the flu seasons 2015/16, 2016/17, 2017/18, and 2018/19, respectively. Negative / positive values indicate an earlier / later forecast. Avg. $\delta$-$y_{\text{p}}$ measures the average magnitude difference in the estimate of the peak of the flu season between a forecasting model and CDC. MAE-p is the MAE when the ILI rate is above the seasonal mean plus one standard deviation. SMAPE-p (\%) is the symmetric mean absolute percentage of error for the same time periods.}
    \label{tab:sup_comp_nn_discuss}
\end{table*}

\chapter{Supplementary Information for Chapter \ref{chapterlabel2}}

\section{Supplementary ODE Solver}
\paragraph{Fourth Order Runge-Kutte Method\\}
\label{sec:appendixRK4}
\begin{align*}
k_1 &= f\left(t_n, x_n\right) \\
k_2 &= f\left(t_n + \frac{h}{2}, x_n + h \frac{k_1}{2}\right) \\
k_3 &= f\left(t_n + \frac{h}{2}, x_n + h \frac{k_2}{2}\right) \\
k_4 &= f\left(t_n + h, x_n + h k_3\right) \\
x_{n+1} &= x_n + \frac{h}{6}\left(k_1 + 2k_2 + 2k_3 + k_4\right) \\
t_{n+1} &= t_n + h
\end{align*}

$k_1$, $k_2$, $k_3$ and $k_4$ are intermediate slopes calculated at different points within the step from $t_n$ to $t_{n+1}$, $f$ represents the ODE function.

\section{ Supplementatry SIR Sensitivity Analysis}
\label{sec:sup_sensitivity_analysis}
We evaluate the sensitivity of an SIR model, highlighting the need for precise specification of the initial conditions and model parameters in our later forecasting models. Here we construct an SIR model where $\beta = 2.0$, $\omega = 1.4$ and initial conditions $[s_0, i_0, r_0] = [0.8, 0.001, 0.199]$. The model is integrated from $t=0$ to $t=26$ weeks, where $t$ is represented time in weeks. To analyse the model's sensitivity, the initial conditions and model parameters are individually perturbed by between $1\%$ and $25\%$. The impact that these perturbations have on the output trajectory is measured using: mean absolute percentage error (MAPE), the delay between peaks (Lag), and the percentage difference in peak ILI rate. Lag is included to measure the degree of temporal shift in the peak of the time series caused by parameter variation. 

\begin{figure}[!ht]
    \centering
    \includegraphics[width=0.8\textwidth]{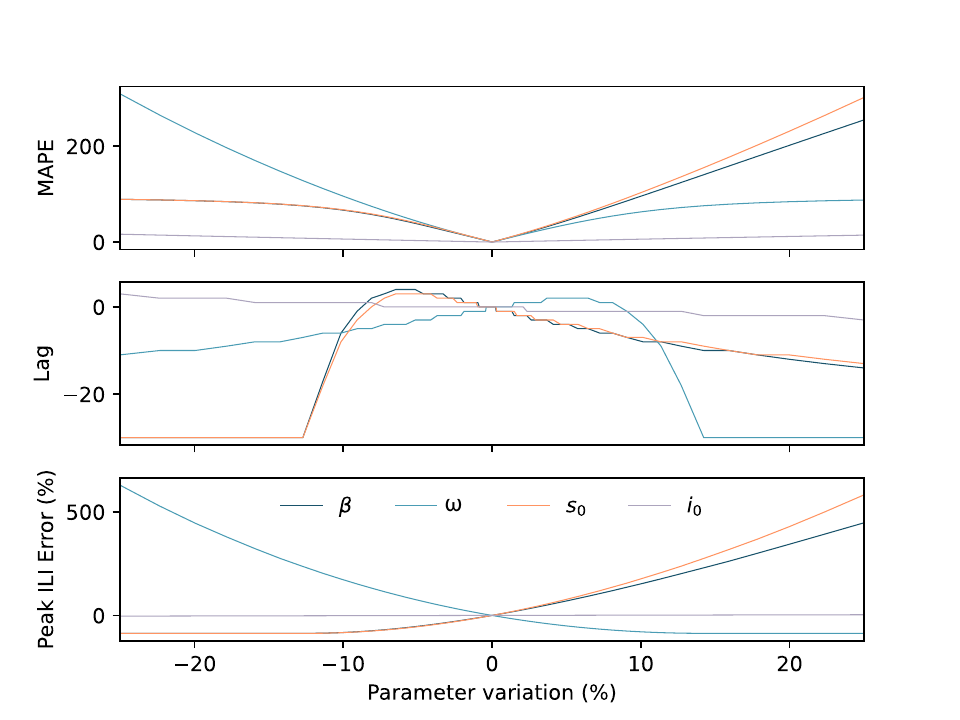}
    \caption[{SIR model parameter-sensitivity-analysis metrics}]{\textbf{SIR model parameter-sensitivity-analysis metrics} \newline 
    SIR sensitivity metrics for $\beta$, $\omega$, $s_0$, and $i_0$. Parameters are varied by $\pm 25\%$ and mean absolute percentage error (MAPE), the delay between forecast peaks (Lag) and the percentage change in the peak ILIpropor(Peak ILI Error (\%))}
    \label{fig:SIR_sensitivity_metrics}
\end{figure}

\begin{figure}
    \centering
    \includegraphics[width=0.8\textwidth]{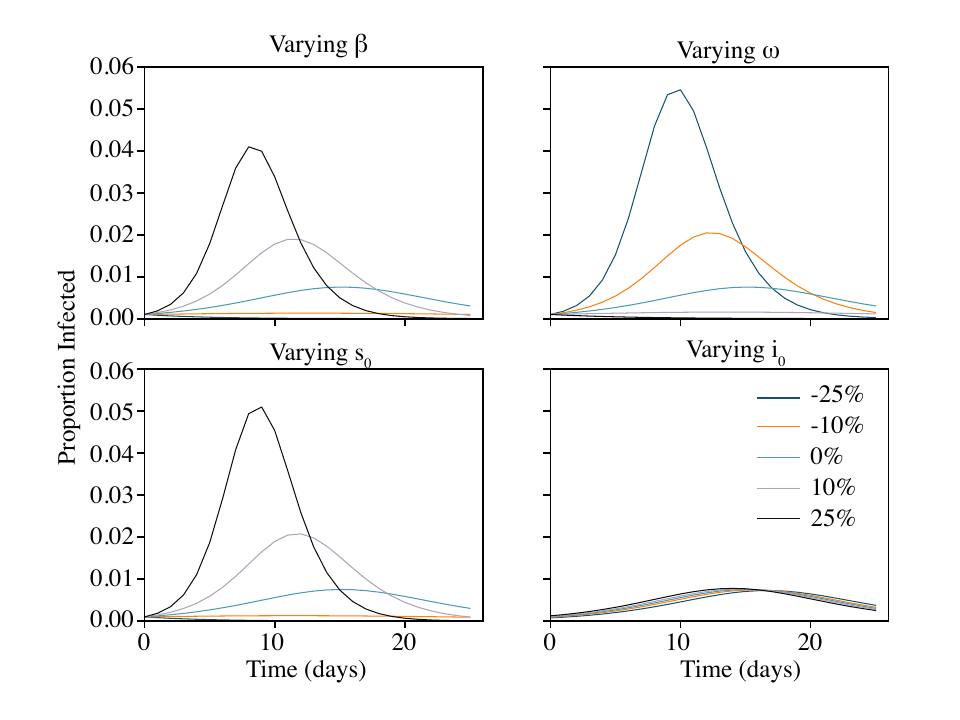}
    \caption[{SIR model parameter-sensitivity-analysis trajectories}]{\textbf{SIR model parameter-sensitivity-analysis trajectories} \newline 
    SIR sensitivity trajectories for $\beta$, $\omega$, $s_0$, and $i_0$. 
    Parameters are varied by $\pm 10
    $ and $\pm25\%$. The epidemic trajectories for each set of parameters are shown.}
    \label{fig:SIR_sensitivity_trajectories}
\end{figure}

Figure~\ref{fig:SIR_sensitivity_metrics} shows the outcomes for sensitivity analysis on each metric. Figure~\ref{fig:SIR_sensitivity_trajectories} shows the epidemic trajectories for the sensitivity analysis.

The SIR model trajectory is highly sensitive to $\beta$ and $\omega$. Increasing $\beta$ by $10\%$ resulted in a $150\%$ increase in the peak infections along with a much earlier peak. Conversely, a $10\%$ decrease in $\omega$ produced a similar effect, with peak infections rising by $175\%$ and occurring sooner. Surprisingly, the initial proportion of infected individuals ($i_0$) had a minimal impact on the infection trajectory. This can be attributed to the fact that the infected population remains relatively small compared to the susceptible population, even at the epidemic's peak. The maximum proportion of infected individuals without perturbation was approximately $0.01$, increasing to approximately $0.05$ for different model parameters. However, the susceptible population always remained significantly larger. 

Modelling epidemics using compartmental models is challenging because of the difficulty in accurately measuring the susceptible population. Although it is possible to approximate the proportion of infected individuals, other factors introduce uncertainties such as an individual's susceptibility to the disease. For example, it remains uncertain whether individuals infected in previous years would be conferred immunity in subsequent years and whether this applies universally or only to specific individuals. A mechanistic model with incorrectly specified parameters or initial conditions will not produce reliable forecasts. However, augmenting the mechanistic model with a non-mechanistic component can help to correct errors in the mechanistic model. We show how a universal differential equation constructed by augmenting an SIR model with a neural ODE can produce forecasts equivalent to an SEIR model.

\section{Supplementary Forecast Plots}
\begin{figure*}[!h]
    \centering
    \includegraphics[width=0.98\linewidth]{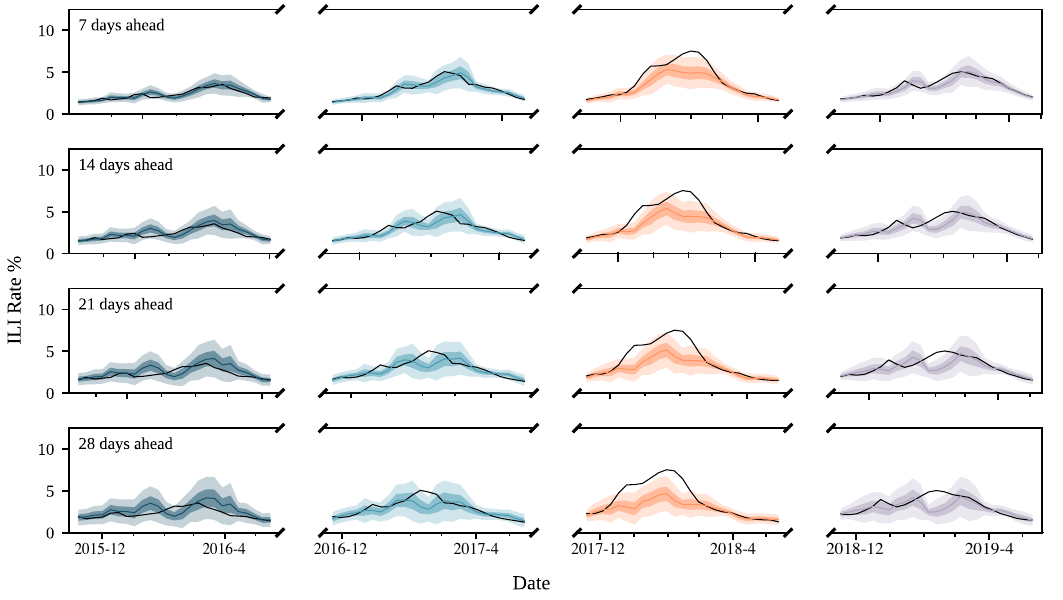}
    \caption[{\FA forecasts}]{\textbf{\FA forecasts}\newline
    }
    \label{fig:FA_forecasts}
\end{figure*}

\begin{figure*}[!h]
    \centering
    \includegraphics[width=0.98\linewidth]{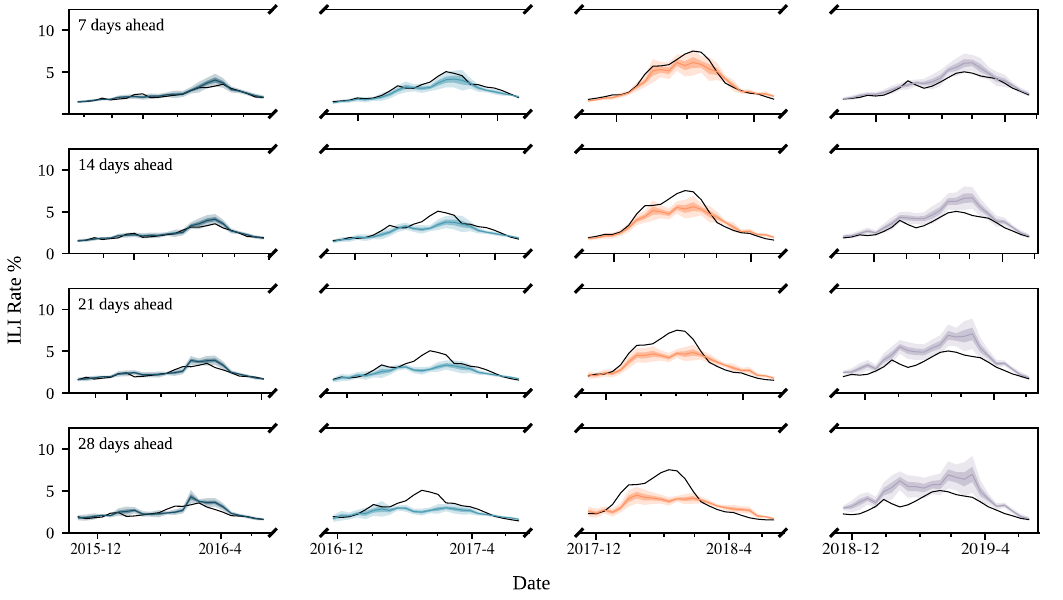}
    \caption[{\FAQS forecasts}]{\textbf{\FAQS forecasts}\newline
    }
    \label{fig:FAQS_forecasts}
\end{figure*}

\begin{figure*}[!h]
    \centering
    \includegraphics[width=0.98\linewidth]{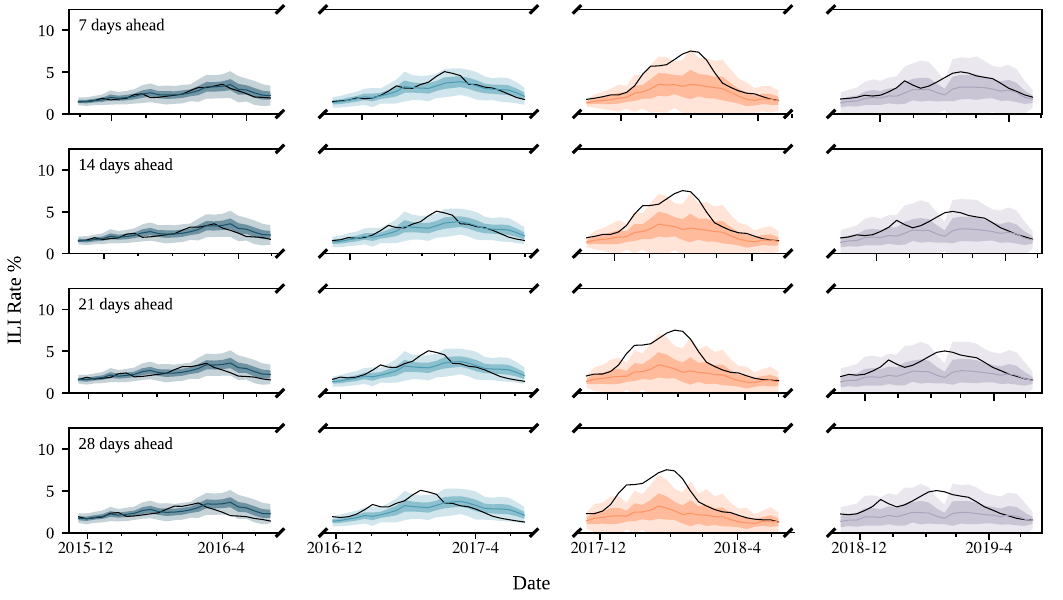}
    \caption[{\SIRSIMPLE forecasts}]{\textbf{\SIRSIMPLE forecasts}\newline
    }
    \label{fig:SIRSIMPLE_forecasts}
\end{figure*}

\begin{figure*}[!h]
    \centering
    \includegraphics[width=0.98\linewidth]{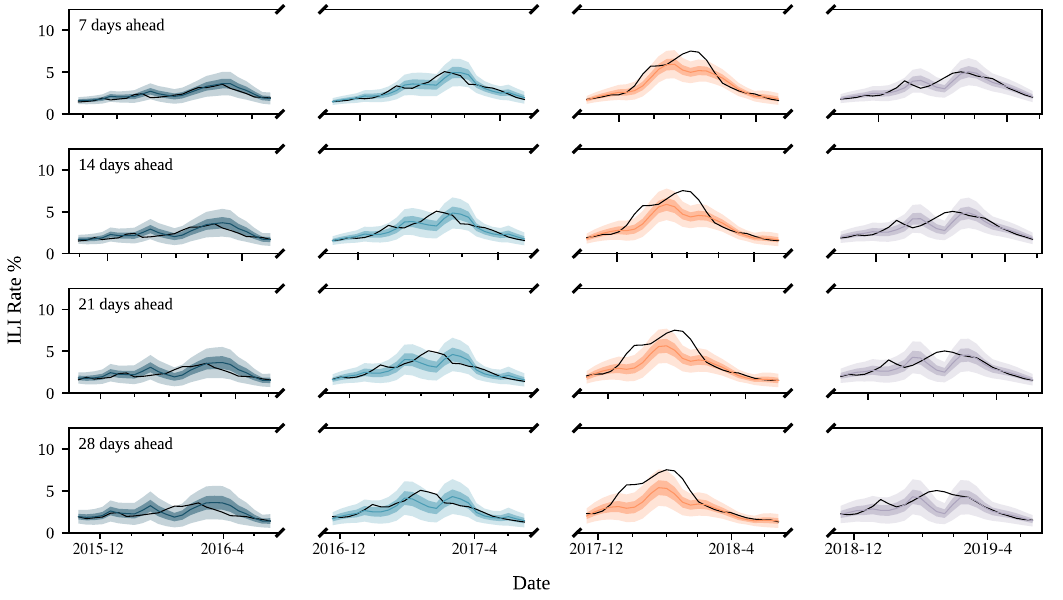}
    \caption[{\SIRNN forecasts}]{\textbf{\SIRNN forecasts}\newline
    }
    \label{fig:SIRNN_forecasts}
\end{figure*}

\begin{figure*}[!h]
    \centering
    \includegraphics[width=0.98\linewidth]{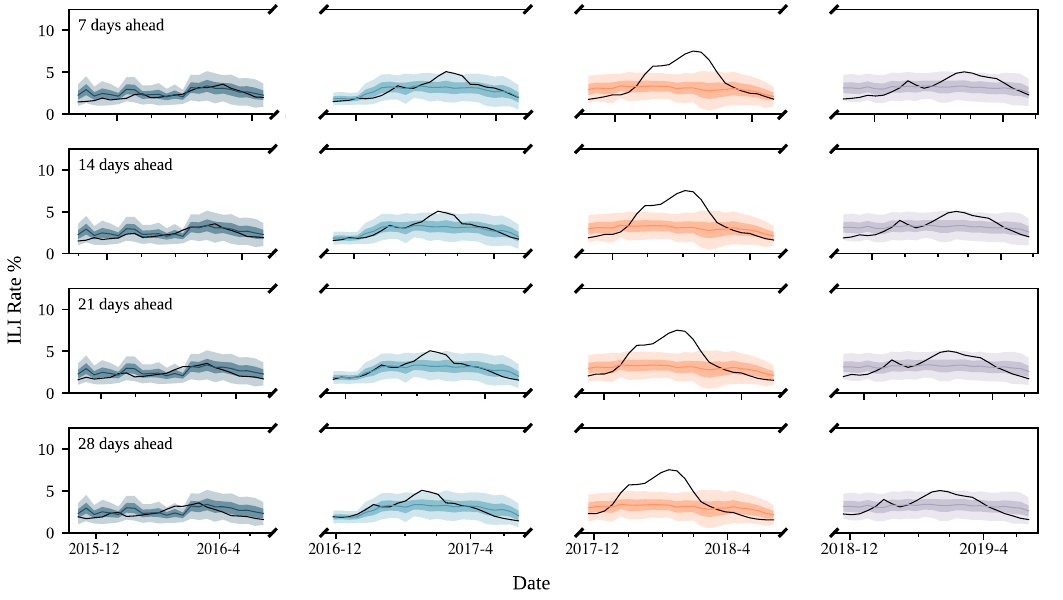}
    \caption[{\SIRQS forecasts}]{\textbf{\SIRQS forecasts}\newline
    }
    \label{fig:SIRQS_forecasts}
\end{figure*}

\begin{figure*}[!h]
    \centering
    \includegraphics[width=0.98\linewidth]{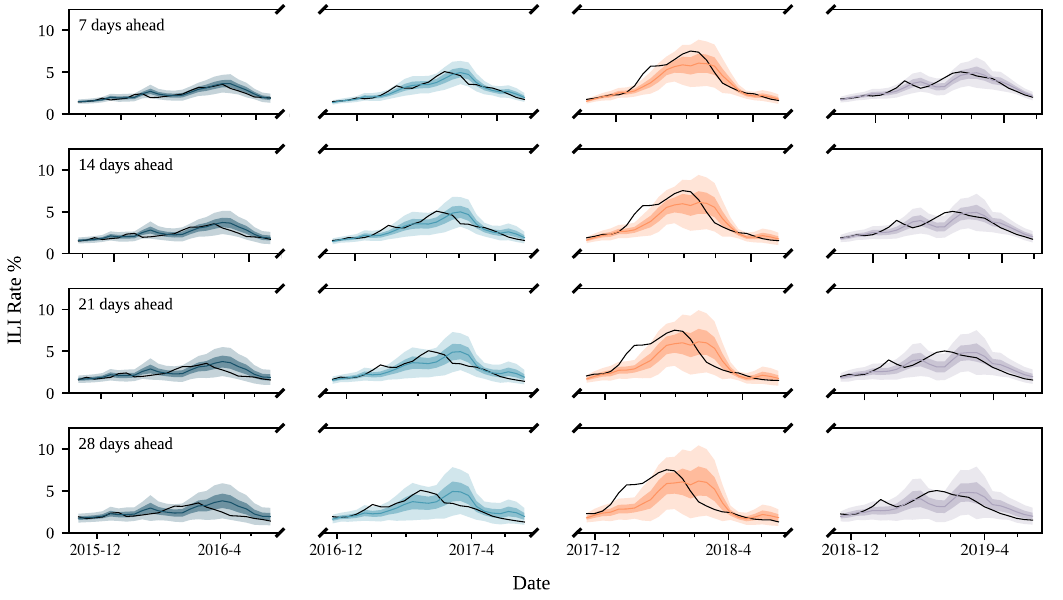}
    \caption[{\SEIRFA forecasts}]{\textbf{\SEIRFA forecasts}\newline
    }
    \label{fig:SEIRFA_forecasts}
\end{figure*}

\clearpage
\section{Supplementary Calibration Plots}

\begin{figure*}[!h]
    \centering
    \includegraphics[width=0.95\linewidth]{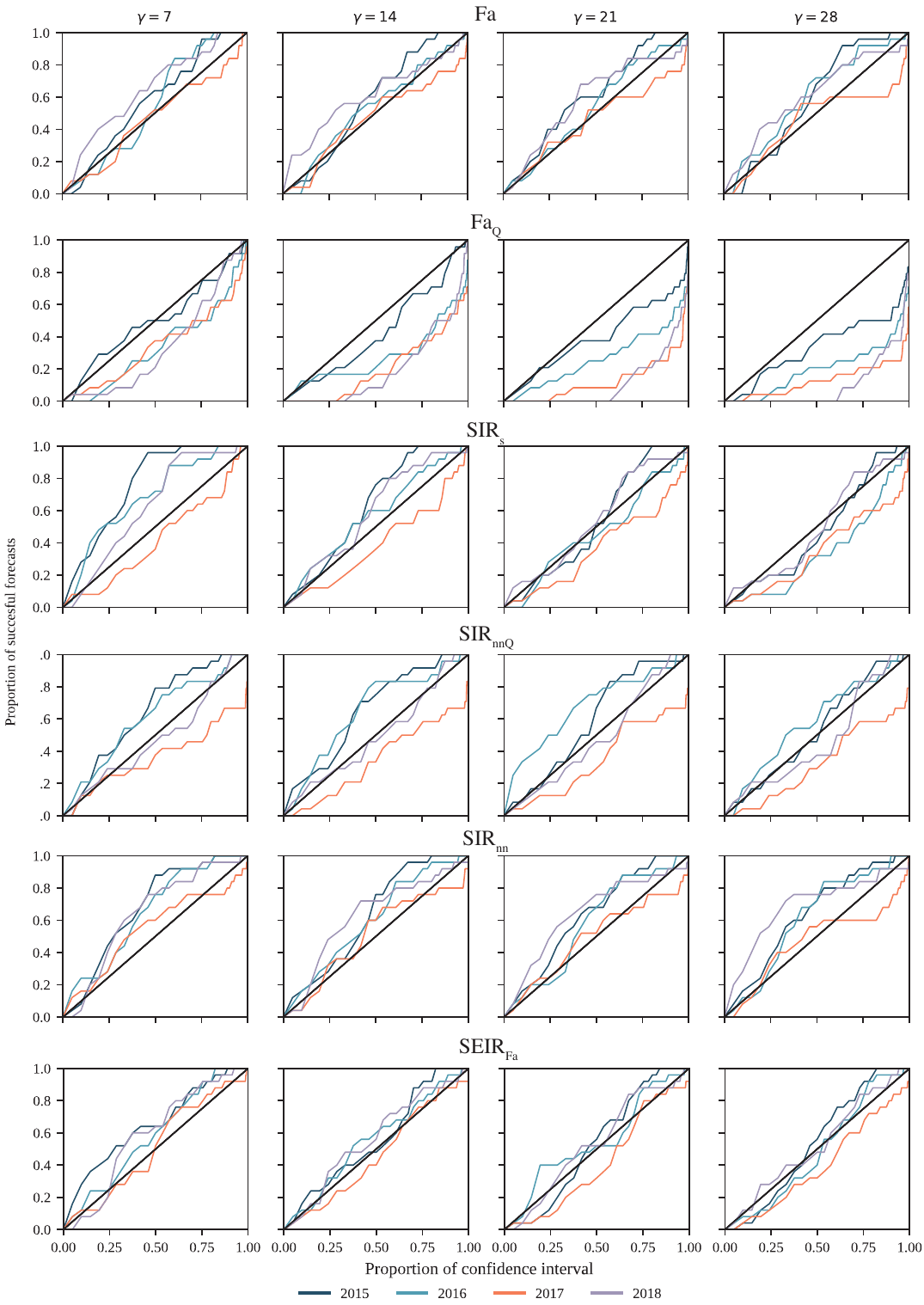}
    \caption[{Supplementary calibration plots for ODE models}]{\textbf{Supplementary calibration plots for ODE models}\newline
    
    Calibration shown for \FA, \FAQS, \SIRSIMPLE, \SIRNN, \SIRQS, and \SEIRFA~for each of the four test periods (2015/16 to 2018/19) and forecasting horizons ($\gamma$). The lines show the how frequently the ground truth falls within a confidence interval (CI) of the same level. To be more precise, a point $(x,y)$ denotes that the proportion $y \in [0, 1]$ of the forecasts when combined with a CI at the $x \times 100\%$ level include the ground truth (successful forecasts). The optimal calibration is shown by the diagonal black line. Points above or below the diagonal indicate an over- or under-estimation of uncertainty, and hence an under- or over-confident model, respectively.}
    \label{fig:sup_ode_calibration}
\end{figure*}

\clearpage
\section{Supplementary Forecast Trajectories}
\begin{figure*}[!h]
    \centering
    \includegraphics[width=0.95\linewidth]{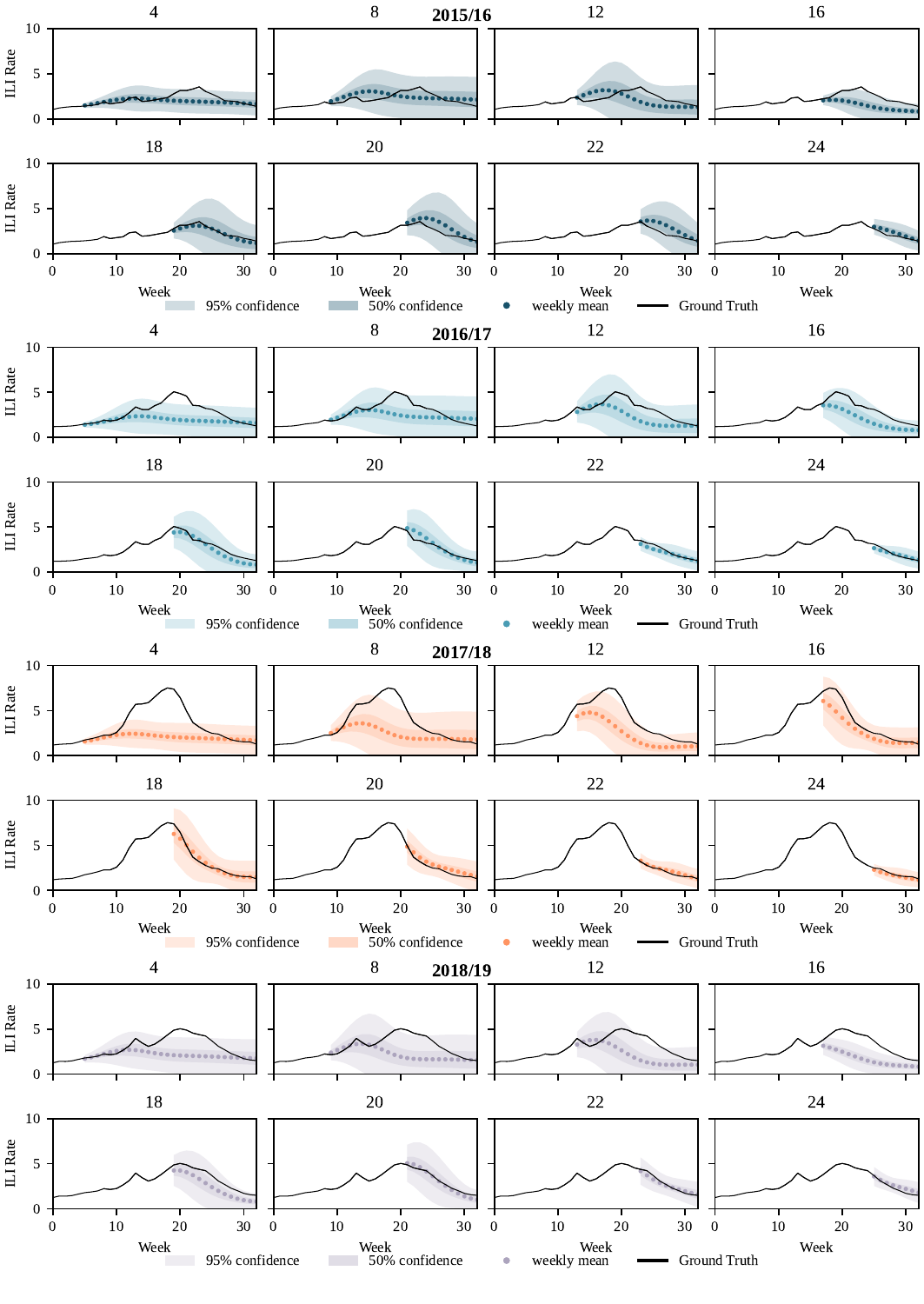}
    \caption[{\FA~trajectories}]{\textbf{\FA~trajectories}\newline
    Forecast trajectories for the \FA~from a given epidemic week (indexed from week $40$ in the year) to the end of the season. Each subplot shows the model's forecast from the given epidemic week (starting a week $40$ in the year). Trajectories shown for the mean, $50\%$ and $90\%$ confidence intervals.
    }
    \label{fig:FA_trajectories}
\end{figure*}

\begin{figure*}[!h]
    \centering
    \includegraphics[width=0.98\linewidth]{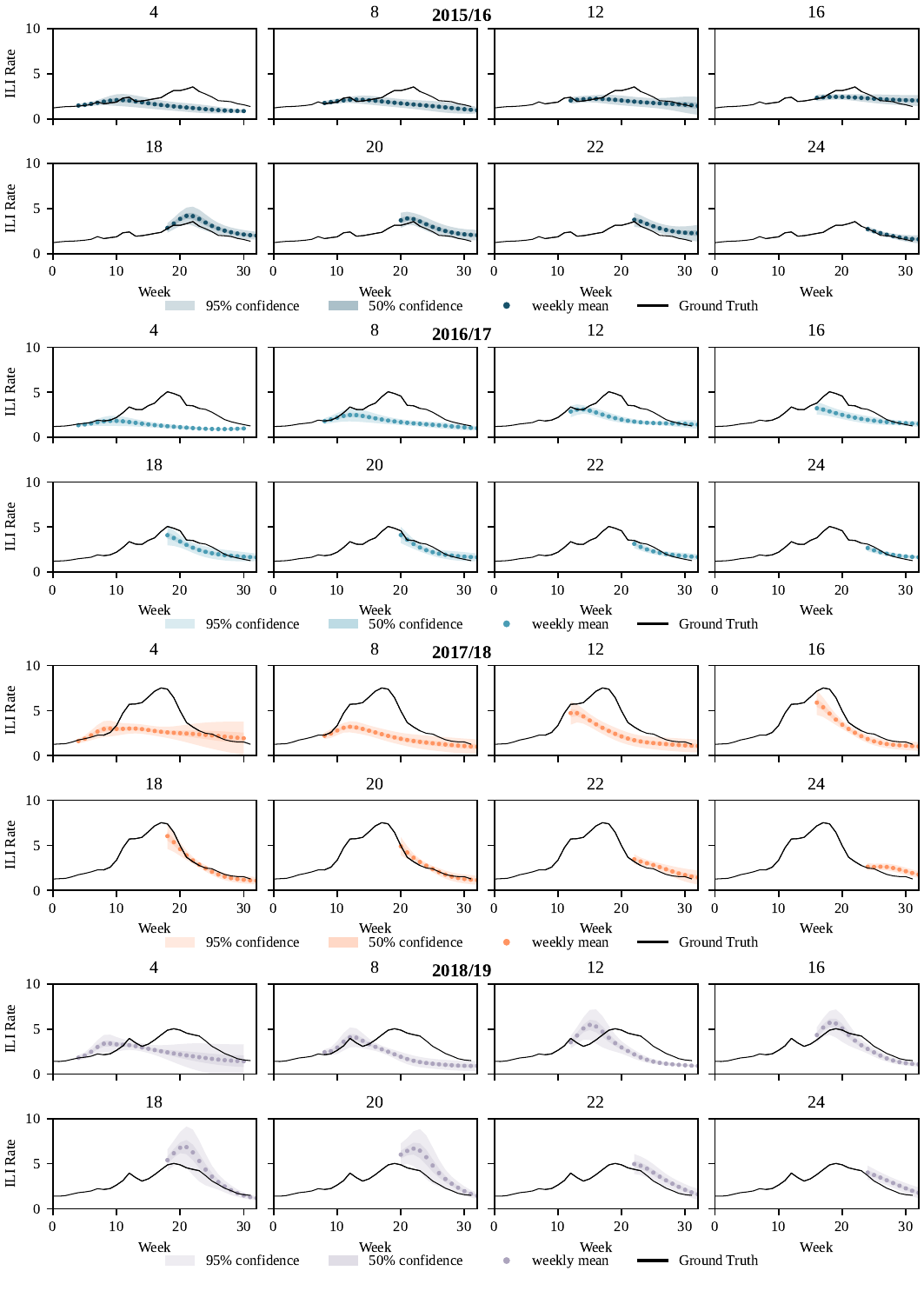}
    \caption[{\FAQS~trajectories}]{\textbf{\FAQS~trajectories}\newline
    Forecast trajectories for the \FAQS~from a given epidemic week (indexed from week $40$ in the year) to the end of the season. Each subplot shows the model's forecast from the given epidemic week (starting a week $40$ in the year). Trajectories shown for the mean, $50\%$ and $90\%$ confidence intervals.
    }
    \label{fig:FAQS_trajectories}
\end{figure*}

\begin{figure*}[!h]
    \centering
    \includegraphics[width=0.98\linewidth]{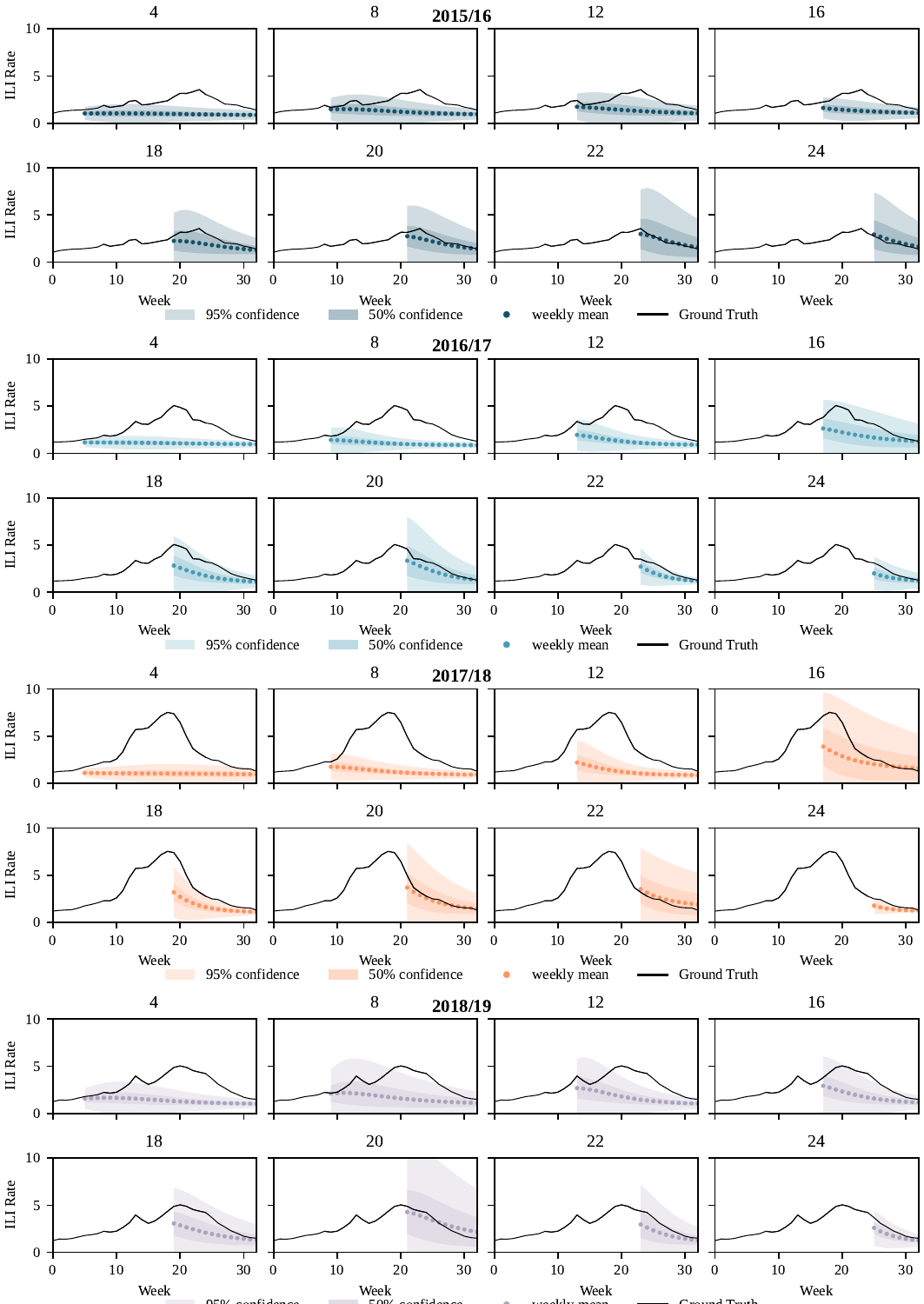}
    \caption[{\SIRSIMPLE~trajectories}]{\textbf{\SIRSIMPLE~trajectories}\newline
    Forecast trajectories for the \SIRSIMPLE~from a given epidemic week (indexed from week $40$ in the year) to the end of the season. Each subplot shows the model's forecast from the given epidemic week (starting a week $40$ in the year). Trajectories shown for the mean, $50\%$ and $90\%$ confidence intervals.
    }
    \label{fig:SIRSIMPLE_trajectories}
\end{figure*}

\begin{figure*}[!h]
    \centering
    \includegraphics[width=0.98\linewidth]{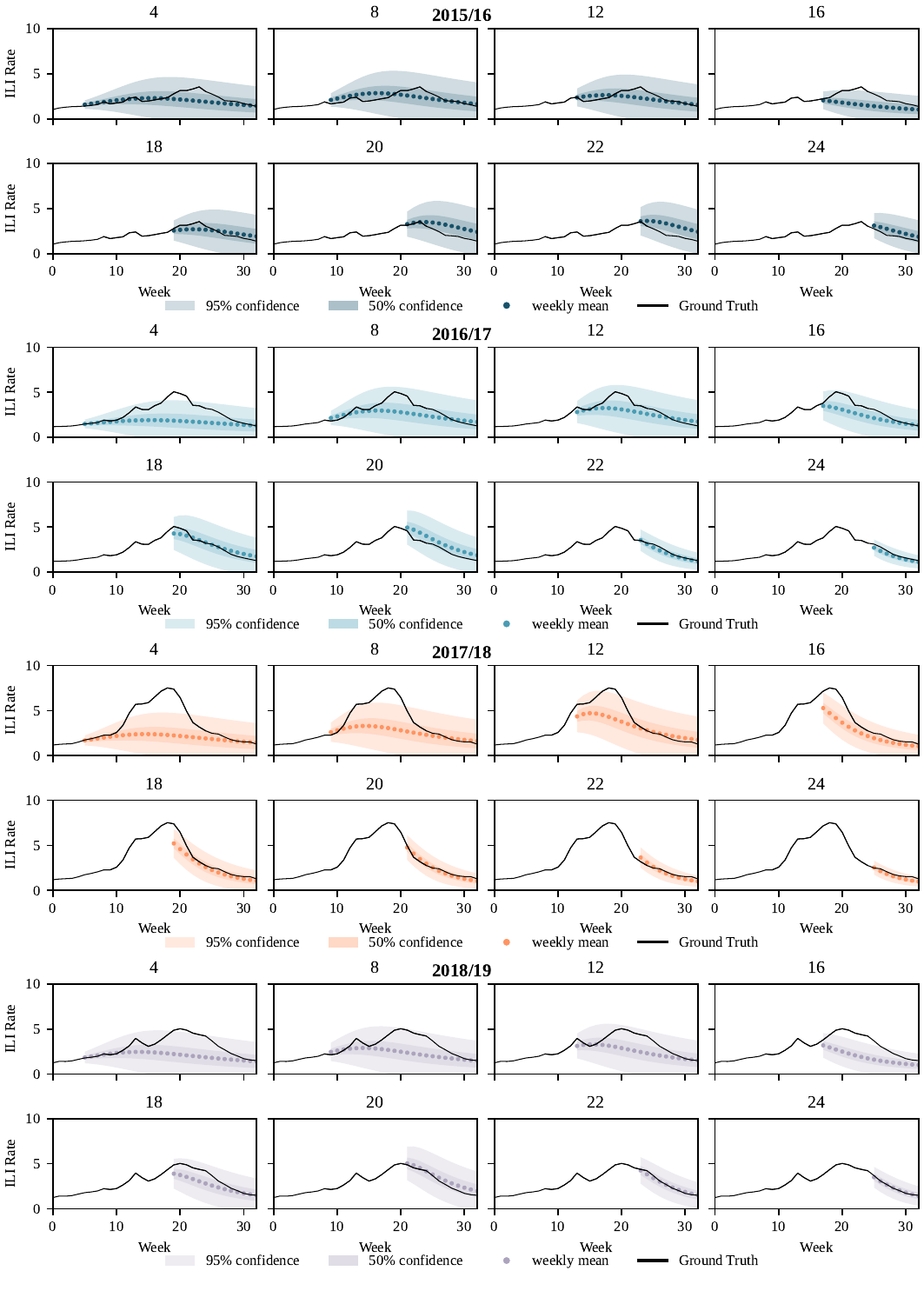}
    \caption[{\SIRNN~trajectories}]{\textbf{\SIRNN~trajectories}\newline
    Forecast trajectories for the \SIRNN~from a given epidemic week (indexed from week $40$ in the year) to the end of the season. Each subplot shows the model's forecast from the given epidemic week (starting a week $40$ in the year). Trajectories shown for the mean, $50\%$ and $90\%$ confidence intervals.
    }
    \label{fig:SIRNN_trajectories}
\end{figure*}

\begin{figure*}[!h]
    \centering
    \includegraphics[width=0.98\linewidth]{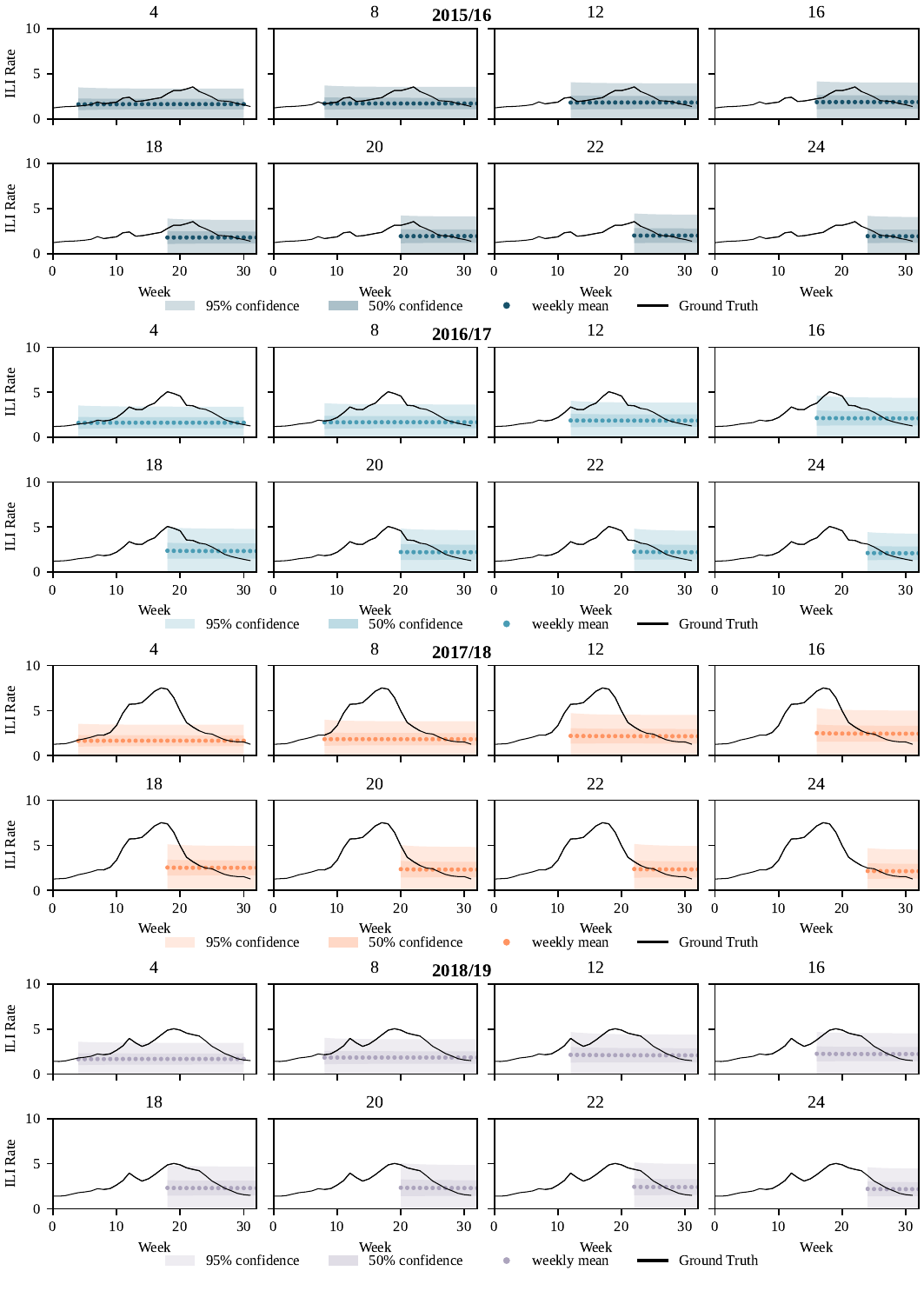}
    \caption[{\SIRQS~trajectories}]{\textbf{\SIRQS~trajectories}\newline
    Forecast trajectories for the \SIRQS~from a given epidemic week (indexed from week $40$ in the year) to the end of the season. Each subplot shows the model's forecast from the given epidemic week (starting a week $40$ in the year). Trajectories shown for the mean, $50\%$ and $90\%$ confidence intervals.
    }
    \label{fig:SIRQS_trajectories}
\end{figure*}

\begin{figure*}[!h]
    \centering
    \includegraphics[width=0.98\linewidth]{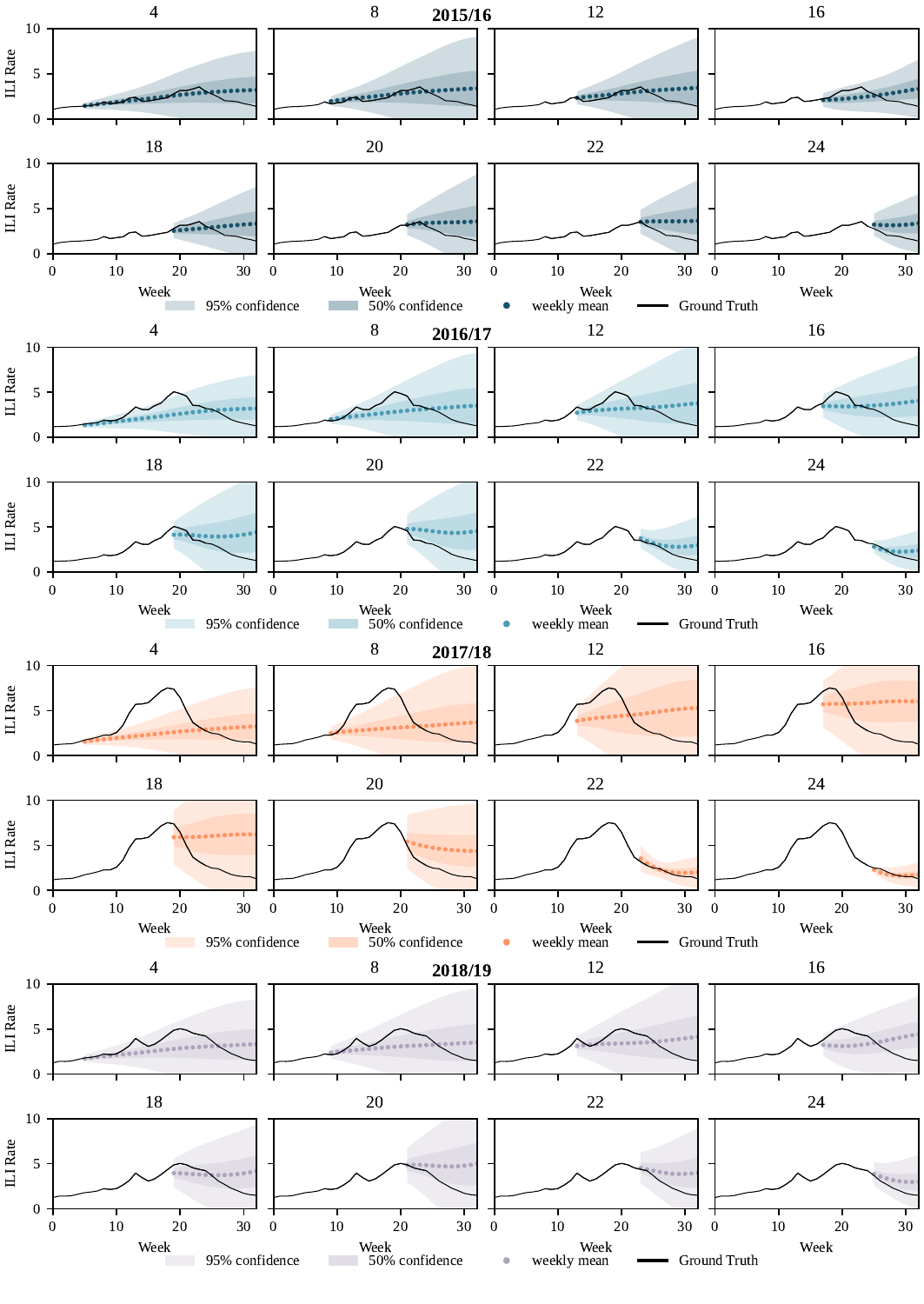}
    \caption[{\SEIRFA~trajectories}]{\textbf{\SEIRFA~trajectories}\newline
    Forecast trajectories for the \SEIRFA~from a given epidemic week (indexed from week $40$ in the year) to the end of the season. Each subplot shows the model's forecast from the given epidemic week (starting a week $40$ in the year). Trajectories shown for the mean, $50\%$ and $90\%$ confidence intervals.
    }
    \label{fig:SEIRFA_trajectories}
\end{figure*}

\clearpage
\section{Supplementary Results}
\begin{table}[h!]
    \centering
    \small
    \setlength{\tabcolsep}{2pt}
    \begin{tabular}{p{1.1cm}p{0.8cm}p{0.8cm}p{0.8cm}p{0.8cm}p{0.8cm}p{0.8cm}p{0.8cm}p{0.8cm}p{0.8cm}}
\ Horizon & Metric  &\multicolumn{8}{c}{Avg (2015-19)} \\ 
% \cmidrule(lr){3-9}\cmidrule(lr){10-16}
% \ Horizon & Metric  &\multicolumn{8}{c}{2015/16} & \multicolumn{8}{c}{2016/17} \\ 
\cmidrule(lr){3-10}
$\gamma$   &     &             \FA &           \FAQS &  \SIRSIMPLE &     \SIRNN &     \SIRQS &          \SIRFA &         \SEIRNN &    \SEIRFA \\
\midrule
7       &  Skill &            0.56 &  \textbf{0.67} &        0.32 &       0.50 &       0.23 &            0.51 &            0.56 &       0.57 \\
     &    NLL &            0.45 &  \textbf{0.37} &        1.15 &       0.68 &       1.58 &            0.61 &            0.46 &       0.46 \\
     &    MAE &            0.39 &  \textbf{0.34} &        0.72 &       0.38 &       0.85 &            0.37 &            0.35 &       0.36 \\
     &   $r$ &            0.93 &  \textbf{0.97} &        0.91 &       0.91 &       0.54 &            0.91 &            0.92 &       0.92 \\
\midrule
14      &  Skill &  \textbf{0.43} &            0.40 &        0.28 &       0.40 &       0.22 &            0.41 &  \textbf{0.43} &       0.43 \\
     &    NLL &  \textbf{0.83} &            1.42 &        1.31 &       0.95 &       1.59 &            0.90 &            0.84 &       0.83 \\
     &    MAE &            0.52 &            0.50 &        0.87 &       0.51 &       0.82 &            0.51 &  \textbf{0.49} &       0.51 \\
     &   $r$ &            0.85 &  \textbf{0.95} &        0.81 &       0.82 &       0.63 &            0.82 &            0.83 &       0.82 \\
\midrule
21      &  Skill &            0.34 &            0.21 &        0.24 &       0.34 &       0.22 &  \textbf{0.35} &  \textbf{0.35} &       0.35 \\
     &    NLL &            1.12 &            3.68 &        1.48 &       1.11 &       1.62 &  \textbf{1.06} &            1.08 &       1.10 \\
     &    MAE &            0.62 &            0.70 &        1.01 &       0.59 &       0.83 &  \textbf{0.58} &            0.59 &       0.66 \\
     &   $r$ &            0.77 &  \textbf{0.90} &        0.71 &       0.75 &       0.66 &            0.75 &            0.74 &       0.71 \\
\midrule
28      &  Skill &            0.29 &            0.14 &        0.21 &       0.32 &       0.21 &  \textbf{0.33} &            0.31 &       0.30 \\
     &    NLL &            1.30 &            6.62 &        1.66 &       1.18 &       1.68 &  \textbf{1.14} &            1.21 &       1.27 \\
     &    MAE &            0.69 &            0.86 &        1.14 &       0.63 &       0.89 &  \textbf{0.62} &            0.66 &       0.80 \\
     &   $r$ &            0.72 &  \textbf{0.83} &        0.62 &       0.72 &       0.60 &            0.73 &            0.70 &       0.60 \\
\bottomrule
\end{tabular}

    \caption[{Neural ODE performance metrics averaged for 2015-19}]{\textbf{Neural ODE performance metrics for 2015/16 and 2016/17} \newline Performance metrics for eight N-ODE models and four forecast horizons ($\gamma = 7$, $14$, $21$, and $28$ days ahead) over two flu seasons. 
    Skill and NLL compare the accuracy weighted by the uncertainty of forecasts. MAE is the mean absolute error, and $r$ is the bivariate correlation between forecasts and reported ILI rates. Best results for each metric and forecast horizon are shown in bold.}
    \label{tab:neuralodecomparison1519}
\end{table}

\begin{sidewaystable}[t]
    \centering
    \small
    \setlength{\tabcolsep}{2pt}
    \begin{tabular}{p{1.1cm}p{1.15cm}p{1.15cm}p{1.15cm}p{1.15cm}p{1.15cm}p{1.15cm}p{1.15cm}p{1.15cm}p{1.15cm}p{1.15cm}p{1.15cm}p{1.15cm}p{1.15cm}p{1.15cm}p{1.15cm}p{1.15cm}p{1.15cm}}
\ Horizon & Metric  &\multicolumn{8}{c}{2015/16} & \multicolumn{8}{c}{2016/17} \\ 
\cmidrule(lr){3-10}\cmidrule(lr){11-18}
% \cmidrule(lr){19-26}
$\gamma$   &     &     \FA &            \FAQS &  \SIRSIMPLE &     \SIRNN &     \SIRQS &     \SIRFA &    \SEIRNN &         \SEIRFA &             \FA &           \FAQS &  \SIRSIMPLE &          \SIRNN &          \SIRQS &          \SIRFA &     \SEIRNN &         \SEIRFA \\
\midrule
7       &  Skill &    0.72 &   \textbf{0.90} &        0.58 &       0.64 &       0.42 &       0.68 &       0.76 &            0.75 &            0.65 &  \textbf{0.66} &        0.45 &            0.56 &            0.36 &            0.56 &        0.61 &            0.63 \\
     &    NLL &    0.20 &  \textbf{-0.28} &        0.51 &       0.39 &       0.93 &       0.28 &       0.10 &            0.12 &  \textbf{0.28} &            0.43 &        0.81 &            0.53 &            1.07 &            0.49 &        0.36 &            0.33 \\
     &    MAE &    0.24 &   \textbf{0.16} &        0.21 &       0.23 &       0.41 &       0.22 &       0.20 &            0.20 &  \textbf{0.30} &            0.33 &        0.38 &            0.32 &            0.54 &            0.31 &        0.31 &  \textbf{0.30} \\
     &   $r$ &    0.90 &   \textbf{0.95} &        0.91 &       0.91 &       0.67 &       0.89 &       0.92 &            0.91 &            0.94 &  \textbf{0.97} &        0.91 &            0.91 &            0.77 &            0.91 &        0.92 &            0.92 \\
\midrule
14      &  Skill &    0.56 &   \textbf{0.90} &        0.54 &       0.54 &       0.42 &       0.56 &       0.60 &            0.61 &  \textbf{0.50} &            0.30 &        0.39 &            0.46 &            0.37 &            0.46 &        0.47 &            0.49 \\
     &    NLL &    0.56 &  \textbf{-0.23} &        0.59 &       0.62 &       0.92 &       0.56 &       0.46 &            0.44 &  \textbf{0.67} &            2.41 &        0.97 &            0.77 &            1.02 &            0.76 &        0.73 &            0.69 \\
     &    MAE &    0.37 &   \textbf{0.19} &        0.33 &       0.33 &       0.39 &       0.35 &       0.30 &            0.33 &  \textbf{0.42} &            0.45 &        0.52 &            0.44 &            0.48 &            0.44 &        0.45 &            0.44 \\
     &   $r$ &    0.82 &   \textbf{0.96} &        0.78 &       0.79 &       0.67 &       0.77 &       0.82 &            0.78 &            0.85 &  \textbf{0.92} &        0.80 &            0.82 &            0.83 &            0.82 &        0.83 &            0.83 \\
\midrule
21      &  Skill &    0.44 &   \textbf{0.80} &        0.51 &       0.47 &       0.40 &       0.47 &       0.49 &            0.52 &  \textbf{0.42} &            0.16 &        0.33 &            0.41 &            0.37 &            0.41 &        0.40 &            0.40 \\
     &    NLL &    0.84 &   \textbf{0.35} &        0.69 &       0.78 &       0.99 &       0.77 &       0.71 &            0.63 &  \textbf{0.89} &            5.24 &        1.19 &            0.90 &            1.01 &            0.90 &        0.92 &            0.92 \\
     &    MAE &    0.47 &   \textbf{0.24} &        0.44 &       0.41 &       0.47 &       0.41 &       0.37 &            0.44 &            0.48 &            0.56 &        0.66 &            0.52 &  \textbf{0.47} &            0.51 &        0.54 &            0.56 \\
     &   $r$ &    0.71 &   \textbf{0.90} &        0.61 &       0.66 &       0.53 &       0.64 &       0.70 &            0.63 &            0.78 &  \textbf{0.85} &        0.67 &            0.76 &            0.83 &            0.76 &        0.76 &            0.73 \\
\midrule
28      &  Skill &    0.38 &   \textbf{0.68} &        0.46 &       0.41 &       0.37 &       0.40 &       0.42 &            0.46 &  \textbf{0.40} &            0.12 &        0.27 &  \textbf{0.40} &            0.35 &            0.39 &        0.37 &            0.35 \\
     &    NLL &    1.03 &             1.03 &        0.82 &       0.92 &       1.08 &       0.97 &       0.89 &  \textbf{0.78} &  \textbf{0.94} &            7.86 &        1.41 &  \textbf{0.94} &            1.08 &            0.96 &        1.01 &            1.11 \\
     &    MAE &    0.58 &   \textbf{0.32} &        0.54 &       0.48 &       0.57 &       0.51 &       0.43 &            0.52 &  \textbf{0.47} &            0.66 &        0.82 &            0.51 &            0.57 &            0.51 &        0.56 &            0.72 \\
     &   $r$ &    0.58 &   \textbf{0.81} &        0.42 &       0.53 &       0.24 &       0.50 &       0.56 &            0.47 &  \textbf{0.78} &            0.77 &        0.53 &  \textbf{0.78} &            0.76 &  \textbf{0.78} &        0.77 &            0.65 \\
\bottomrule
\end{tabular}

    \caption[{Neural ODE performance metrics for 2015/16 and 2016/17}]{\textbf{Neural ODE performance metrics for 2015/16 and 2016/17} \newline Performance metrics for eight N-ODE models and four forecast horizons ($\gamma = 7$, $14$, $21$, and $28$ days ahead) over two flu seasons. 
    Skill and NLL compare the accuracy weighted by the uncertainty of forecasts. MAE is the mean absolute error, and $r$ is the bivariate correlation between forecasts and reported ILI rates. Best results for each metric and forecast horizon are shown in bold.}
    \label{tab:neuralodecomparison1517}
\end{sidewaystable}

\afterpage{
\begin{sidewaystable}[t]
    \centering
    \small
    \setlength{\tabcolsep}{2pt}
    \begin{tabular}{p{1.1cm}p{1.15cm}p{1.15cm}p{1.15cm}p{1.15cm}p{1.15cm}p{1.15cm}p{1.15cm}p{1.15cm}p{1.15cm}p{1.15cm}p{1.15cm}p{1.15cm}p{1.15cm}p{1.15cm}p{1.15cm}p{1.15cm}p{1.15cm}}
\ Horizon & Metric  &\multicolumn{8}{c}{2017/18} & \multicolumn{8}{c}{2018/19} \\ 
\cmidrule(lr){3-10}\cmidrule(lr){11-18}
% \cmidrule(lr){19-26}
$\gamma$   &     &     \FA &           \FAQS &      \SIRSIMPLE &          \SIRNN &     \SIRQS &          \SIRFA &         \SEIRNN &    \SEIRFA &             \FA &           \FAQS &  \SIRSIMPLE &      \SIRNN &      \SIRQS &          \SIRFA &     \SEIRNN &         \SEIRFA \\
\midrule
7       &  Skill &    0.37 &  \textbf{0.53} &            0.19 &            0.33 &       0.06 &            0.35 &            0.42 &       0.39 &            0.59 &  \textbf{0.63} &        0.23 &        0.50 &        0.27 &            0.51 &        0.52 &            0.57 \\
     &    NLL &    0.90 &            0.88 &            1.75 &            1.10 &       2.92 &            1.04 &  \textbf{0.80} &       0.91 &  \textbf{0.40} &            0.45 &        1.52 &        0.67 &        1.41 &            0.62 &        0.59 &            0.50 \\
     &    MAE &    0.72 &  \textbf{0.46} &            1.46 &            0.60 &       1.63 &            0.60 &            0.58 &       0.62 &  \textbf{0.29} &            0.41 &        0.82 &        0.36 &        0.82 &            0.34 &        0.32 &            0.33 \\
     &   $r$ &    0.96 &  \textbf{0.98} &            0.94 &            0.94 &       0.31 &            0.94 &            0.96 &       0.94 &            0.93 &  \textbf{0.97} &        0.86 &        0.89 &        0.43 &            0.90 &        0.89 &            0.91 \\
\midrule
14      &  Skill &    0.26 &            0.25 &            0.15 &            0.25 &       0.06 &            0.26 &  \textbf{0.31} &       0.27 &  \textbf{0.45} &            0.40 &        0.20 &        0.39 &        0.28 &            0.41 &        0.38 &            0.44 \\
     &    NLL &    1.33 &            2.17 &            1.98 &            1.42 &       3.04 &            1.37 &  \textbf{1.20} &       1.37 &  \textbf{0.77} &            1.32 &        1.68 &        0.98 &        1.36 &            0.90 &        0.99 &            0.83 \\
     &    MAE &    0.86 &  \textbf{0.68} &            1.64 &            0.78 &       1.62 &            0.77 &            0.76 &       0.82 &  \textbf{0.43} &            0.69 &        0.98 &        0.47 &        0.79 &            0.46 &        0.45 &            0.46 \\
     &   $r$ &    0.92 &  \textbf{0.96} &            0.92 &            0.90 &       0.47 &            0.90 &            0.90 &       0.87 &            0.81 &  \textbf{0.97} &        0.73 &        0.76 &        0.55 &            0.78 &        0.76 &            0.81 \\
\midrule
21      &  Skill &    0.20 &            0.09 &            0.12 &            0.22 &       0.05 &            0.23 &  \textbf{0.24} &       0.19 &            0.34 &            0.17 &        0.18 &        0.32 &        0.28 &            0.35 &        0.31 &  \textbf{0.36} \\
     &    NLL &    1.61 &            6.17 &            2.24 &            1.57 &       3.13 &  \textbf{1.49} &  \textbf{1.49} &       1.75 &            1.12 &            2.96 &        1.80 &        1.20 &        1.35 &  \textbf{1.07} &        1.20 &            1.08 \\
     &    MAE &    0.99 &            0.93 &            1.80 &  \textbf{0.90} &       1.60 &            0.91 &            0.92 &       1.06 &            0.53 &            1.07 &        1.13 &        0.52 &        0.79 &  \textbf{0.50} &        0.52 &            0.59 \\
     &   $r$ &    0.90 &  \textbf{0.92} &            0.91 &            0.89 &       0.62 &            0.89 &            0.85 &       0.77 &            0.69 &  \textbf{0.94} &        0.65 &        0.67 &        0.65 &            0.71 &        0.66 &            0.71 \\
\midrule
28      &  Skill &    0.17 &            0.04 &            0.09 &            0.22 &       0.05 &  \textbf{0.23} &            0.21 &       0.16 &            0.27 &            0.12 &        0.16 &        0.29 &        0.28 &  \textbf{0.33} &        0.29 &            0.31 \\
     &    NLL &    1.81 &           14.25 &            2.51 &            1.57 &       3.21 &  \textbf{1.48} &            1.64 &       1.97 &            1.42 &            3.35 &        1.89 &        1.30 &        1.36 &  \textbf{1.14} &        1.28 &            1.23 \\
     &    MAE &    1.13 &            1.17 &            1.92 &            1.01 &       1.61 &  \textbf{0.99} &            1.09 &       1.28 &            0.58 &            1.28 &        1.28 &        0.52 &        0.81 &  \textbf{0.48} &        0.55 &            0.69 \\
     &   $r$ &    0.91 &            0.85 &  \textbf{0.92} &            0.91 &       0.71 &            0.91 &            0.80 &       0.65 &            0.62 &  \textbf{0.89} &        0.61 &        0.66 &        0.68 &            0.71 &        0.65 &            0.64 \\
\bottomrule
\end{tabular}

    \caption[{Neural ODE performance metrics for 2017/18 and 2018/19}]{\textbf{Neural ODE performance metrics for 2015/16 and 2016/17} \newline Performance metrics for eight N-ODE models and four forecast horizons ($\gamma = 7$, $14$, $21$, and $28$ days ahead) over two flu seasons. 
    Skill and NLL compare the accuracy weighted by the uncertainty of forecasts. MAE is the mean absolute error, and $r$ is the bivariate correlation between forecasts and reported ILI rates. Best results for each metric and forecast horizon are shown in bold.}
    \label{tab:neuralodecomparison1719}
\end{sidewaystable}
}

\addcontentsline{toc}{chapter}{Bibliography}
\bibliography{bibliography}

\begin{thebibliography}{100}

\bibitem{patz2005impact}
Jonathan~A Patz, Diarmid Campbell-Lendrum, Tracey Holloway, and Jonathan~A Foley.
\newblock Impact of regional climate change on human health.
\newblock {\em Nature}, 438(7066):310--317, 2005.

\bibitem{lampos2017enhancing}
Vasileios Lampos, Bin Zou, and Ingemar~Johansson Cox.
\newblock {Enhancing feature selection using word embeddings: The case of flu surveillance}.
\newblock In {\em Proceedings of the 26th International Conference on World Wide Web}, pages 695--704, 2017.

\bibitem{osthus2019dynamic}
Dave Osthus, James Gattiker, Reid Priedhorsky, and Sara~Y Del~Valle.
\newblock {Dynamic Bayesian influenza forecasting in the United States with hierarchical discrepancy (with discussion)}.
\newblock {\em Bayesian Analysis}, 14(1):261--312, 2019.

\bibitem{rackauckas2020universal}
Christopher Rackauckas, Yingbo Ma, Julius Martensen, Collin Warner, Kirill Zubov, Rohit Supekar, Dominic Skinner, Ali Ramadhan, and Alan Edelman.
\newblock Universal differential equations for scientific machine learning.
\newblock {\em arXiv preprint arXiv:2001.04385}, 2020.

\bibitem{shaman2013real}
Jeffrey Shaman, Alicia Karspeck, Wan Yang, James Tamerius, and Marc Lipsitch.
\newblock {Real-time influenza forecasts during the 2012--2013 season}.
\newblock {\em Nature communications}, 4(1):2837, 2013.

\bibitem{reich2019collaborative}
Nicholas~G Reich, Logan~C Brooks, Spencer~J Fox, Sasikiran Kandula, Craig~J McGowan, Evan Moore, Dave Osthus, Evan~L Ray, Abhinav Tushar, Teresa~K Yamana, et~al.
\newblock {A collaborative multiyear, multimodel assessment of seasonal influenza forecasting in the United States}.
\newblock {\em PNAS}, 116(8):3146--3154, 2019.

\bibitem{zhang2005neural}
G~Peter Zhang and Min Qi.
\newblock {Neural network forecasting for seasonal and trend time series}.
\newblock {\em European journal of operational research}, 160(2):501--514, 2005.

\bibitem{khashei2010artificial}
Mehdi Khashei and Mehdi Bijari.
\newblock {An artificial neural network (p, d, q) model for timeseries forecasting}.
\newblock {\em Expert Systems with applications}, 37(1):479--489, 2010.

\bibitem{kaastra1996designing}
Iebeling Kaastra and Milton Boyd.
\newblock {Designing a neural network for forecasting financial and economic time series}.
\newblock {\em Neurocomputing}, 10(3):215--236, 1996.

\bibitem{volkova2017forecasting}
Svitlana Volkova, Ellyn Ayton, Katherine Porterfield, and Courtney~D Corley.
\newblock Forecasting influenza-like illness dynamics for military populations using neural networks and social media.
\newblock {\em PLOS ONE}, 12(12):e0188941, 2017.

\bibitem{venna2018novel}
Siva~R Venna, Amirhossein Tavanaei, Raju~N Gottumukkala, Vijay~V Raghavan, Anthony~S Maida, and Stephen Nichols.
\newblock {A novel data-driven model for real-time influenza forecasting}.
\newblock {\em IEEE Access}, 7:7691--7701, 2018.

\bibitem{aiken2019towards}
Emily~L Aiken, Andre~T Nguyen, and Mauricio Santillana.
\newblock Towards the use of neural networks for influenza prediction at multiple spatial resolutions.
\newblock {\em arXiv preprint arXiv:1911.02673}, 2019.

\bibitem{der2009aleatory}
Armen Der~Kiureghian and Ove Ditlevsen.
\newblock {Aleatory or epistemic? Does it matter?}
\newblock {\em Structural Safety}, 31(2):105--112, 2009.

\bibitem{centers2019flusight}
Centers for Disease~Control, Prevention, et~al.
\newblock {FluSight: flu forecasting}, 2019.

\bibitem{cdc_flu}
{U.S. influenza surveillance: Purpose and methods}, 2022.

\bibitem{morris2023neural}
Michael Morris, Peter Hayes, Ingemar~J Cox, and Vasileios Lampos.
\newblock Neural network models for influenza forecasting with associated uncertainty using web search activity trends.
\newblock {\em PLOS Computational Biology}, 19(8):e1011392, 2023.

\bibitem{osthus2021multiscale}
Dave Osthus and Kelly~R Moran.
\newblock {Multiscale influenza forecasting}.
\newblock {\em Nature communications}, 12(1):1--11, 2021.

\bibitem{nixon2022real}
Kristen Nixon, Sonia Jindal, Felix Parker, Maximilian Marshall, Nicholas~G Reich, Kimia Ghobadi, Elizabeth~C Lee, Shaun Truelove, and Lauren Gardner.
\newblock {Real-time COVID-19 forecasting: challenges and opportunities of model performance and translation}.
\newblock {\em The Lancet Digital Health}, 4(10):e699--e701, 2022.

\bibitem{Ioannidis2022}
John~P.A. Ioannidis, Sally Cripps, and Martin~A. Tanner.
\newblock {Forecasting for COVID-19 has failed}.
\newblock {\em Int. J. Forecast.}, 38(2):423--438, 2022.

\bibitem{gibson2020real}
Graham~C Gibson, Nicholas~G Reich, and Daniel Sheldon.
\newblock Real-time mechanistic bayesian forecasts of covid-19 mortality.
\newblock {\em medRxiv}, 2020.

\bibitem{tagasovska2019single}
Natasa Tagasovska and David Lopez-Paz.
\newblock {Single-model uncertainties for deep learning}.
\newblock In {\em Advances in Neural Information Processing Systems}, pages 6417--6428, 2019.

\bibitem{shaman2012forecasting}
Jeffrey Shaman and Alicia Karspeck.
\newblock Forecasting seasonal outbreaks of influenza.
\newblock {\em PNAS}, 109(50):20425--20430, 2012.

\bibitem{osthus2022fast}
Dave Osthus.
\newblock {Fast and accurate influenza forecasting in the United States with Inferno}.
\newblock {\em PLOS Comput. Biol.}, 18(1), 2022.

\bibitem{gal2016uncertainty}
Yarin Gal.
\newblock {Uncertainty in deep learning}.
\newblock {\em University of Cambridge}, 2016.

\bibitem{kendall2017uncertainties}
Alex Kendall and Yarin Gal.
\newblock {What uncertainties do we need in Bayesian deep learning for computer vision?}
\newblock In {\em Advances in Neural Information Processing Systems}, pages 5574--5584, 2017.

\bibitem{o2004dicing}
Tony O'Hagan.
\newblock {Dicing with the unknown}.
\newblock {\em Significance}, 1(3):132--133, 2004.

\bibitem{bishop2006pattern}
Christopher~M Bishop.
\newblock {\em {Pattern Recognition and Machine Learning}}.
\newblock Springer, 2006.

\bibitem{rao1971further}
C~Radhakrishna Rao and Sujit~Kumar Mitra.
\newblock Further contributions to the theory of generalized inverse of matrices and its applications.
\newblock {\em Sankhy{\=a}: The Indian Journal of Statistics, Series A}, pages 289--300, 1971.

\bibitem{van1996matrix}
Charles~F Van~Loan and G~Golub.
\newblock Matrix computations (johns hopkins studies in mathematical sciences).
\newblock {\em Matrix Computations}, 5, 1996.

\bibitem{goodfellow2016deep}
Ian Goodfellow, Yoshua Bengio, and Aaron Courville.
\newblock {\em Deep learning}.
\newblock MIT press, 2016.

\bibitem{rumelhart1986learning}
David~E Rumelhart, Geoffrey~E Hinton, and Ronald~J Williams.
\newblock {Learning representations by back-propagating errors}.
\newblock {\em nature}, 323(6088):533--536, 1986.

\bibitem{zhang2003time}
G~Peter Zhang.
\newblock {Time series forecasting using a hybrid ARIMA and neural network model}.
\newblock {\em Neurocomputing}, 50:159--175, 2003.

\bibitem{alom2019state}
Md~Zahangir Alom, Tarek~M Taha, Chris Yakopcic, Stefan Westberg, Paheding Sidike, Mst~Shamima Nasrin, Mahmudul Hasan, Brian~C Van~Essen, Abdul~AS Awwal, and Vijayan~K Asari.
\newblock {A State-of-the-Art Survey on Deep Learning Theory and Architectures}.
\newblock {\em Electronics}, 8(3), 2019.

\bibitem{elman1990finding}
Jeffrey~L Elman.
\newblock Finding structure in time.
\newblock {\em Cognitive science}, 14(2):179--211, 1990.

\bibitem{bengio1994learning}
Yoshua Bengio, Patrice Simard, and Paolo Frasconi.
\newblock Learning long-term dependencies with gradient descent is difficult.
\newblock {\em IEEE transactions on neural networks}, 5(2):157--166, 1994.

\bibitem{hochreiter1997long}
Sepp Hochreiter and J{\"u}rgen Schmidhuber.
\newblock {Long short-term memory}.
\newblock {\em Neural computation}, 9(8):1735--1780, 1997.

\bibitem{cho2014properties}
Kyunghyun Cho, Bart Van~Merri{\"e}nboer, Dzmitry Bahdanau, and Yoshua Bengio.
\newblock {On the properties of neural machine translation: Encoder-decoder approaches}.
\newblock {\em arXiv preprint arXiv:1409.1259}, 2014.

\bibitem{pascanu2013difficulty}
Razvan Pascanu, Tomas Mikolov, and Yoshua Bengio.
\newblock On the difficulty of training recurrent neural networks.
\newblock In {\em International conference on machine learning}, pages 1310--1318. Pmlr, 2013.

\bibitem{gers2001lstm}
Felix~A Gers and E~Schmidhuber.
\newblock Lstm recurrent networks learn simple context-free and context-sensitive languages.
\newblock {\em IEEE transactions on neural networks}, 12(6):1333--1340, 2001.

\bibitem{taylor2000quantile}
James~W Taylor.
\newblock A quantile regression neural network approach to estimating the conditional density of multiperiod returns.
\newblock {\em Journal of Forecasting}, 19(4):299--311, 2000.

\bibitem{shafer2008tutorial}
Glenn Shafer and Vladimir Vovk.
\newblock A tutorial on conformal prediction.
\newblock {\em Journal of Machine Learning Research}, 9(3), 2008.

\bibitem{blei2017variational}
David~M Blei, Alp Kucukelbir, and Jon~D McAuliffe.
\newblock {Variational inference: a review for statisticians}.
\newblock {\em Journal of the American Statistical Association}, 112(518):859--877, 2017.

\bibitem{hershey2007approximating}
John~R Hershey and Peder~A Olsen.
\newblock Approximating the kullback leibler divergence between gaussian mixture models.
\newblock In {\em 2007 IEEE International Conference on Acoustics, Speech and Signal Processing-ICASSP'07}, volume~4, pages IV--317. IEEE, 2007.

\bibitem{srivastava2014dropout}
Nitish Srivastava, Geoffrey Hinton, Alex Krizhevsky, Ilya Sutskever, and Ruslan Salakhutdinov.
\newblock {Dropout: a simple way to prevent neural networks from overfitting}.
\newblock {\em JMLR}, 15(1):1929--1958, 2014.

\bibitem{gal2016dropout}
Yarin Gal and Zoubin Ghahramani.
\newblock {Dropout as a Bayesian approximation: Representing model uncertainty in deep learning}.
\newblock In {\em ICML}, pages 1050--1059, 2016.

\bibitem{gal2016theoretically}
Yarin Gal and Zoubin Ghahramani.
\newblock {A Theoretically Grounded Application of Dropout in Recurrent Neural Networks}.
\newblock In {\em Advances in Neural Information Processing Systems}, volume~29, pages 1019--1027, 2016.

\bibitem{kendall2015Bayesian}
Alex Kendall, Vijay Badrinarayanan, and Roberto Cipolla.
\newblock {Bayesian segnet: Model uncertainty in deep convolutional encoder-decoder architectures for scene understanding}.
\newblock {\em arXiv preprint arXiv:1511.02680}, 2015.

\bibitem{osband2016risk}
Ian Osband.
\newblock {Risk versus uncertainty in deep learning: Bayes, bootstrap and the dangers of dropout}.
\newblock In {\em NIPS Workshop on Bayesian Deep Learning}, volume 192, 2016.

\bibitem{hron2017variational}
Jiri Hron, Alexander G de~G Matthews, and Zoubin Ghahramani.
\newblock {Variational Gaussian dropout is not Bayesian}.
\newblock {\em arXiv preprint arXiv:1711.02989}, 2017.

\bibitem{bishop1994mixture}
Christopher~M Bishop.
\newblock {Mixture Density Networks}.
\newblock Technical report, Aston University, 1994.

\bibitem{graves2011practical}
Alex Graves.
\newblock {Practical Variational Inference for Neural Networks}.
\newblock In {\em Advances in Neural Information Processing Systems}, pages 2348--2356, 2011.

\bibitem{kingma2014adam}
Diederik~P Kingma and Jimmy Ba.
\newblock Adam: A method for stochastic optimization.
\newblock {\em arXiv preprint arXiv:1412.6980}, 2014.

\bibitem{blundell2015weight}
Charles Blundell, Julien Cornebise, Koray Kavukcuoglu, and Daan Wierstra.
\newblock {Weight uncertainty in neural networks}.
\newblock {\em arXiv preprint arXiv:1505.05424}, 2015.

\bibitem{ioffe2015batch}
Sergey Ioffe and Christian Szegedy.
\newblock Batch normalization: Accelerating deep network training by reducing internal covariate shift.
\newblock {\em arXiv preprint arXiv:1502.03167}, 2015.

\bibitem{loshchilov2016sgdr}
Ilya Loshchilov and Frank Hutter.
\newblock {SGDR: Stochastic gradient descent with warm restarts}.
\newblock {\em arXiv preprint arXiv:1608.03983}, 2016.

\bibitem{goyal2017accurate}
Priya Goyal, Piotr Doll{\'a}r, Ross Girshick, Pieter Noordhuis, Lukasz Wesolowski, Aapo Kyrola, Andrew Tulloch, Yangqing Jia, and Kaiming He.
\newblock {Accurate, large minibatch SGD: Training imagenet in 1 hour}.
\newblock {\em arXiv preprint arXiv:1706.02677}, 2017.

\bibitem{osawa2019practical}
Kazuki Osawa, Siddharth Swaroop, Mohammad Emtiyaz~E Khan, Anirudh Jain, Runa Eschenhagen, Richard~E Turner, and Rio Yokota.
\newblock {Practical deep learning with Bayesian principles}.
\newblock In {\em Advances in Neural Information Processing Systems}, volume~32, pages 4287--4299, 2019.

\bibitem{koenker2001quantile}
Roger Koenker and Kevin~F Hallock.
\newblock {Quantile regression}.
\newblock {\em Journal of economic perspectives}, 15(4):143--156, 2001.

\bibitem{gasthaus2019probabilistic}
Jan Gasthaus, Konstantinos Benidis, Yuyang Wang, Syama~Sundar Rangapuram, David Salinas, Valentin Flunkert, and Tim Januschowski.
\newblock Probabilistic forecasting with spline quantile function rnns.
\newblock In {\em The 22nd international conference on artificial intelligence and statistics}, pages 1901--1910. PMLR, 2019.

\bibitem{wen2017multi}
Ruofeng Wen, Kari Torkkola, Balakrishnan Narayanaswamy, and Dhruv Madeka.
\newblock A multi-horizon quantile recurrent forecaster.
\newblock {\em arXiv preprint arXiv:1711.11053}, 2017.

\bibitem{box2015time}
George~EP Box, Gwilym~M Jenkins, Gregory~C Reinsel, and Greta~M Ljung.
\newblock {\em {Time series analysis: forecasting and control}}.
\newblock John Wiley \& Sons, 2015.

\bibitem{kwiatkowski1992testing}
Denis Kwiatkowski, Peter~CB Phillips, Peter Schmidt, and Yongcheol Shin.
\newblock {Testing the null hypothesis of stationarity against the alternative of a unit root: How sure are we that economic time series have a unit root?}
\newblock {\em Journal of econometrics}, 54(1-3):159--178, 1992.

\bibitem{soebiyanto2010modeling}
Radina~P Soebiyanto, Farida Adimi, and Richard~K Kiang.
\newblock {Modeling and predicting seasonal influenza transmission in warm regions using climatological parameters}.
\newblock {\em PLOS one}, 5(3):e9450, 2010.

\bibitem{cheung1995lag}
Yin-Wong Cheung and Kon~S Lai.
\newblock {Lag order and critical values of the augmented Dickey--Fuller test}.
\newblock {\em Journal of Business \& Economic Statistics}, 13(3):277--280, 1995.

\bibitem{shaman2010absolute}
Jeffrey Shaman, Virginia~E Pitzer, C{\'e}cile Viboud, Bryan~T Grenfell, and Marc Lipsitch.
\newblock {Absolute humidity and the seasonal onset of influenza in the continental United States}.
\newblock {\em PLOS biology}, 8(2):e1000316, 2010.

\bibitem{he2018epidemiology}
Zhirui He and Hongbing Tao.
\newblock {Epidemiology and ARIMA model of positive-rate of influenza viruses among children in Wuhan, China: A nine-year retrospective study}.
\newblock {\em International Journal of Infectious Diseases}, 74:61--70, 2018.

\bibitem{ray2017infectious}
Evan~L Ray, Krzysztof Sakrejda, Stephen~A Lauer, Michael~A Johansson, and Nicholas~G Reich.
\newblock {Infectious disease prediction with kernel conditional density estimation}.
\newblock {\em Statistics in medicine}, 36(30):4908--4929, 2017.

\bibitem{ray2018prediction}
Evan~L Ray and Nicholas~G Reich.
\newblock {Prediction of infectious disease epidemics via weighted density ensembles}.
\newblock {\em PLOS computational biology}, 14(2):e1005910, 2018.

\bibitem{brooks2018nonmechanistic}
Logan~C Brooks, David~C Farrow, Sangwon Hyun, Ryan~J Tibshirani, and Roni Rosenfeld.
\newblock {Nonmechanistic forecasts of seasonal influenza with iterative one-week-ahead distributions}.
\newblock {\em PLOS computational biology}, 14(6):e1006134, 2018.

\bibitem{ibe2013elements}
Oliver~C Ibe.
\newblock {\em Elements of random walk and diffusion processes}.
\newblock John Wiley \& Sons, 2013.

\bibitem{ginsberg2009detecting}
Jeremy Ginsberg, Matthew~H Mohebbi, Rajan~S Patel, Lynnette Brammer, Mark~S Smolinski, and Larry Brilliant.
\newblock {Detecting influenza epidemics using search engine query data}.
\newblock {\em Nature}, 457(7232):1012--1014, 2009.

\bibitem{santillana2014can}
Mauricio Santillana, D~Wendong Zhang, Benjamin~M Althouse, and John~W Ayers.
\newblock What can digital disease detection learn from (an external revision to) google flu trends?
\newblock {\em American journal of preventive medicine}, 47(3):341--347, 2014.

\bibitem{lampos2015advances}
Vasileios Lampos, Andrew~C Miller, Steve Crossan, and Christian Stefansen.
\newblock Advances in nowcasting influenza-like illness rates using search query logs.
\newblock {\em Scientific reports}, 5(1):1--10, 2015.

\bibitem{culotta2010towards}
A.~Culotta.
\newblock {Towards Detecting Influenza Epidemics by Analyzing Twitter Messages}.
\newblock In {\em Proc. of the 1st Workshop on Social Media Analytics}, pages 115--122, 2010.

\bibitem{aramaki2011twitter}
Eiji Aramaki, Sachiko Maskawa, and Mizuki Morita.
\newblock {Twitter catches the flu: detecting influenza epidemics using Twitter}.
\newblock In {\em Proc. of the 2011 Conference on empirical methods in natural language processing}, pages 1568--1576, 2011.

\bibitem{paul2014twitter}
Michael~J Paul, Mark Dredze, and David Broniatowski.
\newblock {Twitter improves influenza forecasting}.
\newblock {\em PLOS Curr.}, 6, 2014.

\bibitem{wagner2018added}
Moritz Wagner, Vasileios Lampos, Ingemar~J Cox, and Richard Pebody.
\newblock {The added value of online user-generated content in traditional methods for influenza surveillance}.
\newblock {\em Scientific Reports}, 8(13963), 2018.

\bibitem{Lampos2017WWW}
Vasileios Lampos, Bin Zou, and Ingemar~J. Cox.
\newblock {Enhancing feature selection using word embeddings: The case of flu surveillance}.
\newblock In {\em Proc. of the 26th International World Wide Web Conference}, pages 695--704, 2017.

\bibitem{lampos2021covid}
Vasileios Lampos, Maimuna~S. Majumder, Elad Yom-Tov, Michael Edelstein, Simon Moura, Yohhei Hamada, Molebogeng~X. Rangaka, Rachel~A. McKendry, and Ingemar~J. Cox.
\newblock {Tracking COVID-19 using online search}.
\newblock {\em npj Digital Medicine}, 4(17), 2021.

\bibitem{shahriari2015taking}
Bobak Shahriari, Kevin Swersky, Ziyu Wang, Ryan~P Adams, and Nando De~Freitas.
\newblock {Taking the human out of the loop: A review of Bayesian optimization}.
\newblock {\em Proc. IEEE}, 104(1):148--175, 2015.

\bibitem{jospin2020hands}
Laurent~Valentin Jospin, Hamid Laga, Farid Boussaid, Wray Buntine, and Mohammed Bennamoun.
\newblock {Hands-On Bayesian Neural Networks—A Tutorial for Deep Learning Users}.
\newblock {\em IEEE CIM}, 17(2):29--48, 2022.

\bibitem{zou2005regularization}
Hui Zou and Trevor Hastie.
\newblock Regularization and variable selection via the elastic net.
\newblock {\em Journal of the royal statistical society: series B (statistical methodology)}, 67(2):301--320, 2005.

\bibitem{Lampos2015GFT}
Vasileios Lampos, Andrew~C. Miller, Steve Crossan, and Christian Stefansen.
\newblock {Advances in nowcasting influenza-like illness rates using search query logs}.
\newblock {\em Sci. Rep.}, 5:12760, 2015.

\bibitem{Rasmussen2006}
Carl~E. Rasmussen and Christopher K.~I. Williams.
\newblock {\em {Gaussian Processes for Machine Learning}}.
\newblock MIT Press, 2006.

\bibitem{Zou2018www}
Bin Zou, Vasileios Lampos, and Ingemar~J. Cox.
\newblock {Multi-Task Learning Improves Disease Models from Web Search}.
\newblock In {\em WWW}, pages 87--96, 2018.

\bibitem{nsoesie2013forecasting}
Elaine Nsoesie, Madhav Mararthe, and John Brownstein.
\newblock {Forecasting peaks of seasonal influenza epidemics}.
\newblock {\em PLOS Curr.}, 5, 2013.

\bibitem{Wagner2018}
Moritz Wagner, Vasileios Lampos, Ingemar~J. Cox, and Richard Pebody.
\newblock {The added value of online user-generated content in traditional methods for influenza surveillance}.
\newblock {\em Sci. Rep.}, 8(13963), 2018.

\bibitem{hernandez2015probabilistic}
Jos{\'e}~Miguel Hern{\'a}ndez-Lobato and Ryan Adams.
\newblock Probabilistic backpropagation for scalable learning of bayesian neural networks.
\newblock In {\em ICML}, pages 1861--1869, 2015.

\bibitem{yang2015inference}
Wan Yang, Marc Lipsitch, and Jeffrey Shaman.
\newblock {Inference of seasonal and pandemic influenza transmission dynamics}.
\newblock {\em PNAS}, 112(9):2723--2728, 2015.

\bibitem{baltrusaitis2018evaluation}
Kristin Baltrusaitis, Kathleen Noddin, Colleen Nguyen, Adam Crawley, John~S Brownstein, and Laura~F White.
\newblock {Evaluation of approaches that adjust for biases in participatory surveillance systems}.
\newblock {\em Online J. Public Health Inform.}, 10(1), 2018.

\bibitem{UKHSA2023}
""~UK~Health Security~Agency.
\newblock {Weekly national Influenza and COVID-19 surveillance reports}.
\newblock {\em Official Statistics (UKHSA)}, 2023.

\bibitem{clemente2019improved}
Leonardo Clemente, Fred Lu, Mauricio Santillana, et~al.
\newblock {Improved real-time influenza surveillance: using internet search data in eight Latin American countries}.
\newblock {\em JPHS}, 5(2):e12214, 2019.

\bibitem{Zou2019}
Bin Zou, Vasileios Lampos, and Ingemar~J. Cox.
\newblock {Transfer Learning for Unsupervised Influenza-like Illness Models from Online Search Data}.
\newblock In {\em WWW}, pages 2505--2516, 2019.

\bibitem{ning2019accurate}
Shaoyang Ning, Shihao Yang, and SC~Kou.
\newblock {Accurate regional influenza epidemics tracking using Internet search data}.
\newblock {\em Sci. Rep.}, 9(5238), 2019.

\bibitem{kingma2013auto}
Diederik~P Kingma and Max Welling.
\newblock {Auto-encoding variational bayes}.
\newblock {\em arXiv preprint arXiv:1312.6114}, 2013.

\bibitem{karpatne2017physics}
Anuj Karpatne, William Watkins, Jordan Read, and Vipin Kumar.
\newblock {Physics-guided neural networks (pgnn): An application in lake temperature modeling}.
\newblock {\em arXiv preprint arXiv:1710.11431}, 2, 2017.

\bibitem{burnham2004multimodel}
Kenneth~P Burnham and David~R Anderson.
\newblock Multimodel inference: understanding aic and bic in model selection.
\newblock {\em Sociological methods \& research}, 33(2):261--304, 2004.

\bibitem{chang2021mobility}
Serina Chang, Emma Pierson, Pang~Wei Koh, Jaline Gerardin, Beth Redbird, David Grusky, and Jure Leskovec.
\newblock Mobility network models of covid-19 explain inequities and inform reopening.
\newblock {\em Nature}, 589(7840):82--87, 2021.

\bibitem{chen2018neural}
Ricky~TQ Chen, Yulia Rubanova, Jesse Bettencourt, and David Duvenaud.
\newblock {Neural ordinary differential equations}.
\newblock {\em arXiv preprint arXiv:1806.07366}, 2018.

\bibitem{shaman2009absolute}
Jeffrey Shaman and Melvin Kohn.
\newblock Absolute humidity modulates influenza survival, transmission, and seasonality.
\newblock {\em Proceedings of the National Academy of Sciences}, 106(9):3243--3248, 2009.

\bibitem{ferguson2006strategies}
Neil~M Ferguson, Derek~AT Cummings, Christophe Fraser, James~C Cajka, Philip~C Cooley, and Donald~S Burke.
\newblock {Strategies for mitigating an influenza pandemic}.
\newblock {\em Nature}, 442(7101):448--452, 2006.

\bibitem{ferguson2020impact}
Neil~M Ferguson, Daniel Laydon, Gemma Nedjati-Gilani, Natsuko Imai, Kylie Ainslie, Marc Baguelin, Sangeeta Bhatia, Adhiratha Boonyasiri, Zulma Cucunub{\'a}, Gina Cuomo-Dannenburg, et~al.
\newblock Impact of non-pharmaceutical interventions (npis) to reduce covid-19 mortality and healthcare demand. 2020.
\newblock {\em DOI}, 10:77482, 2020.

\bibitem{keeling2005networks}
Matt~J Keeling and Ken~TD Eames.
\newblock Networks and epidemic models.
\newblock {\em Journal of the royal society interface}, 2(4):295--307, 2005.

\bibitem{kermack1927contribution}
William~Ogilvy Kermack and Anderson~G McKendrick.
\newblock A contribution to the mathematical theory of epidemics.
\newblock {\em Proceedings of the royal society of london. Series A, Containing papers of a mathematical and physical character}, 115(772):700--721, 1927.

\bibitem{weiss2013sir}
Howard~Howie Weiss.
\newblock The sir model and the foundations of public health.
\newblock {\em Materials matematics}, pages 0001--17, 2013.

\bibitem{anderson1992infectious}
Roy~M Anderson, B~Anderson, and Robert~M May.
\newblock {\em Infectious diseases of humans: dynamics and control}.
\newblock Oxford university press, 1992.

\bibitem{bjornstad2020modeling}
Ottar~N Bj{\o}rnstad, Katriona Shea, Martin Krzywinski, and Naomi Altman.
\newblock Modeling infectious epidemics.
\newblock {\em Nat. Methods}, 17(5):455--456, 2020.

\bibitem{yang2014comparison}
Wan Yang, Alicia Karspeck, and Jeffrey Shaman.
\newblock {Comparison of filtering methods for the modeling and retrospective forecasting of influenza epidemics}.
\newblock {\em PLOS computational biology}, 10(4):e1003583, 2014.

\bibitem{balcan2009seasonal}
Duygu Balcan, Hao Hu, Bruno Goncalves, Paolo Bajardi, Chiara Poletto, Jose~J Ramasco, Daniela Paolotti, Nicola Perra, Michele Tizzoni, Wouter Van~den Broeck, et~al.
\newblock {Seasonal transmission potential and activity peaks of the new influenza A (H1N1): a Monte Carlo likelihood analysis based on human mobility}.
\newblock {\em BMC medicine}, 7(1):1--12, 2009.

\bibitem{tizzoni2012real}
Michele Tizzoni, Paolo Bajardi, Chiara Poletto, Jos{\'e}~J Ramasco, Duygu Balcan, Bruno Gon{\c{c}}alves, Nicola Perra, Vittoria Colizza, and Alessandro Vespignani.
\newblock {Real-time numerical forecast of global epidemic spreading: case study of 2009 A/H1N1pdm}.
\newblock {\em BMC medicine}, 10(1):165, 2012.

\bibitem{poli2020hypersolvers}
Michael Poli, Stefano Massaroli, Atsushi Yamashita, Hajime Asama, and Jinkyoo Park.
\newblock Hypersolvers: Toward fast continuous-depth models.
\newblock {\em Advances in Neural Information Processing Systems}, 33:21105--21117, 2020.

\bibitem{yang2019pointflow}
Guandao Yang, Xun Huang, Zekun Hao, Ming-Yu Liu, Serge Belongie, and Bharath Hariharan.
\newblock Pointflow: 3d point cloud generation with continuous normalizing flows.
\newblock In {\em Proceedings of the IEEE/CVF international conference on computer vision}, pages 4541--4550, 2019.

\bibitem{rubanova2019latent}
Yulia Rubanova, Ricky~TQ Chen, and David Duvenaud.
\newblock {Latent odes for irregularly-sampled time series}.
\newblock {\em arXiv preprint arXiv:1907.03907}, 2019.

\bibitem{de2019gru}
Edward De~Brouwer, Jaak Simm, Adam Arany, and Yves Moreau.
\newblock Gru-ode-bayes: Continuous modeling of sporadically-observed time series.
\newblock {\em Advances in neural information processing systems}, 32, 2019.

\bibitem{NEURIPS2020_4a5876b4}
Patrick Kidger, James Morrill, James Foster, and Terry Lyons.
\newblock Neural controlled differential equations for irregular time series.
\newblock In H.~Larochelle, M.~Ranzato, R.~Hadsell, M.F. Balcan, and H.~Lin, editors, {\em Advances in Neural Information Processing Systems}, volume~33, pages 6696--6707. Curran Associates, Inc., 2020.

\bibitem{yildiz2019ode2vae}
Cagatay Yildiz, Markus Heinonen, and Harri Lahdesmaki.
\newblock {ODE2VAE: Deep generative second order ODEs with Bayesian neural networks}.
\newblock {\em Advances in Neural Information Processing Systems}, 32, 2019.

\bibitem{portwood2019turbulence}
Gavin~D Portwood, Peetak~P Mitra, Mateus~Dias Ribeiro, Tan~Minh Nguyen, Balasubramanya~T Nadiga, Juan~A Saenz, Michael Chertkov, Animesh Garg, Anima Anandkumar, Andreas Dengel, et~al.
\newblock {Turbulence forecasting via neural ode}.
\newblock {\em arXiv preprint arXiv:1911.05180}, 2019.

\bibitem{xie2019neural}
Xiang Xie, Ajith~Kumar Parlikad, and Ramprakash~Srinivasan Puri.
\newblock {A neural ordinary differential equations based approach for demand forecasting within power grid digital twins}.
\newblock In {\em 2019 IEEE International Conference on Communications, Control, and Computing Technologies for Smart Grids (SmartGridComm)}, pages 1--6. IEEE, 2019.

\bibitem{srivastava2015unsupervised}
Nitish Srivastava, Elman Mansimov, and Ruslan Salakhudinov.
\newblock {Unsupervised learning of video representations using lstms}.
\newblock In {\em International conference on machine learning}, pages 843--852. PMLR, 2015.

\bibitem{lotter2016deep}
William Lotter, Gabriel Kreiman, and David Cox.
\newblock {Deep predictive coding networks for video prediction and unsupervised learning}.
\newblock {\em arXiv preprint arXiv:1605.08104}, 2016.

\bibitem{li2018disentangled}
Yingzhen Li and Stephan Mandt.
\newblock {Disentangled sequential autoencoder}.
\newblock {\em arXiv preprint arXiv:1803.02991}, 2018.

\bibitem{locatello2019challenging}
Francesco Locatello, Stefan Bauer, Mario Lucic, Gunnar Raetsch, Sylvain Gelly, Bernhard Sch{\"o}lkopf, and Olivier Bachem.
\newblock {Challenging common assumptions in the unsupervised learning of disentangled representations}.
\newblock In {\em international conference on machine learning}, pages 4114--4124. PMLR, 2019.

\bibitem{daxberger2019bayesian}
Erik Daxberger and Jos{\'e}~Miguel Hern{\'a}ndez-Lobato.
\newblock Bayesian variational autoencoders for unsupervised out-of-distribution detection.
\newblock {\em arXiv preprint arXiv:1912.05651}, 2019.

\bibitem{clevert2015fast}
Djork-Arn{\'e} Clevert, Thomas Unterthiner, and Sepp Hochreiter.
\newblock Fast and accurate deep network learning by exponential linear units (elus).
\newblock {\em arXiv preprint arXiv:1511.07289}, 2015.

\bibitem{baker2018mechanistic}
Ruth~E Baker, Jose-Maria Pena, Jayaratnam Jayamohan, and Antoine J{\'e}rusalem.
\newblock Mechanistic models versus machine learning, a fight worth fighting for the biological community?
\newblock {\em Biology letters}, 14(5):20170660, 2018.

\bibitem{maino2016mechanistic}
James~L Maino, Jacinta~D Kong, Ary~A Hoffmann, Madeleine~G Barton, and Michael~R Kearney.
\newblock Mechanistic models for predicting insect responses to climate change.
\newblock {\em Current opinion in insect science}, 17:81--86, 2016.

\bibitem{gneiting2007strictly}
Tilmann Gneiting and Adrian~E Raftery.
\newblock {Strictly proper scoring rules, prediction, and estimation}.
\newblock {\em JASA}, 102(477):359--378, 2007.

\bibitem{bracher2019multibin}
Johannes Bracher.
\newblock {On the multibin logarithmic score used in the FluSight competitions}.
\newblock {\em PNAS}, 116(42):20809--20810, 2019.

\end{thebibliography}

% All done. \o/
\end{document}